\documentclass[twoside,11pt]{article}

\usepackage{jmlr2e}
\usepackage{color, colortbl}
\usepackage{graphicx}  
\usepackage[font=small]{caption}
\usepackage[font=scriptsize]{subfig}
\usepackage{subfloat}
\usepackage{wrapfig}
\usepackage{tabularx}
\usepackage{graphicx}
\usepackage{epstopdf} 
\usepackage{natbib}
\usepackage{array}
\usepackage{enumitem}
\usepackage{fmtcount} 
\usepackage{multirow}
\usepackage{booktabs,xcolor,siunitx}
\definecolor{lightgray}{gray}{0.9}
\usepackage{ragged2e}
\usepackage{dsfont}

\usepackage[utf8]{inputenc}
\usepackage[english]{babel}
\usepackage{amsmath}

\usepackage{lineno}
\usepackage[hidelinks]{hyperref}
\modulolinenumbers[5]

\usepackage{algorithm}
\usepackage[noend]{algpseudocode}

\definecolor{LightCyan}{rgb}{0.88,1,1}

\jmlrheading{X}{2017}{00-00}{00/00}{00/00}{Nadia Figueroa and Aude Billard}

\ShortHeadings{Transform-Invariant Non-Parametric Clustering and Segmentation}{Figueroa and Billard}
\firstpageno{1}

\begin{document}

\title{Transform-Invariant Non-Parametric Clustering of Covariance Matrices and its Application to Unsupervised Joint Segmentation and Action Discovery}

\author{\name Nadia Figueroa \email nadia.figueroafernandez@epfl.ch \\
       \name Aude Billard \email aude.billard@epfl.ch \\
       \addr Learning Algorithms and Systems Laboratory (LASA)\\
       	\'Ecole Polytechnique F\'ed\'erale de Lausanne (EPFL)\\
       	Lausanne, Switzerland - 1015.}
\editor{X}
\maketitle

\begin{abstract}

In this work, we tackle the problem of \textit{transform-invariant unsupervised learning} in the space of Covariance matrices and applications thereof. We begin by introducing the Spectral Polytope Covariance Matrix (SPCM) Similarity function; a similarity function for Covariance matrices, invariant to any type of transformation. We then derive the SPCM-CRP mixture model, a \textit{transform-invariant} non-parametric clustering approach for Covariance matrices that leverages the proposed similarity function, spectral embedding and the distance-dependent Chinese Restaurant Process (dd-CRP) \citep{Blei:JMLR:2011}. The scalability and applicability of these two contributions is extensively validated on real-world Covariance matrix datasets from diverse research fields. Finally, we couple the SPCM-CRP mixture model with the Bayesian non-parametric Indian Buffet Process (IBP) - Hidden Markov Model (HMM) \citep{Fox:NIPS:2009}, to jointly segment and discover \textit{transform-invariant} action primitives from complex sequential data. Resulting in a topic-modeling inspired hierarchical model for unsupervised time-series data analysis which we call ICSC-HMM (IBP Coupled SPCM-CRP Hidden Markov Model). The ICSC-HMM is validated on kinesthetic demonstrations of uni-manual and bi-manual cooking tasks; achieving unsupervised human-level decomposition of complex sequential tasks.\\
\end{abstract}


\section{Introduction}
The \textit{Gaussian distribution} is one of the most widely-used representations of data in any field of science and engineering. The reason for its popularity can be explained from an information-theoric point of view. Given only its first and second moments (i.e. the mean and Covariance) one can describe a distribution of data-points with maximum entropy and minimal assumptions \citep{Cover:EIT:1991}. To recall, it is defined by a probability density function over a random vector $\mathbf{x} \in \mathds{R}^{N}$ whose density $\mathcal{N}(\mathbf{\mu},\mathbf{\Sigma})$ can be estimated with a mean $\mathbf{\mu} \in \mathds{R}^{N}$ and a \underline{Covariance matrix} $\mathbf{\Sigma} \in \mathds{R}^{NxN}$ as follows,
\begin{equation}
\label{eq:gaussian}
f_X(\mathbf{x};\mathbf{\mu},\mathbf{\Sigma}) = \frac{1}{(2\pi)^{d/2}|\mathbf{\Sigma}|^{1/2}} \exp \left\lbrace -\frac{1}{2}(\mathbf{x} - \mu)^{T}\mathbf{\Sigma}^{-1}(\mathbf{x} - \mathbf{\mu}) \right\rbrace.
\end{equation} 
where $\mathbf{\mu} = \mathbb{E}[\mathbf{x}]$ is the formalization of the average value and the Covariance matrix $\mathbf{\Sigma} = \mathbb{E}\left[ (\mathbf{x} - \mathbf{\mathbb{E}[\mathbf{x}]})(\mathbf{x} - \mathbf{\mathbb{E}[\mathbf{x}]})^T \right]$ is the generalization of Covariance in $N$-dimensional space. If we assume the data is centered $\sum_{i=1}^{M}\mathbf{x}_i = 0$ for $M$ samples, the information compressed in the Covariance matrix $\mathbf{\Sigma}$ is sufficient to describe the variance of the distribution in all spatial directions; and consequently, the distribution itself:
\begin{equation}
\label{eq:zero-mean-gaussian}
f_X(\mathbf{x};0,\mathbf{\Sigma}) = \frac{1}{(2\pi)^{d/2}|\mathbf{\Sigma}|^{1/2}} \exp \left\lbrace -\frac{1}{2}\mathbf{x}^{T}\Sigma^{-1}\mathbf{x} \right\rbrace.
\end{equation}  
This has lead researchers to either: (i) fit/assume their data is distributed by \eqref{eq:gaussian} or \eqref{eq:zero-mean-gaussian} or (ii) use the sample Covariance matrix $\mathbf{C}$ directly as a descriptive representation of data:
\begin{equation}
\label{eq:sample_cov}
\mathbf{C} = \frac{1}{M} \sum_{i=1}^{M} (\mathbf{x}_i - \bar{\mathbf{x}} )(\mathbf{x}_i - \bar{\mathbf{x}} )^T,
\end{equation}
where $\bar{\mathbf{x}}$ is the sample mean. The use of \eqref{eq:gaussian}, \eqref{eq:zero-mean-gaussian} and \eqref{eq:sample_cov} is inherent in a myriad of \textit{computer vision, action recognition, medical imaging} and \textit{robotics} applications. For example, in \textit{computer vision} and \textit{action recognition}, Covariance matrices are used to characterize statistics and fuse features for texture/action/face recognition, people/object tracking, camera calibration, skeleton-based human action recognition from motion capture data, to name a few \citep{Tuzel:ECCV:2006,Tosato:ECCV:2010,Vemulapalli:CVPR:2013,Wang:ICCV:2015,Cavazza:ICPR:2016}. In the \textit{medical imaging} community, the analysis and interpolation of Covariance is extremely important as Diffusion Tensors (a type of SPD matrices\footnote{Covariance matrices are Symmetric Positive Definite (SPD). In this work, we consider any dataset of SPD matrices as datasets of Covariance matrices.}) are used to identify regions of similar biological tissue structures through the diffusion of water particles in the brain \citep{Dryden:AAS:2009,Cherian:TPAMI:2013}. 

In \textit{robotics}, the geometrical representation of the Covariance matrix (i.e. an $N$-dimensional hyper-ellipsoid) is widely used to represent uncertainties in sensed data for planning, localization and recognition tasks \citep{Thrun:PR:2005,Miller:CASE:2015}. Furthermore, $N$-dimensional hyper-ellipsoids are also used in robotic manipulation applications to represent parametrizations for tasks involving interaction or variable impedance learning \citep{Friedman:Cortex:2007}. In \citet{Kronander:ICRA:2012,Ajoudani:ICRA:2015}, \textit{stiffness} ellipsoids, representing the stiffness of the robot's end-effector, are used to characterize impedance controllers. Further, in \citet{Li:JRA:1988,El-Khoury:RAS:2015}, ellipsoids are used to model tasks in the wrench space of a manipulated object or tool\footnote{Otherwise known as the Task Wrench Space (TWS) \citep{Li:JRA:1988}.} , to represent the principal directions of the forces/torques exerted on an object to achieve a manipulation task. Finally, in many  Learning from Demonstration (LfD) \citep{Argall:RAS:2009,Billard:SHR:2008} approaches,  Gaussian Mixture Models (GMM) and Hidden Markov Models (HMM), are regularly used for motion modeling, action recognition and segmentation \citep{Calinon:SMC:2007,Khansari:TRO:2011,Calinon:ISR:2015}.

Estimating such Covariance matrices (or Gaussian-\textit{distributed} probabilistic models), for any of the previously mentioned applications, relies on careful \textit{acquisition} and \textit{pre-processing} of data. Often, data acquisition procedures generate data subject to a relative \textit{transformation}, due to changes in rotation, translation, scaling, duration (in the case of time-series), shearing, etc. Covariance matrices, although highly descriptive, do not explicitly decouple such transformations. For example, spatial transformations can be exhibited in Covariance features from image sets, where objects/faces/textures are translated, rotated and sheared wrt. the center of the image or camera origin. DTI (Diffusion Tensor Images) can also display spatial transformations, depending on the diffusivity scale used to generate them or the difference in subject tissue texture. Moreover, in continuous data, where GMMs or HMMs are used to model human activities or manipulation tasks, the data might exhibit \textit{transformations}; due to changes in initial positions, reference frames or contexts. 

To exploit the exponential increase of data available on-line, it is crucial to have algorithms that are robust to such \textit{transformations}, i.e. are \textit{transform-invariant}. The most typical approach to deal with these nuisance transformations is to manually \textit{pre-process} the collected data, \textit{prior} to applying any machine learning algorithm. This, however, requires for previous knowledge of the corresponding \textit{transformations} and much human intervention. Another approach is to jointly estimate the transformations while learning the structure of the dataset, these approaches require either (i) providing a list of possible transformations and approximating the \textit{optimal} parameters through EM-like algorithms \citep{Frey:TPAMI:2003,Frey:NIPS:2001} or (ii) sampling the transformations from a specified parameter-space through Bayesian non-parametric approaches \citep{Sudderth:NIPS:2005,Hu:ICML:2012}. 

\begin{figure*}[!t]
\centering
\begin{minipage}{\linewidth}
	\centering
	\includegraphics[trim={2cm 1cm 2cm 0.5cm},clip,width=0.49\linewidth]{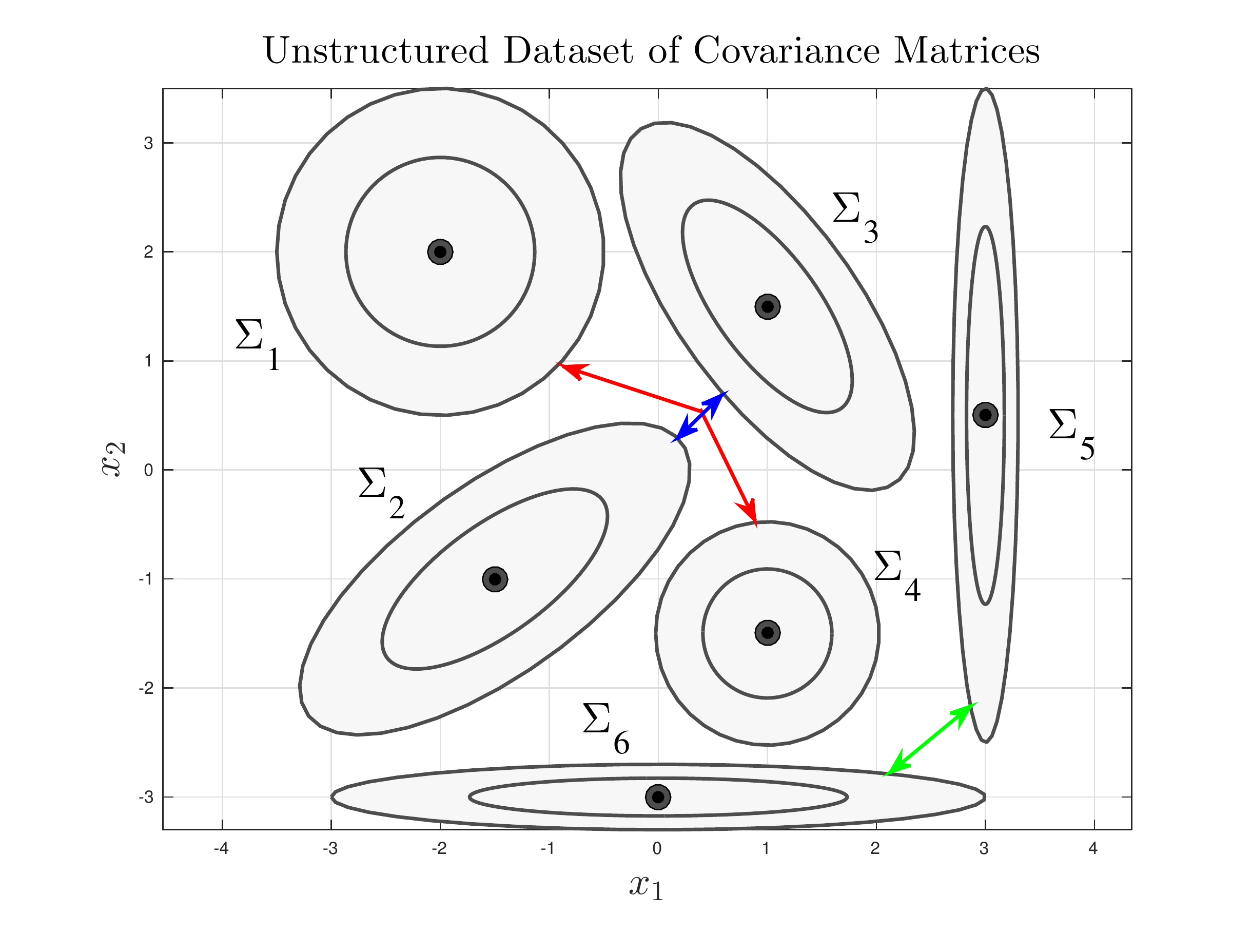}	
	\includegraphics[trim={2cm 1cm 2cm 0.5cm},clip,width=0.49\linewidth]{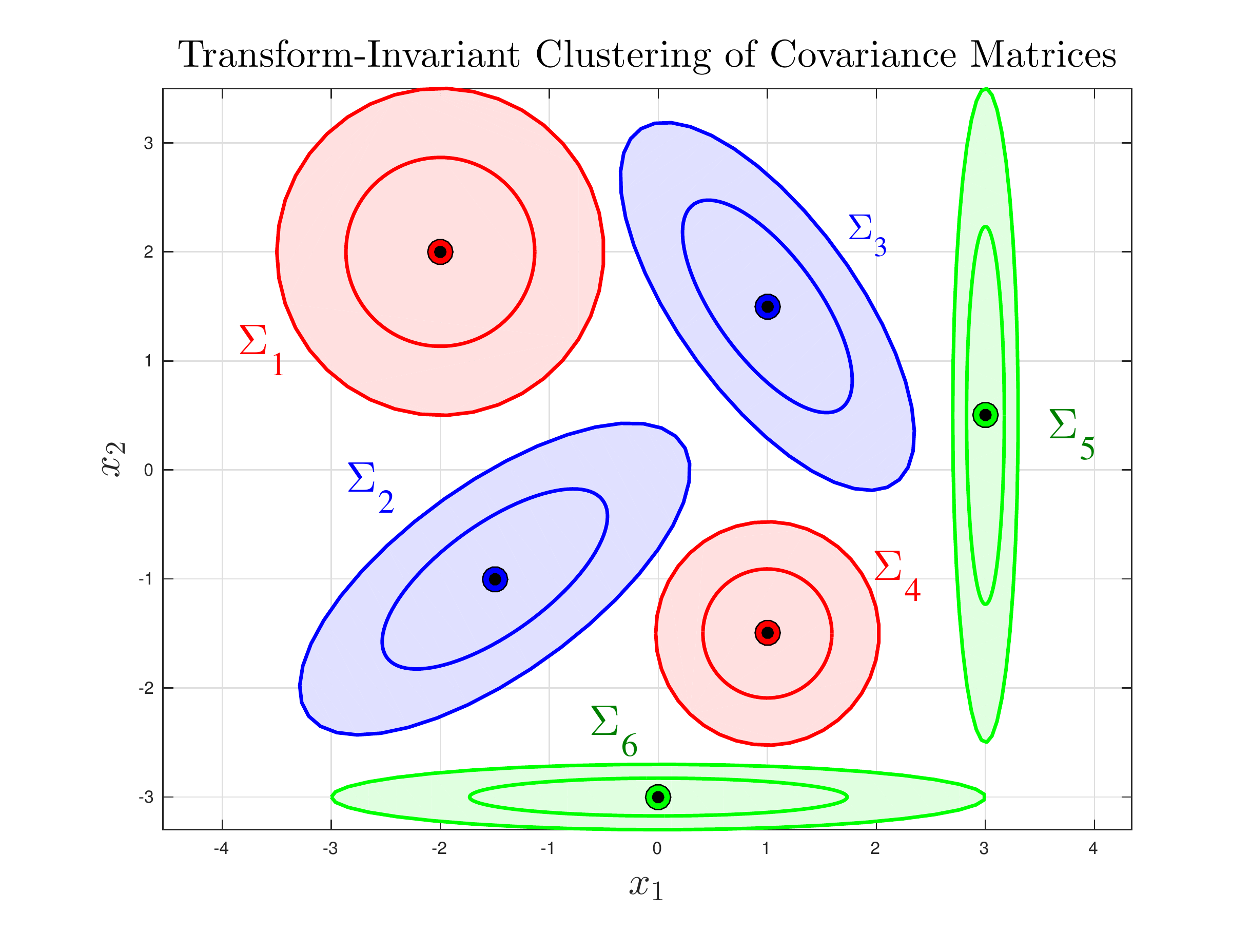}
	\vspace{-5pt}
	\caption{\small \textit{Transform-Invariant Non-Parametric Clustering of Covariance Matrices}: \textbf{(left)} Given a dataset of $M$ Covariance matrices; $\mathbf{\Theta} = \{\mathbf{\Sigma}_1, \dots, \mathbf{\Sigma}_M\}$ where  $\mathbf{\Sigma}_i \in \mathds{R}^{N\times N}$ describing data in $\mathbf{x} \in \mathds{R}^{N}$; \textit{with} or \textit{without} a corresponding location $\mu = \{\mu_1, \dots, \mu_M\}$, we assume \textit{structural similarity} between the Covariance matrices (denoted by the colored arrows)
	\textbf{(right)} The dataset can be described with $K (3)$ transform-invariant clusters: $\mathbf{\Theta} = \{\textcolor{black}{\mathbf{\Sigma}_2, \mathbf{\Sigma}_3} | \textcolor{black}{\mathbf{\Sigma}_1,\mathbf{\Sigma}_4} |\textcolor{black}{\mathbf{\Sigma}_5,\mathbf{\Sigma}_6}\}$; i.e. clustering is \textit{invariant to transformations (scaling, translation, rotation)}.	\label{fig:problem1}}
\end{minipage}
\end{figure*}

The \textit{latter} category, models \textit{transformed} groups of features in images using the transformed Indian Buffet Process (tIBP) and transformed Dirichlet Process (tDP) priors for latent feature models and mixture models, respectively. These approaches, although elegant, rely on sampling simple 2D transformation parameters, such as pixel/feature locations, which are not scalable to $N$-dimensional Covariance matrices. Moreover, they are limited to transformations in an image-to-image sense, not within images. Hence, one must know \textit{a priori} that the data-points/features come from a different source/context. In our work, rather than assuming that we know \textit{a priori} which data-points come from a different source/context and how many sources/contexts exist in the dataset, we seek to discover this from the data itself. In other words, we are interested in tackling the problem of \textit{transform-invariant} \textit{non-parametric} clustering of Covariance matrices (see Figure \ref{fig:problem1}).

\subsection{Transform-Invariance and Non-Parametric Clustering of Covariance Matrices}
The problem illustrated in Figure \ref{fig:problem1} is posed as follows; given a dataset of $M$ Covariance matrices; $\mathbf{\Theta} = \{\mathbf{\Sigma}_1, \dots, \mathbf{\Sigma}_M\}$ where  $\mathbf{\Sigma}_i \in \mathds{R}^{N\times N}$ describing data in $\mathbf{x} \in \mathds{R}^{N}$; \textit{with} or \textit{without} a corresponding location $\mathbf{\mu} = \{\mathbf{\mu}_1, \dots, \mathbf{\mu}_M\}$ (resulting in a Gaussian distribution if given), we seek to describe the dataset with $K$ \textit{transform-invariant} clusters. Hence, such clustering is \textit{invariant} to any type of transformations \textit{(scaling, translation, rotation)} on the Covariance matrices. Moreover, instead of setting the desired cluster value $K$ or the possible transformations between clusters, we want to \textit{discover} this from the dataset. 

One can imagine a plethora of applications that would benefit from such an algorithm; e.g. (i) finding groups or clusters of objects/texture/faces in unstructured/unlabeled datasets, (ii) segmenting DTI images without prior knowledge of the brain region, (iii) discovering \textit{task primitives} from large datasets of task ellipsoids recorded from different tools or subjects, (iv) clustering streams of continuous transformed data, etc. In order to achieve this desiderata, we must address two problems: (i) transform-invariant similarity of Covariance matrices and (ii) non-parametric clustering over (dis)-similarities. 

\subsubsection{Transform-Invariant Covariance Matrix Similarity}
The advantages of using Covariance matrices to represent data can be over-shadowed by their non-Euclidean topology. Most machine learning algorithms (be it \textit{supervised} or \textit{unsupervised}) rely on computing distances/norms/similarities in feature space, assuming the features are i.i.d. from an underlying distribution in $\mathds{R}^N$ Euclidean space. Unfortunately, SPD (and consequently Covariance) matrices lie on a special Riemannian manifold, $\mathbf{\Sigma} \in \mathcal{S}_{++}^{N}$, and should be treated as such. For example, baseline learning algorithms, such as K-NN (Nearest Neighbor) and K-Means, rely on a pair-wise distance $\Delta(\mathbf{x},\mathbf{x}')$ between features; typically a Minkowski metric $L_p(\mathbf{x},\mathbf{x}') = \left(\sum_{i=1}^{N} |x_i - x_i'|^p\right)^{1/p}$. When using such Euclidean geometry, the distance between two Covariance matrices $\mathbf{\Sigma}_i$ and $\mathbf{\Sigma}_j$ would be, $\Delta_{ij}(\mathbf{\Sigma}_i,\mathbf{\Sigma}_j) = || \mathbf{\Sigma}_i - \mathbf{\Sigma}_j ||_2 = \sqrt{\text{trace}(\mathbf{\Sigma}_i - \mathbf{\Sigma}_j)^T(\mathbf{\Sigma}_i - \mathbf{\Sigma}_j)}$.
where $||\mathbf{X}||_2=\sqrt{\text{trace}(\mathbf{X}^T\mathbf{X})}$ is the Euclidean (Frobenius) norm, the typical distance metric used on matrices of $\mathbf{X} \in \mathds{R}^{N\times N}$. Research has shown,  that using such standard $L_p$-norms on $\mathcal{S}_{++}^{N}$ results in very poor accuracy \citep{Dryden:AAS:2009,Cherian:TPAMI:2013}. Moreover, when trying to interpolate or compute Geometric means, Euclidean metrics are prone to swelling and can lead to non-positive semi-definite estimates \citep{Arsigny:JMAA:2006}. Several approaches have been proposed to over-come this, which we categorize into two families of approaches: (i) metric substitution and (ii) kernel methods. 

\textit{Metric substitution} approaches rely on using tailor-made non-Euclidean and Riemannian distances or metrics together with classical learning algorithms. Several non-Euclidean measures of similarity that deal with the unique structure of Covariance matrices have been proposed, for the sake of brevity, we introduce four of the most widely used similarity functions (Table \ref{tab:metrics}). The Affine Invariant Riemannian Metric (AIRM) \citep{Pennec:IJCV:2006} and the Log-Euclidean Riemannian Metric (LERM) \citep{Arsigny:MRM:2006} are the most commonly used metrics, as they calculate a distance analogous to the Frobenius norm while taking into account the curvature of the Riemannian manifold. Whereas, the Jensen-Bregman LogDet Divergence (JBLD) \citep{Cherian:TPAMI:2013} is a matrix form of the symmetrized Jensen-Bregman divergence. The Kullback-Leibler Divergence Metric (KLDM) \citep{Moakher:VPRF:2006}, on the other hand, uses symmetrized $f$-divergences to compute distances between covariance matrices. This list of similarity functions have both advantages and drawbacks, regarding accuracy and computational efficiency\footnote{For a thorough comparison refer to \citet{Cherian:TPAMI:2013} and \citet{Dryden:AAS:2009}}. They all posses most of the desired properties a similarity metric should have, such as non-negativity, definiteness, symmetry, affine-invariance, triangle inequality, etc. However, none of them explicitly ensure the property of \textit{transform-invariance} 
\begin{equation}
\Delta_{ij}(\mathbf{\Sigma}_i,\mathbf{\Sigma}_j) = 0 \hspace{5pt} \Leftrightarrow \hspace{5pt} \mathbf{\Sigma}^{(j)} = \mathbf{R}\mathbf{V}^{(i)}(\gamma\mathbf{\Lambda}^{(i)})(\mathbf{R}\mathbf{V}^{(i)})^{T}    
\end{equation}
where $\gamma \in \Re$ is a scaling factor, $\mathbf{R} \in \mathds{R}^{N \times N}$ is a rotation matrix and $\mathbf{\Sigma}_i = \mathbf{V}^{(i)}\mathbf{\Lambda}^{(i)}(\mathbf{V}^{(i)})^{T}$ is the Eigenvalue decomposition of $\mathbf{\Sigma}_i$. 
\begin{table}[!tbp]	
	\small
	\centering
	\begin{tabular}{|c|c|}
		\hline
		Type & $\Delta_{ij}(\mathbf{\Sigma}_i,\mathbf{\Sigma}_j)$ \\
		\hline
		AIRM \citep{Pennec:IJCV:2006} & $\left|\left|\log(\mathbf{\Sigma}_i^{-1/2}\mathbf{\Sigma}_j\mathbf{\Sigma}_i^{-1/2})\right|\right|$ \\
		LERM \citep{Arsigny:MRM:2006} & $\left|\left|\log(\mathbf{\Sigma}_i) - \log(\mathbf{\Sigma}_j)\right|\right|$ \\
		KLDM \citep{Moakher:VPRF:2006} & $\frac{1}{2}\texttt{trace}(\mathbf{\Sigma}_i^{-1}\mathbf{\Sigma}_j + \mathbf{\Sigma}_j^{-1}\mathbf{\Sigma}_i-2I)$\\
		JBLD \citep{Cherian:TPAMI:2013} & $\log  \left|\frac{\mathbf{\Sigma}_i + \mathbf{\Sigma}_j}{2}\right| - \frac{1}{2}\log\left|\mathbf{\Sigma}_i\mathbf{\Sigma}_j\right| $  \\\hline
	\end{tabular}
		\caption{Standard Covariance Matrix Similarity Functions \label{tab:metrics}}
\end{table}
The \textit{latter} approaches involve \textit{kernel methods} which project the SPD matrices to a higher dimensional feature space via defined kernels or kernels learnt from the data \citep{Vemulapalli:CVPR:2013,Jayasumana:CVPR:2013,Huang:ICML:2015}. Although successful, most of these approaches are either only applicable in a supervised learning setting or are not robust to \textit{transformations}. 

In this work, we focus on the Spectra of Covariance matrices and propose a \textit{transform-invariant} similarity function which is inspired by Spectral Graph Theory and a geometrical intuition of the embedded subspace of the eigenvectors and eigenvalues. We introduce it as the Spectral Polytope Covariance Matrix \textbf{(SPCM)} similarity function in Section \ref{sec:full_spcm}.

\subsubsection{Bayesian Non-Parametric Clustering over Similarities} 
Given a \textit{transform-invariant} similarity function, we must find an appropriate \textbf{clustering} algorithm for our task. Since our dataset consists of abstract objects, i.e. Covariance matrices, we could use similarity-based clustering algorithms such as Spectral Clustering \citep{Ng:NIPS:2001}, kernel K-means \citep{Dhillon:KDD:2004}, Affinity Propagation \citep{Frey:Science:2007}, among others. Although robust in nature, these algorithms require some heavy parameter tuning; typically through \textit{model selection} or \textit{grid search}. An alternative to this, is to adopt \textit{non-parametric} models. We do not refer to \textit{non-parametric} as methods with ``no parameters", rather to models whose complexity is inferred from the data automatically, allowing for the number of parameters to grow w.r.t. the number of data points. \textit{Bayesian non-parametric} methods provide such an approach. When clustering with mixture models, instead of doing model selection to find the \textit{optimal} number of clusters, a Dirichlet processes ($\mathcal{DP}$) prior (or one of its representations\footnote{The $\mathcal{DP}$ can be constructed through several representations. In this work, we adopt the the Chinese Restaurant Process ($\mathcal{CRP}$) construction.}) is used to construct an infinite mixture model. By evaluating an \textit{infinite} mixture model on a \textit{finite} set of samples, one can estimate the number of clusters needed from the data itself, whilst allowing new data-points to form new clusters \citep{Gershman:JMP:2012}. 

To deal with the fact that our data-points (Covariance matrices $\mathbf{\Sigma}_i$) cannot be treated as points in Euclidean space, we use spectral dimensionality reduction to map Covariance matrices $\mathbf{\Sigma}_i$ into a lower-dimensional Euclidean space, i.e. $\mathbf{y}_i = f(\mathbf{\Sigma}_i)$ where $f(\cdot):\mathcal{S}_{++}^{N}\rightarrow\mathds{R}^{P}$, induced by their similarity matrix $\mathbf{S} \in \mathds{R}^{M \times M}$, where $M$ is the number of Covariance matrices in our dataset. In standard spectral clustering approaches, the dimensionality ($P$) of this mapping is typically set by the user or through \textit{model selection}. In this work, we propose a spectral non-parametric variant of this approach, which relaxes the assumption that the number of leading eigenvectors $P$ is equivalent to the number of $K$ clusters in the data; i.e. $K=P$. This assumption is too restrictive, as it is only truly valid in ideal cases where we have sparse similarity matrices $\mathbf{S}$, which is never the case in real-world data \citep{Poon:UAI:2012}. By relaxing this assumption, this leads us to two new subproblems: 1) \textit{What is the appropriate dimensionality of the spectral embedding?} and 2) \textit{What are the optimal clusters in the lower-dimensional space?}\footnote{The general problem of finding the correct dimensionality and number of clusters in Spectral Clustering approaches has been referred to as \textit{rounding} \citep{Poon:UAI:2012}.}. 

Problem 1) is tackled by proposing an unsupervised spectral embedding algorithm based on a probabilistic analysis of the eigenvalues of the Similarity matrix (Section \ref{sec:spcm_crp}). We then address problem 2) by applying a \textit{Bayesian non-parametric} clustering algorithm on the spectral subspace projections of the Covariance matrices, while exploiting the similarity information on the original (Covariance matrix) space. We achieve this by adapting the distance-dependent Chinese Restaurant Process (dd-$\mathcal{CRP}$) \cite{Blei:JMLR:2011} \footnote{A distribution over partitions that allows for dependencies (from space, time and network connectivity) between data-points.} to the proposed Spectral Polytope Covariance Matrix \textbf{(SPCM)} similarity function and the points on the spectral sub-space ($\mathbf{Y} \in \mathds{R}^{M\times P}$), leading to a SPCM \textit{similarity}-dependent Chinese Restaurant Process, which we refer to as the SPCM-$\mathcal{CRP}$ mixture. 

\begin{figure*}[!t]
	\centering
	\begin{minipage}{0.53\textwidth}
		\centering
		\includegraphics[trim={0.2cm 0.75cm 1.75cm 0.35cm},clip,width=\linewidth]{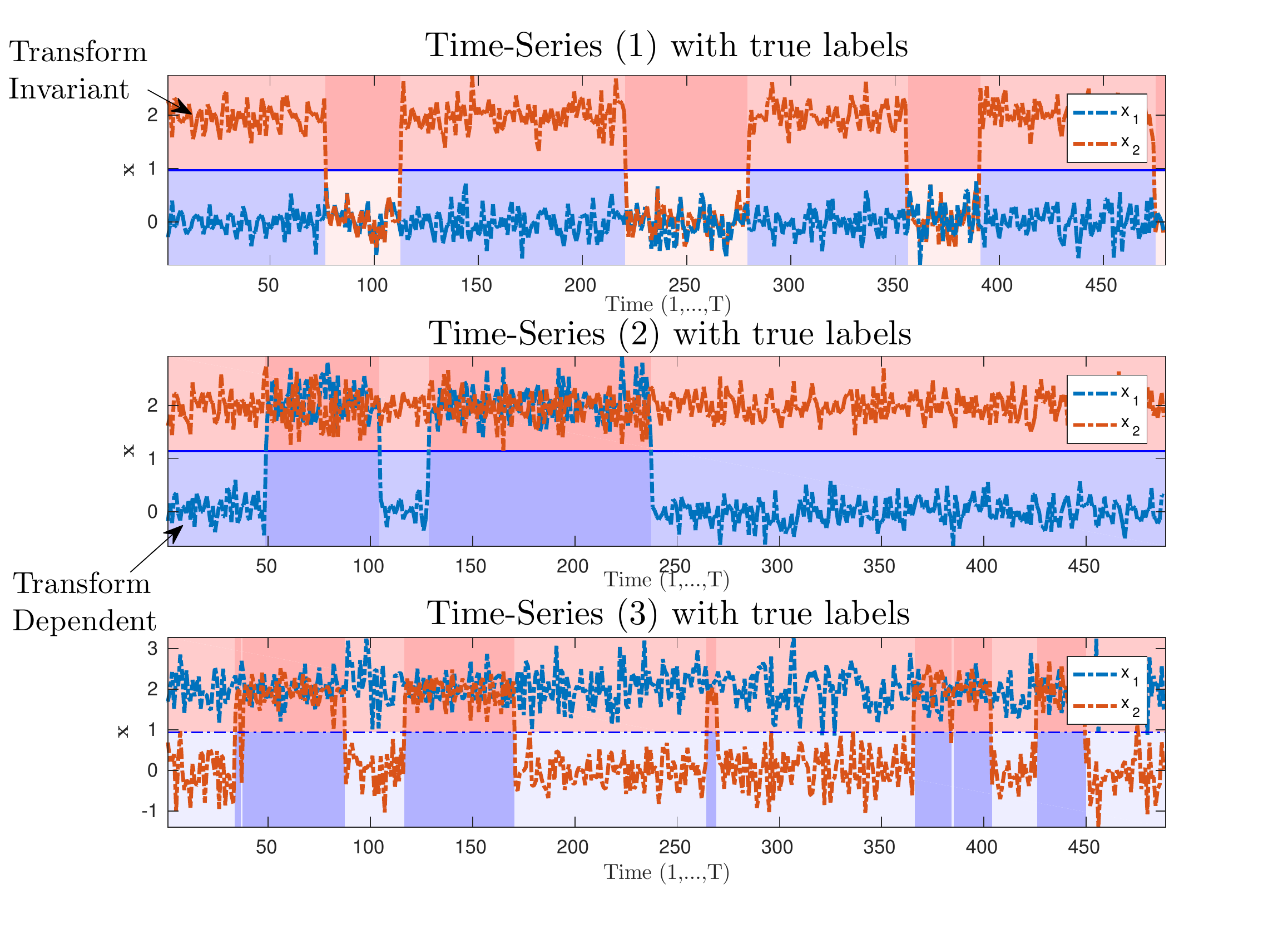}
	\end{minipage}\begin{minipage}{0.465\textwidth}
	\centering
	\includegraphics[trim={0.5cm 0cm 1.5cm 0cm},clip,width=\linewidth]{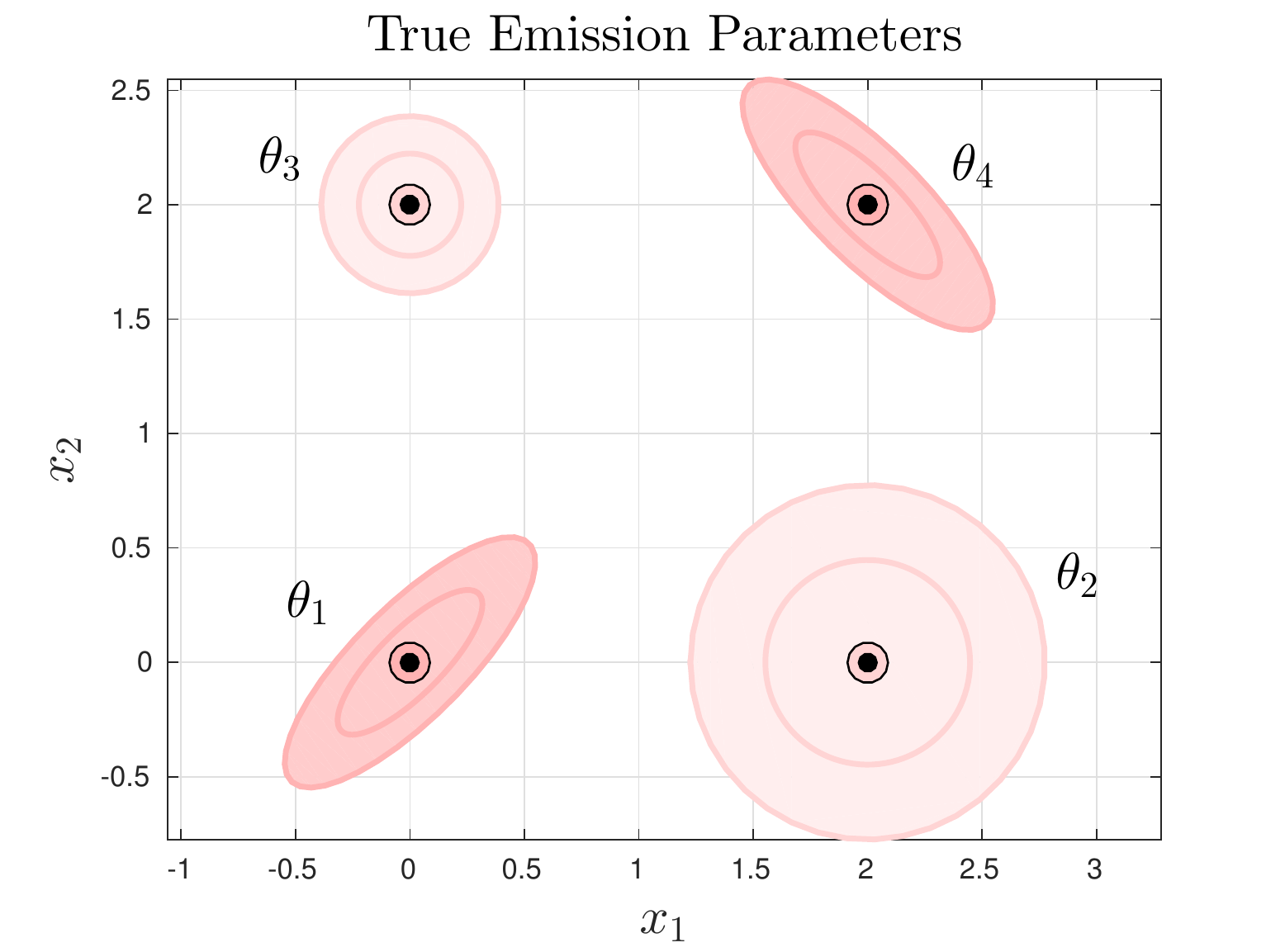}
\end{minipage}
\caption{\small \textit{Joint Segmentation and Action Discovery} of a 2D dataset composed of 3 time-series sampled from 2 \textbf{unique} emission models $\theta_1 = \left\{\mathbf{\mu}_1,\mathbf{\Sigma}_1 \right\},\theta_2 = \left\{\mathbf{\mu}_2,\mathbf{\Sigma}_2 \right\} $ subject to transformations $f_1(\mu, \mathbf{\Sigma};\mathbf{t},\lambda) = \left\{\mathbf{\mu} + \mathbf{t},\mathbf{V}(\lambda\mathbf{\Lambda})\mathbf{V}^T \right\},f_2(\mu, \mathbf{\Sigma}; \mathbf{t}, \mathbf{R}) =  \left\{\mathbf{\mu} + \mathbf{t},(\mathbf{R}\mathbf{V})\mathbf{\Lambda}(\mathbf{R}\mathbf{V})^T \right\} $, for $\mathbf{\Sigma} = \mathbf{V}\mathbf{\Lambda}\mathbf{V}^T$, resulting in a set of transform-dependent emission models  $\Theta = \{\theta_1,\theta_2,\theta_3 = f_1(\theta_2),\theta_4 = f_2(\theta_3)\}$. The goal is to jointly extract the \textit{transform-dependent} segmentation results \textbf{(left)} (shown as the bottom colors in the shaded times-series) and group them into \textit{transform-invariant} clusters; i.e. the top colors in the shaded time-series, which correspond to the colors of the true emission models \textbf{(right)}.
	\label{fig:problem2} }
\end{figure*}

\subsection{Segmentation and Transform-Invariant Action Discovery}
The problem of \textit{transform-invariance} is also exhibited in streams of continuous data. Imagine a set of time-series, corresponding to streams of motion sensors (e.g. kinect information used to control video games or teleoperate robots) or motion/interaction signals from the robotics domain, such as position, velocity, orientation, forces and torques of an end-effector or a hand; representing a human (or robot) executing a complex sequence of actions. Typically, the challenge while analyzing such streams of data is to segment or decompose them into meaningful actions or states. However, such time-series might be subject to \textit{transformations}, due to changes of Reference Frame, change of context or even changes in the execution of the task itself (i.e. changes of position and orientation of the user). We posit that \textit{transform-invariance} is an interesting problem in such time-series analysis scenarios. Following we describe the challenges we tackle in this work in a 2D illustrative example.\\

\noindent \texttt{Illustrative example}: Assume a set of $M$ 2D time-series with varying length $T= \{T^{(1)}, \cdots, T^{(M)}\}$  and switching dynamics $\pi= \{\pi^{(1)}, \cdots, \pi^{(M)}\}$, sampled from 2 \textbf{unique} Gaussian emission models  $\theta_1,\theta_2$ subject to transformations $f_1(\cdot),f_2(\cdot)$ resulting in a set of transform-dependent emission models  $\Theta = \{\theta_1,\theta_2,\theta_3 = f_1(\theta_2),\theta_4 = f_2(\theta_3)\}$, as shown in Figure \ref{fig:problem2}. The problem that we would like to tackle is to jointly: (i) recover the segmentation points of all time-series, (ii) recover the \textit{transform-dependent} emission models $\Theta$ and (iii) capture the similarities across these models to uncover the \textbf{true} sub-set of \textit{transform-invariant} emission models. All of this, without explicitly knowing or modeling the transformations that have caused the variation in the observed time-series nor the expected number of hidden states/emission models. 

Within the \textit{robotics} community, two probabilistic approaches to tackle the segmentation problem prevail: (i) Gaussian Mixture Models \citep{Kruger:ICRA:2012,Lee:AR:2015} and (ii) Hidden Markov Models \citep{Takano:HUM:2006,Kulic:TRO:2009, Butterfield:HUM:2010, Niekum:IROS:2013}. In the \textit{former}, segmentation points are extracted by fitting a (\textit{finite} or \textit{non-parametric}) GMM to the set of demonstrations directly, indicating that each component is an \textit{action} in the sequence. This approach is ill-suited to our problem setting as it segments the trajectories purely on a spatial sense, disregarding state dynamics, transition dynamics or correlations between the trajectories. For this reason, the \textit{latter} approach is more successful in segmentation problems, as transition dynamics and action sequencing are explicitly handled through the \textit{Markovian} state chain. Two types of HMM variants have been mostly used in these approaches: (1) online HMMs \citep{Takano:HUM:2006,Kulic:TRO:2009} and (2) Bayesian Non-Parametric HMMs \citep{Butterfield:HUM:2010, Niekum:IROS:2013}. In this work, we favor the \textit{latter} approaches as they are well-suited for sets of time-series with an unknown number of states and \textit{shared} or \textit{common} emission models. Such approaches include the Hierarchical Dirichlet Process Hidden Markov Model (HDP-HMM \citep{Teh:ASA:2006}) and the Beta Process - Hidden Markov Model (BP-HMM) \citep{Fox:NIPS:2009}. The HDP-HMM is nothing but a collection of Bayesian HMMs (each modeling one time-series) with the same number of states following exactly the same sequence. This is a major set-back for our desiderata, as we do not assume a fixed number of actions per trajectory (see Figure \ref{fig:problem2}). On the contrary, we would like to avoid such restricting assumptions. The BP-HMM, relaxes these assumptions, by instead using the Indian Buffet Process (IBP), induced by the beta process (BP), to sample the states of all HMMs as a shared set of features. This allows for the set of Bayesian HMMs to have \textit{partially} shared states and independent switching dynamics. 

The $\mathcal{IBP}$-HMM\footnote{The terms BP-HMM and $\mathcal{IBP}$-HMM have been used interchangeably in literature \citep{Fox:PhD:2009}. To continue with the whimsical restaurant analogies, in this work we adopt the \textit{latter} term.} has been formulated with autoregressive, multinomial and Gaussian emission models \citep{Fox:NIPS:2009, Hughes:NIPS:2012}. None of which are capable of handling \textit{transform-invariance}. To this end, we propose the IBP Coupled SPCM-$\mathcal{CRP}$ - Hidden Markov Model (ICSC-HMM). In the ICSC-HMM, we couple an $\mathcal{IBP}$-HMM (with Gaussian emission models) together with our SPCM-$\mathcal{CRP}$ mixture model to jointly \textbf{segment} multiple time-series (with partially shared parameters) and extract groups of  \textbf{transform-invariant} emission models (Section \ref{sec:tgau_bp_hmm}). This results in a \textit{topic-model like} hierarchical model capable of addressing the multiple problems posed in Figure \ref{fig:problem2}.

\subsection{Contributions and Paper Organization} 
\noindent The contributions of this manuscript are three-fold:
\begin{enumerate}[leftmargin=*]
	\item We offer a novel \textit{transform-invariant} \underline{similarity function} on the space of Covariance matrices $\mathcal{S}_{++}^N$ which we refer to as the  SPCM (Spectal Polytope Covariance Matrix) similarity function, presented in Section \ref{sec:full_spcm}.
	\item We derive a \textit{transform-invariant} \underline{clustering} approach for Covariance matrices which elegantly leverages spectral clustering and Bayesian non-parametrics. We refer to this approach as the SPCM-$\mathcal{CRP}$ mixture model; introduced in Section \ref{sec:spcm_crp}.
	\item Finally, we couple the proposed Covariance matrix clustering approach with a Bayesian non-parametric formulation of a Hidden Markov Model (HMM) to \underline{jointly segment and} \underline{cluster} transform-invariant states of an HMM for time-series data analysis. We refer to the coupled model as the $\mathcal{IBP}$ Coupled SPCM-$\mathcal{CRP}$ - Hidden Markov Model (ICSC-HMM) and present it in detail in Section \ref{sec:tgau_bp_hmm}).
\end{enumerate}
These sections are followed by a thorough experimental evaluation on simulated and real-world datasets (Section \ref{sec:results}).

\section{Transform-Invariant Covariance Matrix Similarity}
\label{sec:full_spcm}
In this work, we introduce a Covariance similarity function that explicitly holds the property of \textit{transform-invariance} and \textit{boundedness}. The derivation of this similarity function relies on a statistical and geometrical analysis of the eigen-decomposition of a Covariance matrix. Any Covariance matrix $\mathbf{\Sigma} \in \mathcal{S}_{++}^N$ is symmetric, semi-positive definite and full rank. Due to these properties, they have eigenvectors $\mathbf{V} = [V_1,V_2,\dots,V_N]$ with corresponding positive eigenvalues $\mathbf{\Lambda} = \text{diag}(\lambda_1,\lambda_2,\dots,\lambda_N)$, which form an orthonormal basis $\mathbf{\Sigma} = \mathbf{V}\mathbf{\Lambda} \mathbf{V}^T$. While the eigenvalues express the amount of variance corresponding to its respective eigenvector; the eigenvectors are uncorrelated linear combinations of the random variables that produced the covariance matrix. This eigen representation of the Covariance matrix yields an invariant embedding of the structure (i.e. shape) of the data underlying the Covariance matrix. For this reason, we explicitly work on the eigen-decomposition of the Covariance matrix to provide the property of \textit{transform-invariance} in our similarity function (Section \ref{sec:spcm}), while the property of \textit{boundedness} is covered in Section \ref{sec:bound_spcm}.

\subsection{Spectral Polytope Covariance Matrix (SPCM) Similarity}
\label{sec:spcm}
Inspired by spectral graph theory and a geometric intuition of the shapes of convex sets, our similarity function is based on the representation of the covariance matrix as a polytope constructed by the projections of the eigenvalues of $\mathbf{\Sigma}$ on their associated eigenvectors. We refer to it as the Spectral Polytope Covariance Matrix \textit{(SPCM)} similarity function. This non-metric similarity function is based on the assumption that two covariance matrices are indeed similar if their exists a unified homothetic ratio between their spectral polytopes. Following, we elaborate on this assumption.

Spectral algorithms provide an innate representation of the underlying structure of data, derived from the eigenvalues and eigenvectors of a symmetric positive definite (SPD) matrix (e.g. Covariance or Affinity matrix) \citep{Brand:AISTATS:2003}. Let $\mathbf{\Sigma} \in \mathcal{S}_{++}^{N}$ be a SPD Covariance matrix and $\mathbf{\Sigma}=\mathbf{V}\mathbf{\Lambda} \mathbf{V}^T$ the eigenvalue decomposition, where $\mathbf{\Lambda}$ is the diagonal matrix of eigenvalues and $\mathbf{V}$ the matrix of eigenvectors.

\begin{figure}[!t]
	\subfloat[$\mathbf{X}^{(i)}$,$\mathbf{X}^{(j)}$ and Spectral Polytopes $\mathbf{SP}_{i}$ and $\mathbf{SP}_{j}$ \label{fig:polytopes}]{\includegraphics[trim={0.1cm 1.5cm 2cm 0.5cm},clip,width=0.7\linewidth]{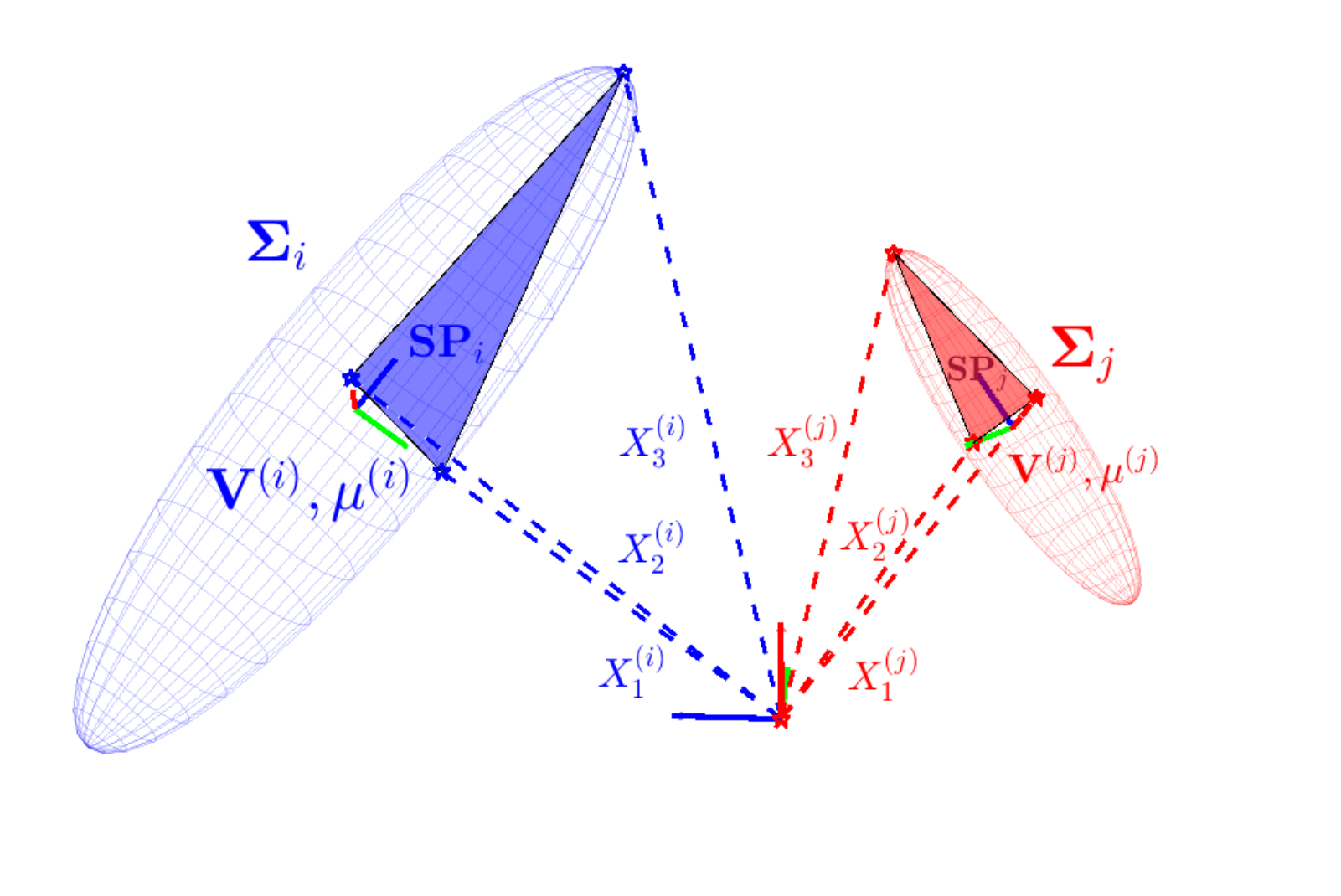}}	
	\subfloat[Homothety in $\mathbf{SP}$s \label{fig:homothety}]{ \includegraphics[width=0.3\linewidth]{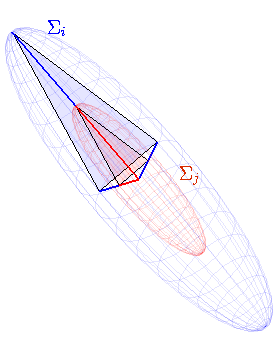}}
	\caption{Illustration of Spectral Polytope Similarity Principle 	    \label{fig:Spectral_polytope}}
\end{figure}

\newtheorem{mydef}{Definition}
\begin{mydef}
	\textbf{Spectral Polytope} ($\mathbf{SP}$): The $\mathbf{SP}$ is a geometrical representation of the invariant structure of a Covariance matrix $\mathbf{\Sigma} \in \mathcal{S}_{++}^N$. It is constructed by taking the convex hull of the set of orthogonal column vectors $\mathbf{X} = \left\{ X_1|X_2|\dots|X_N \right\}$, where $X_i = V_i\lambda_i^{1/2}$, creating a $(N-1)$-polytope representing the invariant shape of $\mathbf{\Sigma}$, as depicted in Figure \ref{fig:polytopes}.
\end{mydef}

\noindent \texttt{Constructing $\mathbf{SP}$ via $\mathbf{X}$:} Assume two 3-$dim$ Covariance matrices $\mathbf{\Sigma}_i$,$\mathbf{\Sigma}_j$ similar in shape, yet different in rotation $\mathbf{R}$ and scale $\gamma$; i.e.  $\mathbf{\Sigma}_j=\mathbf{R}\mathbf{V}^{(i)}(\gamma\mathbf{\Lambda}^{(i)})(\mathbf{R}\mathbf{V}^{(i)})^{T}$ (as in Figure \ref{fig:polytopes}). In 3-$dim$ space, the hyper-spheres of $\mathbf{\Sigma}_i,\mathbf{\Sigma}_j$ form ellipsoids centered at $\mu^{(i)},\mu^{(j)}$ with axes $\mathbf{V}^{(i)},\mathbf{V}^{(j)}$ and axis lengths $\mathbf{\Lambda}^{(i)},\mathbf{\Lambda}^{(j)}$. Each Covariance matrix has a corresponding \textit{invariant} set of column vectors $\mathbf{X} = \left\{X_1|X_2|\dots|X_N\right\}$ where $X_i = V_i\lambda_i^{1/2}$ is a projection on the $i$-th eigenvector scaled by its respective eigenvalue. Such a set encodes the linear correlations of the distribution of points embedded on the surface of a hypersphere, producing a scale and rotation invariant representation of the original dataset used to construct $\mathbf{\Sigma}$. As shown in Figure \ref{fig:polytopes}, the convex hull of the endpoints of the sets of column vectors $\mathbf{X}^{(i)}=\left\{ X_{1}^{(i)}|X_2^{(i)}|X_3^{(i)}\right\}$ and $\mathbf{X}^{(j)}=\left\{X_{1}^{(j)}|X_2^{(j)}|X_3^{(j)}\right\}$ generate the spectral $N-1$-polytopes $\mathbf{SP}_{i}$ and $\mathbf{SP}_{j}$, depicted by the shaded regions in Figure \ref{fig:polytopes}.
\noindent \textit{Transfom-invariance through homothety:} By stripping down the ellipsoids from their orientation ($\mathbf{V}$) and translation ($\mathbf{\mu}$) constraints (as shown in Figure \ref{fig:homothety}), we observe that $\mathbf{SP}_i$ is an \textit{enlargement} or \textit{magnification} of $\mathbf{SP}_j$\footnote{One can also state that $\mathbf{SP}_j$ is a shrinkage or contraction of $\mathbf{SP}_i$.}, in other words it is subject to a \textit{homothetic transformation}. Since $\mathbf{X}^{(i)}$ and $\mathbf{X}^{(j)}$ are  convex sets of vectors in $\mathds{R}^N$ used to construct the polytopes $\mathbf{SP}_i,\mathbf{SP}_j$, we can analogously state that, the nonempty set $\mathbf{X}^{(j)}$ is a \textit{homothetic projection} of $\mathbf{X}^{(i)}$ via the following theorem:
\newtheorem{mytheor}{Theorem}
\begin{mytheor}
	Two nonempty sets $\mathbf{X}^{(i)}$ and $\mathbf{X}^{(j)}$ in Euclidean space $\mathds{R}^N$ are homothetic provided $\mathbf{X}^{(i)} = \mathbf{z} + \gamma \mathbf{X}^{(j)}$ for a suitable point $z \in \mathds{R}^N$ and a scalar $\gamma \in \Re_{\neq 0}$, i.e. the homothety ratio.
	\begin{proof}
		Provided in \citet{Soltan:BAG:2010}
	\end{proof}
\end{mytheor}
\noindent In other words, if we can represent the column vectors of $\mathbf{X}^{(i)}$ as $X_k^{(i)} = z + \gamma X_k^{(j)}$ $\forall  k=\{1,\dots,N\}$, \textit{homothety} is the linear projection that preserves collinearity across the points aligned on $\mathbf{X}^{(i)}$ and $\mathbf{X}^{(j)}$. Hence, assuming $z = \emptyset_N$ and $\gamma \in \Re_{+}$, the \textit{positive homothety ratio}, $\gamma$, represents the scaling factor between the two convex sets $\mathbf{X}^{(i)}$ and $\mathbf{X}^{(j)}$, and consequently between two Covariance matrices $\mathbf{\Sigma}_i$ and $\mathbf{\Sigma}_j$.
\begin{mydef}
	\textbf{Homothetic similarity:} $\mathbf{\Sigma}_i$ and $\mathbf{\Sigma}_j$ are similar if their associated convex sets $\mathbf{X}^{(i)}$ and $\mathbf{X}^{(j)}$ are homothetic.
\end{mydef}
\noindent We derive our similarity function from the following corrolary:
\newtheorem{mycorro}{Corrolary}
\begin{mycorro}
	Given $\mathbf{\Sigma}_i$,$\mathbf{\Sigma}_j \in \mathcal{S}_{++}^{N}$ and their associated convex sets $\mathbf{X}^{(i)}=\left\{X_1^{(i)}|\dots|X_N^{(i)}\right\}$, $\mathbf{X}^{(j)}=\left\{X_1^{(j)}|\dots|X_N^{(j)}\right\}$, if a homothety ratio $\gamma \in \Re_{\neq 0}$ exists such that $X_k^{(i)} =\gamma X_k^{(j)}$ $\forall  k=\{1,\dots,N\}$, $\mathbf{\Sigma}_i$ and $\mathbf{\Sigma}_j$ are deemed similar, subject to a homothetic scaling $\gamma$.
\end{mycorro}

\noindent Our proposed similarity function, thus, relies on the existence of a single homothety ratio, $\gamma$, that holds for all pairs of column vectors $X_k^{(j)}\rightarrow X_k^{(i)}$ $\forall k=\{1,\dots,N\}$. We approximate this ratio by taking the vector-norms, namely the $L_2$ norm, of each element in the convex set and computing the element-wise division between the two sets $\mathbf{X}^{(i)}$ and $\mathbf{X}^{(j)}$, as follows:
\begin{equation}
\hat{\mathbf{\gamma}}\left(\mathbf{X}^{(i)},\mathbf{X}^{(j)}\right)= \left[\frac{||X_1^{(i)}||}{||X_1^{(j)}||}, \dots, \frac{||X_k^{(i)}||}{||X_k^{(j)}||}, \dots, \frac{||X_N^{(i)}||}{||X_N^{(j)}||}   \right].
\label{eq:ratios}
\end{equation}
This yields a vector of $N$  \textit{approximate homothety ratios} $\hat{\mathbf{\gamma}}\left(\mathbf{X}^{(i)},\mathbf{X}^{(j)}\right)=[\hat{\gamma}_1,\dots,\hat{\gamma}_N]$ corresponding to each $k$-th dimension of $\mathbf{\Sigma} \in \mathds{R}^{N\times N}$. In the ideal case, where $\mathbf{\Sigma}_i \equiv \mathbf{\Sigma}_j$, all elements of \eqref{eq:ratios} are equivalent, i.e. $\hat{\gamma}_1 = \dots = \hat{\gamma}_k = \dots = \hat{\gamma}_N$. The overall approximate homothety ratio $\bar{\hat{\gamma}}$ can then be computed as the mean of all ratios $\hat{\gamma}_k$ approximated for each dimension $k$,
\begin{equation}
\bar{\hat{\gamma}}\left(\mathbf{X}^{(i)},\mathbf{X}^{(j)}\right) = \frac{1}{N}\sum_{k=1}^{N} \hat{\gamma}_k.
\label{eq:mean_gamma}
\end{equation}
Consequently, the variance of the approximate homothety ratios, 
\begin{equation}
\sigma^2\left(\hat{\gamma}\left(\mathbf{X}^{(i)},\mathbf{X}^{(j)}\right)\right) = \frac{1}{N}\sum_{k=1}^{N} (\hat{\gamma}_k - \bar{\hat{\gamma}})^2, 
\label{eq:var_gamma}
\end{equation}
represents the variation of approximate homothetic ratios between the two convex sets $\mathbf{X}^{(i)},\mathbf{X}^{(j)}$ and thus, provides a measure of \textit{homothetic similarity} between Covariance matrices $\mathbf{\Sigma}_i, \mathbf{\Sigma}_j$. Hence, $\sigma^2\left(\hat{\gamma}\left(\mathbf{X}^{(i)},\mathbf{X}^{(j)}\right)\right) = 0$ when $\mathbf{\Sigma}_i \equiv \mathbf{\Sigma}_j$ and $\sigma^2\left(\hat{\gamma}\left(\mathbf{X}^{(i)},\mathbf{X}^{(j)}\right)\right) \rightarrow \Re_{+}$ otherwise.

As per \eqref{eq:ratios}, the proposed measure of similarity will approximate either a ratio of \textit{magnification} (if $\texttt{area}(\mathbf{SP}_i)>\texttt{area}(\mathbf{SP}_j)$) or \textit{contraction} (if $\texttt{area}(\mathbf{SP}_i)<\texttt{area}(\mathbf{SP}_j)$) in the direction of $\mathbf{X}^{(i)} \rightarrow \mathbf{X}^{(j)}$, i.e. it is uni-directional. In order to provide a bi-directional measure of similarity and consider only the \textit{magnification} ratio between the two sets $\mathbf{X}^{(i)}$ and $\mathbf{X}^{(j)}$, we formulate our similarity function as follows,

\begin{equation}
\begin{split}
\Delta_{ij}(\mathbf{\Sigma}_i,\mathbf{\Sigma}_j) & =
H(\delta_{ij})\sigma^2\left(\hat{\gamma}\left(\mathbf{X}^{(i)},\mathbf{X}^{(j)}\right)\right) + \left(1-H(\delta_{ij})\right)\sigma^2\left(\hat{\gamma}\left(\mathbf{X}^{(j)},\mathbf{X}^{(i)}\right)\right) \\
\text{with} & \quad \delta_{ij} = \bar{\hat{\gamma}}\left(\mathbf{X}^{(i)},\mathbf{X}^{(j)}\right) - \bar{\hat{\gamma}}\left(\mathbf{X}^{(j)},\mathbf{X}^{(i)}\right),
\end{split}
\label{eq:spcm}
\end{equation}
where $H(\delta_{ij})$ is the heavyside step function, which can be approximated as $H(\delta_{ij}) = \frac{1}{2}[1 + \text{sign}(\delta_{ij})]$. Hence, \eqref{eq:spcm} represents the variation of approximate \textit{magnifying} homothetic ratios between the two convex sets $\mathbf{X}^{(i)},\mathbf{X}^{(j)}$. This similarity function is indeed not a \textit{proper} metric for (dis)similarity. However, it can be considered, a \textit{semimetric}, as it exhibits most of the required properties of a legitimate distance function or metric: 
\begin{enumerate}[leftmargin=*]
	\item $\Delta_{ij}(\mathbf{\Sigma}_i,\mathbf{\Sigma}_j)\geq 0$ \hfill Non-negativity 
	\item  $\Delta_{ij}(\mathbf{\Sigma}_i,\mathbf{\Sigma}_j) = 0 \Leftrightarrow  \mathbf{\Sigma}_i  = \mathbf{\Sigma}_j$  \hfill Positive Definiteness
	\item $\Delta_{ij}(\mathbf{\Sigma}_i,\mathbf{\Sigma}_j) = \Delta_{ji}(\mathbf{\Sigma}_j,\mathbf{\Sigma}_i)$ \hfill Symmetry 
	\item $\Delta_{ij}(\mathbf{\Sigma}_i,\mathbf{\Sigma}_j)=0 \Leftrightarrow \mathbf{\Sigma}_j = \mathbf{R}\mathbf{V}^{(i)}(\gamma\mathbf{\Lambda}^{(i)})(\mathbf{R}\mathbf{V}^{(i)})^{T}$   \hfill Transform-Invariance
\end{enumerate}
\noindent \eqref{eq:spcm} is thus, a \textit{semimetric}
on $\mathbf{\Sigma} \in \mathcal{S}_{++}^{N}$, as it is a function $\Delta : \mathbf{\Sigma} \times \mathbf{\Sigma} \rightarrow \Re$ that satisfies the first three axioms (1-3) of a metric except the axiom of \textit{triangle inequality}. Strict \textit{triangle inequality} is only crucial for distance-dependent learning algorithms, such as $K$-Means, GMM or $K$-NN. In spectral algorithms, it is \textit{sufficient} for a similarity/affinity measure to hold axioms (1-3).  Moreover, it has been shown that applying distance-dependent algorithms directly in the space of $\mathcal{S}_{++}^N$ is not a straight-forward procedure \citep{Cherian:TPAMI:2013}. Thus, the lack of this property is not detrimental to our targeted applications, as we will focus on a spectral-based mixture model variant, described in Section \ref{sec:spcm_crp}. Following we introduce a formulation of \eqref{eq:spcm} that provides the property of \textit{boundedness}.
\begin{figure}[!t]
	\centering
	\includegraphics[trim={0cm 0cm 0.5cm 0cm},clip,width=0.54\linewidth]{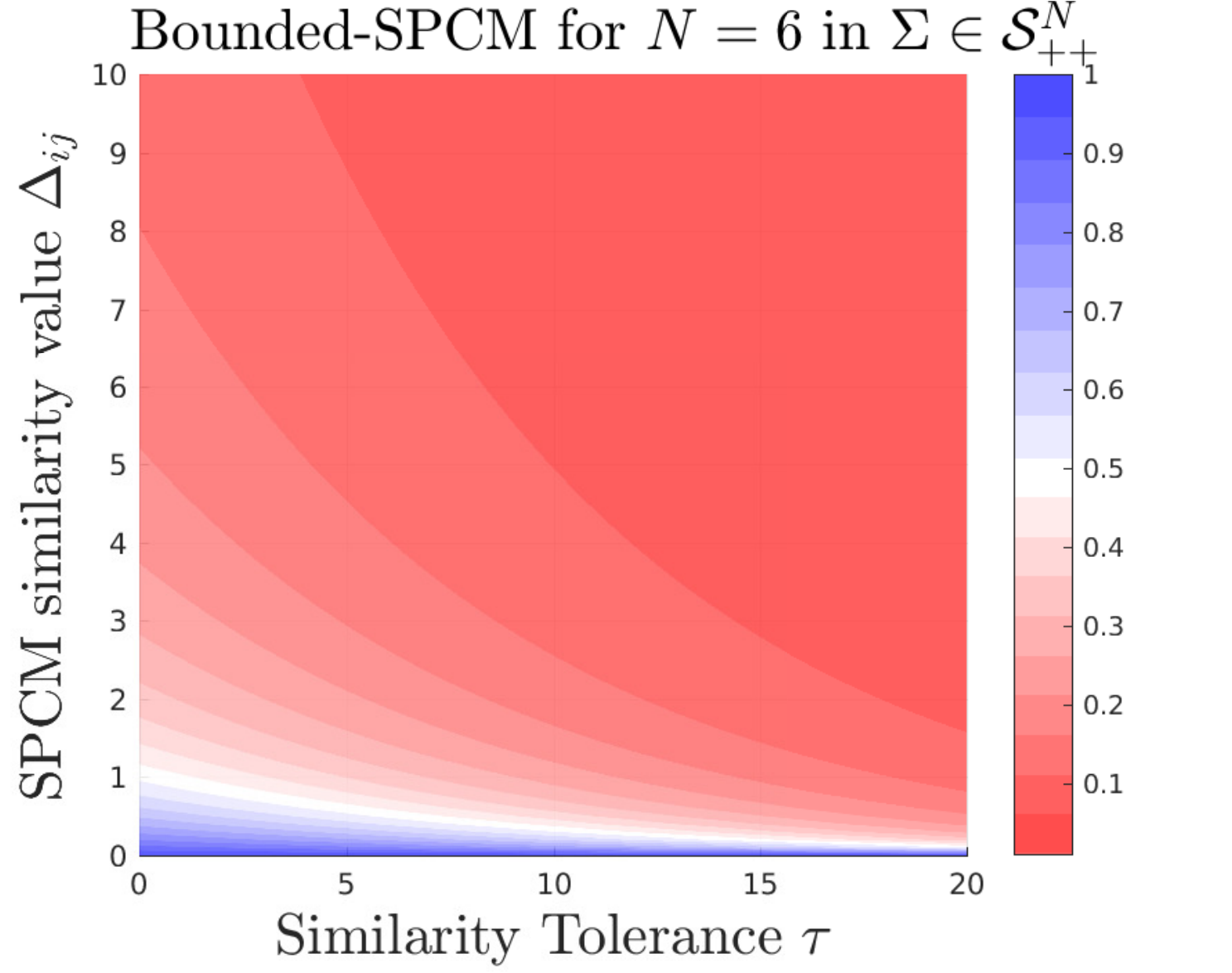}\includegraphics[trim={0.75cm 0cm 0.75cm 0cm},clip,width=0.46\linewidth]{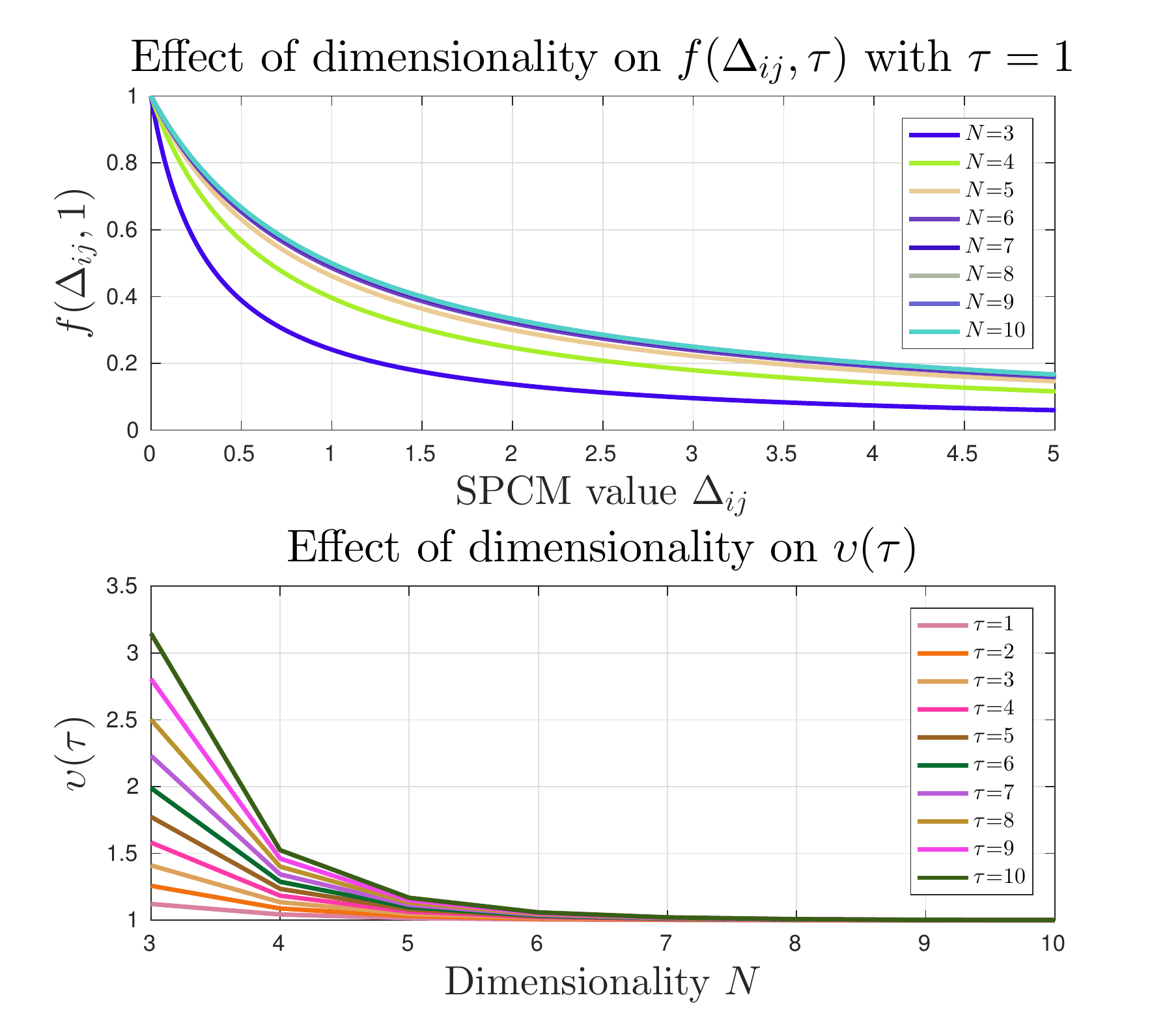}
	\caption{Effects of B-SPCM hyper-parameters. \label{fig:b_spcm}}
\end{figure} 

\subsection{Bounded Decaying SPCM Similarity function}
\label{sec:bound_spcm}
\eqref{eq:spcm} yields values in the range of $\Delta : \mathbf{\Sigma} \times \mathbf{\Sigma} \rightarrow [0,\infty)$. Having such an unbounded metric can be a nuisance. Hence, we formulate a bounded function for SPCM similarity, as $f(\cdot)$, which is a monotonic decay function bounded between $[1,0]$, where $f(\cdot)=1$ represents definite similarity, i.e. $\mathbf{\Sigma}_i \equiv \mathbf{\Sigma}_j$ whilst  $f(\cdot)=0$ is absolute dissimilarity:

\begin{equation}
f(\Delta_{ij},\tau) = \frac{1}{1+\upsilon(\tau) \Delta_{ij}(\mathbf{\Sigma}_i,\mathbf{\Sigma}_j)}
\label{eq:bspcm}
\end{equation}

where $\Delta_{ij}=\Delta_{ij}(\mathbf{\Sigma}_i,\mathbf{\Sigma}_j)$ is the SPCM similarity value given by \eqref{eq:spcm} and $\upsilon$ is a scaling function determined by the following equation:
\begin{equation}
\upsilon(\tau) = 10^{(\tau e^{-N})}
\end{equation}
where $N$ is the dimensionality of $\mathbf{\Sigma} \in \mathcal{S}_{++}^N$ and $\tau$ is a design tolerance hyper-parameter. The \textit{B-SPCM} similarity function \eqref{eq:bspcm} was designed so that it could hold the following properties:
\begin{enumerate}[leftmargin=*]
	\item $f(\cdot)$ decreases as positive $\Delta_{ij}$ increases
	\item $f(0)=1$
	\item $f(\cdot) \rightarrow 0$ as $\Delta_{ij} \rightarrow +\infty$ 
\end{enumerate}
\normalsize
In Figure \ref{fig:b_spcm} we show the effect of the tolerance hyper-parameter, $\tau$, and the dimensionality of the data, $N$, on the B-SPCM similarity function.  The rationale behind the scaling function $\upsilon$ being dependent on the $N$ is to introduce a semantic distinction between similarity values in different dimensions. For example, a similarity value $\Delta_{ij}=1$ for $\mathbf{\Sigma}_i,\mathbf{\Sigma}_j \in \mathcal{S}_{++}^3$ should not yield the same B-SCPM value if $\mathbf{\Sigma}_i,\mathbf{\Sigma}_j \in \mathcal{S}_{++}^6$. 
When this happens, according to Eq. \ref{eq:spcm}, it means that the variance on the homothetic ratios is the same. However, if we have the same variance for 3-$dim$ as in 6-$dim$, the relative deviation between them is not the same. Hence, the scaling function $\upsilon(\tau)$ is a form of dimensional scaling to account for this behavior. This is evident in the span of dimensions between $N = [3,9]$. For Covariance matrices with $N>9$, the effect of the scaling function is trivial. Regarding the tolerance value, $\tau$, it can take any non-negative value in $\Re$, and is merely an amplification factor that can be tuned for datasets which we know are quite noisy, typical values that we have used in our dataset lie in the range of $[1,10]$.

\begin{figure}[!h]
		\centering 
		\includegraphics[trim={1.5cm 3.65cm 0.5cm 3cm},clip,width=0.8\linewidth]{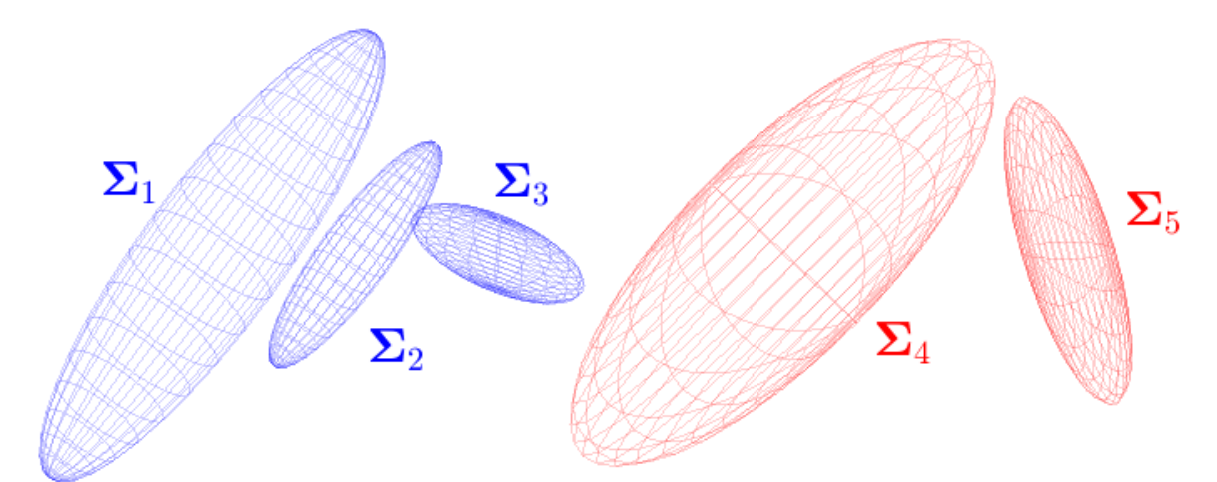}
		\caption{\small \textit{Toy-ellipsoids} dataset of transformed 3D Covariance matrices (color corresponds to similar matrices). \label{fig:ellipsoids}}
		\vspace{-10pt}
\end{figure}

\subsection{Comparison to Standard Similarity Functions}
\label{sec:comparison-standard}
We highlight the advantage of the SPCM function to find similarity in \textit{transformed} Covariance matrices over the standard similarity functions (presented in Table \ref{tab:metrics}) on a toy example.

\paragraph{Toy Example} Consider a small dataset of Covariance matrices $\mathbf{\Theta} = \left\lbrace \mathbf{\Sigma},\mathbf{\Sigma}_2,\mathbf{\Sigma}_3,\mathbf{\Sigma}_4, \mathbf{\Sigma}_5 \right\rbrace$ where $\mathbf{\Sigma} \in \mathcal{S}_{++}^3$. The dataset is generated by two \textit{distinct} Covariance matrices, namely $\mathbf{\Sigma}_1$ and $\mathbf{\Sigma}_4$: 
\begin{equation*}
\textcolor{black}{\mathbf{\Sigma}_1 = \begin{bmatrix}
b&a&a\\
a&b&a\\
a&a&b
\end{bmatrix}},\hspace{5pt}
\textcolor{black}{\mathbf{\Sigma}_4 = \begin{bmatrix}
a&d&e\\
d&b&f\\
e&f&c
\end{bmatrix}}
\end{equation*}
where $a-f$ take on real values in $\Re$, whose signs are constrained to yield a SPD matrix. Moreover, $\textcolor{black}{\mathbf{\Sigma}_2}$ is a scaled and noisy version of  $\textcolor{black}{\mathbf{\Sigma}_1}$, $\textcolor{black}{\mathbf{\Sigma}_3}$ is a rotated version of $\textcolor{black}{\mathbf{\Sigma}_1}$ and $\textcolor{black}{\mathbf{\Sigma}_5}$ is a rotated and scaled version of $\textcolor{black}{\mathbf{\Sigma}_4}$, such that $\mathbf{\Theta} = \{\textcolor{black}{\mathbf{\Sigma}_1,\text{rot}(\mathbf{\Sigma}_1),\text{scaled}(\mathbf{\Sigma}_1)}, \textcolor{black}{\mathbf{\Sigma}_4, \text{rot}(\text{scaled}(\mathbf{\Sigma}_4))} \}$. In Figure \ref{fig:ellipsoids}, we illustrate these Covariance matrices as 3D ellipsoids; ellipsoids with the same color correspond to similar Covariance matrices. Thus, if we want to group these matrices based on \textit{transform-invariant} similarity, we should end up with two clusters $C = \{\textcolor{black}{c_1}, \textcolor{black}{c_2} \}$ where $\textcolor{black}{c_1 = \{\mathbf{\Sigma}_1, \mathbf{\Sigma}_2,\mathbf{\Sigma}_3 \}}$ and $\textcolor{black}{c_2 = \{\mathbf{\Sigma}_4,\mathbf{\Sigma}_5 \}}$. \\

\begin{figure}[!t]
	\begin{minipage}{\linewidth}
		\centering 
		\includegraphics[trim={3cm 1cm 3cm 0.2cm},clip,width=0.35\linewidth]{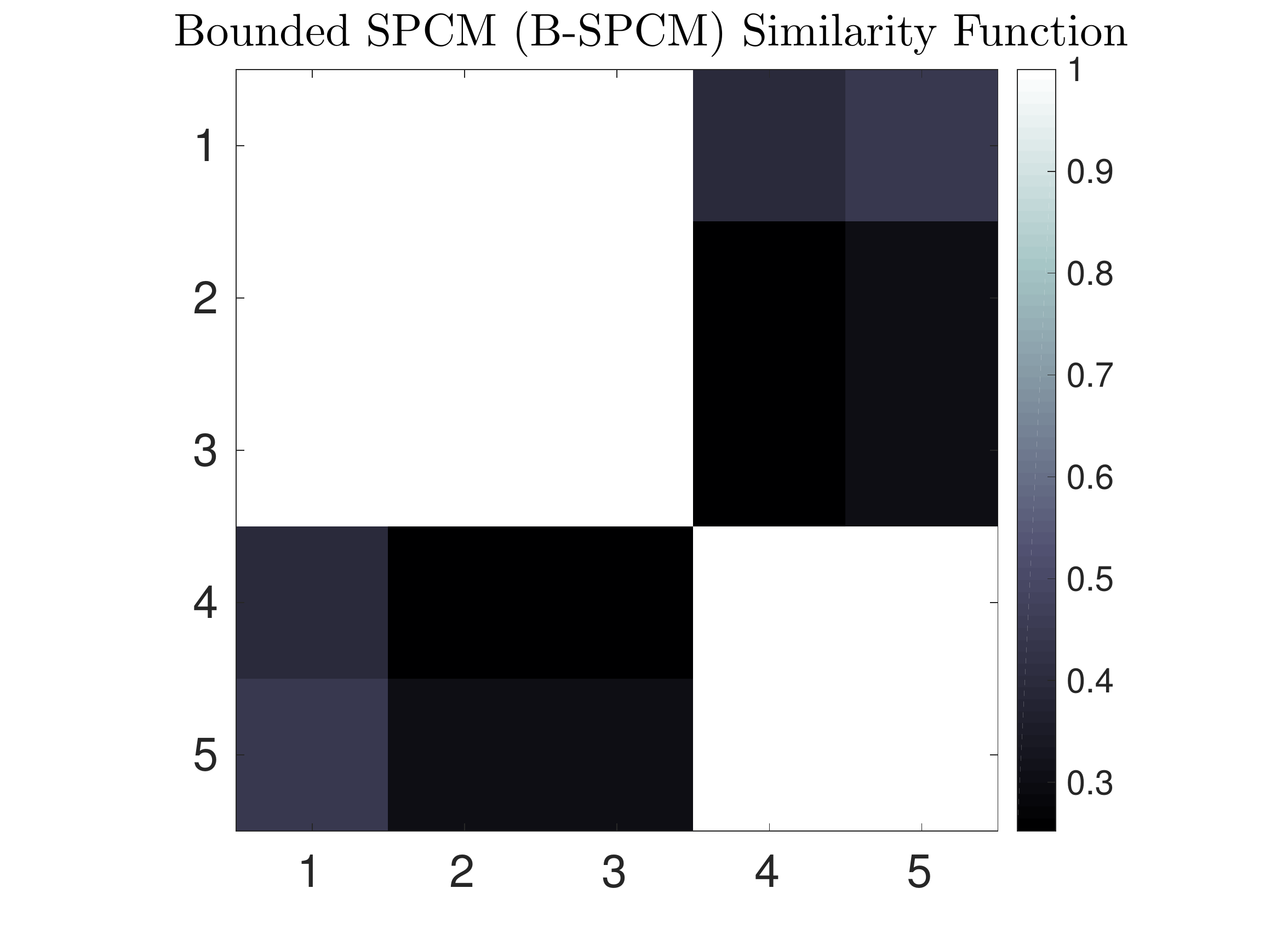}\hspace{5pt}\includegraphics[trim={3cm 1cm 3cm 0.2cm},clip,width=0.35\linewidth]{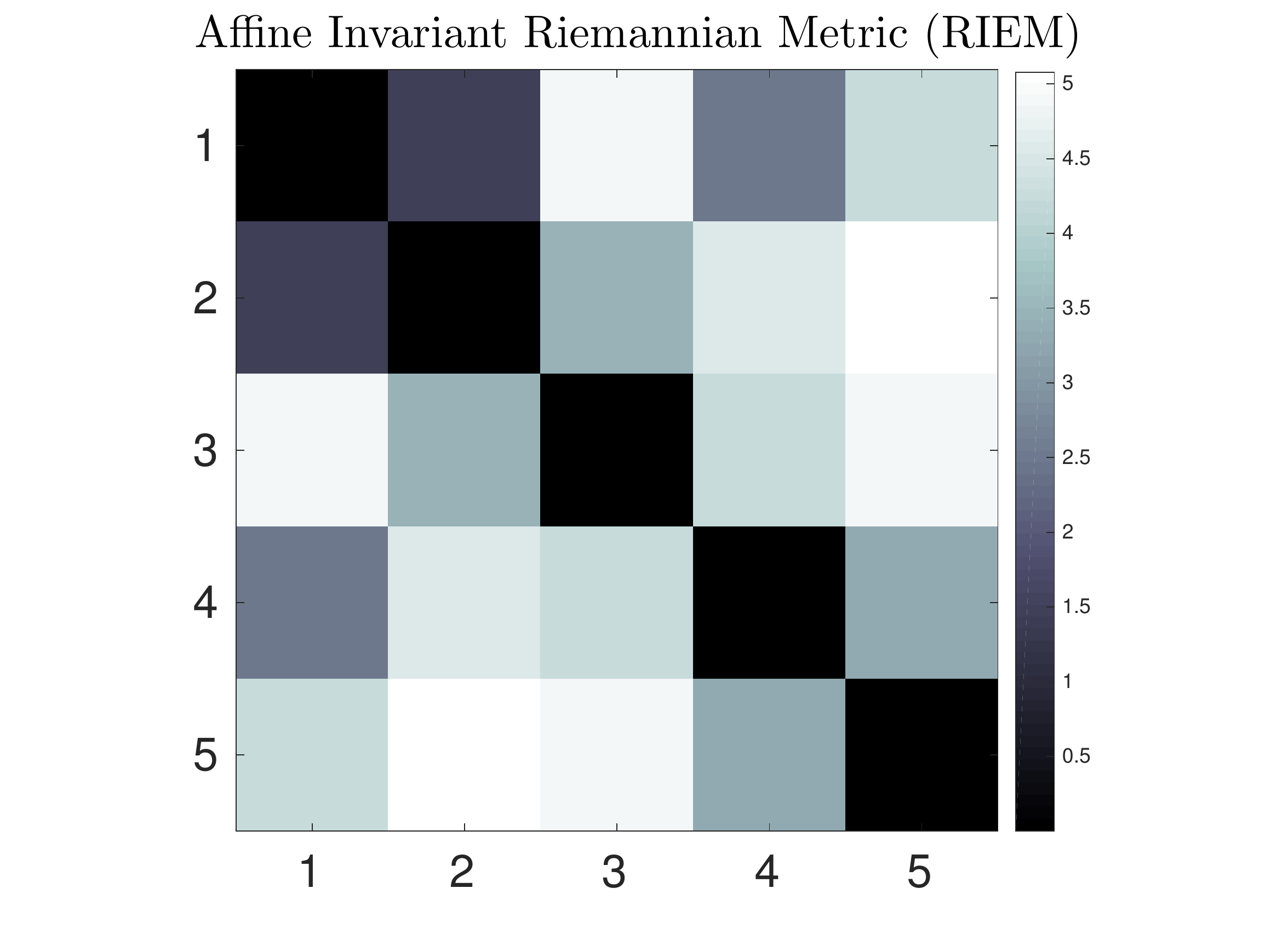}
		
		\vspace{5pt}
		\includegraphics[trim={3cm 1cm 3cm 0.2cm},clip,width=0.333\linewidth]{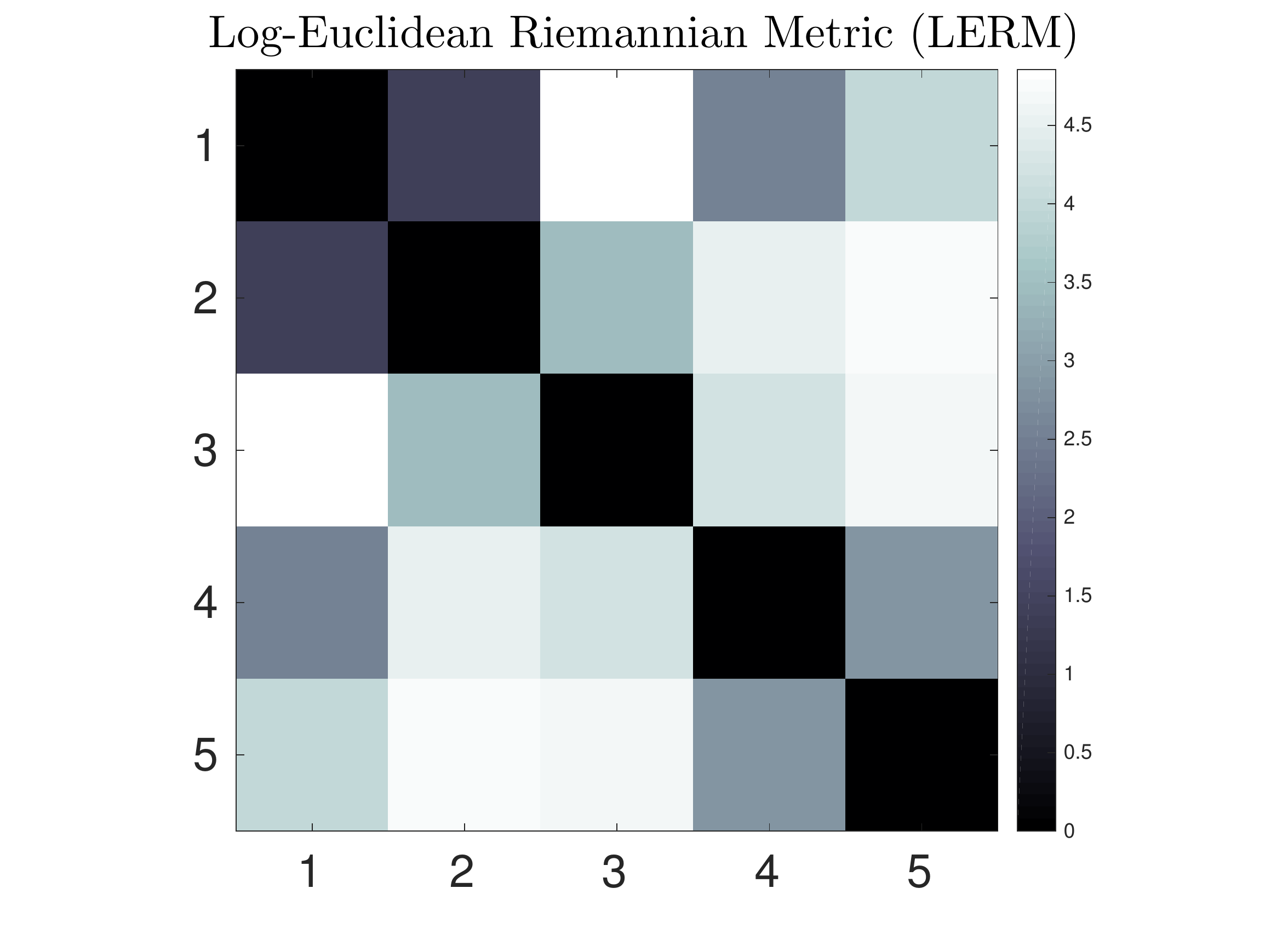}\includegraphics[trim={3cm 1cm 3cm 0.2cm},clip,width=0.333\linewidth]{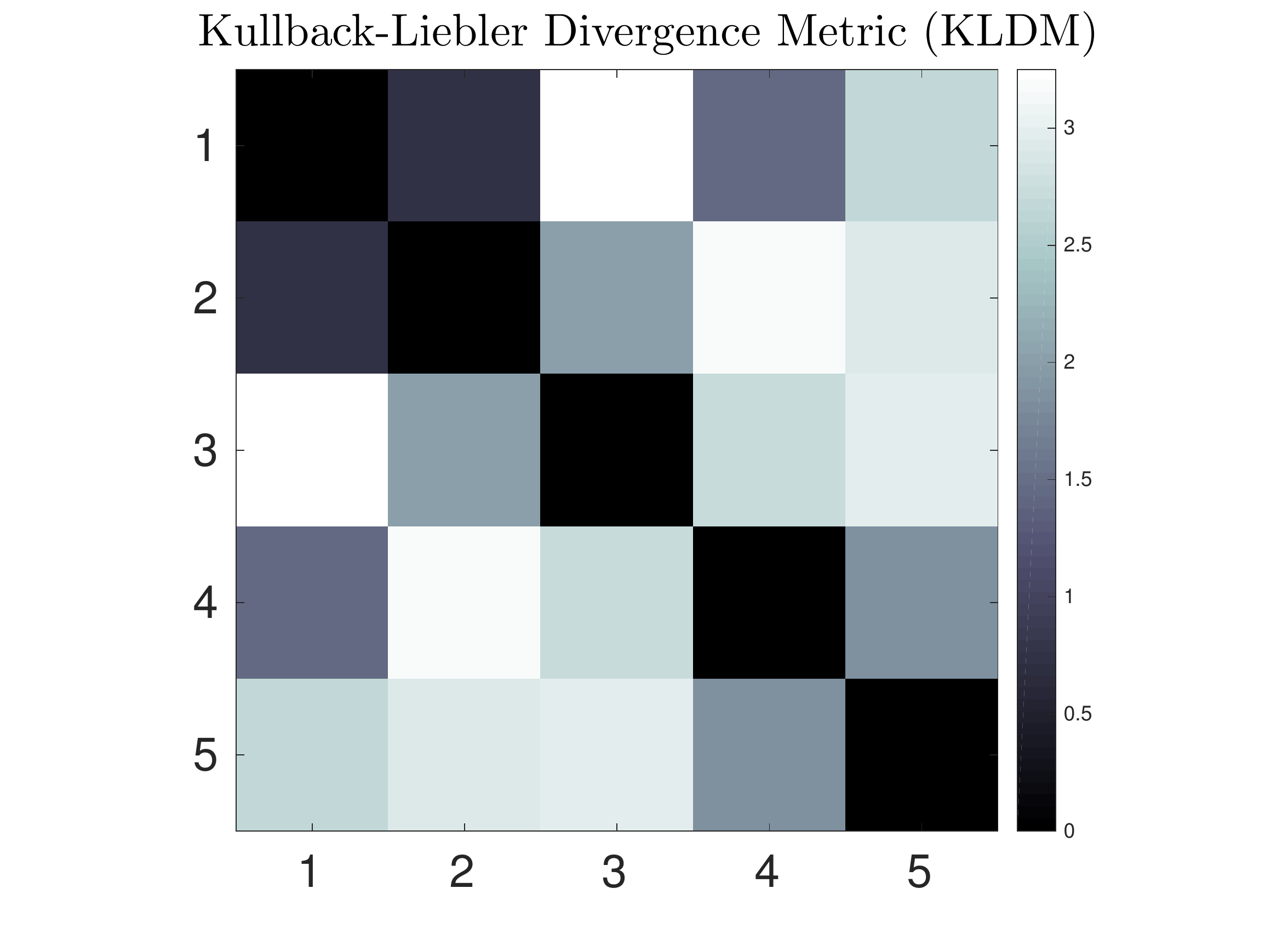}\includegraphics[trim={3cm 1cm 3cm 0.2cm},clip,width=0.333\linewidth]{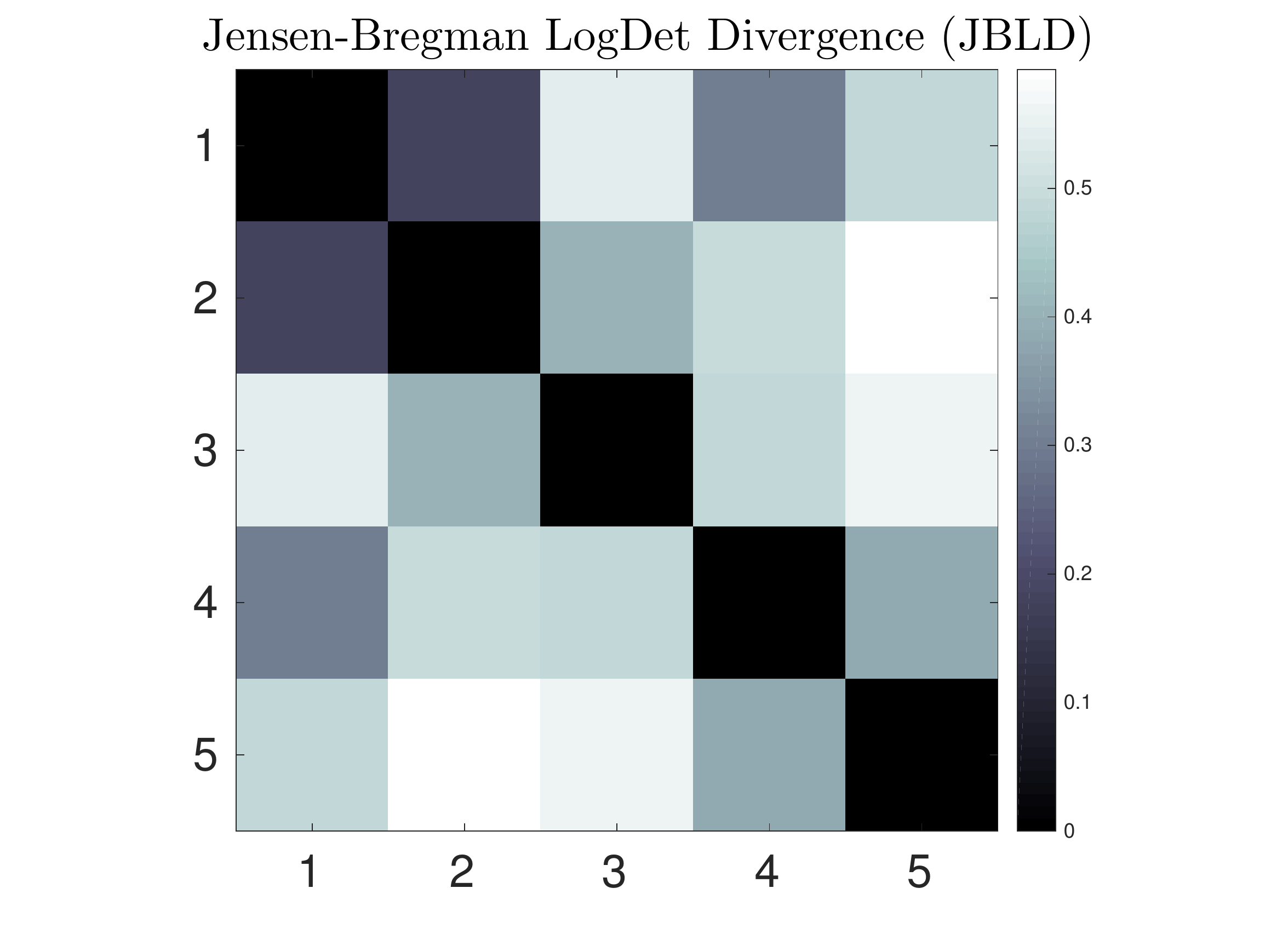}
		\caption{ \small Similarity Matrices sorted by cluster labels for \textbf{B-SCPM} (top-left), \textbf{RIEM} (top-center), \textbf{LERM} (top-right), \textbf{KLDM} (bottom-left) and \textbf{JBLD} (bottom-right) on the \textit{Toy-Ellipsoids} dataset \label{fig:metrics_toy}}
	\end{minipage}
\end{figure}

In Figure \ref{fig:metrics_toy}, we show the similarity matrices generated by the B-SPCM and the standard Covariance matrix similarity functions. As can be seen, the latter fail to provide discriminative values for the true partitioning of the dataset. Whereas, the B-SPCM function, which focuses on the analysis of the spectral polytope, makes for a robust \textit{transform-invariant} similarity function. This is due to the fact that none of the standard similarity functions hold the rotation and scale invariance property \textit{explicitly}. Nonetheless, it is commonly known that even though partitions from a confusion matrix are not easily identified, we can still recover them with similarity-based clustering algorithms. In Section \ref{sec:results}, we provide an \textit{exhaustive} quantitative evaluation of the performance of the B-SPCM similarity function compared to the standard similarity functions using two popular similarity-based clustering algorithms. As will be discussed later on, none of the standard similarity functions are able to recover the two clusters from our toy dataset with neither of the clustering algorithms. Whereas, applying both algorithms to the B-SPCM similarity matrix yields the true labels. 

\newpage
\section{Spectral Non-Parametric Clustering of Covariance Matrices}
\label{sec:spcm_crp}
Given the similarity matrix $\mathbf{S} \in \mathds{R}^{M \times M}$ of $M$ Covariance matrices, where each entry $s_{ij}=f(\Delta_{ij},\tau)$ is computed by \eqref{eq:bspcm}, we wish to find a partition that recovers a natural \textit{transform-invariant} grouping of \textit{similar} Covariance matrices. One of the most popular approaches to solve this problem is \textit{Spectral Clustering}, which relies on graph Laplacian matrices and the study thereof (i.e. spectral graph theory) \citep{Ng:NIPS:2001, Luxburg:TSC:2007}. Spectral clustering determines relationships across data-points embedded in a graph induced by the similarity matrix $\mathbf{S}$. In other words,  we would find a mapping of the Covariance matrices $\mathbf{\Sigma}_i$ onto a lower-dimensional Euclidean space, $\mathbf{y}_i = f(\mathbf{\Sigma}_i)$ where $f(\cdot):\mathcal{S}_{++}^{N}\rightarrow\mathds{R}^{P}$, induced by their similarity matrix $\mathbf{S} \in \mathds{R}^{M \times M}$. This mapping is found by constructing the graph Laplacian matrix $\mathbf{L} = \mathbf{D} - \mathbf{S}$; where $\mathbf{D}$ is a diagonal weighting matrix whose entries are column sums of $\mathbf{S}$. After performing an Eigenvalue decomposition of $\mathbf{L} = \mathbf{V}\mathbf{\Lambda}\mathbf{V}^T$ and ordering the eigenvectors in ascending order wrt. their eigenvalues $\lambda_0 = 0 \leq \lambda_1 \leq \dots \leq \lambda_{M-1}$; a $P$-dimensional \textit{spectral embedding} is constructed by selecting the first $P$ eigenvectors of the graph Laplacian $\mathbf{Y} \in \mathds{R}^{P \times M}$ as $\mathbf{Y} = [V^{(1)}, \dots, V^{(P)}]^T$. We would then perform $K$-Means on the spectral embedding $\mathbf{Y} \in \mathds{R}^{P \times M}$, which represent the clusters in the original space corresponding to the set of Covariance matrices $\mathbf{\Theta} = \{\mathbf{\Sigma}_1, \dots, \mathbf{\Sigma}_M\}$\footnote{For an in-depth tutorial and derivation of spectral clustering and variants, the authors recommend reading \citet{Luxburg:TSC:2007}.}.

One limitation is that performance relies on a good choice of $P$, the sub-space dimensionality, and $K$, the number of expected clusters. In this work, we provide an algorithm that leverages spectral clustering and Bayesian non-parametrics in one single model in order to automatically estimate $P$ and $K$. Hence, we tackle the following two main challenges:\\

\noindent \textit{(1) \underline{Sub-Space Dimensionality Selection}:} 
In its original form, the \textit{spectral clustering} algorithm assumes that the number of clusters $K$ is equivalent to the sub-space dimensionality $P$. The value of $P,K$ could be chosen either through model selection or by analyzing the eigenvalues of $\mathbf{L}$. Such a strong assumption comes from the theorem of connected graphs, which states that the multiplicity of the eigenvalue $\lambda=0$ determines the number of $K$ connected graphs \citep{Luxburg:TSC:2007}. This theorem holds only for sparse similarity matrices $\mathbf{S}$. When $\mathbf{S}$ is fully connected solely one eigenvalue is $\lambda_0=0$, however, the contiguous eigenvalues are $\lambda_{1,\dots,P} \rightarrow 0$. Thus, we can determine $P,K$, by choosing the $K$-th and $K+1$-th pair whose gap is the largest. This approach, however, can be quite sensitive as different datasets can yield different patterns of distribution of the eigenvalues \citep{Zelnik:NIPS:2004}. To alleviate this, we first propose a relaxation on the assumption that $P=K$, which has been strongly supported by previous work \citep{Poon:UAI:2012}. We then introduce an unsupervised dimensionality selection algorithm for the \textit{spectral embedding} construction. We treat the eigenvalues as weights and apply a global normalization layer using the softmax function \citep{Bishop:PRM:2006}, to find the contiguous set of $P$-eigenvectors that best describes the dataset. By applying the softmax function on the eigenvalues, we are performing a nonlinear transformation into a range of normalized values between $[0,1]$, where the effect of extreme values is reduced; i.e. we are ``squashing" the eigenvalues. Thus, for eigenvalues $\lambda_0,\dots,\lambda_M$ of the Laplacian matrix, we can compute their weighted eigenvalues as follows,
\begin{equation}
\label{eq:softmax}
\rho\left(\mathbf{\lambda}\right)_i = \frac{\exp\left(\lambda_i\right)}{\sum_{j=1}^{M} \exp\left(\lambda_j\right)} \hspace{5pt} \text{for} \hspace{5pt} i=\left\{1,\dots,M\right\}
\end{equation}
, for $M$ number of samples in our dataset. To find the continuous set of leading $P$ eigenvalues, which we will now call the \textit{primary} eigenvalues, we re-normalize the weights computed from Eq. \ref{eq:softmax} to lie between the range of $[-1,1]$. By normalizing the values in this range, the weighted eigenvalues $\rho(\mathbf{\lambda})_i < 0$ correspond to the \textit{$P$-primary} eigenvalues. We list the full unsupervised spectral embedding approach in Algorithm \ref{alg:spect_dim_red}.\\  

\begin{algorithm}[!t]
	\renewcommand{\algorithmicrequire}{\textbf{Input:}}
	\renewcommand{\algorithmicensure}{\textbf{Output:}}
	\caption{Unsupervised Spectral Embedding}
	\label{alg:spect_dim_red}
	\footnotesize
	\begin{algorithmic}[1]
		\Require A positive-definite pair-wise similarity matrix $\mathbf{S} \in \mathds{R}^{MxM}$ of a dataset $\mathbf{\Theta} = \{\mathbf{\Sigma}_1,\dots,\mathbf{\Sigma}_M \}$ where $X_i \in \mathcal{S}_{++}^{N}$.
		\Ensure  Spectral embedding of $\mathbf{\Theta}$ as $\mathbf{Y} \in \mathds{R}^{P\times M}$ for $P<N$
		\Procedure{UnsupervisedSpectralEmbedding}{$\mathbf{S}$}		
		
		\Statex \texttt{Compute Degree Matrix} $\mathbf{D}$
		\State $D_{ii} = \sum_{j=1}^{n}S_{ij} $
		\Statex \texttt{Compute Symmetric Normalized Laplacian}
		\State $\mathbf{L} = \mathbf{D} - \mathbf{S} $ \Comment{Unnormalized Laplacian}
		\State $\mathbf{L}_{sym} = \mathbf{D}^{-1/2}\mathbf{L}\mathbf{D}^{-1/2}  = \mathbf{I} - \mathbf{D}^{-1/2}\mathbf{S}\mathbf{D}^{-1/2}$ 
		\Statex \texttt{Eigendecomposition of} $\mathbf{L}_{sym}$
		\State $\mathbf{L}_{sym} = \mathbf{V}\mathbf{\Lambda} \mathbf{V}^T$ 
		\Statex \texttt{Order Eigenvectors wrt. $\lambda_0=0 \leq \lambda_1 \leq \dots \leq \lambda_{M-1}$}
		\State $\mathbf{V} = [V^{(0)},V^{(1)},\dots,V^{(M)}]$ \Comment{Ordered-columnwise}
		
		\Statex \texttt{Apply SoftMax function on Eigenvalues}
		\State	$\rho(\mathbf{\lambda})_i \leftarrow$ Equation \ref{eq:softmax}
		\Statex \texttt{Normalize weights to} $[-1,1]$
		\State $\overline{\rho(\mathbf{\lambda})_i} = \texttt{normalize}(\rho(\mathbf{\lambda})_i) $
		\Statex \texttt{Find $P$-Primary Eigenvalues}
		\State $P \leftarrow \sum\limits_{i=1}^{N} \delta_i$ where $\delta_i=\begin{cases}
		1, & \text{if $\rho(\mathbf{\lambda})_i < 0$}.\\
		0, & \text{otherwise}.
		\end{cases}$ 
		\Statex \texttt{Construct datapoints on the $P$-dimensional manifold}
		\State $\mathbf{V}^{p} = [V^{(0)},V^{(1)},\dots,V^{(P)}]$ \Comment{keep first $P$-th columns}
		\State $\mathbf{V}^{p}(i,j) = \mathbf{V}^{p}(i,j)/\left(\sum_j\mathbf{V}^{p}(i,j)^2\right)^{1/2}$ \Comment{re-normalize rows}
		\State $\mathbf{y}_i = \mathbf{V}^{p}(i,:) \hspace{5pt} \forall i= \{1,\dots,M\}$ \Comment{vector corresponding to $i$-th row of $\mathbf{V}^{p}$} 
		\EndProcedure
	\end{algorithmic}
\end{algorithm}

\noindent \textit{(2) \underline{Cardinality}:} Once we have automatically determined a $P$-dimensional sub-space, $\mathbf{Y} \in \mathds{R}^{P \times M}$, corresponding to the set of Covariance matrices $\mathbf{\Theta} = \{\mathbf{\Sigma}_1, \dots, \mathbf{\Sigma}_M\}$, we can now apply a clustering approach on $\mathbf{Y}$ in order to find the $K$ expected clusters in $\mathbf{\Theta}$. When using EM-based algorithms such as $K$-Means or Gaussian Mixture Models (GMM), the most appropriate way to find the optimal $K$ is through model selection or heuristics. An alternative approach is to use a Bayesian non-parametric (BNP) model, namely the $\mathcal{DP}$-GMM to automatically estimate the optimal cluster $K$ from $\mathbf{Y}$. While this solves the \textit{cardinality} problem, such distance-based approaches can perform poorly when the distribution of the points of the \textit{spectral embedding}, $\mathbf{Y}$,  exhibits idiosyncrasies such as high curvatures, non-uniformities, etc\footnote{This statement is justified empirically in Section \ref{sec:results}}. In such cases, one could benefit from the similarity values $\mathbf{S} \in \mathds{R}^{M \times M}$ in order to bias the clustering results and provide a robust algorithm to irregular point distributions that might be constructed from the \textit{spectral embedding}. A prior that is capable of including such side-information in a Bayesian non-parametric model is the distance-dependent Chinese Restaurant Process (dd-$\mathcal{CRP}$) \citep{Blei:JMLR:2011}. It is a distribution over partitions that allows for dependencies between data-points, by relaxing the \textit{exchangeability} assumption of the standard $\mathcal{CRP}$ (see Table \ref{tab:bnp_clust_prelims}). It can be applied to any non-temporal settings by using an appropriate distance function on the prior. It has been successfully used with arbitrary distances \citep{Socher:JMLR:2011}, as well as spatial distances between pixels, triangular meshes, voxels \citep{Ghosh:NIPS:2011,Ghosh:NIPS:2012,Janssen:Neuro:2015}. 
In this work, we propose to use the dd-$\mathcal{CRP}$ prior with our similarities $\mathbf{S} \in \mathds{R}^{M\times M}$ (computed from the original Covariance matrix space) \underline{to bias clustering} on a non-parametric mixture model applied on the sub-space of the \textit{spectral embedding} $\mathbf{Y} \in \mathds{R}^{P\times M}$. 

\begin{table}[!t]
\footnotesize
\colorbox{violet!3}{
\resizebox{\textwidth}{!}{\begin{tabular}{p{\linewidth}}

\textbf{Chinese Restaurant Process ($\mathcal{CRP}$)} The $\mathcal{CRP}$ is a distribution over partitions of integers described by a culinary metaphor of a Chinese restaurant with an infinite number of tables. It defines a sequence of probabilities for the incoming customers (observations) to sit at specific tables (clusters) \citep{Jordan:NIPS:2005}. The first customer sits at the first table. The $i$-th customer chooses to either sit at a table with a probability proportional to the number of customers sitting at that table; otherwise she/he sits alone at a new table with a probability proportional to the hyper-parameter $\alpha$ (known as the concentration parameter). This process is summarized as follows:
\begin{equation}
p(z_i = k \hspace{2pt}|\hspace{2pt} Z_{-i}, \alpha) =  
\begin{cases}\!
\frac{M_{(k,-i)}}{\alpha + M - 1} & \text{for}  \hspace{10pt} k \leq K \\
\frac{\alpha}{\alpha + M - 1} & \text{for}  \hspace{10pt} k = K + 1 \\
\end{cases}
\label{eq:crp_prior}
\end{equation}
 $M_{(k,-i)}$ is the number of customers sitting at table $k$, excluding the $i$-th customer and $M$ is the total number of customers. \\ \\

\textbf{Distance-Dependent Chinese Restaurant Process (dd-$\mathcal{CRP}$) \citep{Blei:JMLR:2011}} The dd-$\mathcal{CRP}$ focuses on the probability of customers sitting \textit{with other customers} based on an external measure of distance.  In this case, the culinary metaphor becomes \textit{customer centric}. Now, the $i$-th customer has two choices, she/he can either sit with the $j$-th customer with a probability proportional to a decreasing function of a distance between them $f(d_{ij})$, or sit alone with probability proportional to $\alpha$ as follow,  
\begin{equation}
p(c_i = j \hspace{2pt}|\hspace{2pt}\mathbf{D}, \alpha) \propto  
\begin{cases}\!
f(d_{ij}) & \text{if}  \hspace{10pt} i \neq j \\
\alpha & \text{if}  \hspace{10pt} i = j \\
\end{cases}
\label{eq:ddcrp_prior}
\end{equation}
where $\mathbf{D} \in \mathds{R}^{M\times M}$ is a matrix of pairwise distances between $M$ customers. The smaller the distance between the customers, the more likely they are to sit together and vice versa. This can only hold if the decay function $f: \mathds{R}^{+} \rightarrow \mathds{R}^{+}$ is non-increasing and $f(\infty)=0$. Moreover, the set of \underline{customer} seating assignments $C = \{c_{1},\dots,c_{N}\}$ sampled from the dd-$\mathcal{CRP}$, can be mapped to $Z = \{z_{1},\dots,z_{N}\}$, via $Z = \mathbf{Z}(C)$, which is a recursive mapping function that gathers all linked customers. Hence, the \underline{table} (cluster) assignments emerge from the customer seating assignments. 
\end{tabular}}}
\caption{Bayesian Non-parametric Priors for Partitions\label{tab:bnp_clust_prelims}}
\end{table}

\subsection{SPCM Similarity Dependent - Chinese Restaurant Process ($\mathcal{SPCM-CRP}$) Mixture Model}
\label{sec:ssd-spcm}

\paragraph{Non-parametric Prior} Following the definition of the dd-$\mathcal{CRP}$ in Table \ref{tab:bnp_clust_prelims}, it is clear to see that we can directly use our B-SPCM similarity function to generate a prior distribution over partitions by substituting $f(d_{ij})$, which should be a decaying function with \eqref{eq:bspcm}. This then yields a multinomial over \textit{customer seating} assignments conditioned on B-SPCM similarities $\mathbf{S} \in \mathds{R}^{M\times M}$, where $s_{ij}=f(\Delta_{ij},\tau)$ is generated from \eqref{eq:bspcm} for $i,j=1,\dots M$. Our SPCM-dependent $\mathcal{CRP}$ prior then becomes:
\begin{equation}
p(c_i = j \hspace{2pt}|\hspace{2pt} \mathbf{S}, \alpha) \propto  
\begin{cases}\!
s_{ij} & \text{if}  \hspace{10pt} i \neq j \\
\alpha & \text{if}  \hspace{10pt} i = j \\
\end{cases}
\label{eq:sdcrp_prior}
\end{equation}
We refer to \eqref{eq:sdcrp_prior} as the $\mathcal{SPCM-CRP}$ prior. To summarize, this prior indicates the probability of $c_i = j$; in other words the probability of customer (observation) $i$ sitting together (being in the same cluster) with customer (as observation) $j$, given a deterministic similarity measure between them (B-SPCM).

\begin{figure}[!t]
	\centering
	\begin{minipage}{0.35\textwidth}
	\centering
	\includegraphics[trim={0.15cm 0.85cm 2.2cm 0.5cm},clip,width=\linewidth]{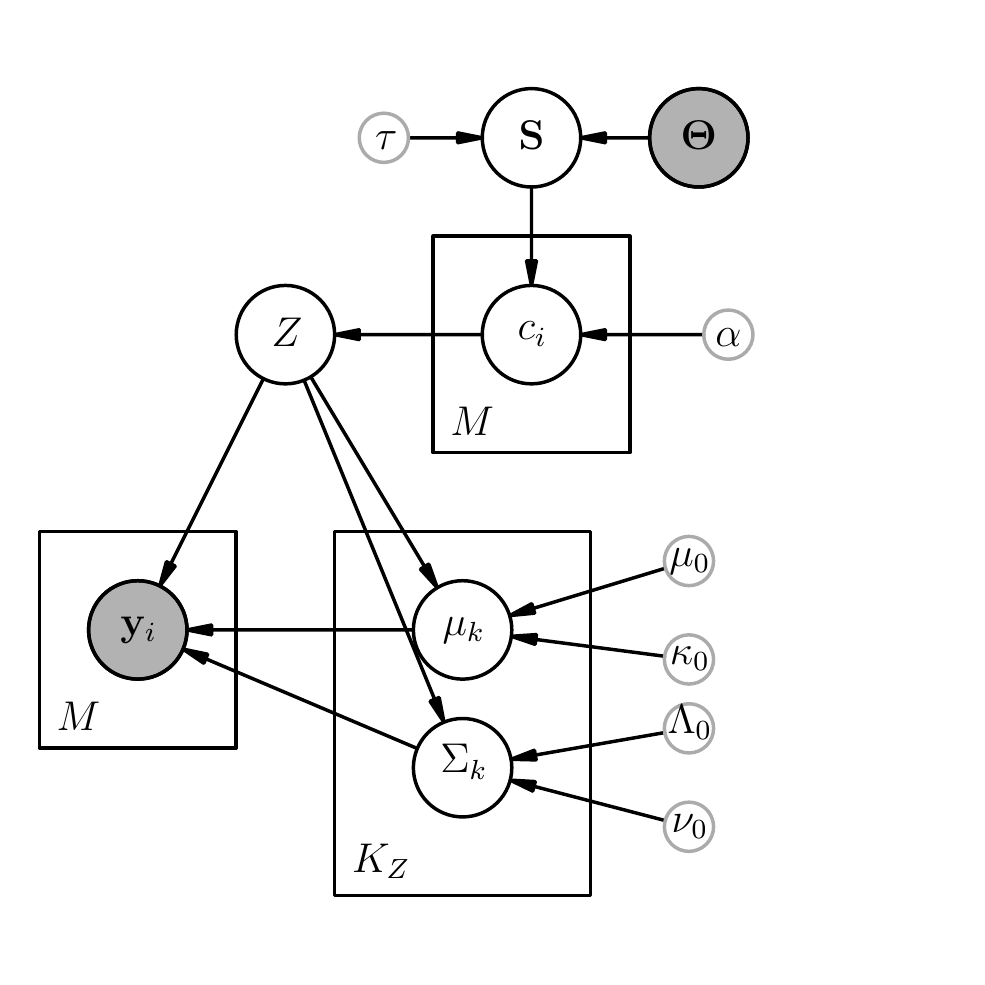}
	\end{minipage}
	\caption{\small Graphical representation of \textbf{our proposed clustering and segmentation approaches} the \textbf{SPCM-CRP} Mixture model. \textbf{Colored gray} nodes correspond to observed variables, \underline{black nodes} correspond to latent variables and \underline{small gray nodes} correspond to hyper-parameters.\label{fig:my_models} }
\end{figure}

\paragraph{Mixture Model} The $\mathcal{SPCM-CRP}$ prior is inherently imposing clusters on a parameter space, we thus formulate a mixture model with the emissions modeled from observations on the spectral sub-space $\mathbf{Y}\in \mathds{R}^{P \times M}$. The $\mathcal{SPCM-CRP}$ mixture model, with generative distribution $ \mathcal{N}(.|\theta)$ and base distribution $ \mathcal{NIW}(\lambda_0)$, can thus be constructed as follows: 
\begin{equation}
\begin{aligned}
c_i & \sim \mathcal{SPCM-CRP}(\mathbf{S},\alpha)\\
z_i & = \mathbf{Z}(c_i)\\
\theta_k & \sim \mathcal{NIW}(\lambda_0)\\
 \mathbf{y}_i|z_i = k & \sim \mathcal{N}(\theta_k)
\end{aligned}
\label{eq:spcm-crp-mm}
\end{equation}

where each $\theta_k=(\mu_k,\Sigma_k)$ are the parameters of a Gaussian distribution, $\mathcal{N}$. For each $k$-th cluster, its parameters are drawn from a $\mathcal{NIW}$ distribution, with hyper-parameter $\{\mu_0,\kappa_0,\Lambda_0, \nu_0 \}$. A graphical representation of this proposed mixture model is illustrated in Figure \ref{fig:my_models}.  In the standard $\mathcal{CRP}$ mixture model a partition $\mathbf{z}$ is directly drawn from the table assignments in \eqref{eq:crp_prior}. In this model, the prior is in terms of customer assignments, $c_i$. Nevertheless, these indirectly determine the cluster assignment $z_i$, through the mapping function $\mathbf{Z}(C):C \rightarrow Z$, which recovers the connections of $c_i$, i.e. the emerged table assignments. It must be noted that this mixture model is not strictly a \textit{Bayesian} non-parametric model, as the $\mathcal{SPCM-CRP}$ is not generated from a \textit{random measure}. However, it still imposes a distribution over infinite partitions, thus keeping the \textit{non-parametric} nature of the $\mathcal{CRP}$ \citep{Blei:JMLR:2011}. For our clustering application, due to conjugacy, we can integrate out the model parameters $\Theta$ and estimate the posterior distribution of the customer assignments $p(C | \mathbf{Y}, \mathbf{S}, \alpha, \lambda)$,
\begin{equation}
\label{eq:full-post}
p(C | \mathbf{Y}, \mathbf{S}, \alpha, \lambda) = \frac{p(C \hspace{2pt}|\hspace{2pt} \mathbf{S}, \alpha) p(\mathbf{Y}| \mathbf{Z}(C), \lambda)}{\sum_C p(C \hspace{2pt}|\hspace{2pt} \mathbf{S}, \alpha) p(\mathbf{Y}| \mathbf{Z}(C), \lambda)}.
\end{equation}
The prior probability $p(C|\mathbf{S},\alpha)$ is determined by the $\mathcal{SPCM-CRP}$ \eqref{eq:sdcrp_prior} and can be computed as follows, 
\begin{equation}
p(C \hspace{2pt}|\hspace{2pt} \mathbf{S}, \alpha) = \prod_{i=1}^{M} p(c_i = j \hspace{2pt}|\hspace{2pt}\mathbf{S}, \alpha),
\end{equation}
where,
\begin{equation}
p(c_i = j \hspace{2pt}|\hspace{2pt} \mathbf{S}, \alpha) =  
\begin{cases}\!
\frac{s_{ij}}{\sum_{j=1}^{M}s_{ij} + \alpha} & \text{if}  \hspace{10pt} i \neq j \\
\frac{\alpha}{M + \alpha} & \text{if}  \hspace{10pt} i = j \\
\end{cases}
\label{eq:spcmcrp_prior}
\end{equation}
The likelihood of the partition $Z=\mathbf{Z}(C)$ is computed as the product of the probabilities of the customers $\mathbf{Y}$ sitting at their assigned tables $Z$, 
\begin{equation}
p(\mathbf{Y}|\mathbf{Z}(C), \lambda) = \prod_{k=1}^{|\mathbf{Z}(C)|} p (\mathbf{Y}_{\mathbf{Z}(C)=k}|\lambda)
\label{eq:spcm-crp_lik}
\end{equation}
where $|\mathbf{Z}(C)|$ denotes the number of unique tables emerged from $\mathbf{Z}(C)$; i.e. $K$ in a finite mixture model, and $\mathbf{Z}(C)=k$ is the set of customers assigned to the $k$-th table. Further, each marginal likelihood in  \eqref{eq:spcm-crp_lik} has the following form,
\begin{equation}
\label{eq:integral}
p (\mathbf{Y}_{\mathbf{Z}(C)=k}|\lambda) = \int\limits_{\theta}^{} \left( \prod_{i \in \mathbf{Z}(C)=k} p\left( \mathbf{y}_{i} \hspace{2pt}|\hspace{2pt} \theta \right) \right)  p\left( \theta\hspace{2pt}|\hspace{2pt}\lambda \right)d\theta.
\end{equation}
Since $p(\mathbf{y}_i \hspace{2pt} |\hspace{2pt} \theta) = \mathcal{N}(\mathbf{y}_i \hspace{2pt} |  \mu, \Sigma)$ and $p(\theta \hspace{2pt} |\hspace{2pt} \lambda)  = \mathcal{NIW}(\mu, \Sigma \hspace{2pt} | \hspace{2pt} \lambda)$,  \eqref{eq:integral} has an analytical solution which can be derived from the posterior $p(\mu, \Sigma |\mathbf{Y})$. The full posterior, \eqref{eq:full-post}, is, however, intractable, as the combinatorial sum in the denominator increases exponentially wrt. $M$. 

Nevertheless, it can be approximated via Gibbs sampling, where latent variables $C$, are sampled from the following posterior distribution,

\begin{equation}
\label{eq:cond_spcm_crp}
\begin{aligned}
p(c_i = j \hspace{2pt}|\hspace{2pt} C_{-i},\mathbf{Y},\mathbf{S}, \alpha, \lambda) \hspace{2pt}
\propto\hspace{2pt}& \\  \textcolor{red}{\underbrace{p(c_i = j \hspace{2pt}|\hspace{2pt}\mathbf{S}, \alpha) }_{\text{\shortstack{Similarities in \\ Original Space}}}} & \textcolor{blue}{\underbrace{p(\mathbf{Y}\hspace{2pt}|\hspace{2pt}\mathbf{Z}(c_i = j \cup C_{-i}), \lambda)}_{\text{\shortstack{Observations in \\ Spectral Embedding}}}} 
\end{aligned}
\end{equation}

with $C_{-i}$ indicating the customer seating assignments for all customers except the $i$-th. \eqref{eq:cond_spcm_crp} holds some particularities as opposed to typical \underline{collapsed conditionals} and typical \underline{mixture models}. First of all, we can see that the prior is represented through customer assignments, while the likelihood is in terms of table assignments. Moreover, due to our adaptation of the dd-$\mathcal{CRP}$, the \textcolor{red}{prior} uses \textcolor{red}{similarity measures from the original space} of the Covariance matrices, while the \textcolor{blue}{likelihood} is computed solely on the observations which live in the \textcolor{blue}{lower-dimensional spectral embeddding} constructed from the pair-wise Similarity matrix $\mathbf{S}$ of Covariance matrices. Finally, the likelihood is computed for all points $\mathbf{Y}$, rather than just for the sampled point $\mathbf{y}_i$, this is due to the fact that the partition of the dataset depends on customer assignments and not table assignments. In the $\mathcal{CRP}-MM$, sampling for new \underline{table} assignments does not affect the overall partition of the data because we assume that each customer is conditionally independent of the other customers' assignment. In the $\mathcal{SPCM-CRP}$ mixture this is not the case, sampling for new \underline{customer} assignments directly affects the partition in several ways; e.g. a table could be split or two tables could be merged; all because the customers are not conditionally independent of the other customers' assignments. Since the $\mathcal{SPCM-CRP}$ has the same form as the dd-$\mathcal{CRP}$, we adapt the Collapsed Gibbs sampler proposed by \cite{Blei:JMLR:2011} for the original dd-$\mathcal{CRP}$ model which takes these special cases into consideration. 

\begin{algorithm}[!t]
	\renewcommand{\algorithmicrequire}{\textbf{Input:}}
	\renewcommand{\algorithmicensure}{\textbf{Output:}}
	\caption{Spectral Non-Parametric Clustering of Covariance Matrices}
	\footnotesize
	\label{alg:nonparam-clustering}
	\begin{algorithmic}[1]
		\Require $\mathbf{\Sigma} = \{\Sigma_1, \dots, \Sigma_N\}$ where $\Sigma \in \mathds{R}^{D \times D}, \Sigma \succeq 0, \Sigma =\Sigma^T $ \Comment{Data}
		\Statex \hspace{15pt} $\tau, \alpha, \lambda = \{\mu_0, \kappa_0, \Lambda_0, \nu_0\}$ \Comment{Hyper-parameters}			
		\Ensure $\Psi=\{K,C,Z,\Theta\}$ \Comment{Inferred Clusters and Cluster indicators}
		\Statex Compute pair-wise B-SPCM similarity values (Eq.\ref{eq:bspcm})
		\State 	$\mathbf{S}\in\mathds{R}^{NxN} \leftarrow s_{ij} = f(\Sigma_i, \Sigma_j, \tau) \quad \forall i,j \in \{1,\dots,N\}$ 
		\Statex Unsupervised Spectral Embedding (Alg.\ref{alg:spect_dim_red})
		\State $\mathbf{Y} \in \mathds{R}^{P \times N} \leftarrow$ \texttt{UnSpectralEmb($\mathbf{S}$)}
		\Procedure{SPCM-CRP-Gibbs-Sampler}{$\mathbf{Y}, \mathbf{S}, \alpha, \lambda$}
		\State Set  $\Psi^{t-1} = \{C,K,Z\}$ where $c_i = i$ for $C = \{c_1, \dots, c_N\}$
		\For{\texttt{iter t = 1 to T}}
		\State Sample a random permutation $\tau(\cdot)$ of integers $\{1, \dots, N \}$.
		\For{\texttt{obs i = $\tau(1)$ to $\tau(N)$}}	
		\State \textbf{Remove} customer assignment $c_i$ from the partition
		\If {$\mathbf{Z}(C_{-i}) \neq \mathbf{Z}(C)$}
		\State \textbf{Update} likelihoods according to Eq. \ref{eq:spcm-crp_lik}  
		\EndIf
		\State \textbf{Sample} new cluster assignment 
		\State \texttt{$c_i^{(i)} \sim p(c_i = j  | C_{-i}, \mathbf{Y}_{-i},   \mathbf{S}, \alpha)$} (Eq. \ref{eq:spcmcrp_cond_final})
		\If {$\mathbf{Z}(C_{-i}) \neq \mathbf{Z}(c_i =j \cup C_{-i})$}
		\State \textbf{Update} table assignments $Z$.
		\EndIf
		\EndFor
		\State \textbf{Resample} table parameters $\Theta$ from $\mathcal{NIW}$ posterior 
		\State update equations \eqref{eq:niw_updates}.
		\EndFor
		\EndProcedure		
	\end{algorithmic}
\end{algorithm}

\subsection{Collapsed Gibbs Sampler for SPCM-CRP Mixture Model} 
\noindent The conditional in \eqref{eq:cond_spcm_crp} is sampled via a two-step procedure:\\
\noindent \underline{Step 1.} The $i$-th customer assignment is removed from the current partition $\mathbf{Z}(C)$. If this causes a change in the partition; i.e. $\mathbf{Z}(C_{-i}) \neq \mathbf{Z}(C)$; the customers previously sitting at $\mathbf{Z}(c_i)$ are split and the likelihood must be updated via \eqref{eq:spcm-crp_lik}.\\
\noindent \underline{Step 2.}  A new customer assignment $c_i$ must be sampled, by doing so a new partition  $\mathbf{Z}(c_i =j \cup C_{-i})$ is generated. This new customer assignment might change (or not) the current partition $\mathbf{Z}(C_{-i})$. If $\mathbf{Z}(C_{-i}) = \mathbf{Z}(c_i =j \cup C_{-i})$, the partition was unchanged and the $i$-th customer either joined an existing table or sat alone. If $\mathbf{Z}(C_{-i}) \neq \mathbf{Z}(c_i =j \cup C_{-i})$, the partition was changed, specifically $c_i = j$ caused two tables to merge, table $l$ which is where the $i$-th customer was sitting prior to step 1 and table $m$ is the new table assignment emerged from the new sample $\mathbf{Z}(c_i = j)$. Due to these effects on the partition, instead of explicitly sampling from Eq. \ref{eq:cond_spcm_crp}, \cite{Blei:JMLR:2011} proposed to sample from the following distribution,
\begin{equation}
p(c_i = j \hspace{2pt}|\hspace{2pt} C_{-i},\mathbf{Y},\mathbf{S}, \alpha, \lambda) \propto
\begin{cases}\!
p(c_i = j | \mathbf{S}, \alpha)\Lambda(\mathbf{Y}, C, \lambda) & \text{if} \hspace{2pt} \texttt{cond} \\
p(c_i = j | \mathbf{S}, \alpha) &  \text{otherwise}, \\
\end{cases}
\label{eq:spcmcrp_cond_final}
\end{equation}
where \texttt{cond} is the condition of $c_i = j \hspace{2pt} \text{merges tables} \hspace{2pt} m \hspace{2pt} \text{and} \hspace{3pt} l$ and $\Lambda(\mathbf{Y}, C, \lambda)$ is equivalent to,
\begin{equation}
\Lambda(\mathbf{Y}, C, \lambda) = \frac{p(\mathbf{Y}_{(\mathbf{Z}(C)=m \hspace{2pt} \cup \hspace{2pt} \mathbf{Z}(C)=l )}|\lambda)}{p(\mathbf{Y}_{\mathbf{Z}(C)=m}|\lambda)p(\mathbf{Y}_{\mathbf{Z}(C)=l}|\lambda)}.
\end{equation}

\noindent This procedure is iterated $T$ times, once it converges, we can sample the table parameters $\Theta = \{\theta_1, \dots, \theta_K\}$ through the posterior of the $\mathcal{NIW}$ distribution \citep{Sudderth:PhD:2006}, refer to Appendix \ref{app:Sample_NIW} for the exact equations. The complete non-parametric clustering algorithm is summarized in Algorithm \ref{alg:nonparam-clustering}, detailing the Collapsed Gibbs sampler steps\footnote{A MATLAB implementation of this clustering approach can be found in \underline{\url{https://github.com/nbfigueroa/SPCM-CRP}}.}. 


\section{Unsupervised Joint Segmentation and Transform-Invariant Action Discovery}\label{sec:tgau_bp_hmm}
\vspace{-5pt}
In this section, we propose a coupled model that addresses the problem of joint segmentation and \textit{transform-invariant} action discovery. To recall, given a dataset of $M$ demonstrations of continuous $N$-dimensional trajectories (i.e. time-series) composed by sequences of multiple \textbf{common} actions, we seek to individually segment each trajectory, while discovering the \textbf{common (i.e. transform-invariant)} actions that describe each segment. We adopt a probabilistic modeling approach and use Hidden Markov Models (HMM) to extract such information. Formulated as an HMM, each trajectory is considered as a sequence of observations $\mathbf{x} = \{\mathbf{x}_t\}_{t=0}^{T-1}$ for $\mathbf{x}_t \in \mathds{R}^{N}$ over $T$ discrete time steps, that are independently sampled and conditioned on an underlying hidden state $\mathbf{s} = \{s_t\}_{t=0}^{T-1}$, that evolves through a first-order temporal Markov process, modeled through a transition probability matrix $\mathbf{\pi} \in \mathds{R}^{K \times K}$. Such hidden states indicate the \textbf{actions} present in each trajectory parametrized by $K$ emission models $\Theta = \{\theta_1, \dots, \theta_K\}$, while the transitions between hidden states $\mathbf{s}_t \rightarrow \mathbf{s}_{t+1}$ denote the \textbf{segmentation points}. 

\subsection{Challenges in HMM-based segmentation}
\vspace{-2pt}
Due to their temporal Markovian assumption and flexibility of modeling different stochastic processes HMMs have become the staple method for analyzing time-series. Nevertheless, applying them to unstructured, unlabeled and transformed data such as our target application becomes quite challenging. We list the three main issues and our proposed solution:

\vspace{5pt}
\noindent \textit{(1) \underline{Cardinality}:} As in the previous section, finding the optimal number of hidden states $K$, through the classic HMM EM-based estimation approach relies on model selection and heuristics. For the specific case of HMMs, such external model fitting approaches tend to over-fit the time-series; i.e. either over-segment or under-segment.

\vspace{5pt}
\noindent \textit{(2) \underline{Fixed Switching Dynamics}:} When modeling \textit{multiple} related time-series with a HMM, the main assumption is that the time series are tied together with the same set of transition dynamics and emission parameters. This might come as a nuisance when we have multiple time-series which are indeed related, but do not necessarily follow the same switching dynamics or use the same emission models in each time-series.

\vspace{5pt}
\noindent \textit{(3) \underline{Transform-Invariance}:} The emission models, $\Theta$, of an HMM are always assumed to be unique. In other words, they are not expected to have any correlations, nor are they invariant to transformations or variations. Moreover, the standard HMM assumption is that the generative distribution parameters associated to the hidden state $s_t$ (i.e. $\theta_k=\{\mu_k,\Sigma_k\}$ for $s_t=k$) are identical between time series.\\

\noindent Challenge (1) is known to be solved by formulating an HMM with the Bayesian Non-Parametric (BNP) treatment. A BNP formulation of an HMM sets an infinite prior on $\mathbf{\pi}$, namely the Hierarchical Dirichlet Process (HDP) \citep{Teh:HDP:2004} or its state-persistent variant the sticky HDP-HMM \citep{Fox:ICML:2008}; and consequently, priors on the associated emission model parameters $\Theta$. This model, however, follows the fixed switching dynamics and hence, cannot account for challenge (2). This strong assumption was then relaxed by \cite{Fox:PhD:2009}, who proposed using an \textit{infinite feature model}, namely the Indian Buffet Process ($\mathcal{IBP}$) (see Table \ref{tab:ibp}), as a prior on the transition distribution,  resulting in a collection of $M$ Bayesian HMMs with independent transition distributions $\mathbf{\pi} =\{\pi^{(1)}, \dots, \pi^{(M)}\}$ with partially shared emission models. Regarding challenge (3), to the best of our knowledge, there is no HMM variant that has addressed it, nor that has addressed challenges (1-2-3) in a joint fashion. 
\begin{figure}[!t]
	\centering
	\begin{minipage}{0.53\linewidth}
		\centering
		\includegraphics[trim={0.75cm 0cm 1.5cm 0cm},clip,width=\linewidth]{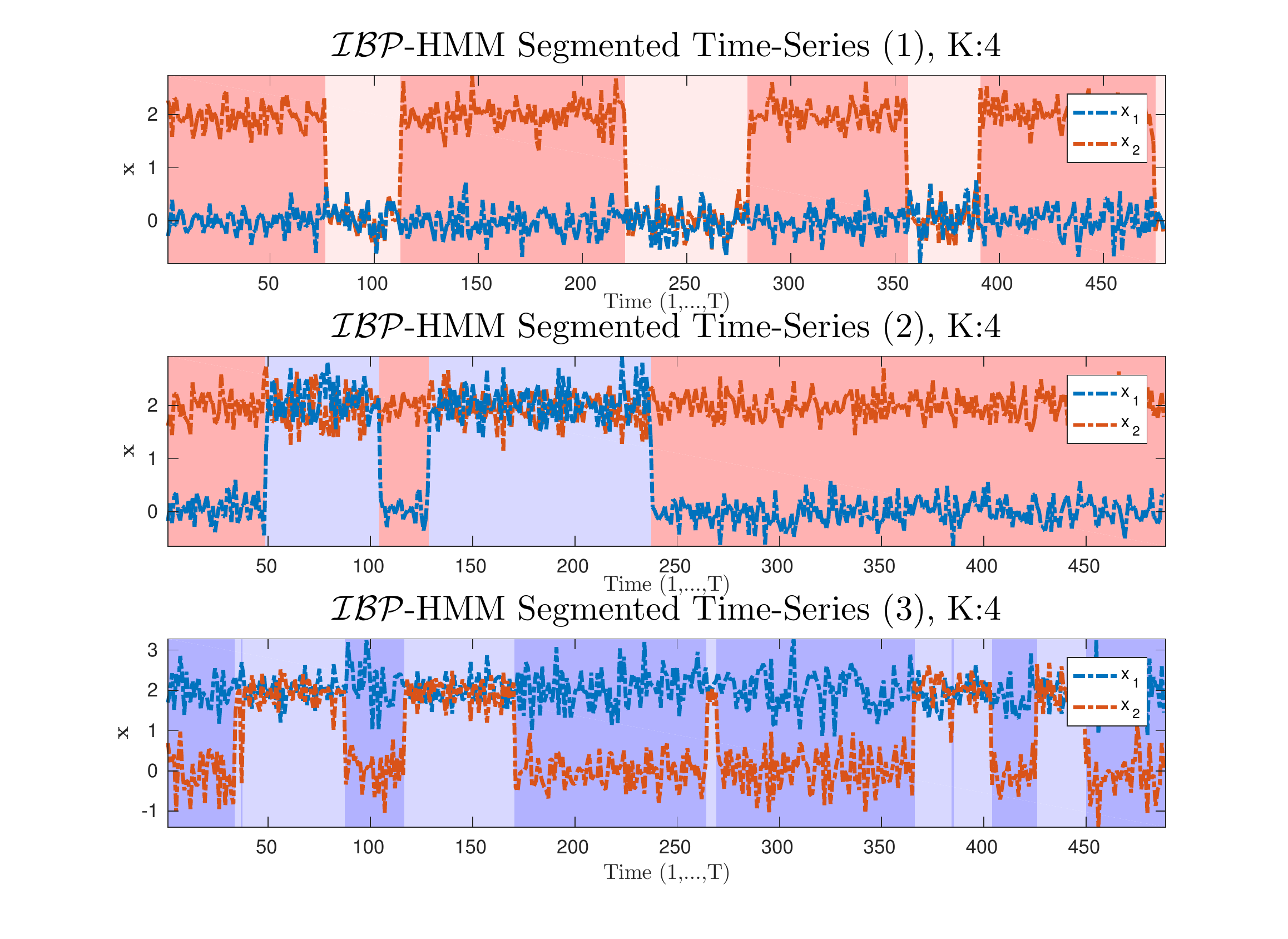}
	\end{minipage}\begin{minipage}{0.48\linewidth}
	\centering
	\includegraphics[trim={0.5cm 0cm 1cm 0cm},clip,width=\linewidth]{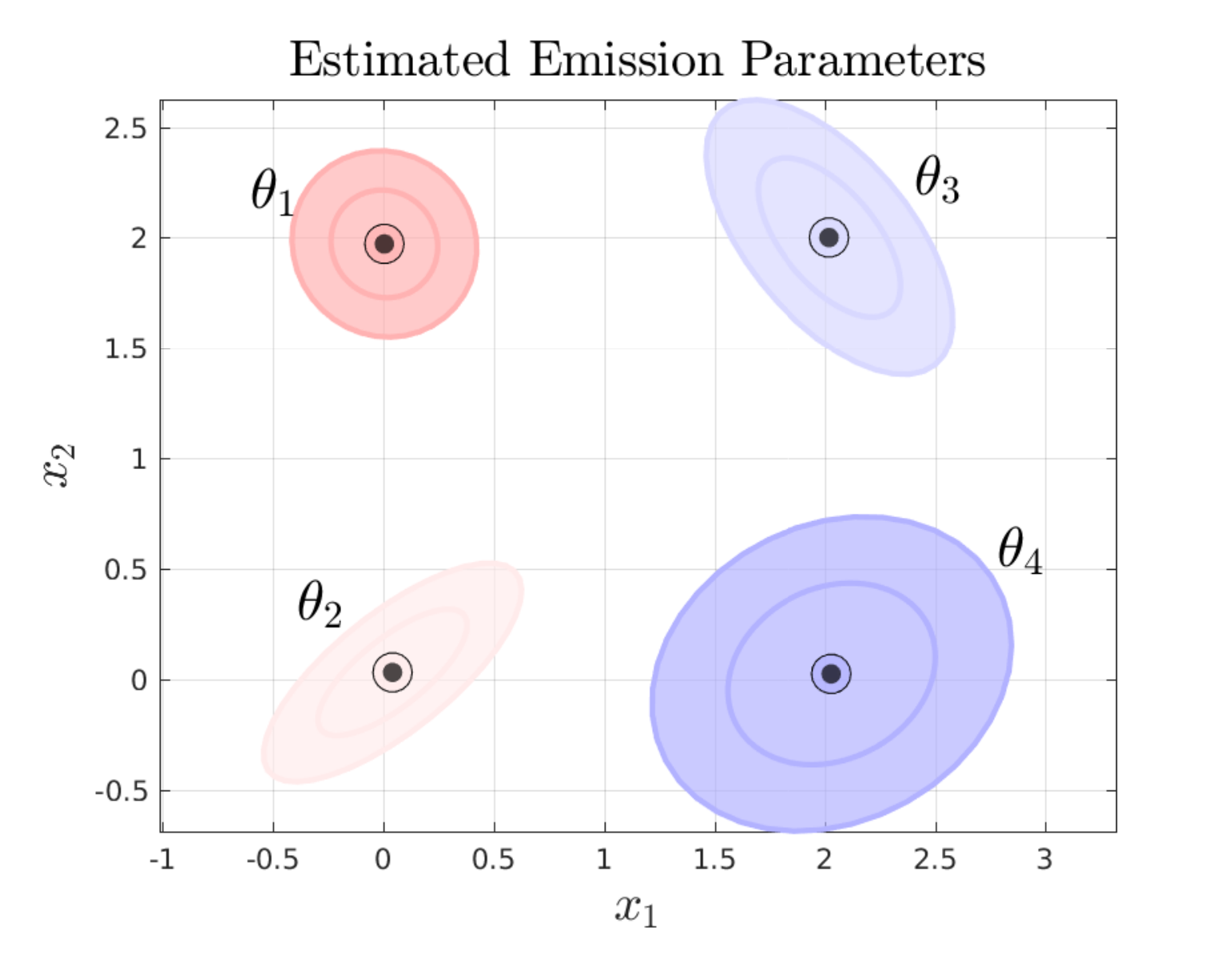}
\end{minipage}
\vspace{-10pt}
\caption{\footnotesize Segmentation of the 2D dataset from Figure \ref{fig:problem2} with $\mathcal{IBP}$-HMM. Colors correspond to feature labels. As can be seen,  $\mathcal{IBP}$-HMM is capable of extracting correct segmentation and \textit{transform-dependent} but cannot associate the \textit{transform-invariant} features; i.e. $\theta_1 \leftrightarrow \theta_4$ and $\theta_2 \leftrightarrow \theta_3$.
	\label{fig:ibp-ts-segmentation} }
\vspace{-15pt}
\end{figure}
The toy example described in Figure \ref{fig:problem2} is a clear motivation for the need of an HMM variant that can handle such challenges. Which is not only relevant for our target application, but for time-series analysis of sensor data in general. Imagine that each of these time-series are observations from spatial tracking sensors. If the sensors are subject to any type of motion and are not properly fixed to the surface, they are surely going to be subject to some sort of transformation, such as translations or rotations. Any variant of an HMM, be it classic HMM with EM-based estimation, $\mathcal{HDP}$-HMM or $\mathcal{IBP}$-HMM, might recover the correct segmentation points, yet, the number of emission models $K$ and consequently the number of hidden states will be \textit{transform-dependent}, as seen show in Figure \ref{fig:ibp-ts-segmentation}\footnote{This claim is supported empirically in the testing scripts provided in the accompanying code:\\ \url{https://github.com/nbfigueroa/ICSC-HMM}}. Even though the latter approach is capable of capturing the inclusion of each emission model in each time-series, it is not equipped with finding the similarities between them. Thankfully, finding these similarities was the desiderata of the approaches presented in Sections \ref{sec:full_spcm} and \ref{sec:spcm_crp}. Hence, in this section, \textbf{our third and final contribution}, we focus on providing \textit{transform-invariance} to the already flexible $\mathcal{IBP}$-HMM by coupling it with our proposed SPCM-$\mathcal{CRP}$ mixture model. Following we summarize a naive approach towards providing \textit{transform-invariance} for the $\mathcal{IBP}$-HMM which motivates our proposed $\mathcal{IBP}$ Coupled SPCM-CRP Hidden Markov Models (ICSC-HMM).

\begin{table}[!t]
	\footnotesize
	\centering
	\colorbox{violet!3}{
		\resizebox{\textwidth}{!}{\begin{tabular}{p{\linewidth}}
			\noindent \textbf{Beta-Bernoulli Process} $\mathcal{BP}-BeP(M,c, \gamma)$: The beta-Bernoulli process is a stochastic process which provides a BNP prior for models involving collections of infinite binary-valued features.  A draw from the beta process yields an infinite collection of probabilities in the unit interval, $B \sim \mathcal{BP}(c, B_0 )$ for a concentration parameter $c > 0$ and base measure $B_0$ on $\Theta$ with total mass $B_0(\Theta) = \gamma$. In other words, it provides an infinite collection of coin-tossing probabilities, that are represented by a set of global weights that determine the potentially infinite number of features, $B = \sum_{k=1}^{\infty} \omega_k \delta_{\theta_k}$,  represented with $\theta_k$ and $\omega_k \in (0,1)$  indicate the inclusion of the $k$-th feature. The draw $B$ is then linked to $M$ Bernoulli process draws to generate the binary-valued feature matrix $\mathbf{F}$; this process is summarized as follows: 
			\begin{equation}
			\begin{aligned}
			\label{eq:BP}
			B | B_0 & \sim \mathcal{BP}(c, \gamma B_0 )\\
			X_i | B & \sim \text{BeP}(B) \quad i = 1, \cdots , M
			\end{aligned}
			\end{equation}
			The draw $B$ provides the set of global weights for the potentially infinite number of features. Then, for each $i$-th time series, an $X_i = \sum_{k=1} f_{ik} \delta_{\theta_k}$, is drawn from a Bernoulli Process (BeP). Each $X_i$ is thus used to construct the binary vector $\mathbf{f}_i$ through independent Bernoulli draws $f_{ik} \sim \text{Bernoulli}(\omega_k)$. We can denote \eqref{eq:BP} as $\mathbf{F} \sim \mathcal{BP}-BeP(M, c, \gamma)$. \\ \\
			
			\noindent \textbf{Indian Buffet Process} $\mathcal{IBP}$: As proven in \cite{Thibaux}, marginalizing out $B$ and setting $c=1$ induces the predictive distribution of the Indian Buffet Process ($\mathcal{IBP}$). It is described through a culinary metaphor consisting of an infinitely long buffet line of dishes, or \textit{features}. The first arriving customer (i.e. object) chooses Poisson($\gamma$) dishes. The following $i$-th customers select a previously tasted dish $k$ with probability $\frac{m_k}{i}$ proportional to the number of previous customers $m_k$ that have tried it, and also samples Poisson($\frac{\gamma}{i}$) new dishes. Assuming that the $i$-th object is the last customer, the following conditional distribution can be derived for the $\mathcal{IBP}$:
			\begin{equation}
			p(f_{ik} = 1 \hspace{2pt}|\hspace{2pt} \mathbf{f}_{-i,k}) =  
			\begin{cases}\!
			\frac{m_{-i,k}}{M} & \text{for}  \hspace{10pt} k \quad \text{with} \quad m_{-i,k}>0 \\
			\end{cases}
			\label{eq:ibp_prior}
			\end{equation} 
			Moreover, the number of new features associated with the $i$-th object is drawn from a
			$\text{Poisson}(\gamma/M)$ distribution.
		\end{tabular}}}
			\caption{Bayesian Non-Parametric Priors for Feature Models 	\label{tab:ibp}}
	\end{table}

\subsection{Naive Transform-Invariance for the $\mathcal{IBP}$-HMM }
\noindent \textit{$\mathcal{IBP}$-HMM.} Before describing a \textit{naive} approach to impose \textit{transform-invariance} on the $\mathcal{IBP}$-HMM, we summarize it briefly. As mentioned earlier, the $\mathcal{IBP}$-HMM is a collection of $M$ Bayesian HMMs, each with its own independent transition distribution $\pi^{(i)}$ for $i=\{1,\dots,M\}$ time-series defined over an unbounded set of shared emission models $\Theta = \{\theta_1,\dots,\theta_K\}$ for $K \rightarrow \infty$. The emission models that are used in each time-series are defined by the binary-valued matrix $\mathbf{F} \in \mathbb{I}^{M\times K}$, sampled from a beta-Bernoulli process $\mathcal{BP}-BeP(M,1, \gamma)$ (see Table \ref{tab:ibp}). The columns of $\mathbf{F}$ correspond to the time-series, while the rows indicate the inclusion of the $k$-the feature (i.e. emission model). Hence, given $\mathbf{f}_i$ we can define a feature-constrained transition distribution $\pi^{(i)}$, by first defining a doubly infinite collection of random variables $\eta_{jk}^{(i)}  \sim \text{Gamma}(\alpha + \kappa\delta_{jk},1)$,
where $\delta_{jk}$ is the Kronecker delta function and $\kappa$ is the sticky hyper-parameter introduced in \cite{Fox:ICML:2008} which places extra expected mass on self-transitions to induce state persistence. The transition variables $\mathbf{\eta}^{(i)}$ are normalized over the set of emission models present in the $i$-th time-series $\mathbf{f}_i$ to generate the corresponding transition distributions $\pi_{jk^{(i)}}$. Once we sample the corresponding transition distributions $\pi^{(i)}$ for each $i$-th HMM, we assign a state $s_{t}^{(i)}=k$ at each $t$-th time step, which determines the model parameters $\theta_k$ that generated $\mathbf{x}_{t}^{(i)}$ (observed data of $i$-th time series at step $t$). The full $\mathcal{IBP}$-HMM is summarized as follows: 
\begin{equation}
\begin{aligned}
\label{eq:IBP-HMM}	
\mathbf{F} & \sim \mathcal{BP}-BeP(M, 1, \gamma)  \\	
\eta_{jk}^{(i)} & \sim \text{Gamma}(\alpha_{b} + \kappa\delta_{jk},1) \\
s_t^{(i)} & \sim \mathbf{\pi}_{s_{t-1}}^{(i)} \qquad  \text{with} \quad \pi_{jk}^{(i)}  = \frac{\eta_{jk}^{(i)}f_{k}^{(i)}}{\sum_{n}  f_{n}^{(i)}\eta_{jn}^{(i)}}\\
\mathbf{x}_{t}^{(i)}|s_{t}^{(i)}& =k  \sim \mathcal{N}(\theta_k) \quad  \text{with} \quad \theta_k \sim \mathcal{NIW}(\lambda)
\end{aligned}
\end{equation}
where each $\theta_k=(\mu_k,\Sigma_k)$ are the parameters of a Gaussian distribution, which are drawn from a $\mathcal{NIW}$ distribution, with hyper-parameter $\lambda =\{\mu_0,\kappa_0,\Lambda_0, \nu_0 \}$. Posterior inference of the latent variables of the $\mathcal{IBP}$-HMM, namely $\{\mathbf{F},\mathbf{S},\mathbf{\Theta}\}$ is performed through a Markov Chain Monte Carlo (MCMC) method that estimates the joint posterior with collapsed $\mathbf{\Theta}$, namely $p(\mathbf{F},\mathbf{S},\mathbf{X})$.
This is achieved by alternating between (i) re-sampling $\mathbf{F}$ given the dynamics parameters $\{\eta^{(i)}, \Theta\} $ and observations $\mathbf{X}$ and (ii) sampling $\{\eta^{(i)}, \Theta\} $ given $\mathbf{F}$ and $\mathbf{X}$, accomplished by interleaving Metropolis-Hastings and Gibbs sampling updates, as listed in Algorithm \ref{alg:nonparam-segmentation}. For sake of simplicity, we have abstained from describing each step from the $\mathcal{IBP}$-HMM sampler, as it is not our contribution and they are thoroughly described in \cite{Fox:PhD:2009} and \cite{Hughes:NIPS:2012}. Instead, we focus on the sampling steps that we add for our proposed coupled model, which are highlighted in \textcolor{blue}{blue} and described in Section \ref{sec:ICSC-HMM}.\\

\noindent \textit{Naive Approach.} To address the problem of \textit{transform-invariance} in the $\mathcal{IBP}$-HMM we can use the proposed B-$\mathcal{SPCM}$ similarity function \eqref{eq:bspcm} to \textit{merge} the features (i.e. emission models $\theta_k$) that are similar and directly generate the \textit{transform-invariant} states labels. To do so, one could use split-merge moves based on sequential allocation, proposed for the $\mathcal{IBP}$-HMM \cite{Hughes:NIPS:2012}. This is a re-sampling step that proposes new candidate configurations for feature matrix $\mathbf{F}$ and state sequences $\{s^{(i)}\}_{i=1}^M$ via accepts/rejects of Metropolis-Hastings updates. The proposals indicate if one should  either merge features which are \textit{similar} or split features which explain the data in a better way separately. The measure of similarity used here is the \textit{marginal likelihood ratio}\footnote{The algorithmic details of this sampling step are presented in Appendix \ref{app:split-merge}.}  One could easily substitute this measure of similarity with our proposed similarity function \eqref{eq:bspcm} in order to force the merging of \textit{transform-dependent} features into 1 \textit{transform-invariant} feature. This approach might seem to solve our problem from the \textit{feature model} point-of-view, as the feature matrix $\mathbf{F}$ will now have a number of columns equal to the \textit{transform-invariant} emission models, however, it has two main disadvantages:\\

\noindent (1) \textit{Computational burden}. The MCMC sampling scheme used to infer the latent variables of the  $\mathcal{IBP}$-HMM involves mostly collapsed sampling steps \citep{Fox:NIPS:2009}. For the split/merge moves, $\Theta$ is collapsed away and solely the data assigned to the features is used to compute the marginal likelihoods. Hence, re-sampling of $\mathbf{F}$ and  $\{s^{(i)}\}_{i=1}^M$ is only dependent on their current values and the data itself, not on the emission model parameters associated with the features. Thus, in order for us to use \eqref{eq:bspcm} we must sample $\Theta$ from the current state sequence $\{s^{(i)}\}_{i=1}^M$ and then apply the split/merge moves. Something that is sought to be avoided in the original MCMC sampler, i.e. $\Theta$ is sampled until the end.

\begin{figure}[!h]
	\centering
	\includegraphics[trim={0.15cm 0.65cm 0.25cm 0.5cm},clip,width=0.45\linewidth]{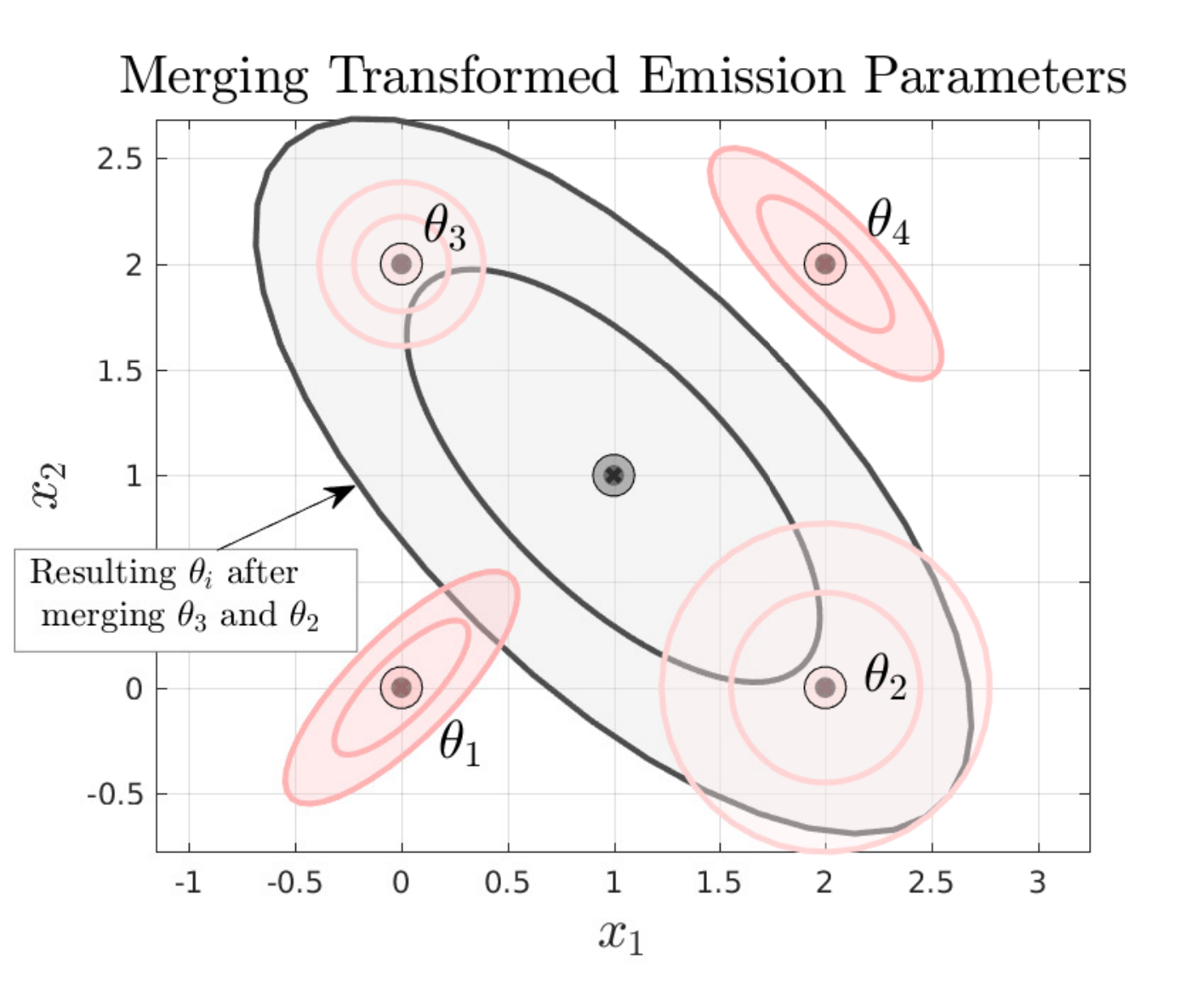}	
	\caption{\small Resulting emission model (depicted by grey ellipse) from \textit{naive merging} of transform-dependent features $\theta_2$ and $\theta_3$ from the toy example.\label{fig:wrong_emission} }
\end{figure}

\noindent (2) \textit{Meaningless Emission Models}. While merging two features which are \textit{transform-dependent}, the new emission model parameters $\Theta$ will not be the \textit{transform-invariant} emission model, but rather a meaningless Gaussian distribution encapsulating the two sets of transformed data as shown in Figure \ref{fig:wrong_emission} for our illustrative example. This might not be a problem if the \textit{transformed} data-points of a similar nature are far from other; however, if they are close, we are vulnerable to merging features that are not similar at all, leading to i) incorrect and meaningless emission models and ii) loss of correct segmentation points. Due to this observation, we posit that such \textit{transform-invariance} cannot be imposed internally in the inference scheme of the HMM parameters, but rather externally, once the emission model parameters $\Theta$ are properly estimated.

\subsection{The IBP Coupled SPCM-CRP Hidden Markov Model}
\label{sec:ICSC-HMM} 		
\begin{figure*}[!t]
	\centering
	\begin{minipage}{0.35\textwidth}
	\centering
	\includegraphics[trim={6cm 13.5cm 4cm 8cm},clip,width=\linewidth]{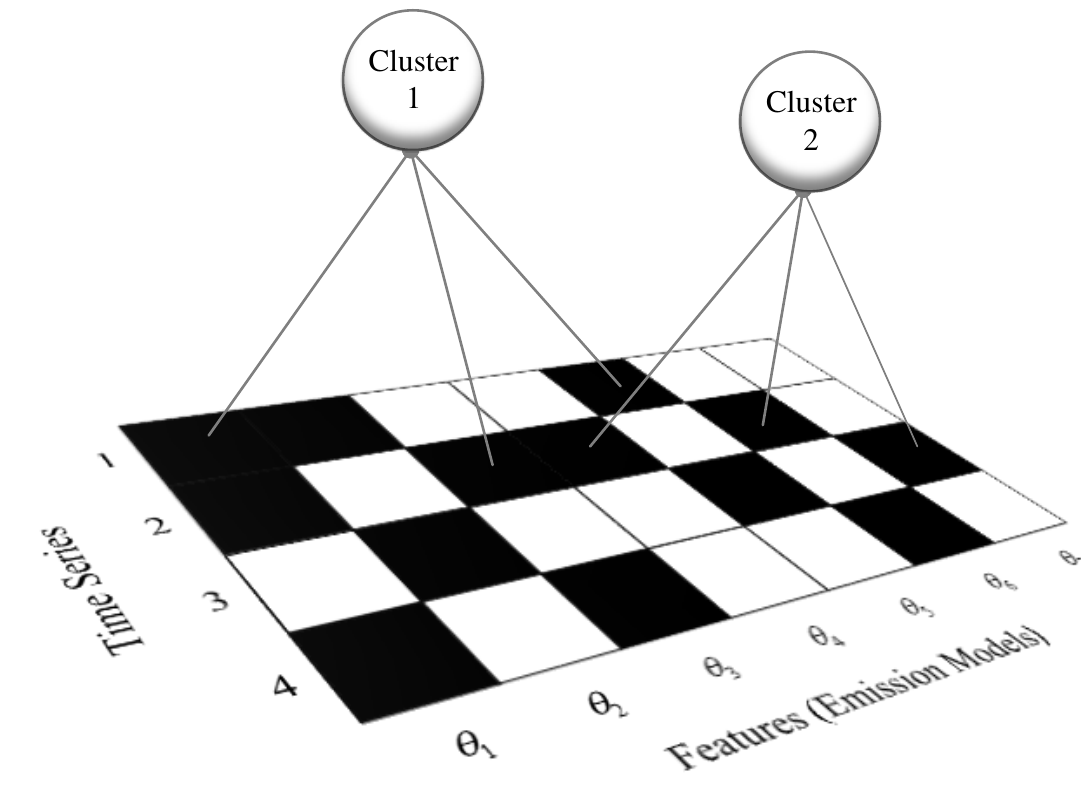}
	\end{minipage}\hspace{10pt}\begin{minipage}{0.6\textwidth}
	`	\includegraphics[trim={0.5cm 0.5cm 1.5cm 3cm},clip,width=\linewidth]{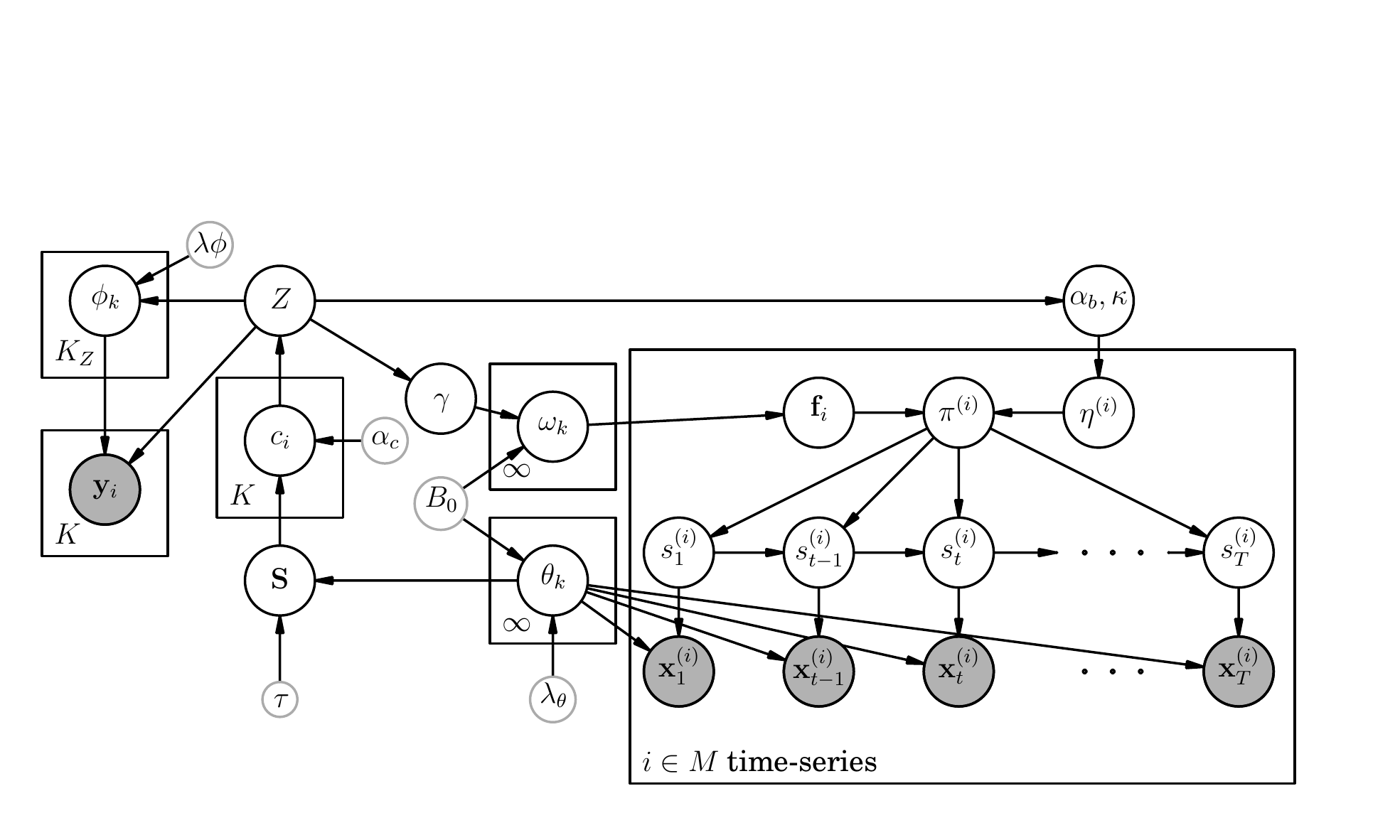}
	\end{minipage}
	\caption{\small (left) Graphical representation of our $\mathcal{IBP}$-HMM feature clustering idea and (right) the $\mathcal{IBP}$ Coupled SPCM-CRP Hidden Markov Models (\textbf{ICSC-HMM}). \textbf{Colored gray} nodes correspond to observed variables, \underline{black nodes} correspond to latent variables and \underline{small gray nodes} correspond to hyper-parameters.\label{fig:something} }
\end{figure*}

We impose \textit{transform-invariance} in the $\mathcal{IBP}$-HMM by proposing a \textit{topic-model-inspired} coupling between the $\mathcal{SPCM-CRP}$ mixture model and the $\mathcal{IBP}$-HMM, as shown in Figure \ref{fig:something} (left). The idea is quite straightforward, we want to jointly estimate the \textit{transform-dependent} feature matrix $\mathbf{F}$ as well as clusters that might arise from similarities in these features; without the two models hindering each others processes. This will lead to \textit{transform-dependent} state sequences $S = \{s^{(i)}\}_{i=1}^M$ as well as \textit{transform-invariant} state sequences $Z = \{z^{(i)}\}_{i=1}^M$, as shown in Figure \ref{fig:problem2}. One might argue that these two sets of \textit{state labels} could be achieved in a decoupled manner; i.e. estimate $\Theta$ with the original $\mathcal{IBP}$-HMM and then cluster the emission models with the $\mathcal{SPCM-CRP}$ mixture model. This is, of course, a valid approach which will be evaluated in Section \ref{sec:results}. By coupling them, we can alleviate the need for setting the hyper-parameters of the $\mathcal{IBP}$-HMM, namely $\alpha_b, \kappa$ and $\gamma$, which correspond to the hyper-parameters for the Dirichlet transition distribution used to sample $\eta^{(i)}$ and the hyper-parameter for the beta-Bernoulli process used to sample $\mathbf{f}_i$; both of which have a direct influence on the feature-constrained transition distributions $\pi^{(i)}$. The proposed coupling results in \textbf{our final contribution} which we refer to as the IBP Coupled SPCM-CRP Hidden Markov Models (ICSC-HMM), shown in Figure \ref{fig:something} (left).  

We begin our model coupling by placing a prior on the $\mathcal{IBP}$ hyper-parameter $\gamma$, parametrized by the number of clusters $K_{Z}$ obtained by sampling $Z$ from the SPCM-$\mathcal{CRP}$ mixture model on $\Theta$. The conditional posterior of $\mathbf{F}$ can be expressed as follows:
\begin{equation}
\label{eq:ibp_cond}
p(\mathbf{F} | \gamma) \propto \gamma^{K_+}\exp\left(-\gamma H_M \right)
\end{equation}
where $H_M$ is the $M$-th harmonic number, defined by $H_M = \sum_{j=1}^{M}\frac{1}{j} $ and $K_+$ is the number of \textit{unique features} (i.e. those that are unique for each time-series). \eqref{eq:ibp_cond} takes form due to the $\mathcal{IBP}$ marginal distribution \citep{Thibaux}. Hence, placing a conjugate Gamma prior on $\gamma \sim \text{Gamma}(a_{\gamma},b_{\gamma})$ yields a closed form solution to its posterior distribution, as follows:
\begin{equation}
\begin{split}
p(\gamma | \mathbf{F}, a_{\gamma},b_{\gamma})  & \propto p(\mathbf{F} | \gamma)p(\gamma | a_{\gamma},b_{\gamma})\\
& \propto \gamma^{K_{+}}\exp\left(-\gamma H_M\right)\frac{\gamma^{a_{\gamma}-1}\exp(-b_{\gamma}\gamma)}{\Gamma(\gamma)}\\
& = \text{Gamma}(a_{\gamma} + K_{+}, b_{\gamma} + H_M)\\
\end{split}
\label{eq:gamma_distsadsd}
\end{equation}
where $K$ is the total number of current features \citep{Fox:PhD:2009}. The task of defining the values for the hyper-priors $a_{\gamma},b_{\gamma}$ is often overlooked in literature and set empirically. In this work, rather than setting the hyper-priors to some empirically fixed values, we adopt a data-driven approach which defines  $a_{\gamma},b_{\gamma}$ at each iteration as follows:
\begin{equation}
\begin{split}
a_{\gamma} =b_{\gamma}=\frac{K}{K_{Z}M} \quad \text{for} \quad K_{Z} = |\mathbf{Z}(C)|
\end{split}
\label{eq:gamma_coupling}
\end{equation}
The intuition behind \eqref{eq:gamma_coupling} is to set $a_{\gamma},b_{\gamma}$ as the ratio of \textit{transform-invariant} features in the total set of features, scaled by the number of time-series $M$. This ratio will implicitly control the number of new features to be sampled for the
$\mathcal{IBP}$ prior. When $K_Z << K$ is low, this indicates that the current sampled features are very similar. In this situation, the ratio is high, inducing $\gamma$ to be higher, in order to increase the number of unique features and consequently the columns of $\mathbf{F}$. Conversely, when $K_Z = K$, it means that all features are different and hence, it is not necessary to push towards sampling more unique features. Regarding the hyper-parameters on $\eta^{(i)}$, namely $\alpha_b$ and $\kappa$, we also place Gamma priors:
\begin{equation}
\begin{split}
\alpha_b & \sim \text{Gamma}(a_{\alpha_b}, b_{\alpha_b})\\
\kappa & \sim \text{Gamma}(a_{\kappa}, b_{\kappa} )\\
\end{split}
\label{eq:kappa_coupling}
\end{equation}

In this case, we couple the models by setting $a_{\alpha_b}, b_{\alpha_b}$
equivalent to \eqref{eq:gamma_coupling} and $a_{\alpha_{\kappa}}, b_{\alpha_{\kappa}} = \frac{K}{K_{Z}}$, as it is the sticky-parameter inducing state-persistence, we force it to be higher than the concentration parameter $\alpha_b$. As discussed in \cite{Fox:PhD:2009}, sampling from $\eta^{(i)}$ is a non-conjugate process and thus does not have a closed form solution. Instead we must apply Metropolis-Hastings steps to iteratively re-sample $\alpha_{b}|\kappa$ and $\kappa|\alpha_{b}$, a detailed description of these steps is provided in \cite{Fox:PhD:2009}. As in the $\mathcal{IBP}$-HMM, posterior inference of the full model, namely $\{\mathbf{F},\mathbf{S},\mathbf{\Theta}, \mathbf{Z},\mathbf{\Phi},\}$ is performed through an MCMC sampler that estimates the joint posterior of the latent variables with collapsed $\mathbf{\Theta}$ and $\mathbf{\Phi}$, namely:
\begin{equation}
\begin{split}
p(\mathbf{F},S,\mathbf{X}, Z, \mathbf{Y}) \propto \textcolor{blue}{\underbrace{p(C \hspace{2pt}|\hspace{2pt} \mathbf{S}, \alpha) p(\mathbf{Y}| \mathbf{Z}(C), \lambda_{\phi})}_{\text{SPCM-$CRP$ Mixture}}} \textcolor{red}{\underbrace{p(\mathbf{F} | \gamma)}_{\mathcal{IBP}}}\\
\textcolor{violet}{\underbrace{\prod_{i=1}^{M}p\left(\mathbf{s}^{(i)}| \mathbf{f}_i, \alpha_b, \kappa \right) \prod_{i=1}^{M}\prod_{t=1}^{T}p\left(s_t^{(i)}|s_{t-1}^{(i)}\right)p(\mathbf{x}_{t}^{(i)}|s_t^{(i)},\theta_{s_t^{(i)}}, \lambda_{\theta})}_{\text{$M$ Bayesian HMM's}}}.
\end{split}
\label{eq:icsc-joint}
\end{equation}

\begin{algorithm}[!t]
	\renewcommand{\algorithmicrequire}{\textbf{Input:}}
	\renewcommand{\algorithmicensure}{\textbf{Output:}}
	\caption{Joint Segmentation and Transform-Invariant Action Discovery MCMC Sampler}
	\footnotesize
	\label{alg:nonparam-segmentation}
	\begin{algorithmic}[1]
		\Require $\mathbf{X} = \{\mathbf{x}^{(i)}\}_{i=1}^{M}$ for $\mathbf{x} = \{\mathbf{x}_t\}_{t=0}^{T-1}$ where $\mathbf{x}_t \in \mathds{R}^{N}$ \Comment{$M$ Time-Series}
		\Statex \hspace{15pt}$\{\tau, \kappa\}$ \Comment{Hyper-parameters}			
		\Ensure $\Psi=\{K,S,\Theta,K_{Z},Z,\Phi\}$ \Comment{Segments \& State Clustering}
		\Procedure{ICSC-HMM-MCMC-Sampler}{}
		\State Set  $\Psi^{t-1} = \{\cdot\}$  \Comment{Initialize Markov Chain}
		\For{\texttt{iter t = 1 to Maxiter}}
		\State Set $\{\eta^{(i)}\} = \{\eta^{(i)}\}^{(t-1)}$ 
		, $\{\mathbf{\mu}_k,\mathbf{\Sigma}_k\}=\{\mathbf{\mu}_k,\mathbf{\Sigma}_k\}^{(t-1)}$ , $\mathbf{F}= \mathbf{F}^{(t-1)}$
		\State From current $\mathbf{F}$, compute count vector $\mathbf{m} = [m_1,\dots,m_{K_{\theta}+}]$ \\ \hfill with $m_k =$ number of time-series possessing feature $k$.
		\State \textbf{Sample} Shared Features $f_{ik}$ with Const. MH Updates \cite{Fox:PhD:2009}
		\State \textbf{Sample} Unique Features $f_{ik}$ with RJ-DD Updates \cite{Fox:PhD:2009}
		\State \textbf{Sample} State Sequence $\{s^{(i)}\}_{i=1}^M$ with Gibbs sampler \cite{Fox:PhD:2009}
		\State \textbf{Re-sample} $\mathbf{F}$ and  $\{s^{(i)}\}_{i=1}^M$ with split/merge moves \cite{Hughes:NIPS:2012}
		\State \textbf{Sample} trans. probabilities $\{\eta^{(i)}\}$ with Gibbs updates \cite{Fox:PhD:2009}
		\State \textbf{Sample} emission parameters $\Theta$ with Gibbs updates \cite{Fox:PhD:2009}
		\State \textcolor{blue}{\textbf{Sample} parameter clusters $\{z_i\}_{i=1}^{K_{Z}}$ with Gibbs updates}
		\State \textcolor{blue}{$C \sim SPCM-\mathcal{CRP}-MM(\Theta)$ (Algorithm \ref{alg:nonparam-clustering})}
        \State \textcolor{blue}{\textbf{Sample} hyper-parameter $\gamma$ with Gibbs updates \eqref{eq:gamma_distsadsd}\eqref{eq:gamma_coupling}}        
        \State \textcolor{blue}{\textbf{Sample} hyper-parameters $\alpha_b,\kappa$ with MH updates \eqref{eq:kappa_coupling} \cite{Fox:PhD:2009}} 
		\EndFor
		\EndProcedure		
	\end{algorithmic}
\end{algorithm}

Given that the couplings between the SPCM-$\mathcal{CRP}$ and the $\mathcal{IBP}$ are linked via deterministic equations that parametrize the hyper-priors, \eqref{eq:icsc-joint} can be estimated by following the same sampling steps as the original $\mathcal{IBP}$-HMM with three main re-sampling steps after sampling $\Theta$.  Namely, we must run the Collapsed sampler for the SPCM-$\mathcal{CRP}$ mixture model listed in Algorithm \ref{alg:nonparam-clustering}. However, instead of letting the chain run for \texttt{Maxiter} steps, the number of iterations depends on the current Markov Chain state of the $\mathcal{IBP}$-HMM $\Psi^{(t)}$. More specifically, if the current estimated features $K^{(t)}$ have not changed from the previous sample $K^{(t-1)}$ we run a limited number of iterations, namely \texttt{iter $\leq$ 5}, and initialize the next chain with the previous cluster assignments $C^{(t-1)}$ Whereas, if the features have changed we reset the chain and let it run for \texttt{iter $\leq$ 10}. Once $C$ has been sampled, we compute $Z = \mathbf{Z}(C)$ and can parametrize the hyper-priors for $\gamma, \alpha_b, \kappa$ for \eqref{eq:gamma_coupling} and \eqref{eq:kappa_coupling}\footnote{MATLAB code of this sampler is provided in: \url{https://github.com/nbfigueroa/ICSC-HMM}}.

\vspace{-10pt}
\section{Evaluation and Applications}
\label{sec:results}
\subsection{Datasets \& Metrics for Similarity Function and Clustering}
To evaluate the proposed similarity function (B-SPCM) (Section \ref{sec:full_spcm}) against standard similarity metrics and the proposed clustering algorithm (SPCM-CRP-MM) (Section \ref{sec:spcm_crp}), we use the following datasets and metrics:\\

\noindent \textbf{Toy 6D Dataset:} This is a synthetic 6-D ellipsoid dataset. It is generated from a set of 3 unique 6-D Covariance matrices. Each unique Covariance matrix has the following values:
\small
\begin{align*}
\textcolor{black}{\Sigma_1} & \textcolor{black}{=\mathbf{I}_6 \lambda_1 \mathbf{I}_6,} \qquad \textcolor{black}{\Sigma_2 = \mathbf{I}_6 \lambda_2 \mathbf{I}_6,} \qquad \textcolor{black}{\Sigma_3 = \mathbf{I}_6 \lambda_3 \mathbf{I}_6}\\
\textcolor{black}{\lambda_1} & \textcolor{black}{= \begin{bmatrix}    
  |\epsilon_1|, |\epsilon_1|,|\epsilon_1|,|\epsilon_1|,|\epsilon_1|,|\epsilon_1|
\end{bmatrix}^T}\\
\textcolor{black}{\lambda_2} &   = \textcolor{black}{\begin{bmatrix}    
  |\epsilon_2|,|10\epsilon_2|,|10\epsilon_2|,|10\epsilon_2|,|\epsilon_2|,|\epsilon_2|
\end{bmatrix}^T}\\
\textcolor{black}{\lambda_3} & = \textcolor{black}{\begin{bmatrix}
    |\epsilon_3|,|10\epsilon_3|,|20\epsilon_3| ,|30\epsilon_3|,|40\epsilon_3|, |50\epsilon_3|
\end{bmatrix}^T}.
\end{align*}
\normalsize

$\epsilon_i$ is a random Gaussian value sampled from $\epsilon_i \sim \mathcal{N}(0,1)$. The full dataset $\mathbf{\Sigma} = \{\textcolor{black}{\Sigma_1,\dots,\Sigma_{20}}|\textcolor{black}{\Sigma_{21},\dots,\Sigma_{40}}|\textcolor{black}{\Sigma_{41},\dots,\Sigma_{60}}\}$
is composed of 60 randomly transformed Covariance matrices (20 per unique Covariance matrix). Each random transformation is generated by sampling a random matrix $\mathbf{A}_j \in \mathds{R}^{6\times6}$ for each $j$-th sample. Then, an orthogonal transformation matrix is extracted through QR decomposition on $\mathbf{A}_j = \mathbf{Q}_j\mathbf{R}_j$. The extracted orthogonal matrix is used to rotate each $i$-th unique Covariance matrix with respect to the $j$-th orthogonal rotation matrix $\mathbf{Q}_j$ as follows $\Sigma_j = \mathbf{Q}_j\mathbf\Sigma_i\mathbf{Q}_j^T \forall i \in [1,3], j \in [1,20]$.
The expected clustering result $\mathbf{K=3}$ groups of 20 samples each.\\
\noindent \textbf{6D Task-Ellipsoid Dataset:} This is a real dataset of 6-D task-ellipsoids. These were collected from human demonstrations of 3 tasks (circle drawing, cutting, screw-driving) with a sensorized tool (i.e. force/torque sensor at the end-effector) \cite{El-Khoury:RAS:2015}. These task-ellipsoids are used to represent the principal directions of the forces \textcolor{black}{$f = \{f_x,f_y,f_z \}$} and torques \textcolor{black}{$\tau = \{\tau_x,\tau_y,\tau_z \}$} exerted on an object to achieve a task. The Covariance matrix representing a task in this Task-Wrench-Space is generated as follows: 
\begin{equation}
\label{eq:task-ellipsoid}
\Sigma= \begin{bmatrix}
\textcolor{black}{\Sigma_{ff}} &  \textcolor{black}{\Sigma_{\tau f}} \\
 \textcolor{black}{\Sigma_{f\tau}} & \textcolor{black}{\Sigma_{\tau\tau}}  
 \end{bmatrix}; \qquad \Sigma_{**} \in \mathds{R}^{3\times 3}. \end{equation}

The dataset is composed of 105 samples of such Covariance matrices \eqref{eq:task-ellipsoid}, which belong to the following groups/classes of tasks: (a) \textit{circle drawing} - 63 samples, (b)\textit{cutting} - 21 and (c) \textit{screw-driving} - 21 samples each, respectively. The expected clustering result is $\mathbf{K=3}$ clusters of (1) 63, (2) 21 and (3) 21.\\

\begin{figure}[!h]
    \centering 
	\includegraphics[trim={0cm 0.5cm 0.5cm 0.5cm},clip,width=0.7\linewidth]{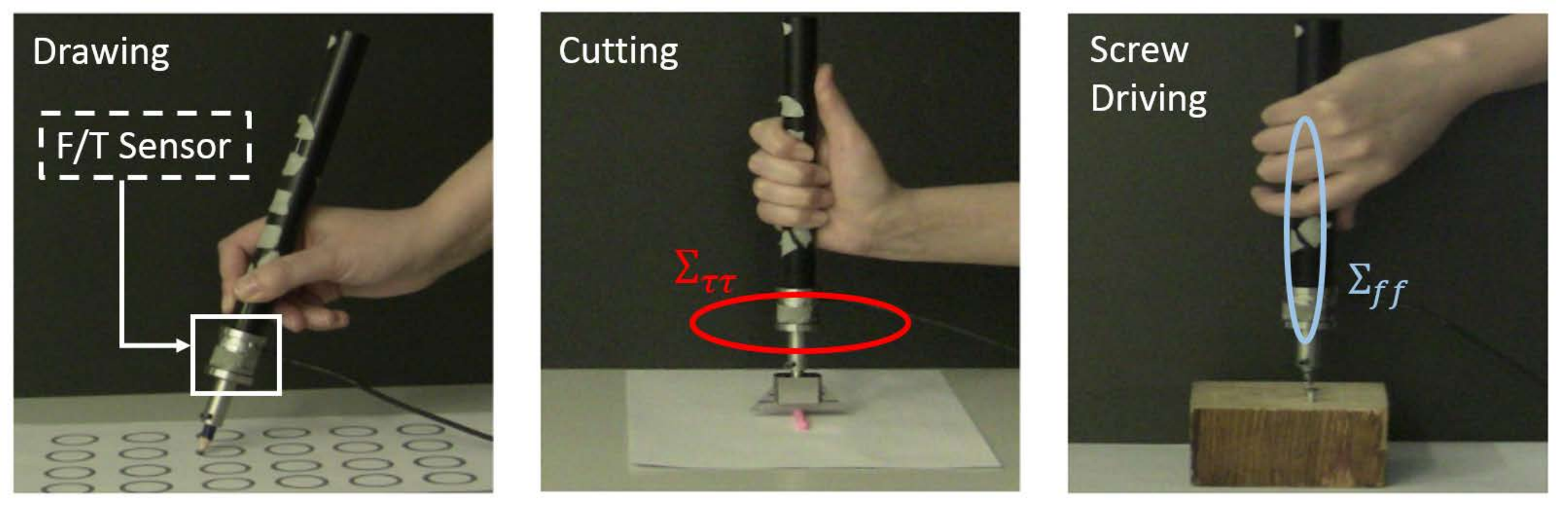}
	\captionof{figure}{\small Experimental Settings of task-ellipsoid data gathering. Each tool is equipped with a Force/Torque (F/T) sensor. After completing each task: (left) drawing (center) cutting (right) screw-driving, the signals of the (F/T) are compressed into an ellipsoid. }
\end{figure}

\noindent \textbf{3D Synthetic Diffusion Tensor Field (DTI) :}
Diffusion Tensors (DT) are widely used to represent the diffusivity of water in brain tissue from Magnetic Resonance Images (MRI). The resulting approximation is named DT-MRI and yields a matrix-valued image (also referred to as Tensor Field), where each element of the image is a DT. Diffusion Tensors (DT) are rank 2 tensors, equivalent to 3-D symmetric positive matrices (SPD). Researchers in the field of medical imaging, use such matrix-valued images to segment areas of the scanned brain into regions that exhibit similar behaviors, to detect injuries or chronic illnesses. Overviews of such segmentation algorithms are presented in \cite{Lenglet:2006:TMI} and \cite{Barmpoutis:TMI:2007}. There are two main approaches to do such segmentation: (i) Convert the DTI to a scalar/vector valued image based on DT properties such as fractional anisotropy or mean diffusivity \cite{Shepherd:2006:NeuroImage} and apply well-established edge-based or level-set based image segmentation algorithms. (ii) or segment the matrix-valued image directly by modified active contour models or statistical clustering frameworks where the topology of SPD matrices is considered \citep{Barmpoutis:TMI:2007}. 
\begin{figure}[!h]
\centering
	\includegraphics[trim={5cm 9cm 5cm 8cm},clip,width=0.26\linewidth]{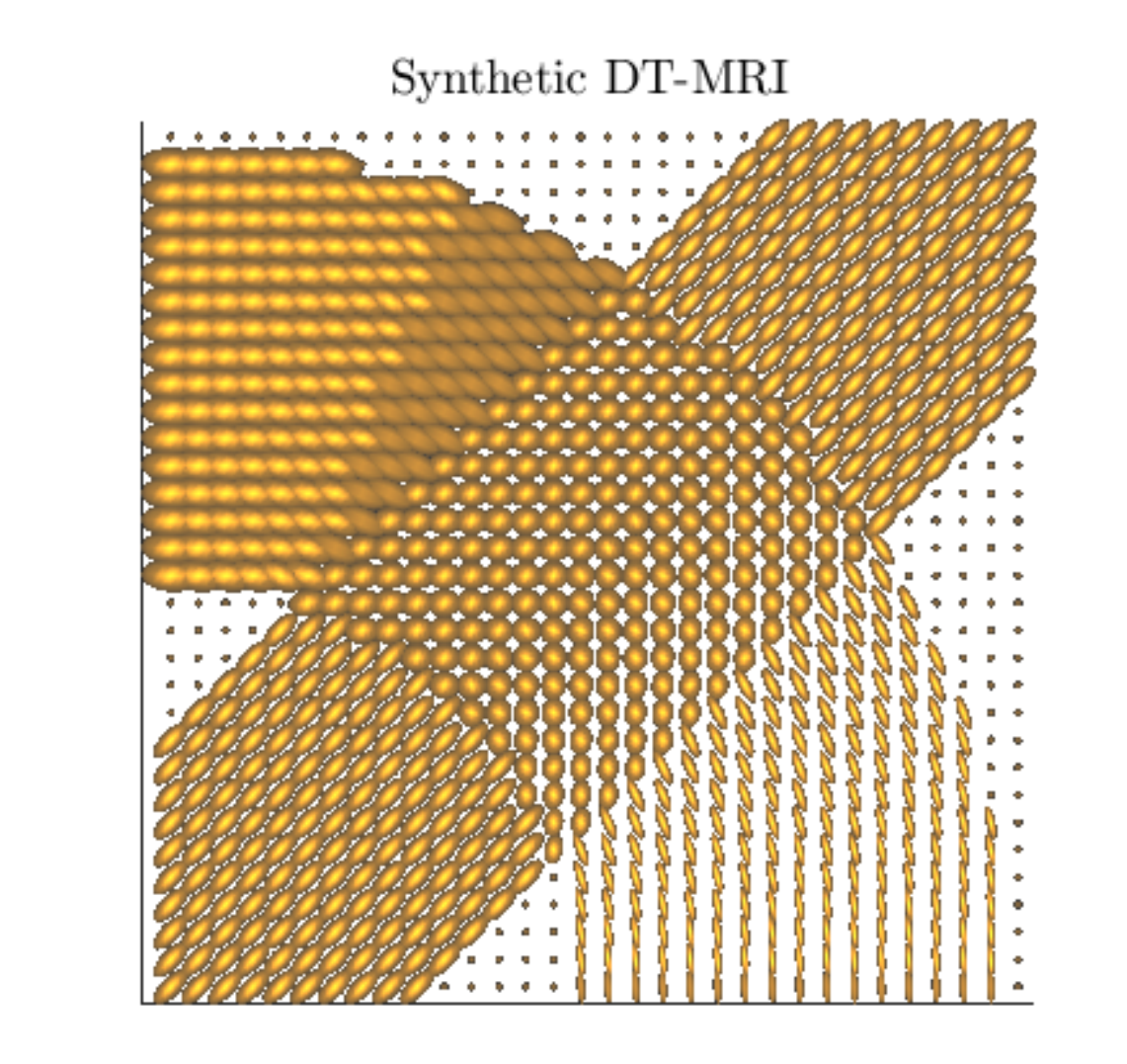}\includegraphics[trim={5cm 9cm 5cm 8cm},clip,width=0.26\linewidth]{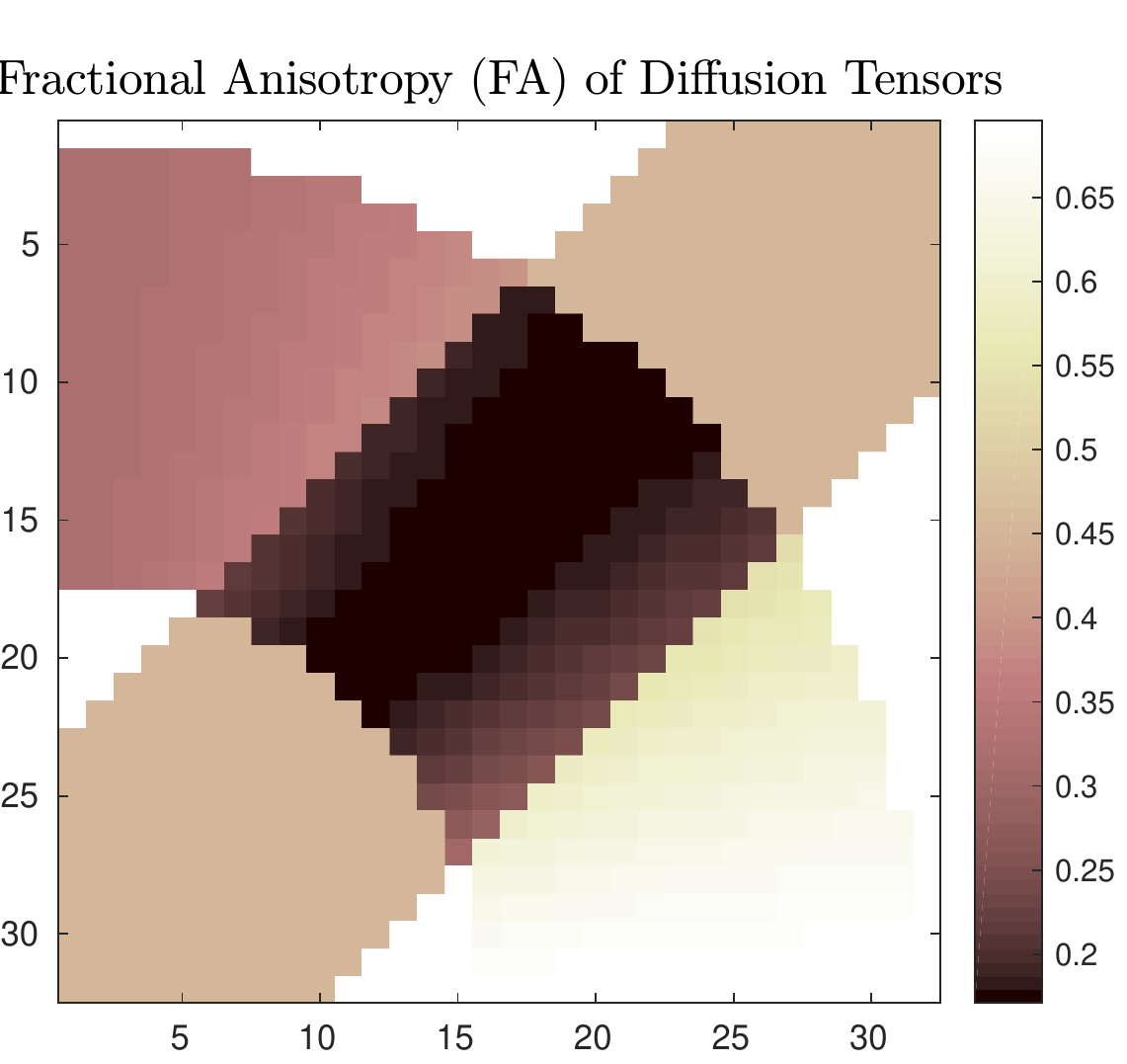}\includegraphics[trim={5cm 9cm 5cm 8cm},clip,width=0.26\linewidth]{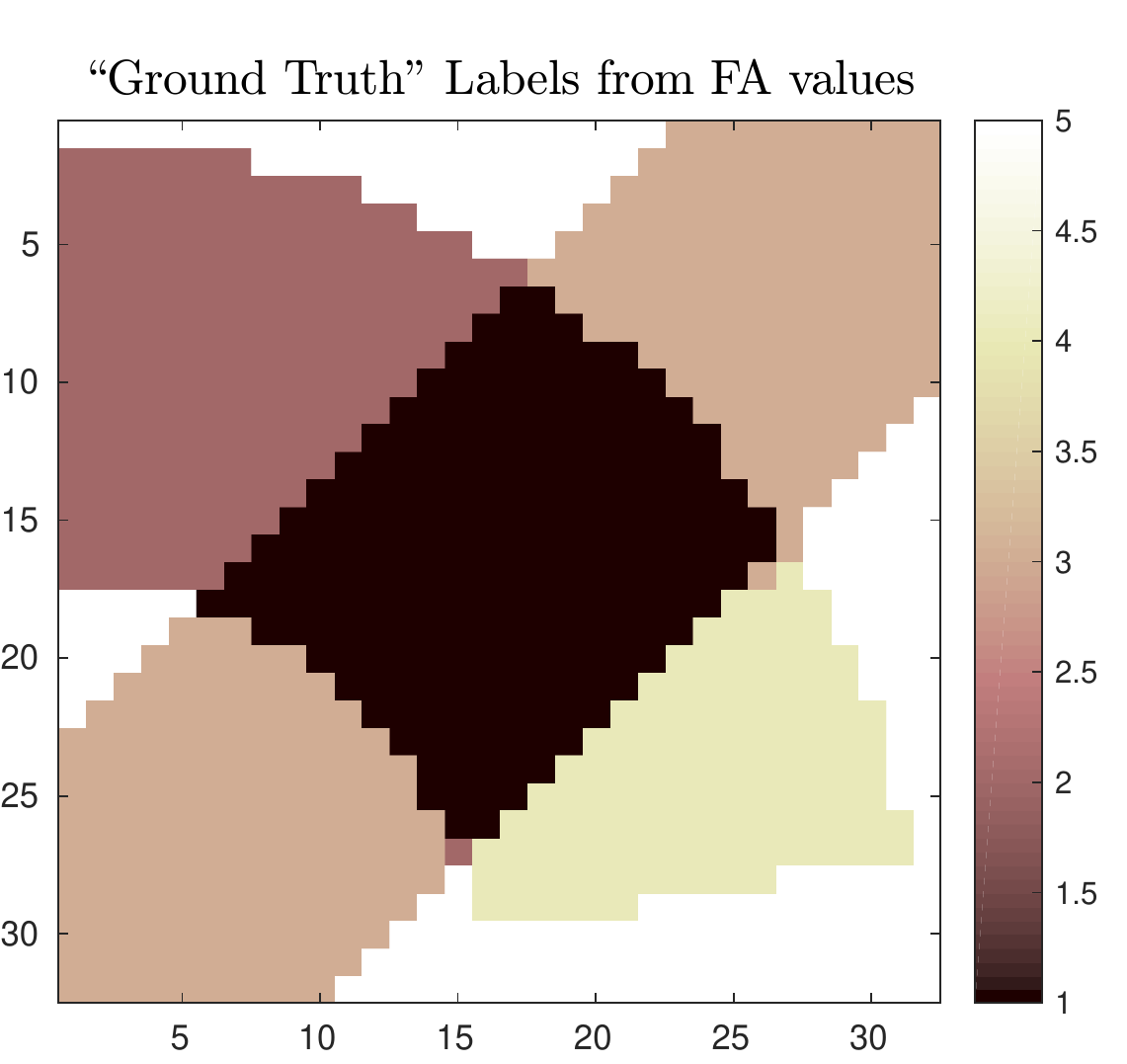}

	\captionof{figure}{\small  Datasets of DT from a Synthetic Diffusion Tensor Field generated from the fanDTasia Matlab Toolbox \citep{Barmpoutis:TMI:2007}. (center) \textit{FA} value of each diffusion tensor (right) Expected Cluster/Segmentation labels from binning the \textit{FA} values \label{fig:DTI-datasets}}
\end{figure}
Our proposed $\mathcal{SPCM}-CRP$ mixture model can be categorized in the latter approaches. As the DTI is a lattice of tensors, we can unroll this lattice and treat it as an unordered set of SPD matrices. Through this dataset, we can show the scalability of our proposed clustering algorithm and demonstrate that it's generic enough to be applied to different fields, with no modification. The Fractional Anisotropy (FA) value of the tensor field is used to evaluate the performance of our algorithm, as it is a popular quantity used to process and analyze DTIs \citep{Lenglet:2006:TMI}. We used the fanDTasia Matlab Toolbox \citep{Barmpoutis:TMI:2007} and the accompanying tutorial \citep{Barmpoutis:2010} to generate a $32\times32$ synthetic DTI, shown in Figure \ref{fig:DTI-datasets}. Since FA is a continuous value between $[0,1]$ we can only use it to visually compare segmentation results. For this reason, we apply an automatic binning procedure on the FA values of the entire image to create \textit{virtual} regions on the image. These regions will act as the ``ground truth" segmentation, albeit being an estimate, it provides a proper quantification of the performance of our approach. From Figure \ref{fig:DTI-datasets}, we can see that we are expected to segment the image into 5 regions. This is equivalent to clustering $1024$ SPD matrices into $\mathbf{K=5}$ clusters of (1) 218, (2) 185, (3) 301, (4) 119 and (5) 201 samples each, respectively.\\

\begin{figure}[!h]
\centering
  	\includegraphics[trim={5cm 9cm 5cm 8cm},clip,width=0.26\linewidth]{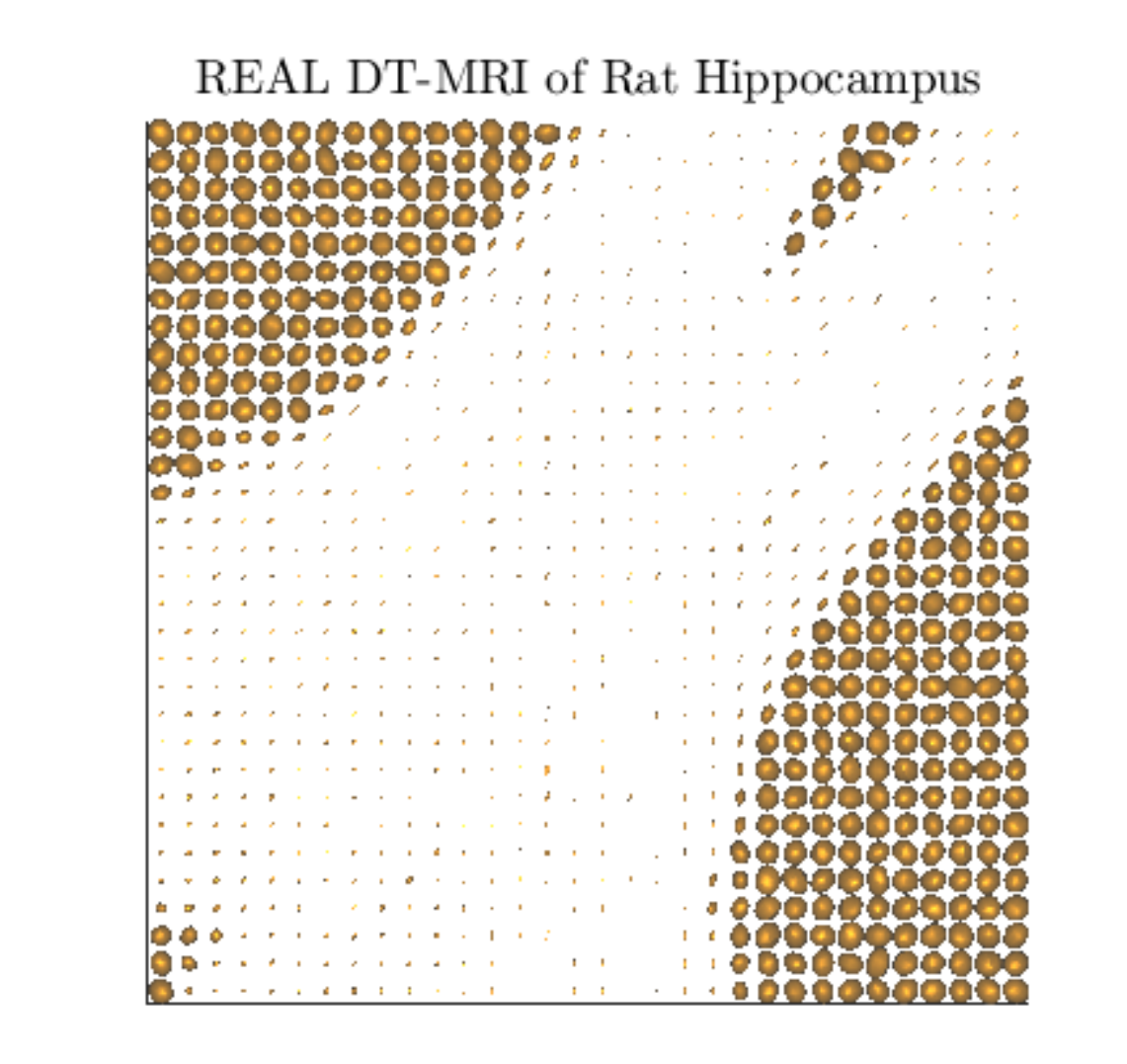}\includegraphics[trim={5cm 9cm 5cm 8cm},clip,width=0.26\linewidth]{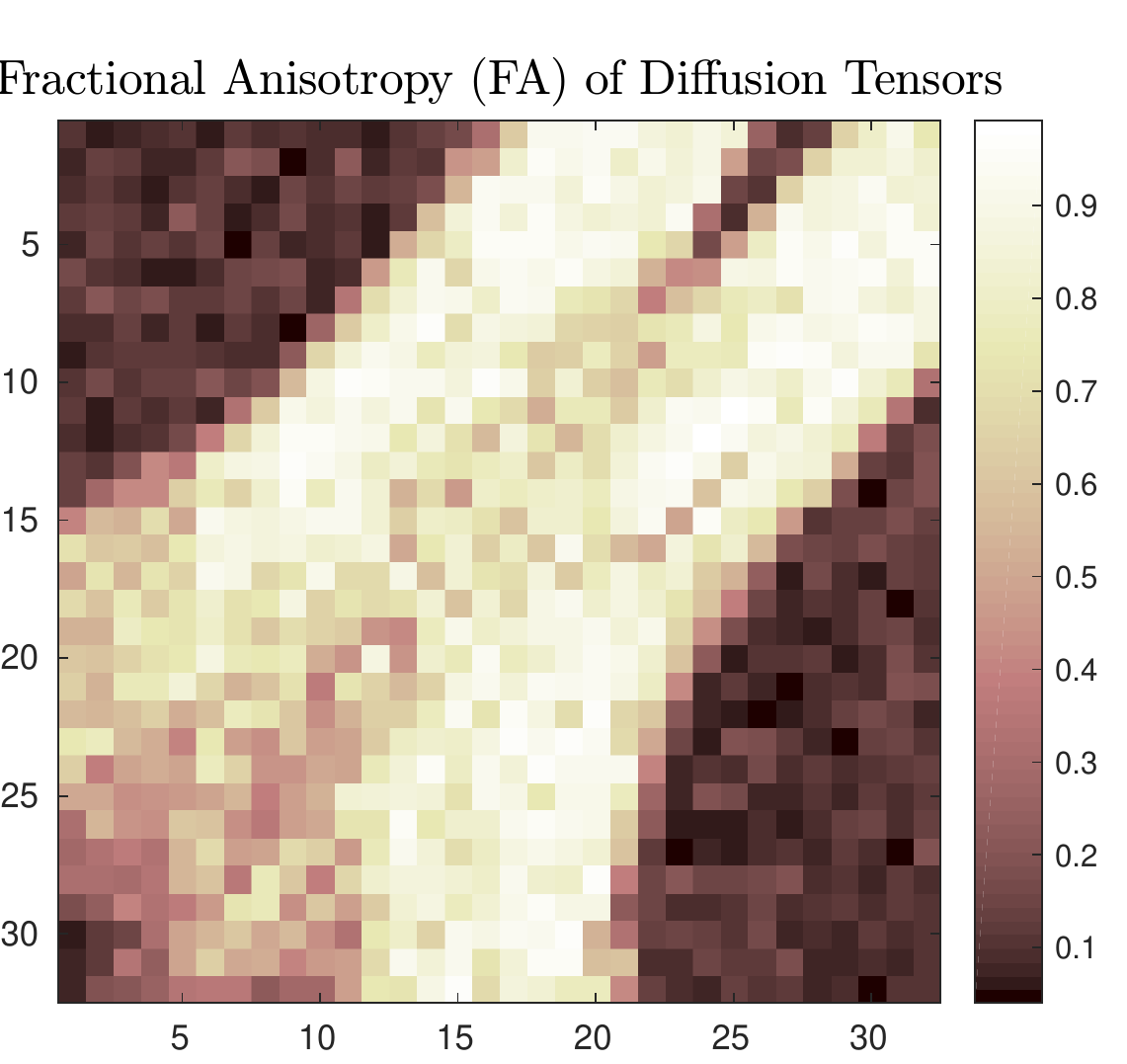}\includegraphics[trim={5cm 9cm 5cm 8cm},clip,width=0.26\linewidth]{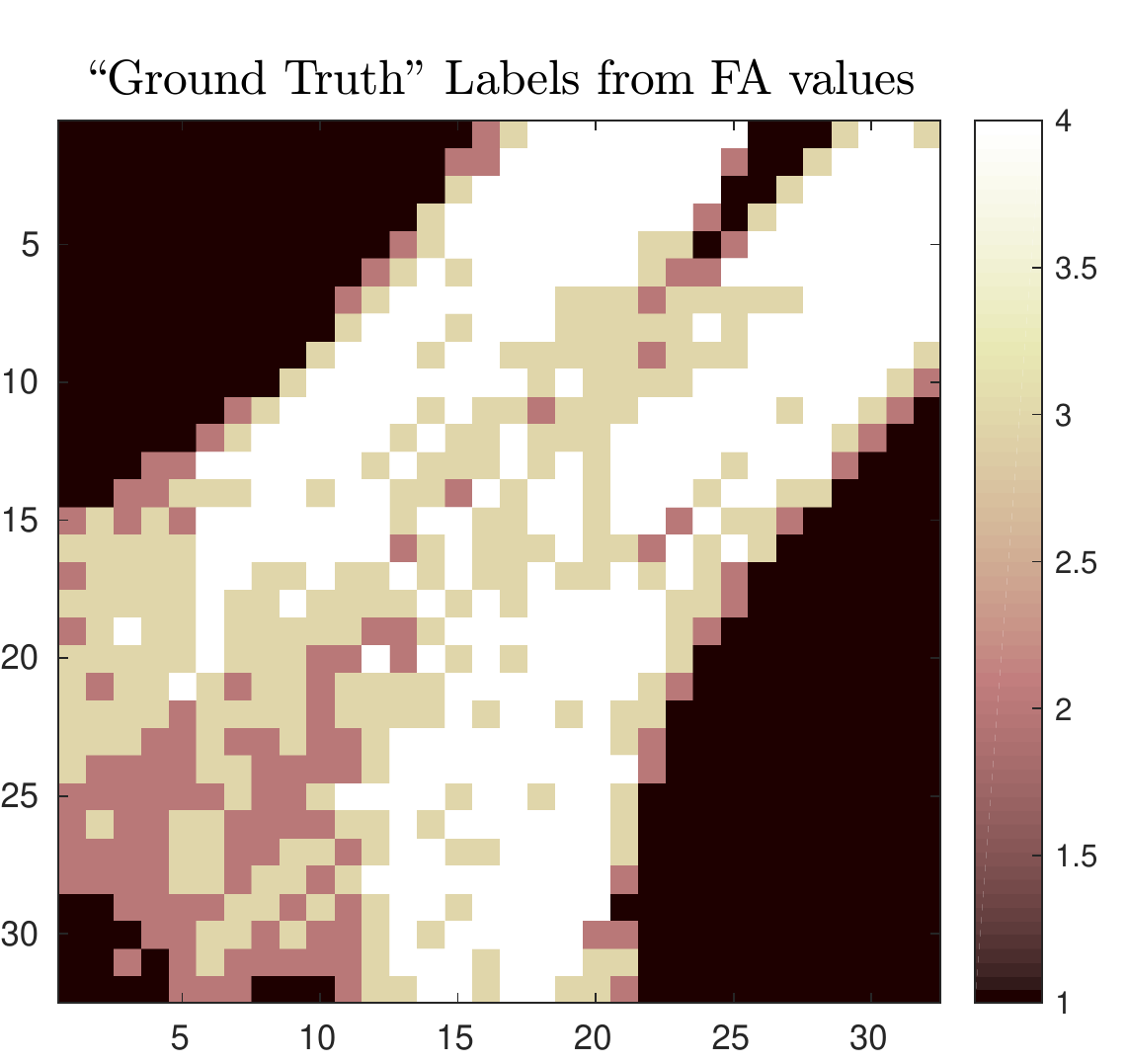}

	\captionof{figure}{\small  Dataset of DT from a Rat's isolated Hippocampus. Generated as Figure \ref{fig:DTI-datasets} \label{fig:DTI-datasets_rat}}
\end{figure}

\noindent \textbf{3D DTI from a Rat's Hippocampus:} This dataset has the same properties and was generated with the same software as \textbf{Dataset 3}, however it represents a slice of a real DTI from an isolated MRI of a rat's hippocampus \citep{Barmpoutis:TMI:2007}. It has the size of $32 \times 32$ and is shown in Figure \ref{fig:DTI-datasets_rat}. The procedure used in \textbf{Dataset 3} was used to produce the expected segmentation regions, and consequently the expected clusters K. We can see that we are expected to segment the image into 4 regions. This is equivalent to clustering $1024$ SPD matrices into $\mathbf{K=4}$ clusters of (1) 336, (2) 118, (3) 239, and (4) 331 samples each, respectively.\\

\noindent \textbf{400D Covariance Features from \href{https://www.mpi-inf.mpg.de/departments/computer-vision-and-multimodal-computing/research/object-recognition-and-scene-understanding/analyzing-appearance-and-contour-based-methods-for-object-categorization/}{ETH-80 Dataset}:} The ETH-80 dataset is an image-set based dataset for object recognition. It has 8 object categories with 10 object instances each. Each object instance is an image-set of 41, $20\times20$ pixel, images of that object from different view. In \cite{Vemulapalli:CVPR:2013}, they collectively represent each image-set with a single Covariance feature. These Covariance features are simply the data Covariance matrix $C = SS^T/(N-1)$ for an image-set $S = \{s_1, \dots, s_N\}$, where $s_i \in \mathds{R}^d$ is the feature vector representing the image, in this case the intensity values of the rolled out image, i.e. $d = 20\times20 = 400$. Each object instance, hence, is represented by a Covariance matrix $C \in \mathds{R}^{400\times400}$. \cite{Vemulapalli:CVPR:2013} use these Covariance features together with a kernel-learning SVM approach and achieve high classification accuracy. In this work, we do not intend to surpass such classification results, as we are not doing supervised learning. We rather test this dataset to demonstrate the scalability and limitations of our proposed method, by clustering a dataset of $80$ samples of $400$-d Covariance features into $\mathbf{K=8}$ with 10 samples.\\

\noindent \textbf{External Clustering Metrics}
Evaluating the performance of clustering algorithms without ground truth labels is still an open problem in machine learning. However, given that we have the true labels for all of our datasets, we use the following external clustering metrics to evaluate our proposed approach: 
\begin{enumerate}[leftmargin=*]
\item \underline{\textit{Purity}} is a simple metric that evaluates the quality of the clustering by measuring the number of clustered data-points that are assigned to the same class \citep{Manning:2008:IIR}.
\item \textit{Normalized Mutual Information \underline{(NMI)}}: is an information-theoretic metric, it measures the trade-off between the quality of the clustering  and the total number of clusters \citep{Manning:2008:IIR}.
\item The \underline{$\mathcal{F}$-measure} is a well-known classification metric which represents the harmonic mean between Precision and Recall  and can be applied to clustering problems.
\end{enumerate}
\normalsize
Refer to Appendix \ref{app:clustering_metrics} for computation details of each metric.

\subsection{Similarity Function Evaluation}
\begin{figure}[!t]
\begin{minipage}{\linewidth}
	\includegraphics[trim={1.5cm 0cm 1.5cm 0cm},clip,width=0.2\linewidth]{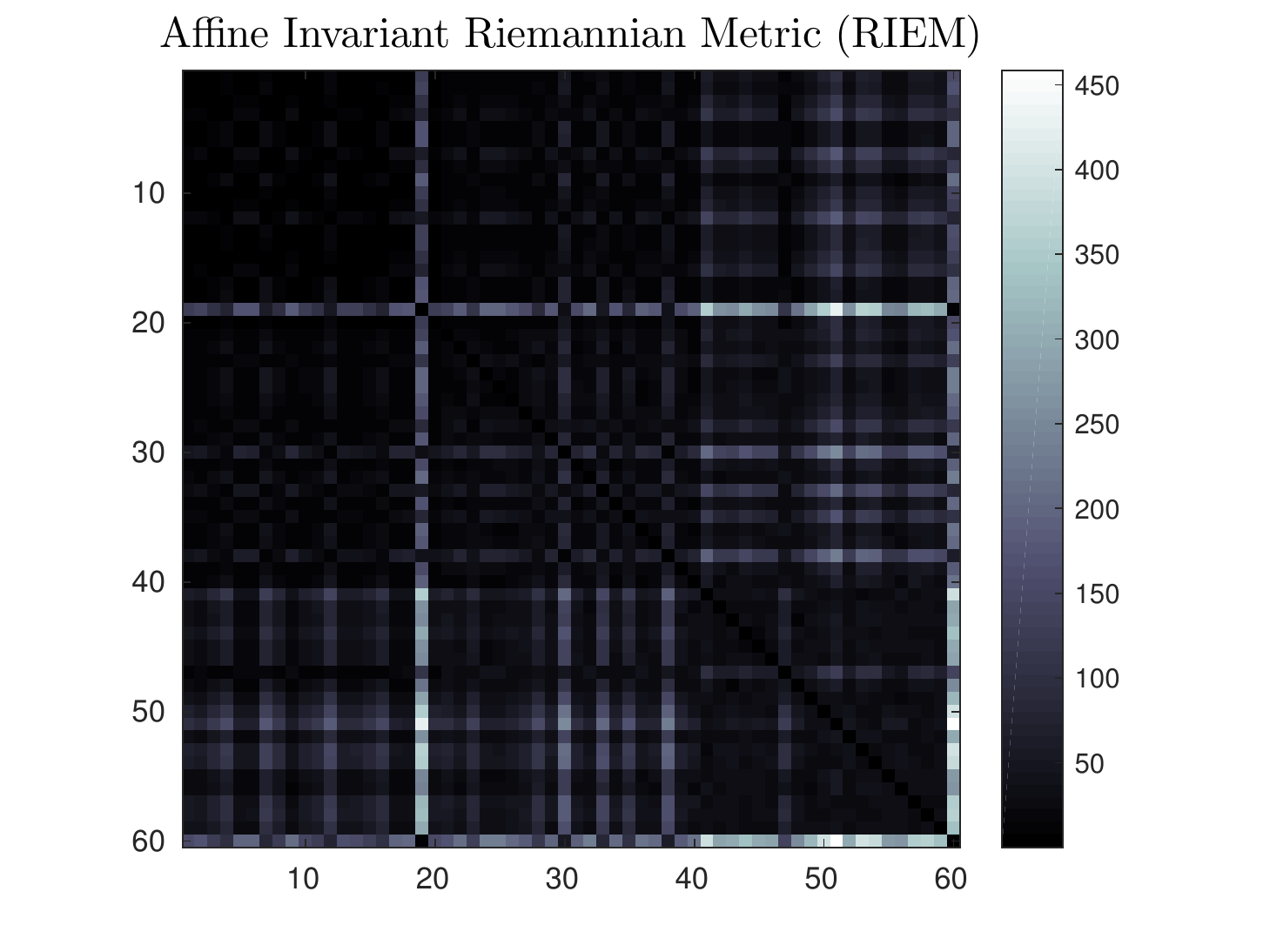}\includegraphics[trim={1.5cm 0cm 1.5cm 0cm},clip,width=0.2\linewidth]{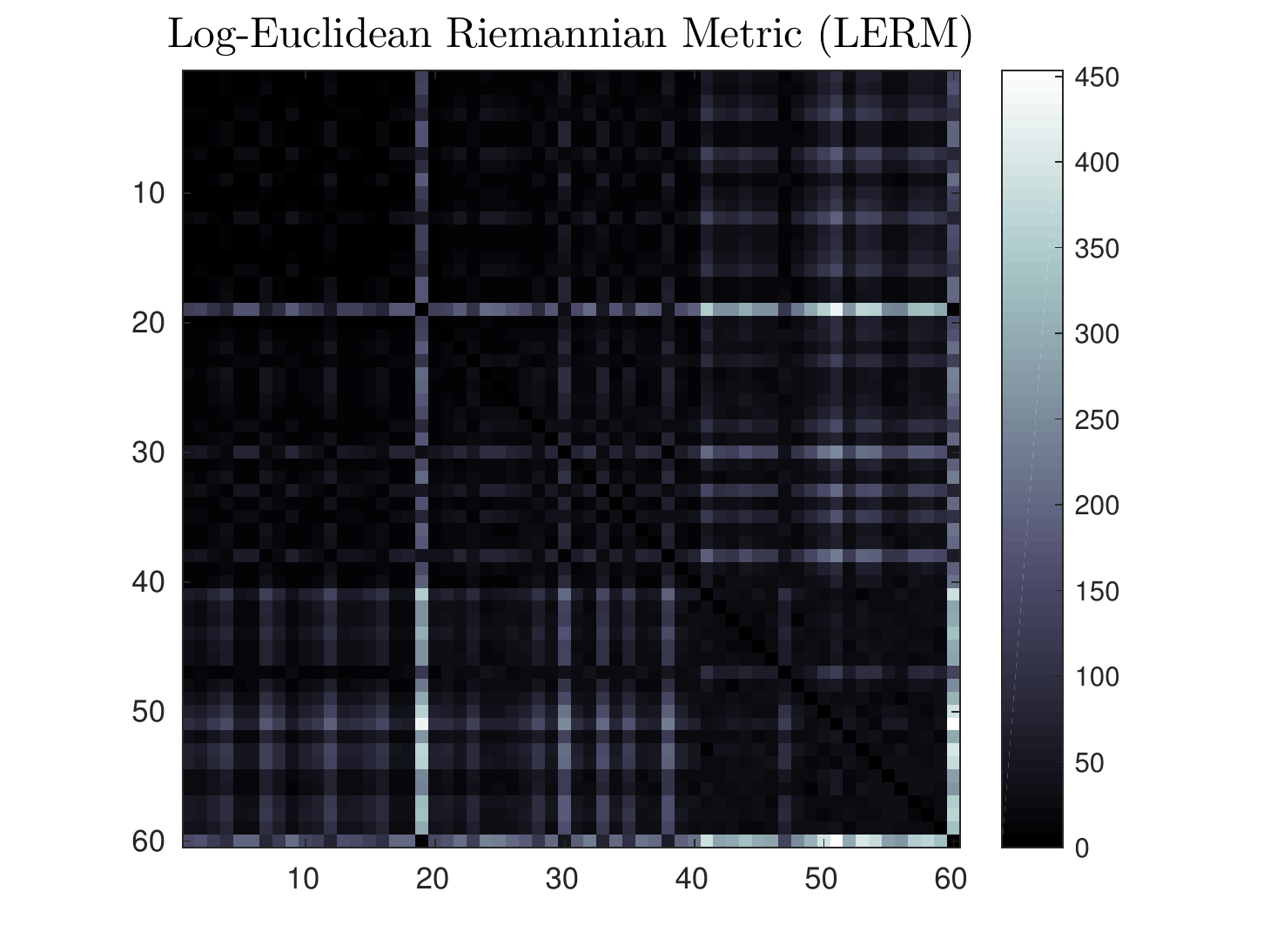}\includegraphics[trim={1.5cm 0cm 1.5cm 0cm},clip,width=0.2\linewidth]{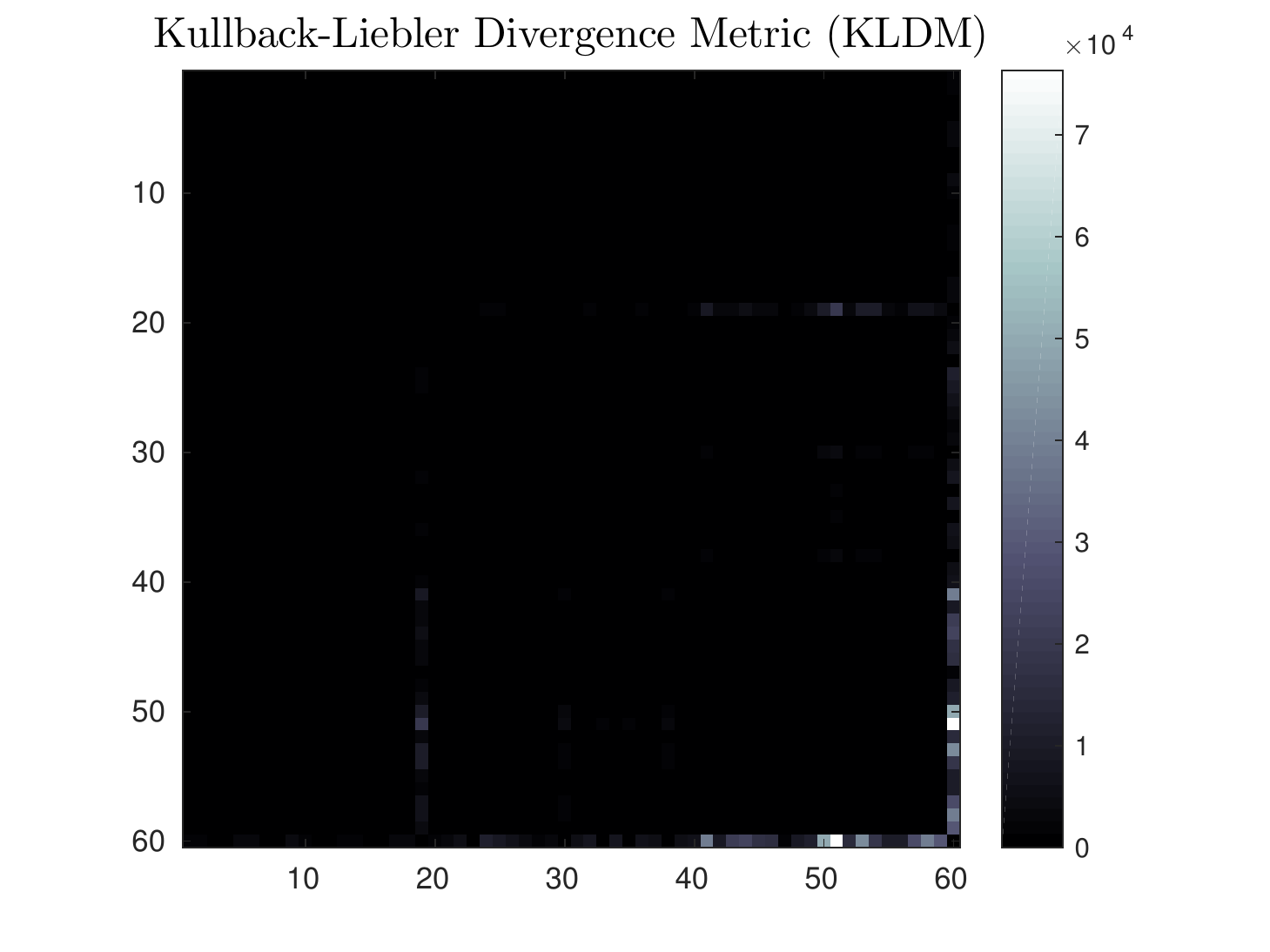}\includegraphics[trim={1.5cm 0cm 1.5cm 0cm},clip,width=0.2\linewidth]{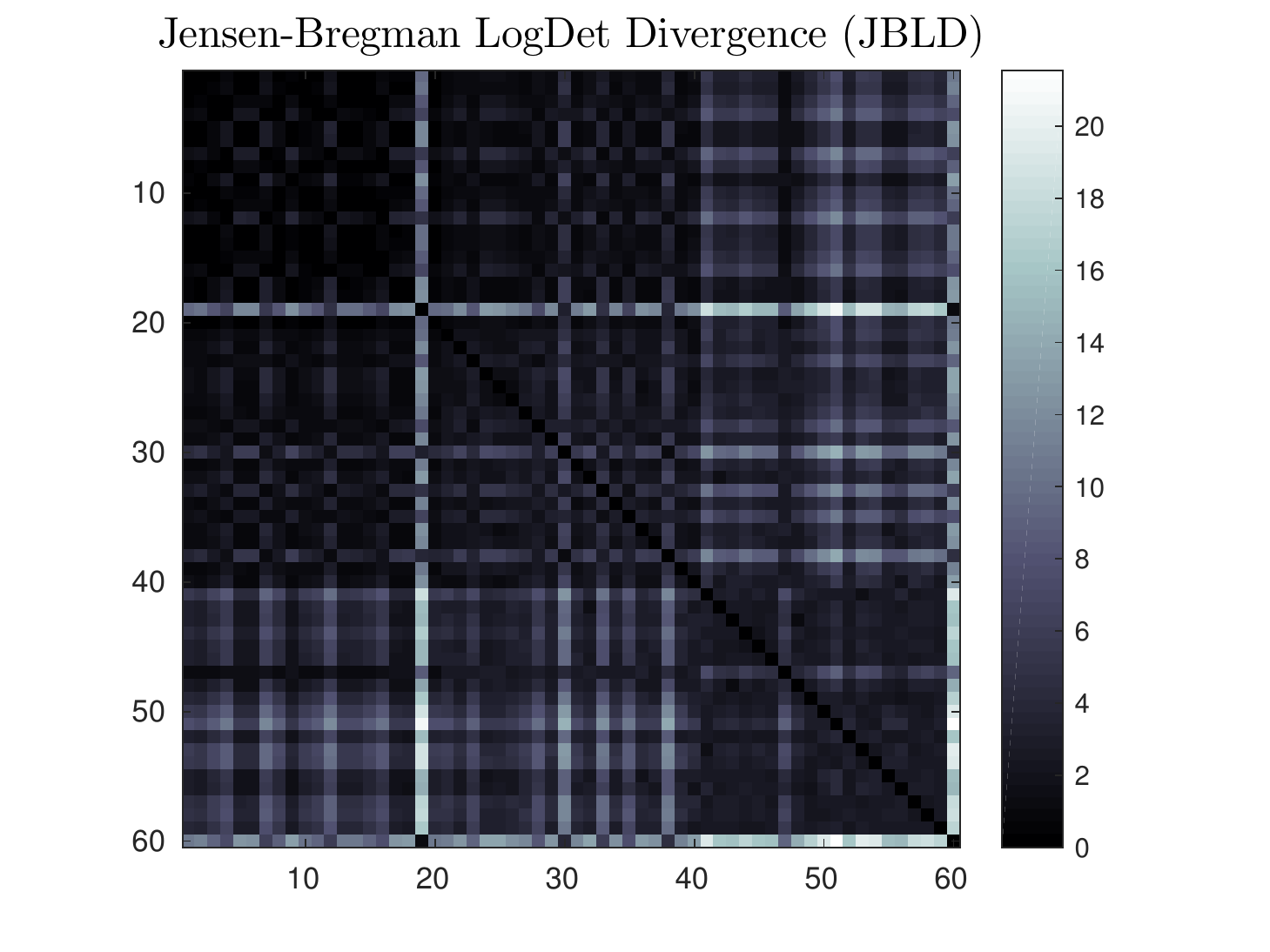}\includegraphics[trim={1.5cm 0cm 1.5cm 0cm},clip,width=0.2\linewidth]{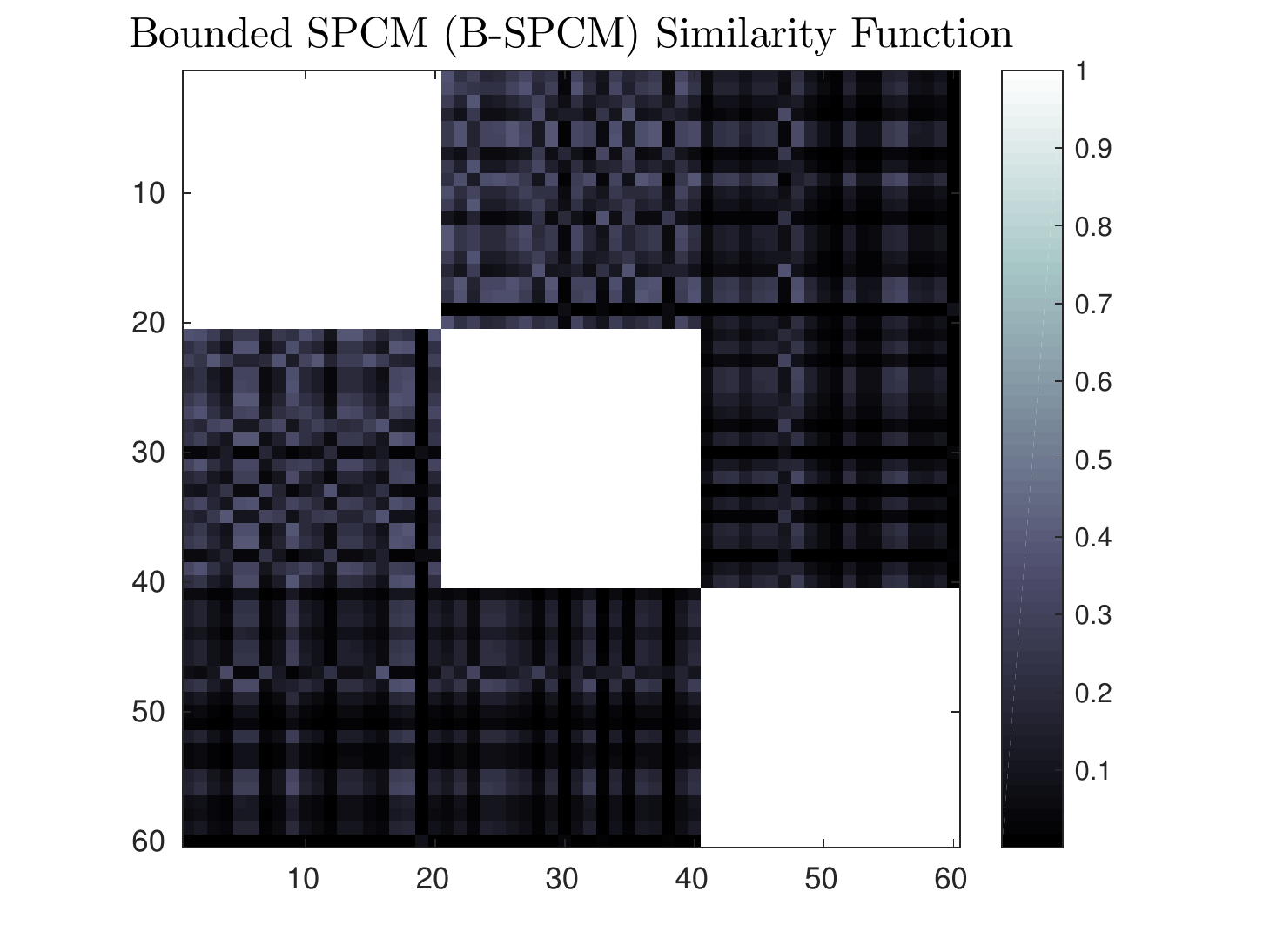}
\end{minipage}

\begin{minipage}{\linewidth}
   \includegraphics[trim={1.5cm 0cm 1.5cm 0cm},clip,width=0.2\linewidth]{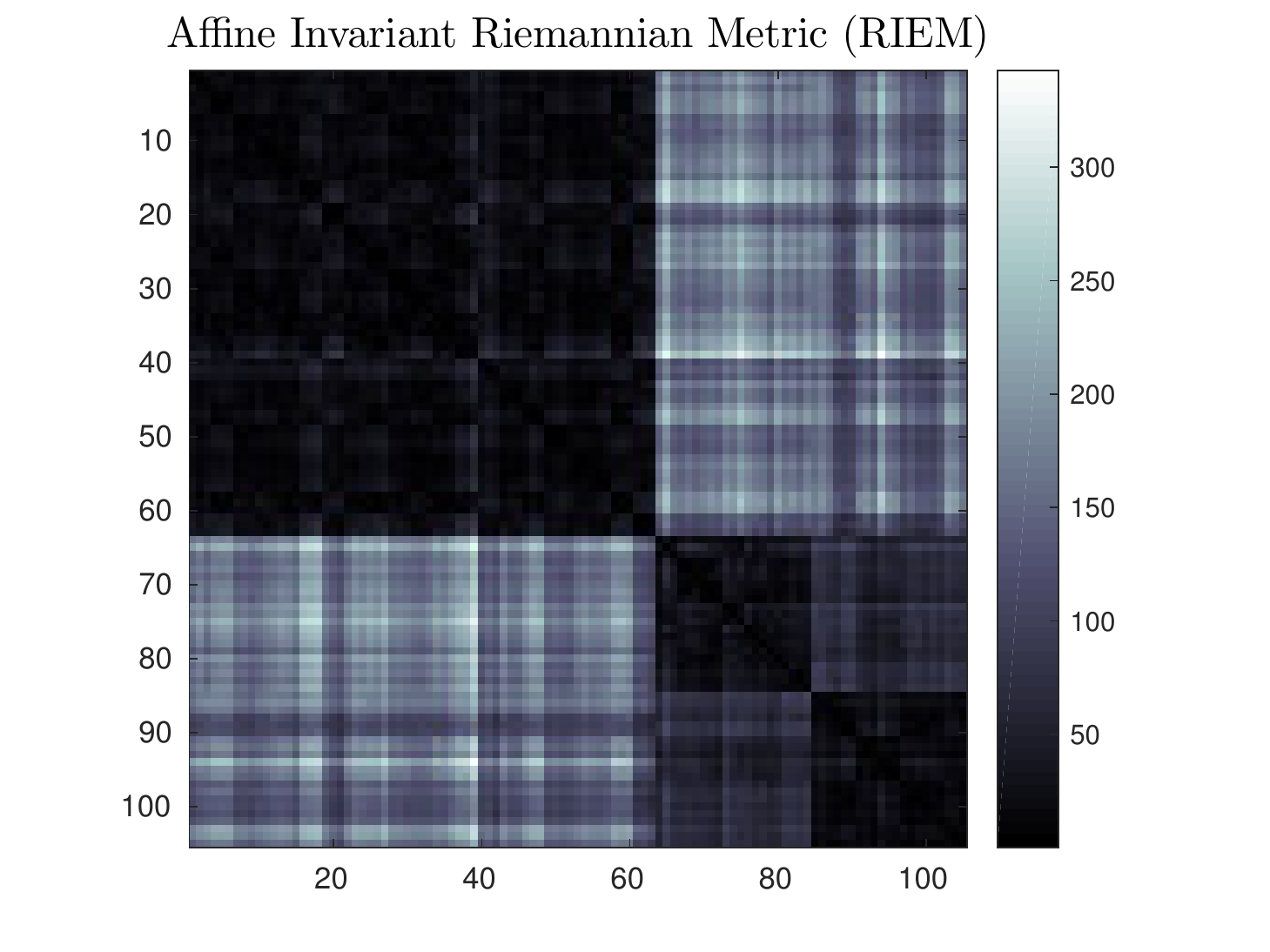}\includegraphics[trim={1.5cm 0cm 1.5cm 0cm},clip,width=0.2\linewidth]{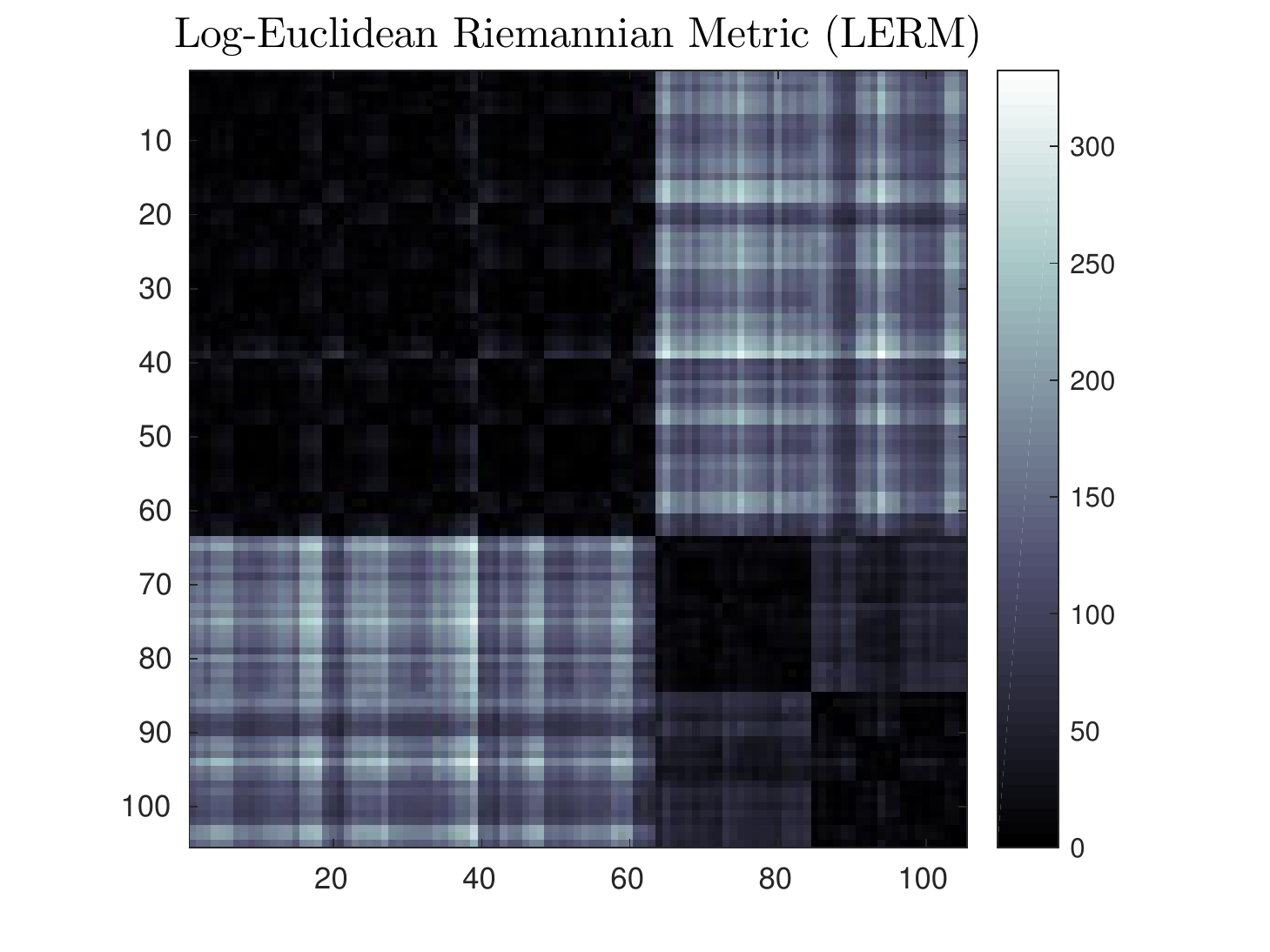}\includegraphics[trim={1.5cm 0cm 1.5cm 0cm},clip,width=0.2\linewidth]{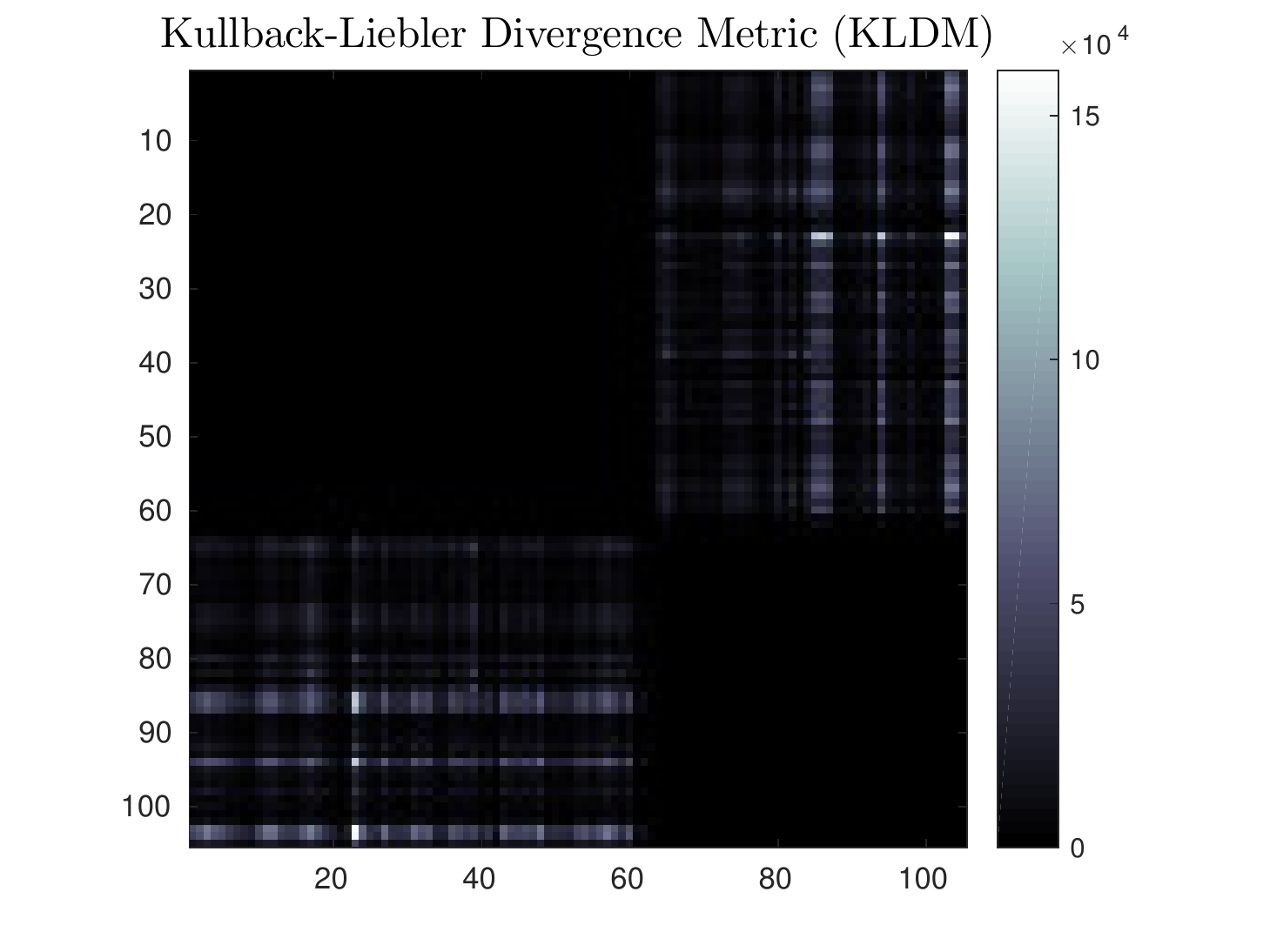}\includegraphics[trim={1.5cm 0cm 1.5cm 0cm},clip,width=0.2\linewidth]{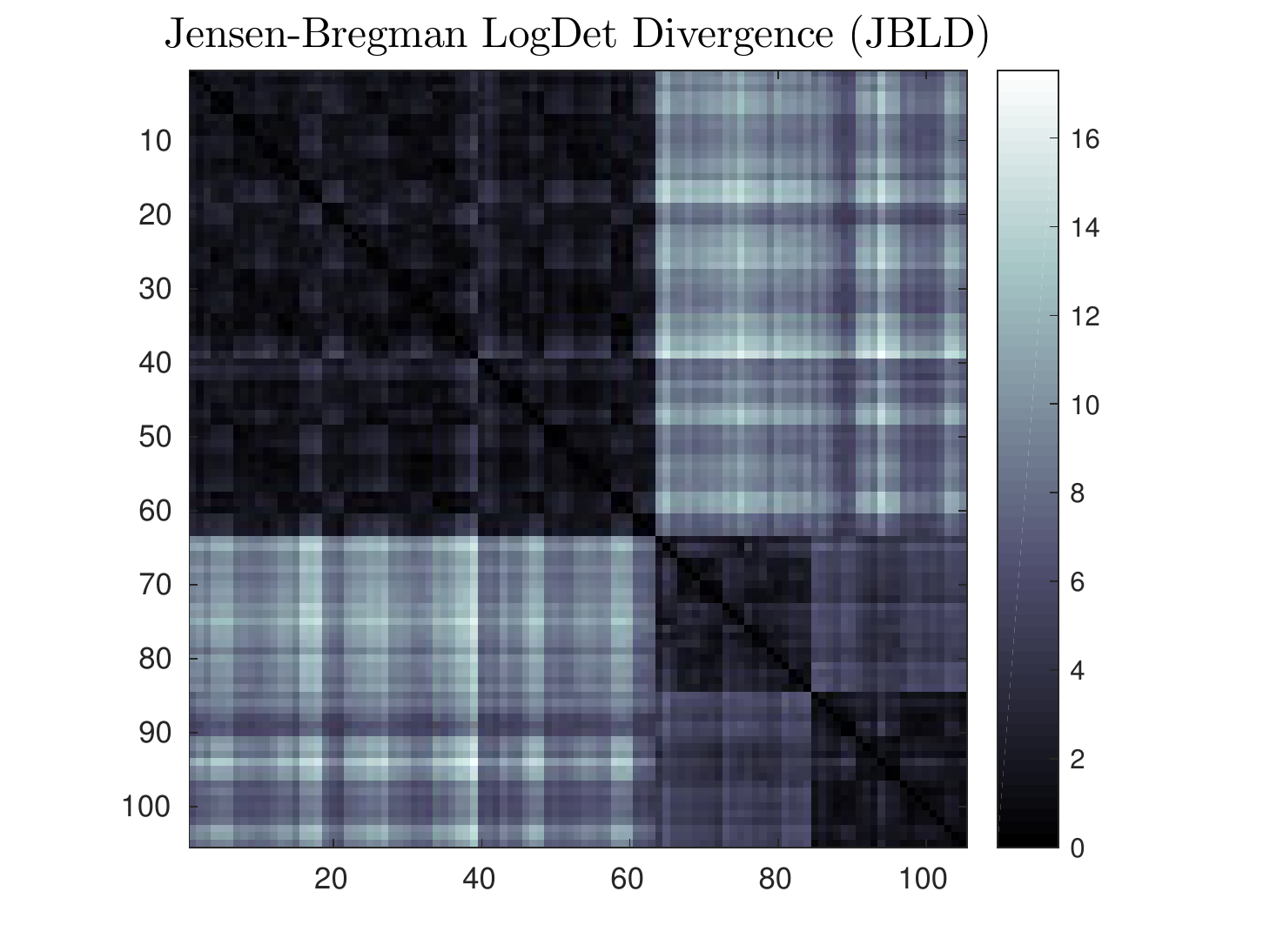}\includegraphics[trim={1.5cm 0cm 1.5cm 0cm},clip,width=0.2\linewidth]{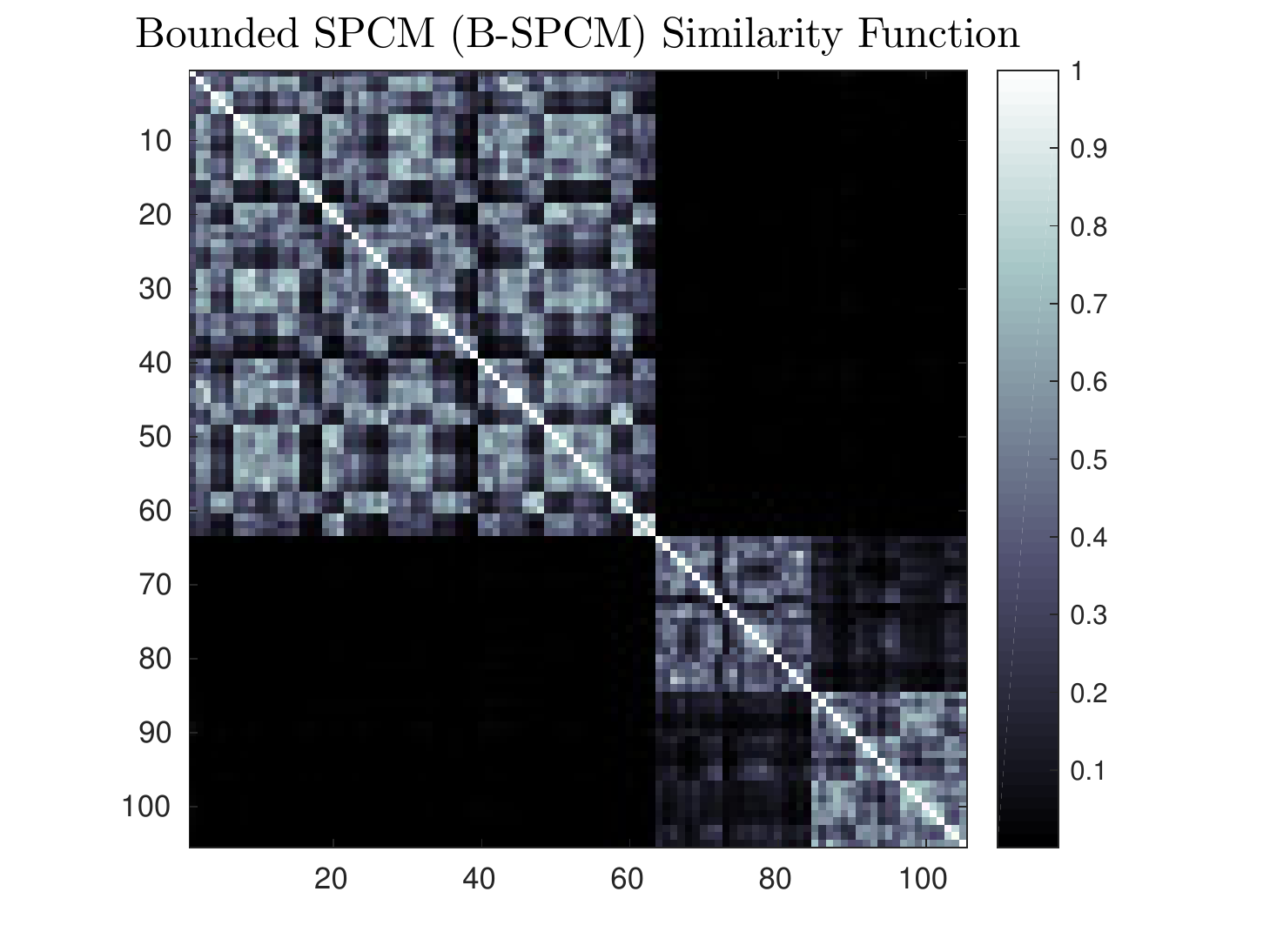}
\end{minipage}

\begin{minipage}{\linewidth}
	\includegraphics[trim={1.25cm 0cm 1.25cm 0cm},clip,width=0.2\linewidth]{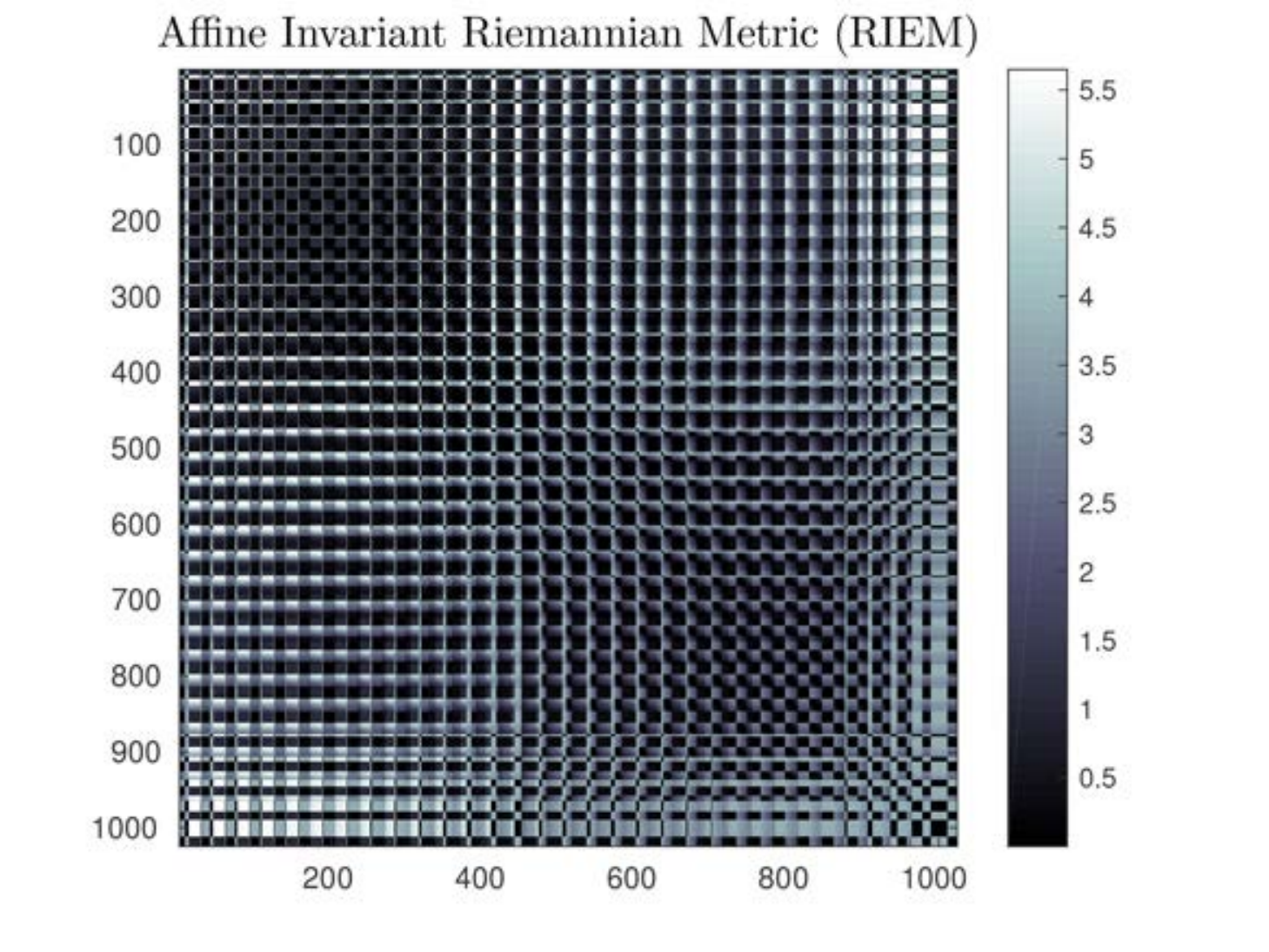}\includegraphics[trim={1.25cm 0cm 1.25cm 0cm},clip,width=0.205\linewidth]{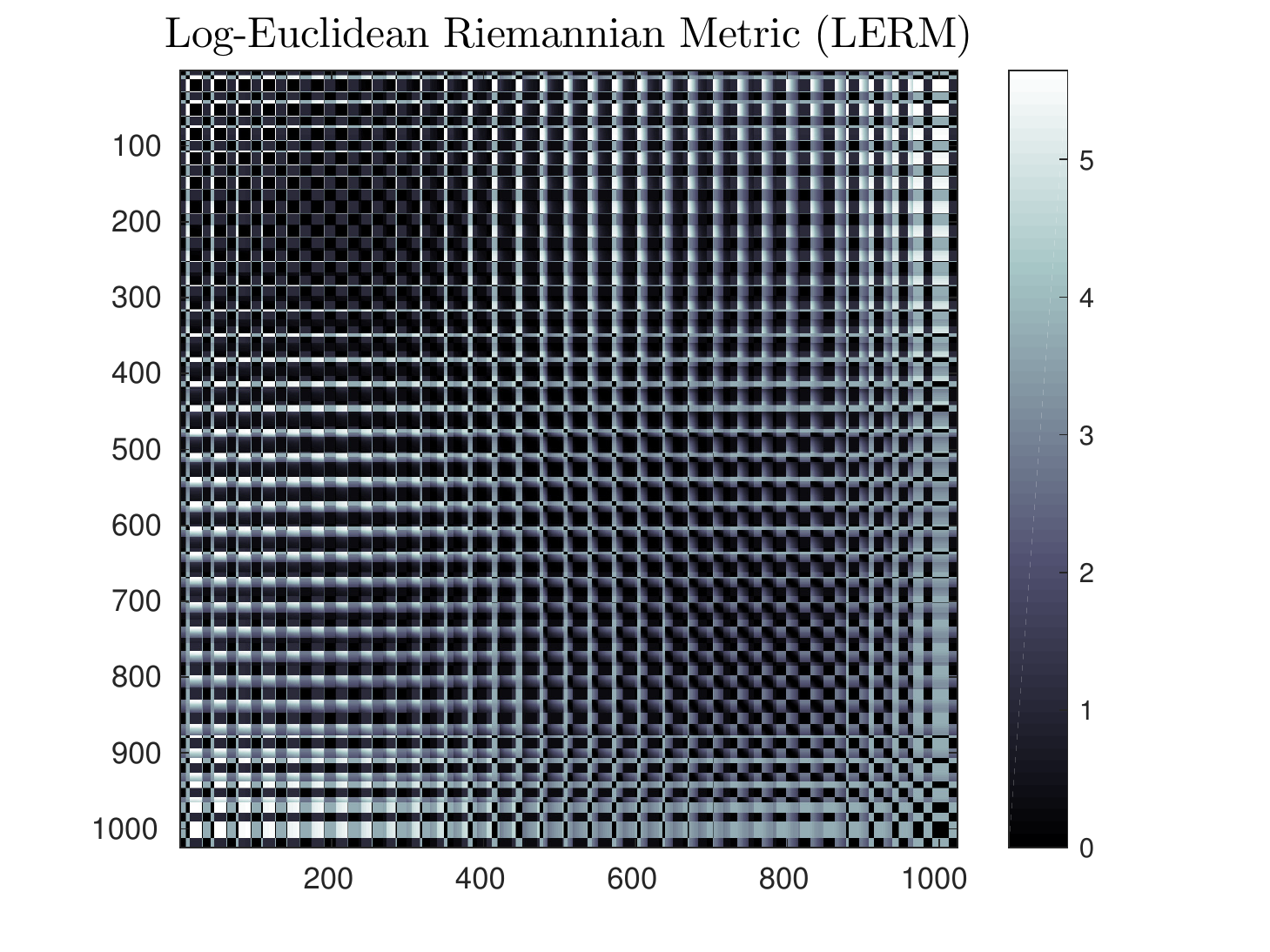}\includegraphics[trim={1.25cm 0cm 1.25cm 0cm},clip,width=0.205\linewidth]{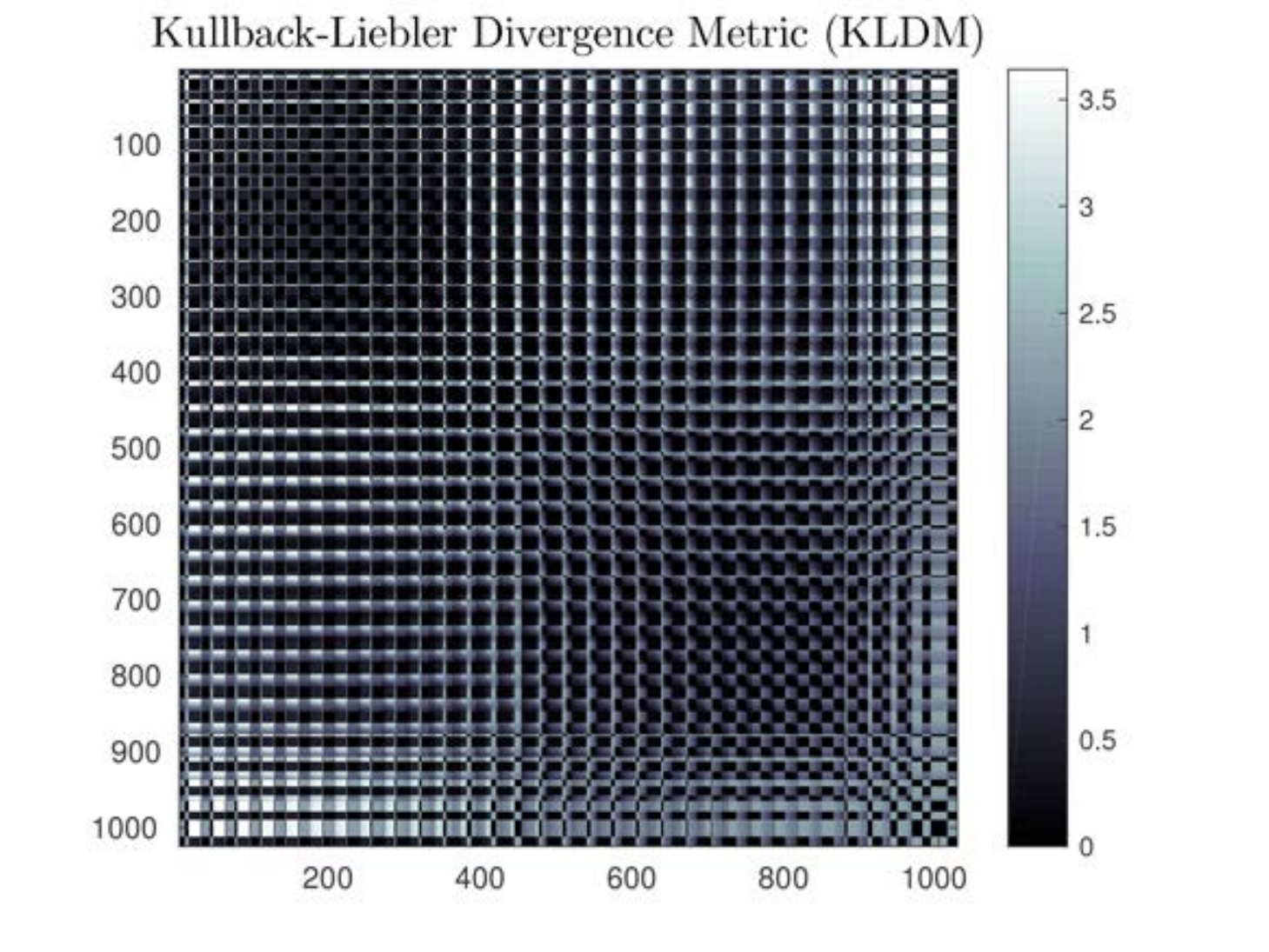}\includegraphics[trim={1.25cm 0cm 1.25cm 0cm},clip,width=0.205\linewidth]{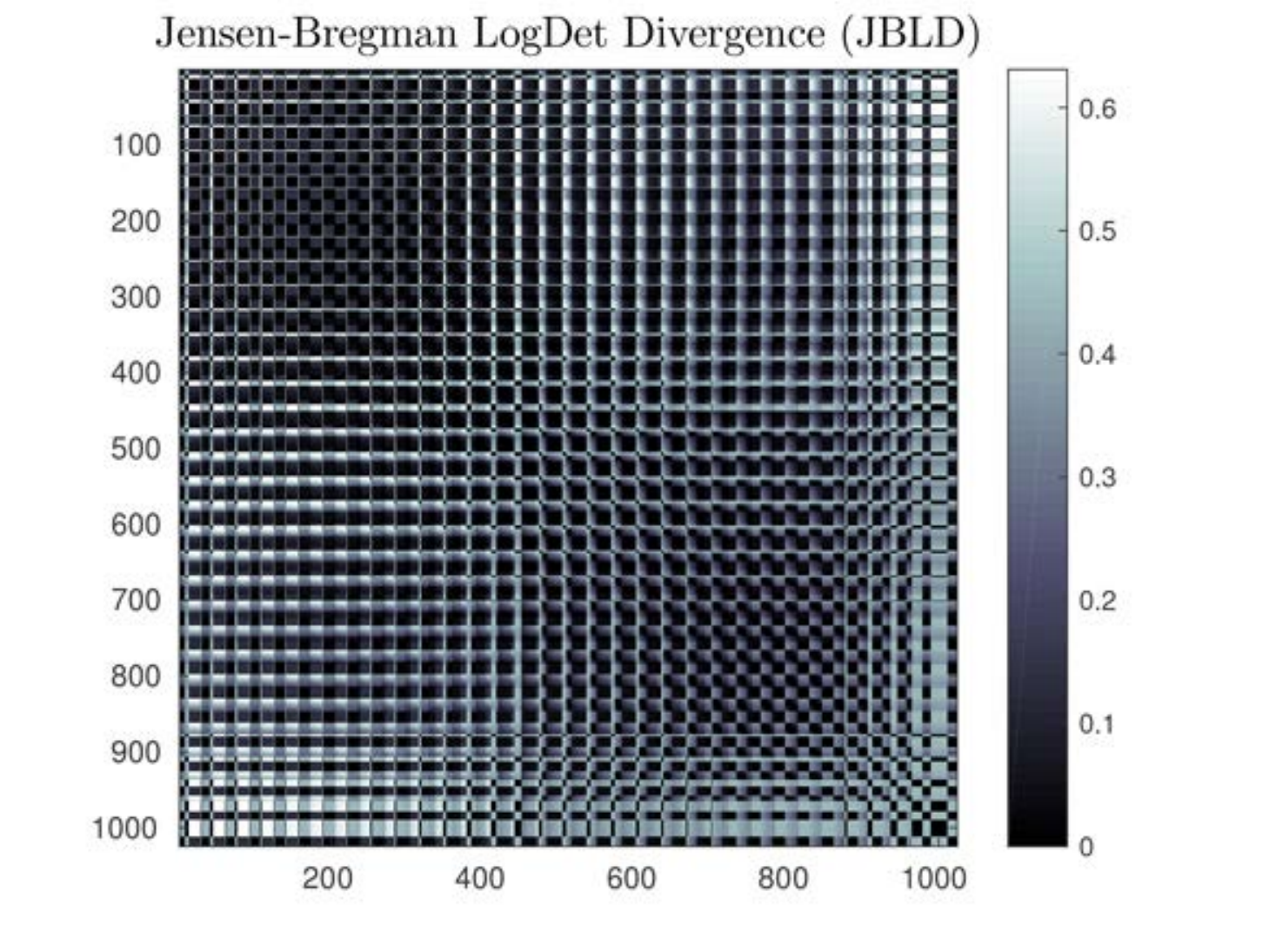}\includegraphics[trim={1.25cm 0cm 1.25cm 0cm},clip,width=0.2\linewidth]{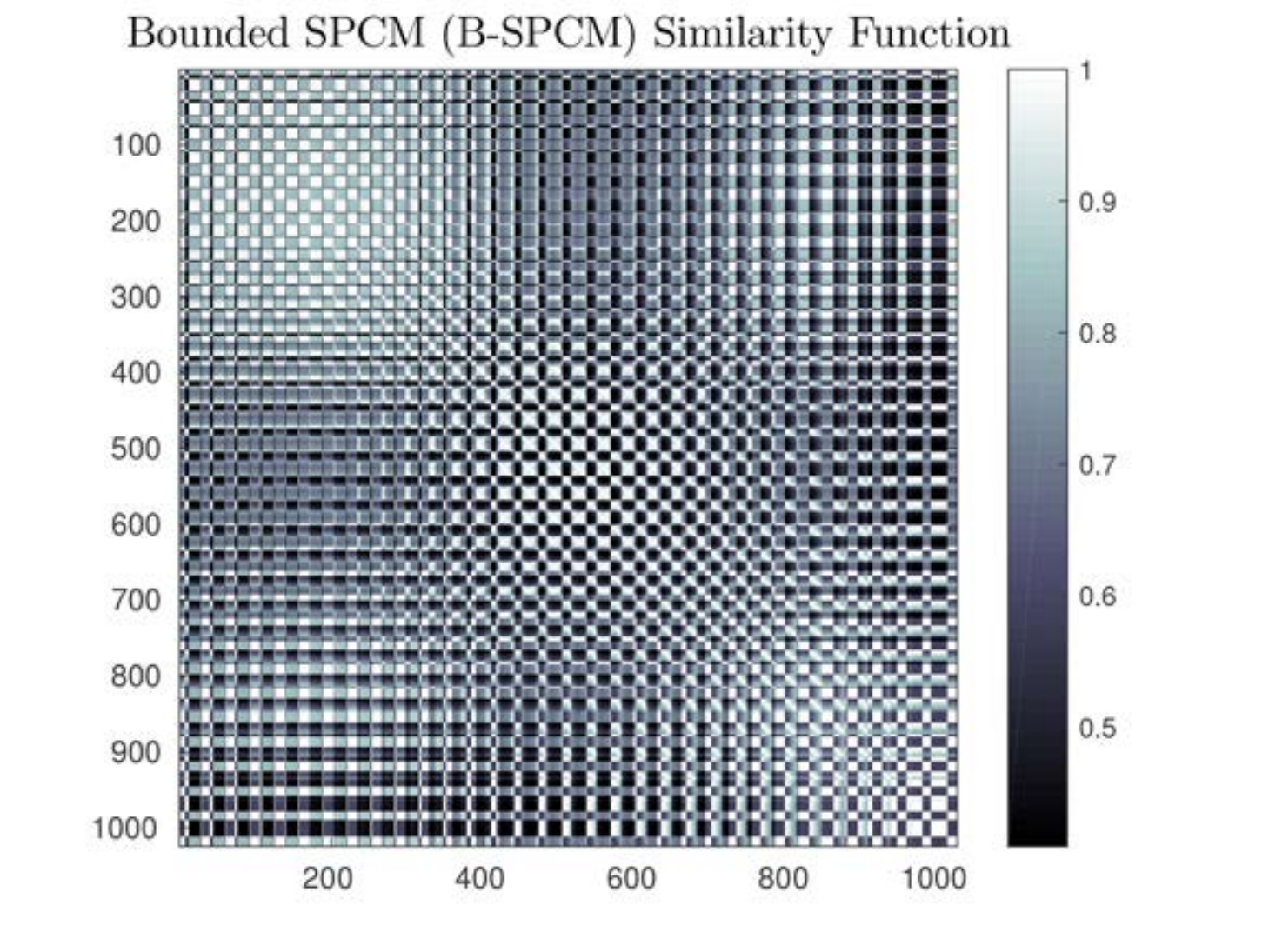}
		\vspace{-15pt}
\end{minipage}
		\captionof{figure}{\small Similarity matrices for \textbf{(top)}  \textbf{Dataset 1 (toy 6D)},  \textbf{(center)} \textbf{Dataset 2 (6D task-ellipsoids)} and  \textbf{(bottom)}   \textbf{Dataset 3 (Synthetic 3D DT-MRI)} \label{fig:dti_syn_matrices}}
\end{figure}

We devised an evaluation strategy to measure the \textit{effectiveness} of our similarity function compared to standard Covariance distances by applying the two main similarity/affinity-base clustering algorithms, namely Spectral Clustering (\textit{SC}) and Affinity Propagation (\textit{AP}). The performance of \textit{SC} and \textit{AP} are evaluated on the obtained pairwise Similarity matrices $\mathbf{S}\in \mathds{R}^{N \times N}$ for \textbf{Dataset 1}(Toy 6D), \textbf{Dataset 2} (6D task-ellipsoid), \textbf{Dataset 3} (3D Synthetic DTI) and \textbf{Dataset 4}(3D DT-MRI Rat's Hippocampus); for each similarity function: RIEM, LERM, KDLM, JBLD and B-SPCM (see Figure \ref{fig:dti_syn_matrices}). For completeness, we include metrics for the 3D toy dataset presented in Section \ref{sec:comparison-standard} and refer to it as \textbf{Dataset 0}. In Table \ref{tab:spcm-compare-algos}, we present performance metrics of each clustering method applied to every similarity matrix for the mentioned datasets. For \textit{SC}, we set $K$ to the true cluster number. On the other hand, for Affinity Propagation (\textit{AP}) we set the damping factor $\lambda$ to optimal values, found empirically, for each dataset. The B-SPCM hyper-parameter $\tau$ is set to 1 for all datasets. As can be seen in Table \ref{tab:spcm-compare-algos}, using the \textit{SC} algorithm, the proposed B-SPCM similarity function outperforms all other standard metrics for Datasets 0-1 and 3-4, which are indeed composed of Covariance matrices with large transformations. On Dataset 2, RIEM and JBLD with \textit{SC} were capable of recovering the true clusters, however B-SPCM still gives higher scores than RIEM and KLDM, these results are understandable as the transformations on this dataset are marginal, compared to the other datasets. On the other datasets, the \textit{AP} algorithm yields dramatically different results when compared to \textit{SC} (Table \ref{tab:spcm-compare-algos}). Except for \textbf{Dataset 0}, the best clustering performance (considering all Similarity metrics) is sub-optimal. 

We found that \textit{AP} is extremely unstable for the type of similarities that we target in this work. First of all, for each dataset we had to adjust the damping factor $\lambda$. With a $\lambda < 0.2$ the standard metrics would yield $K\approx M$, yet with a higher value $\lambda>0.5$ they would at $K=1$. Moreover, for each dataset we also had to find the optimal way of feeding the similarity matrix to the \textit{AP} algorithm, for each dataset we had to transform the similarities; either normalizing or negating, so that the AP algorithm would converge. Such parameter tuning and data transformation is cumbersome for some applications. This led to our motivation of pursuing a non-parametric clustering approach that is robust and converges without the need of heavy parameter tuning.

\begin{table*}[!t]
\begin{center}
\resizebox{\textwidth}{!}{\begin{tabular}{cc|ccccc|ccccc}
    \hline
    \hline
    \multicolumn{1}{c}{\multirow{4}{*}{Dataset}} & \multicolumn{1}{c}{\multirow{4}{*}{Metrics}} & \multicolumn{5}{c}{\multirow{2}{*}{Similarity functions + Affinity Propagation}} & \multicolumn{5}{c}{\multirow{2}{*}{Similarity functions + Spectral Clustering}}\\
	\\    
    & & \multicolumn{1}{c}{\multirow{1}{*}{RIEM}} & \multicolumn{1}{c}{\multirow{1}{*}{LERM}} & \multicolumn{1}{c}{\multirow{1}{*}{KLDM}} & \multicolumn{1}{c}{\multirow{1}{*}{JBLD}} & \multicolumn{1}{c}{\multirow{1}{*}{B-SPCM}} & \multicolumn{1}{c}{\multirow{1}{*}{RIEM}} & \multicolumn{1}{c}{\multirow{1}{*}{LERM}} & \multicolumn{1}{c}{\multirow{1}{*}{KLDM}} & \multicolumn{1}{c}{\multirow{1}{*}{JBLD}} & \multicolumn{1}{c}{\multirow{1}{*}{B-SPCM}}\\
    \cline{3-7}
        \hline
    \multicolumn{1}{c}{\multirow{4}{*}{Toy 3D}} & NMI &  0.65 & 0.65 & 0.65 & 0.65  & \textbf{1.00} & 0.29 (0.1) & 0.26 (0.09) & 0.26 (0.09) & 0.28 (0.09) &  \textbf{1.00} \\
    & Purity & 1.00 & 1.00 & 1.00 & 1.00  & \textbf{1.00} & 0.77 (0.11) &  0.66 (0.10) & 0.66 (0.10) & 0.68 (0.10) &  \textbf{1.00}\\
    & $\mathcal{F}$ &  0.57 & 0.57 & 0.57 & 0.57  & \textbf{1.00} & 0.70 (0.09) & 0.66 (0.08) & 0.66 (0.08) & 0.68 (0.09)  & \textbf{1.00}\\
    $(\lambda=0.15)$& $K(2)$ & 5 & 5 & 5 & 5  & \textbf{2} & 2 & 2 & 2 & 2 & \textbf{2}\\
        \hline
        
    \multicolumn{1}{c}{\multirow{4}{*}{Toy 6D}} & NMI &  0.19 & 0.19 & 0.19 & 0.19  & \textbf{0.40} &  0.12 (0.03) & 0.14 (0.03)  & 0.10 (0.03)  & 0.12 (0.08)  &  \textbf{1.00}  \\
	& Purity & 0.55 & 0.55 & 0.55 & 0.55  &\textbf{0.63} &  0.39 (0.02)  &  0.40 (0.02)  & 0.39 (0.03) & 0.40 (0.06) &  \textbf{1.00} \\
	& $\mathcal{F}$ & 0.58 & 0.58 & 0.58 & 0.58  & \textbf{0.68} & 0.49  & 0.49 & 0.49 (0.01) &  0.51 (0.05) & \textbf{1.00} \\
	$(\lambda=0.5)$ & $K(3)$  & 2 & 2  & 2 & 2  & 2 & 3 &  3 & 3 & 3 & \textbf{3} \\
	\hline
	
	\multicolumn{1}{c}{\multirow{2}{*}{6D Task}} & NMI &  0.73 & 0.73 & 0.73 & 0.73  & 0.64 &  \textbf{1.00}  & 0.03 &  0.07 (0.04) & \textbf{1.00}   &  \textit{0.77 (0.07)}\\   
	\multicolumn{1}{c}{\multirow{3}{*}{Ellipsoids}} & Purity & 0.79 & 0.79 & 0.79 & 0.79  & \textbf{1.00} & \textbf{1.00} & 0.60 &  0.60 & \textbf{1.00} & \textit{0.82 (0.04)}\\
	& $\mathcal{F}$ & 0.85 & 0.85 & 0.85 & 0.85  & 0.53 & \textbf{1.00} & 0.58 & 0.55 (0.02) & \textbf{1.00} & \textit{0.84 (0.05)}\\
    $(\lambda=0.5)$	& $K(3)$  & 2 & 2 & 2 & 2  & 11 & 3 &  3 & 3 & 3 & \textbf{3}\\
	\hline
	
	\multicolumn{1}{c}{\multirow{1}{*}{3D DTI}} & NMI &   0.56  & 0.56   & 0.56  & 0.56   & \textbf{0.62} &  0.12 (0.10)  & 0.25 (0.20)  & 0.14 (0.07)  & 0.08 (0.02) & \textbf{0.46 (0.19)}\\
	\multicolumn{1}{c}{\multirow{2}{*}{Synthetic}} & Purity &   0.55 & 0.55 & 0.55  & 0.55  & \textbf{0.55} & 0.32 (0.02) &  0.40 (0.10)  & 0.35 (0.04)  & 0.32 (0.01) &  \textbf{0.53 (0.11)}\\
    & $\mathcal{F}$ & 0.58  & 0.58  & 0.58 & 0.58  & \textbf{0.59} & 0.38 (0.03)  & 0.46 (0.09) & 0.40 (0.04)  & 0.37 & \textbf{0.53 (0.10)} \\
	$(\lambda = 0.5)$ & $K (5)$  & 2 & 2 & 2 & 2  & 2 & 5 & 5 & 5 & 5 & \textbf{5} \\
	\hline
	
	\multicolumn{1}{c}{\multirow{1}{*}{3D DTI Rat's}} & NMI &  0.17 & 0.15 & 0.00 & 0.20  & \textbf{0.23} &  \textbf{0.08 (0.14)} & 0.05 (0.01) & 0.03  & 0.03 (0.01)  & 0.02\\
	\multicolumn{1}{c}{\multirow{2}{*}{Hippocampus}} & Purity &  0.41 & 0.42 & 0.33 & 0.44  & \textbf{0.49}  & \textbf{0.36 (0.09)} & 0.35 (0.01) & 0.33 & 0.34  & 0.33  \\
	& $\mathcal{F}$ & 0.43 & 0.42 & 0.43 & 0.46 &  \textbf{0.52} & \textbf{0.45 (0.07)} & 0.43 & 0.43 & 0.43  & 0.43 \\
	$(\lambda=0.5)$ & $K (4)$  & 2 & 3 & 1 & 2  & 2 & \textbf{4} & 4 & 4 & 4 & 4 \\
	\hline	
	\hline
\end{tabular}}
\end{center}
\vspace{-10pt}
\caption{Performance Comparison of all Covariance Matrix similarity functions with \textit{Affinity Propagation (AP)} and \textit{Spectral Clustering (SC)} algorithm with K set to the real value \textit{(presenting mean (std) of performance metrics over 10 runs).} \label{tab:spcm-compare-algos}} 
\end{table*}

\subsection{Similarity-based Non-parametric Clustering Evaluation}
Our proposed clustering approach relies on sampling from a posterior distribution \eqref{eq:cond_spcm_crp} with the following hyper-parameters set by the user: $\tau, \alpha, \lambda = \{\mu_0, \kappa_0, \Lambda_0, \nu_0\}$. The first evaluation of our approach will focus on the convergence of the implemented collapsed Gibbs sampler and its robustness to multiple initializations. The second evaluation will focus on its robustness to hyper-parameters. Typically, the hyper-parameters of the $\mathcal{NIW}$, i.e. $\lambda = \{\mu_0, \kappa_0, \Lambda_0, \nu_0\}$, are set to data-driven values. For example, $\mu_0 = \frac{1}{M}\sum_{i=1}^M \mathbf{y}_i$ can be set to the mean value of all data-points or simple zero-mean (if the data is centered), while $\Lambda_0 = \frac{1}{M}\mathbf{Y}\mathbf{Y}^T$ can be set to the sample Covariance of all data-points and $\kappa_0 = 1,\nu_0 = M$. Hence, in reality we only have 2 hyper-parameters to tune: (1) the tolerance value, $\tau$, for the B-SPCM metric  and (2) the concentration parameter, $\alpha$, of the SPCM-$\mathcal{CRP}$ prior. Finally, we will compare the results of the SPCM-$\mathcal{CRP}$ mixture model with two mixture variants: (1) GMM with model selection and (2) $\mathcal{CRP}$ mixture model.
\subsubsection{Collapsed Gibbs Sampler Convergence}
\begin{figure*}[!t]
\begin{minipage}{0.33\linewidth}
\centering
\hspace{5pt}
   \includegraphics[trim={1.5cm 0cm 1.5cm 0cm},clip,width=\linewidth]{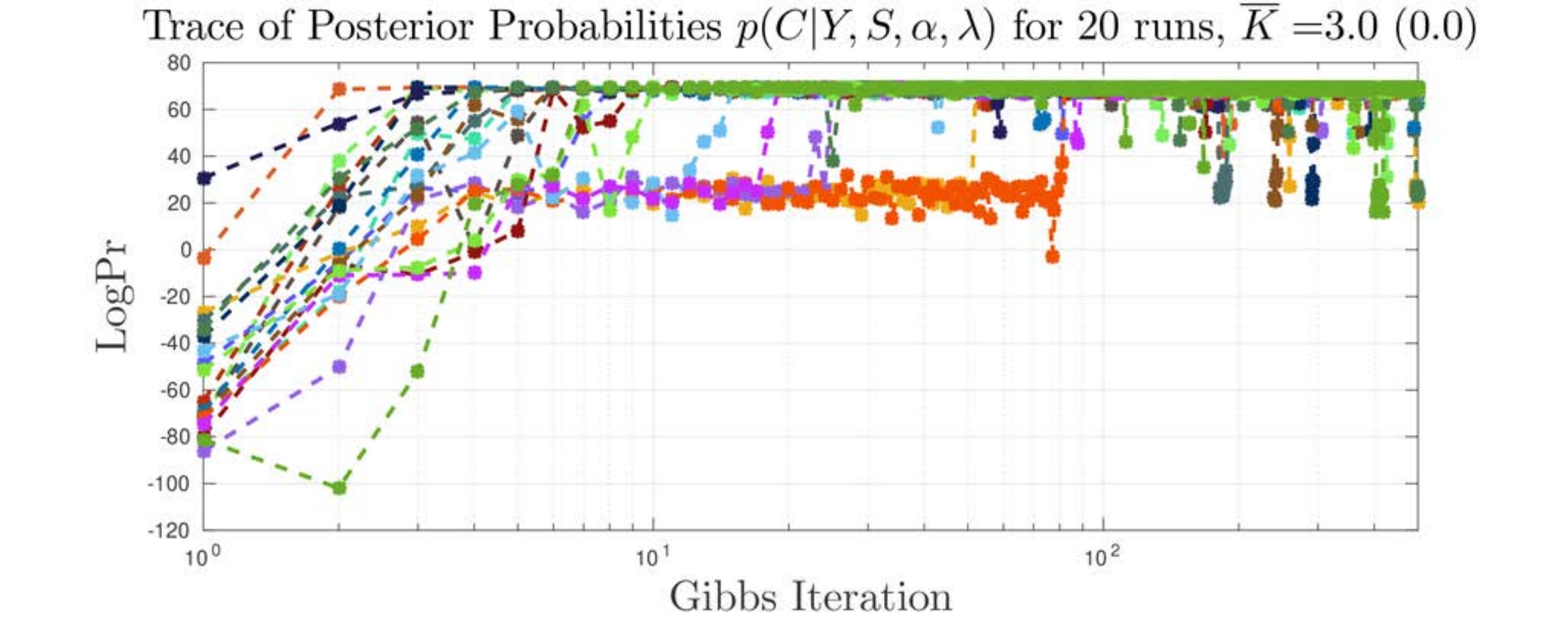}
  \includegraphics[trim={1.5cm 0cm 1.5cm 0cm},clip,width=\linewidth]{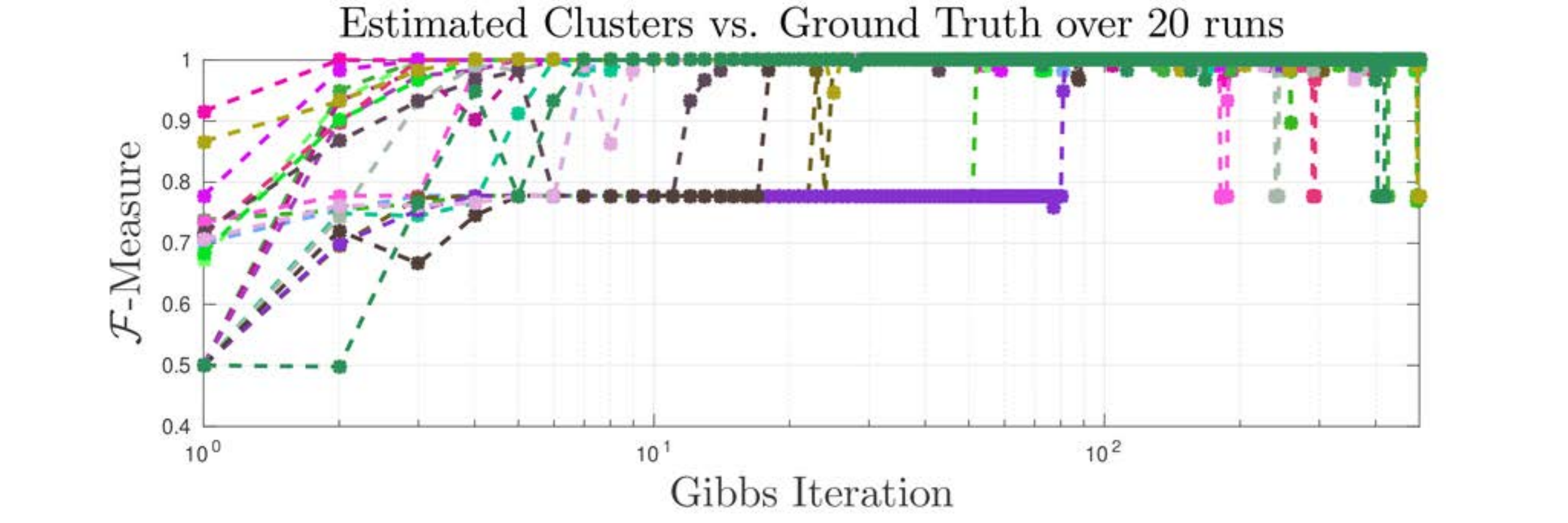}
   \includegraphics[trim={1.5cm 0cm 1.5cm 0cm},clip,width=\linewidth]{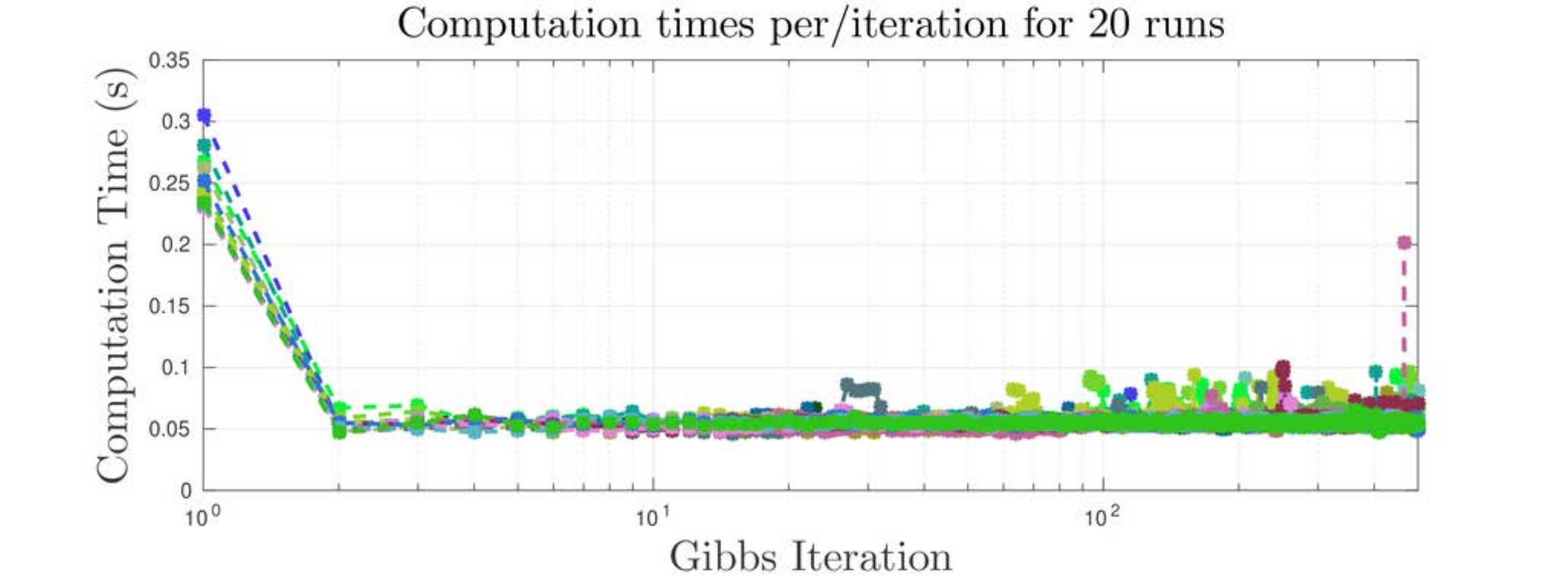}
\end{minipage}\begin{minipage}{0.33\linewidth}
\centering
   \includegraphics[trim={1.5cm 0cm 1.5cm 0cm},clip,width=\linewidth]{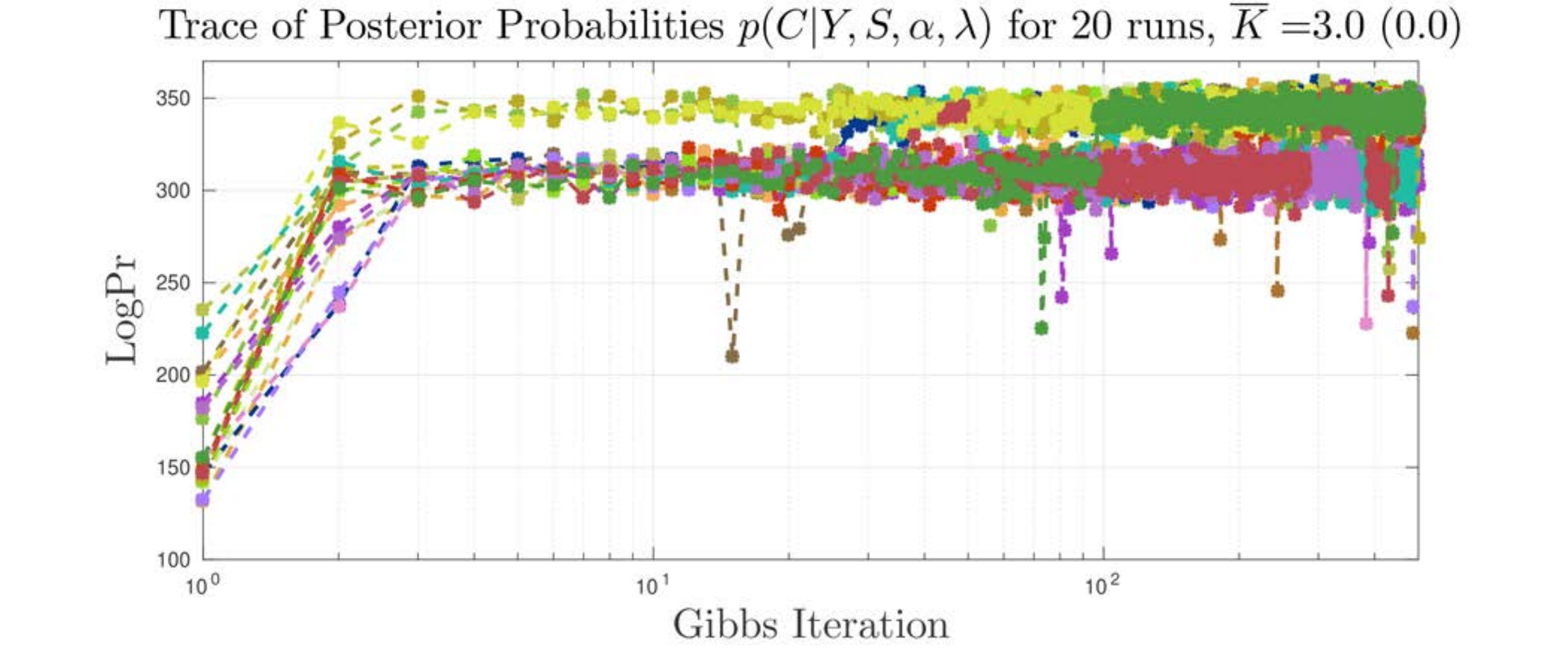}
  \includegraphics[trim={1.5cm 0cm 1.5cm 0cm},clip,width=\linewidth]{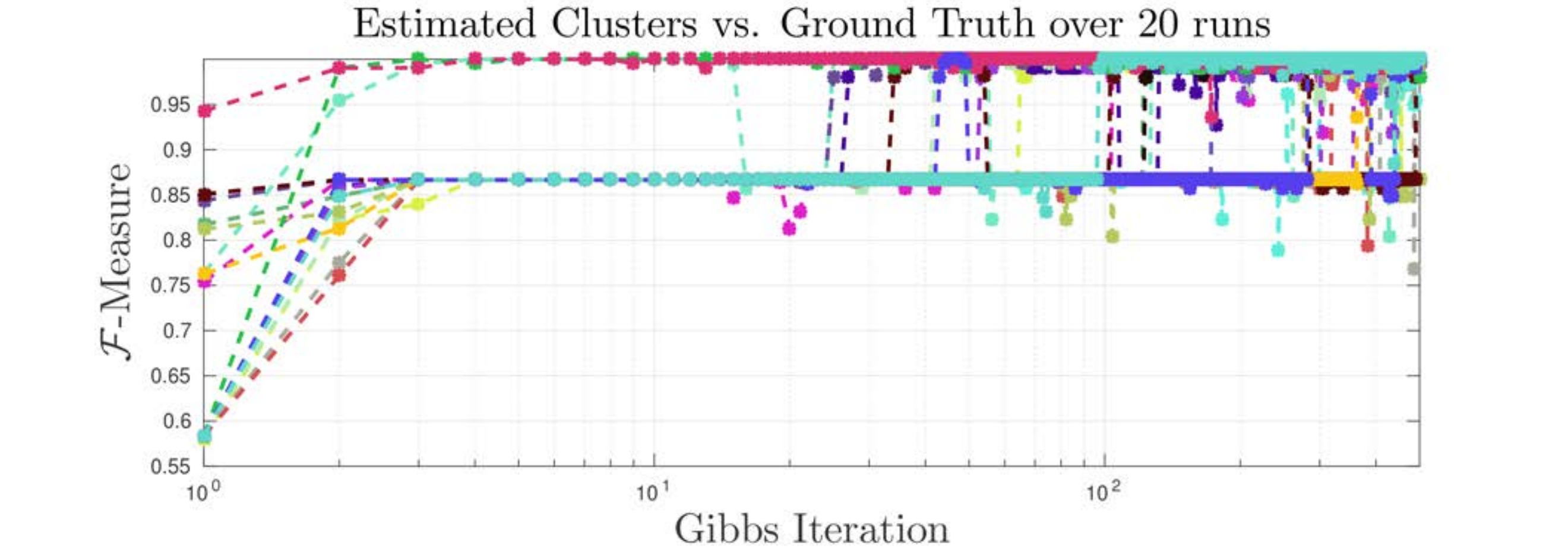}
   \includegraphics[trim={1.5cm 0cm 1.5cm 0cm},clip,width=\linewidth]{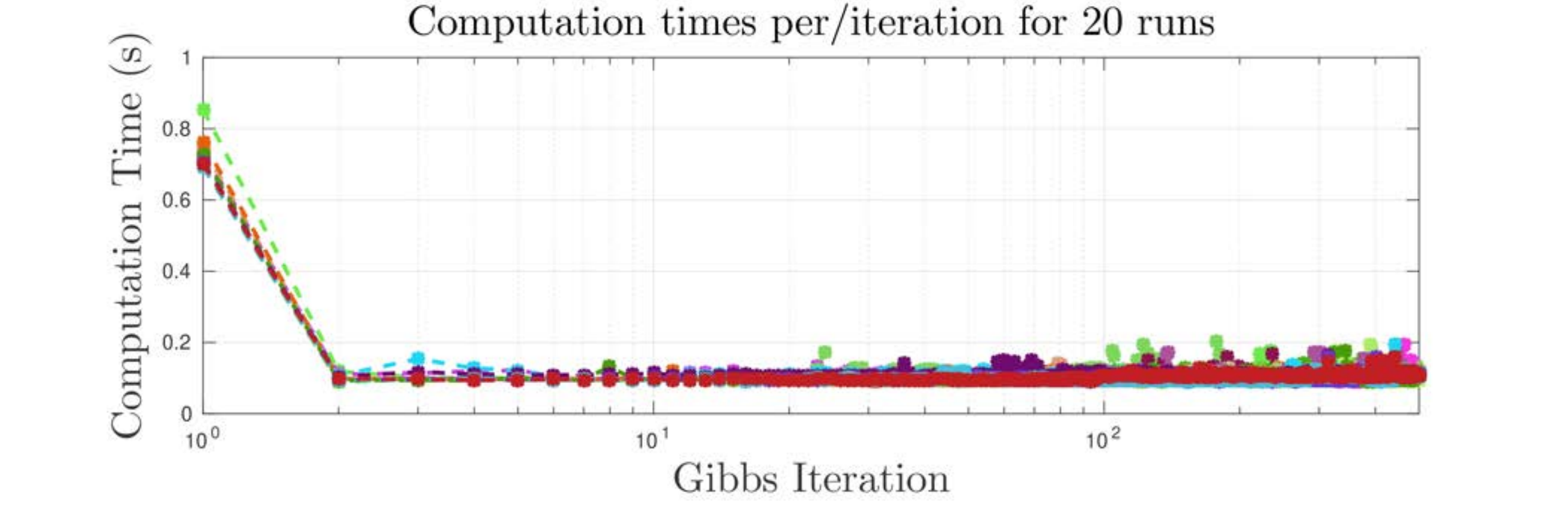}
\end{minipage}\begin{minipage}{0.33\linewidth}
\centering
\vspace{-5pt}
   \includegraphics[trim={1.5cm 0cm 1.5cm 0cm},clip,width=\linewidth]{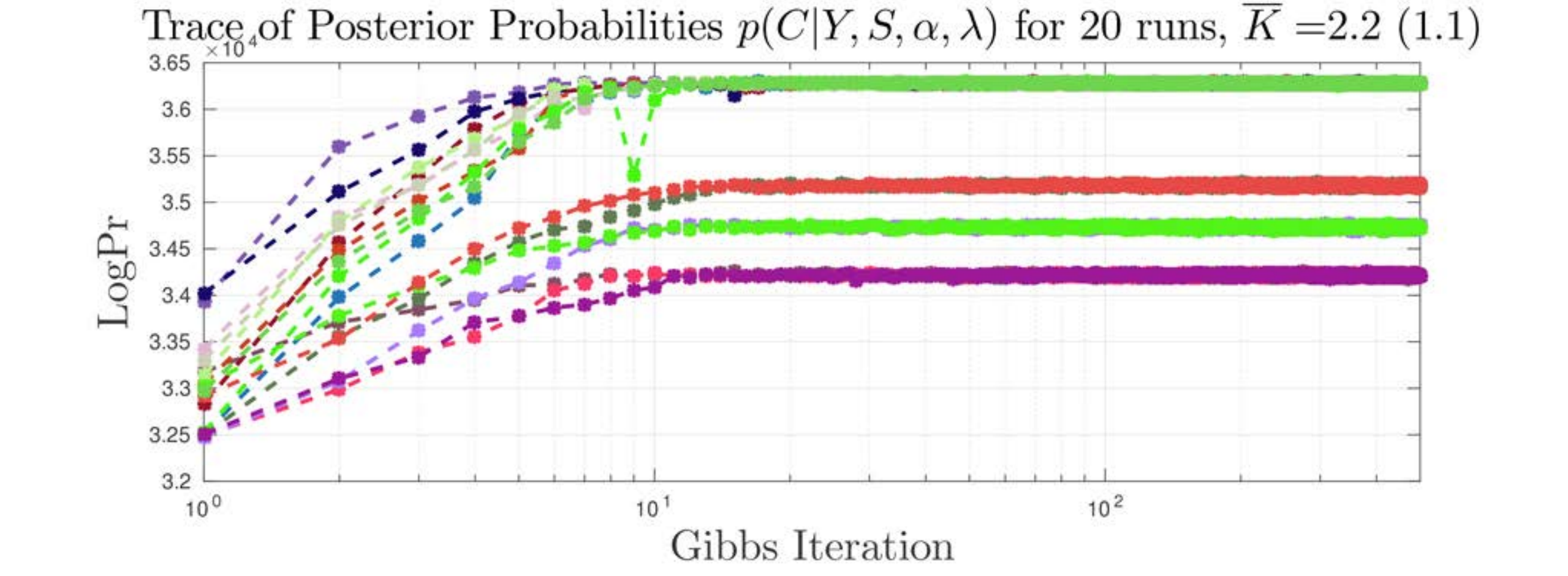}
  \includegraphics[trim={1.5cm 0cm 1.5cm 0cm},clip,width=\linewidth]{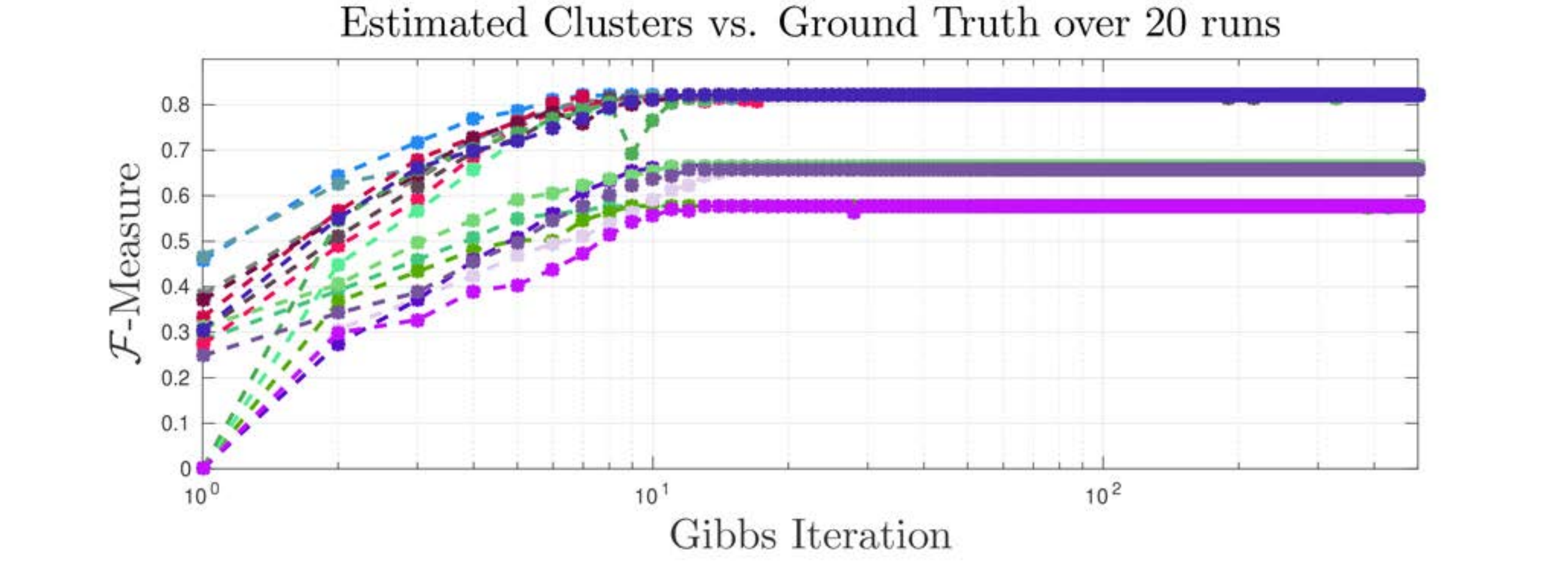}
   \includegraphics[trim={1.5cm 0cm 1.5cm 0cm},clip,width=\linewidth]{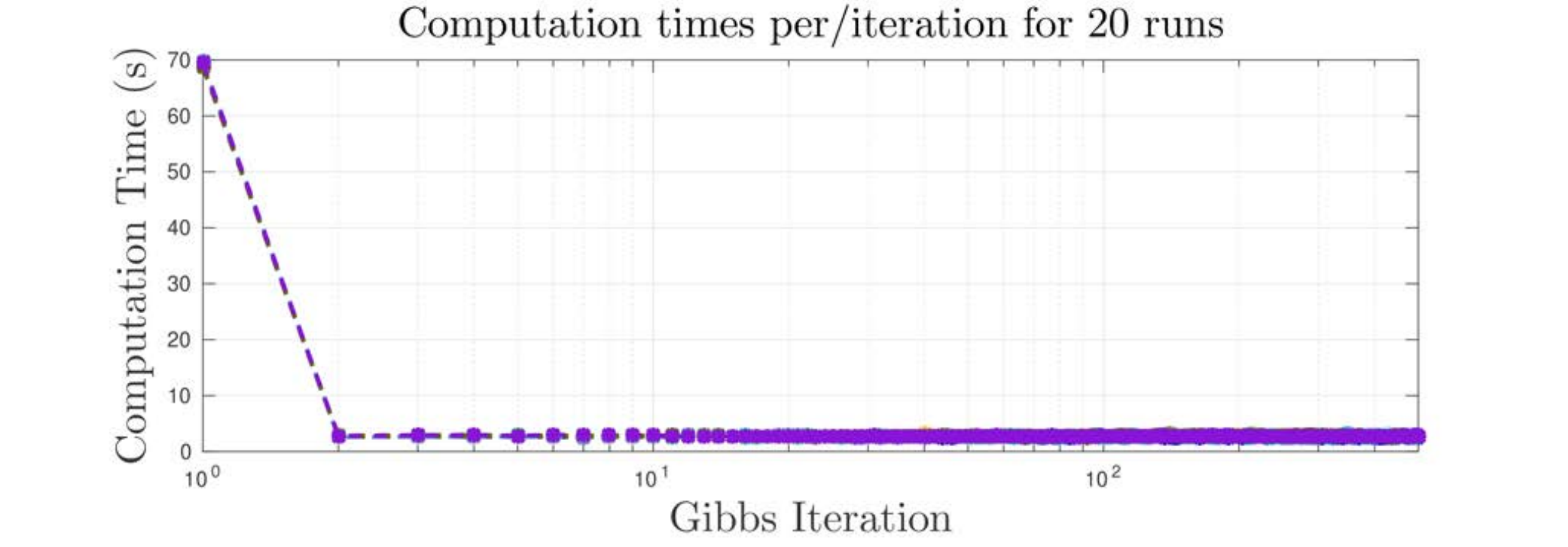}
\end{minipage}
	\captionof{figure}{\small Sampler Tests on (left) \textbf{6D Toy Dataset}, (center) \textbf{6D Real Task Ellipsoid Dataset}  and (right) \textbf{3D Synthethic DTI Dataset}, for 20 random initializations with 500 iterations, setting $\alpha=1, \tau=10 \rightarrow P = 3$, results in average estimated clusters $\overline{K}= 2.7 ~(0.7)$ vs. $\mathbf{K}= 5 $ \label{fig:sampler_DTISYN}}
\end{figure*}

To evaluate the convergence properties of our proposed Collapsed Gibbs Sampler we recorded the trace of the log posterior probabilities, accompanied by the $\mathcal{F}$-measure and computation time per iteration on an Intel® Core™ i7-3770 CPU@3.40GHz$\times$8;  for three of our datasets: \textbf{6D Toy Dataset}, \textbf{6D Real Dataset} and \textbf{3D DTI Synthetic Dataset} (see Figure \ref{fig:sampler_DTISYN}, respectively). 

We devised \textit{sampler tests} where 20 independent chains were run for 500 iterations each. For each dataset a fixed value of $\alpha$ and $\tau$ was defined and all runs begin with each customer sitting by her/himself; i.e. $K=M$. Typically, one's interested in a sampler's capacity for rapid mixing. As can be seen, in all of our tests the Markov chains seem to come close to the steady state distribution in $<100$ iterations. Moreover, as the chains evolve, we can see how the accuracy of our cluster estimates reach \textit{or come closer to} the ground truth (through the $\mathcal{F}$-measure plots). 

Regarding computation cost, the proposed sampler for the SPCM-$\mathcal{CRP}$ is indeed more costly than the classical Collapsed Gibbs sampler for the $\mathcal{CRP}$, as we must recompute the seating assignments for all customers, every time one of them creates or breaks a table. This results in a higher computational cost per iteration, more so, for the first iteration, as all the customers are being assigned to their corresponding tables for the first time, biased by the similarity matrix $\mathbf{S}$. This can be clearly seen in the plots shown in Figure \ref{fig:sampler_DTISYN}. Nevertheless, due to the fact that, in the first iteration, all the seating assignments are explored we see a big jump in both the posterior probabilities and the $\mathcal{F}$-measure, resulting in the rapid mixing capabilities of our sampler.

\subsubsection{Sensitivity to Hyper-parameters}
One of the goals of this work is to alleviate the need for heavy parameter tuning and propose a method that is robust to hyper-parameter changes. As discussed earlier, we have two main hyper-parameters for the SPCM-$\mathcal{CRP}$ mixture model: (1) the tolerance parameter, $\tau$, for the SPCM similarity function and (2) $\alpha$, the concentration parameter for the SPCM dependent $\mathcal{CRP}$ prior. We begin this evaluation by discussing the intuitive effect of each parameter and then show how our clustering algorithm is robust for a different range of values for these parameters.

$\tau$ is a scaling factor for the un-bounded similarity values $\Delta_{ij}$ in \eqref{eq:bspcm}. Generally, this value can be set to $1$, however, when set to values $>1$ the similarity values for lower-dimensional datasets; i.e. $N < 6$ for $C \in \mathds{R}^{N \times N}$; are simply amplified. $\alpha$ plays a role in the computation of the prior probabilities for seating assignments in the SPCM dependent $\mathcal{CRP}$ prior. In \eqref{eq:sdcrp_prior}, we see that the $i$-th observation has a probability of being grouped with the $j$-th observation according to \eqref{eq:bspcm}, which is bounded between [0,1]. The probability of the $i$-th observation belonging to a new singleton cluster is imposed by $\alpha$. Hence, $\alpha$ can be seen as a dispersion parameter, which dictates how tight the clustering must be. Since \eqref{eq:bspcm} is bounded between [0,1], naturally $\alpha$ should take the same values. A lower $\alpha$ would impose less clusters; i.e. more grouped observations; and vice-versa. 

\begin{figure}[!t]
	\begin{minipage}{0.25\textwidth}
		\includegraphics[trim={0cm 0cm 1.5cm 0cm},clip,width=\textwidth]{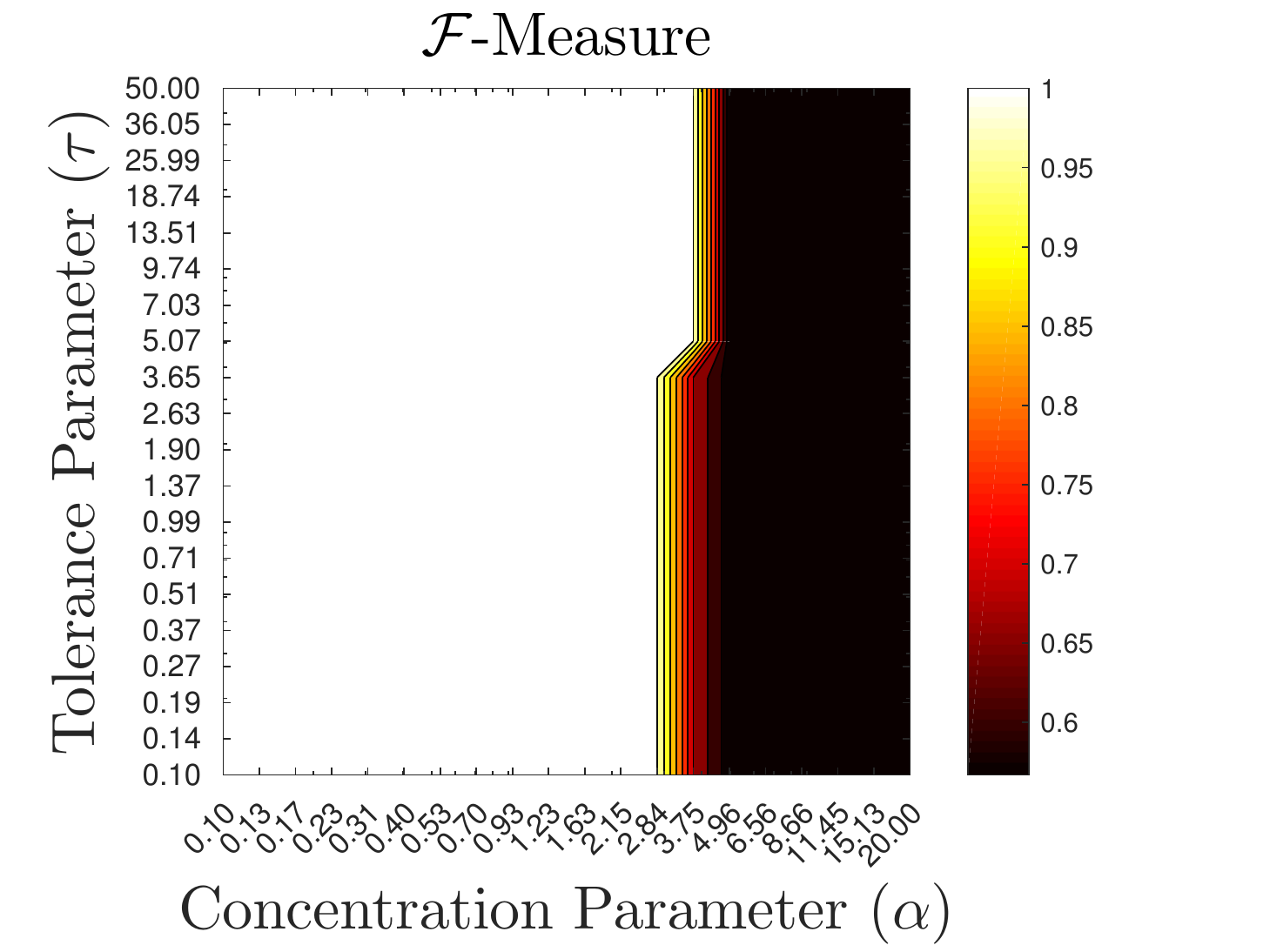}\includegraphics[trim={0cm 0cm 1.5cm 0cm},clip,width=\textwidth]{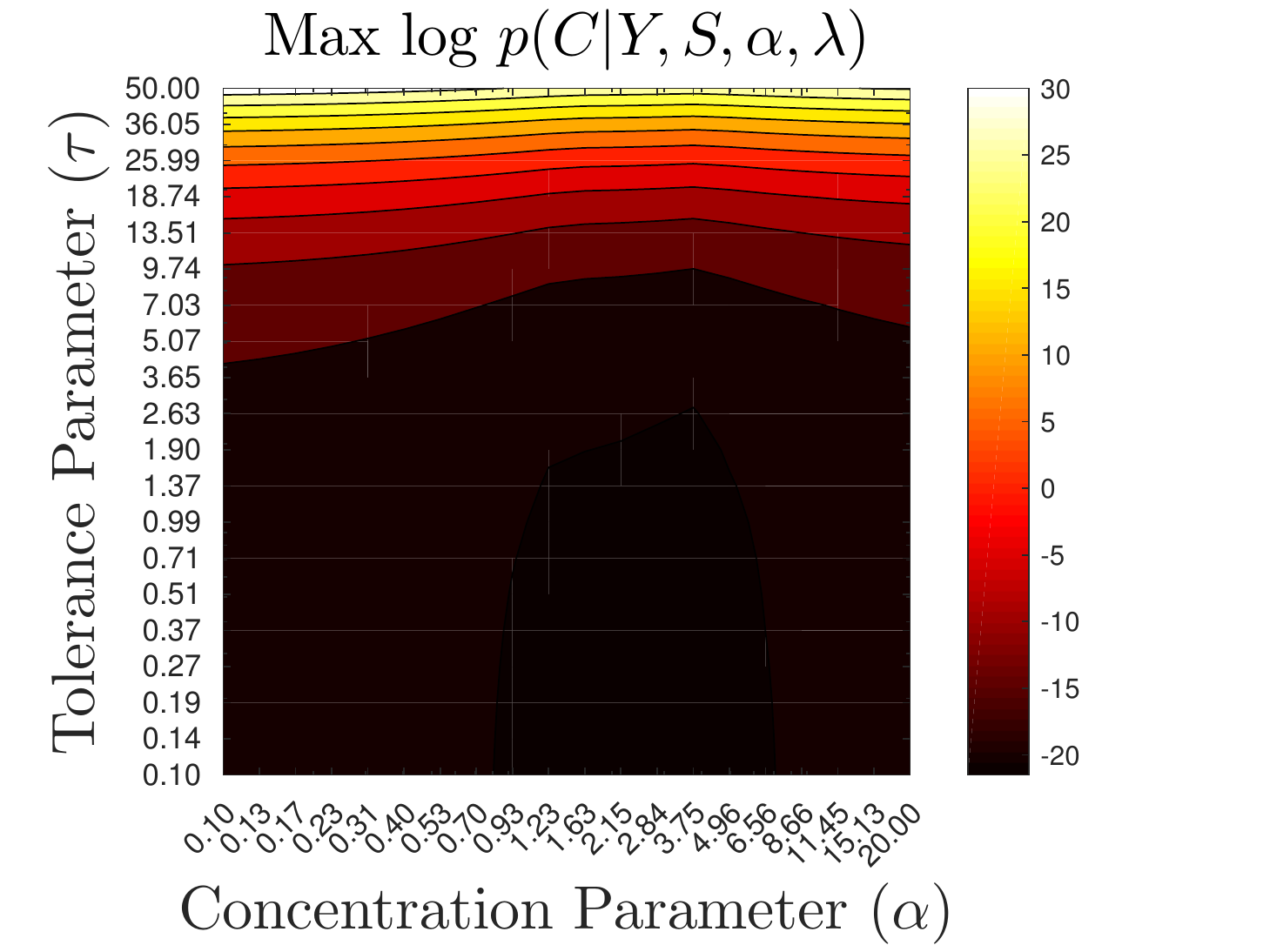}
	\end{minipage}\hspace{110pt}\begin{minipage}{0.25\textwidth}
	\includegraphics[trim={0.5cm 0cm 1.5cm 0cm},clip,width=\textwidth]{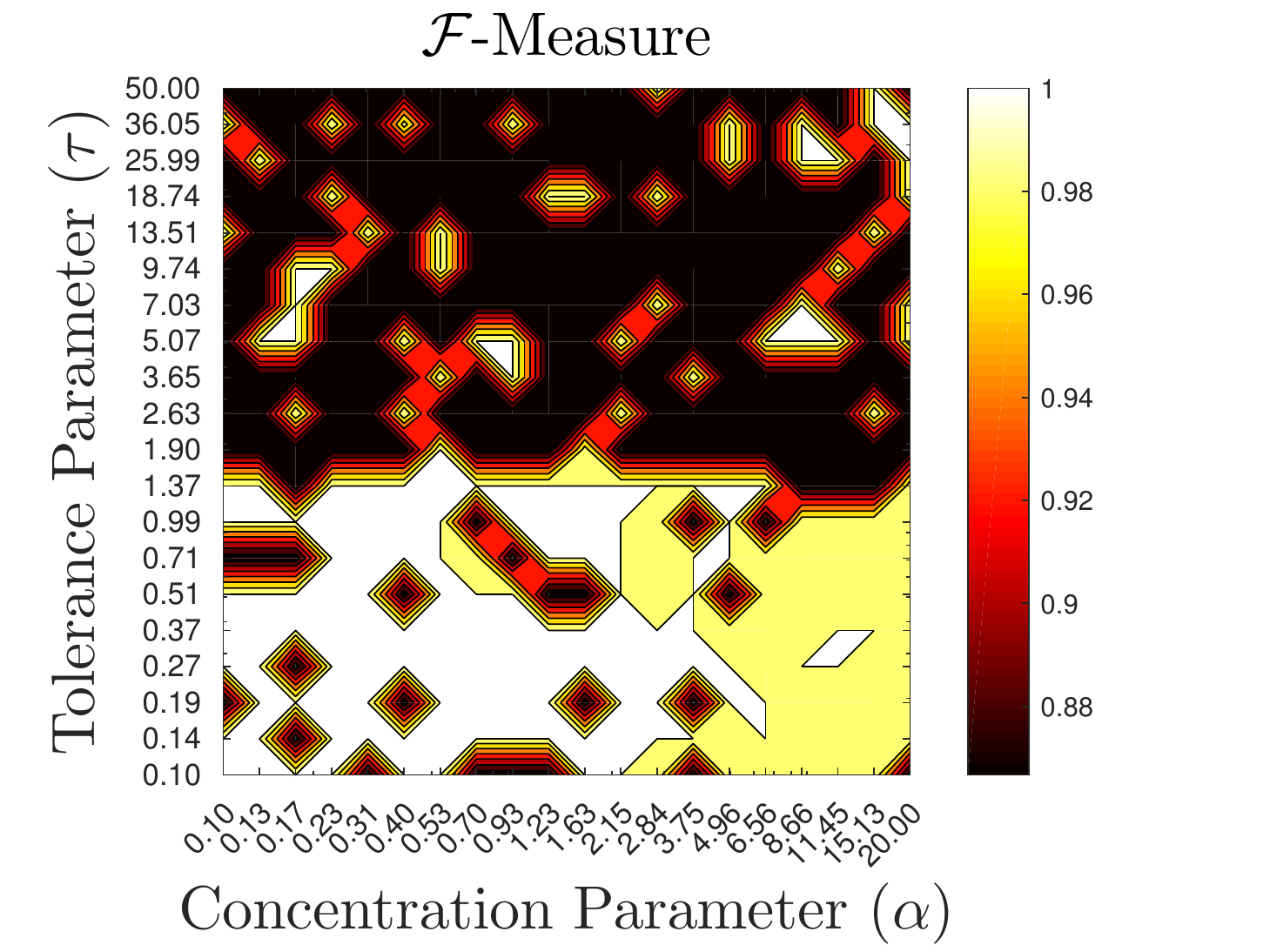}\includegraphics[trim={0.5cm 0cm 1.5cm 0cm},clip,width=\textwidth]{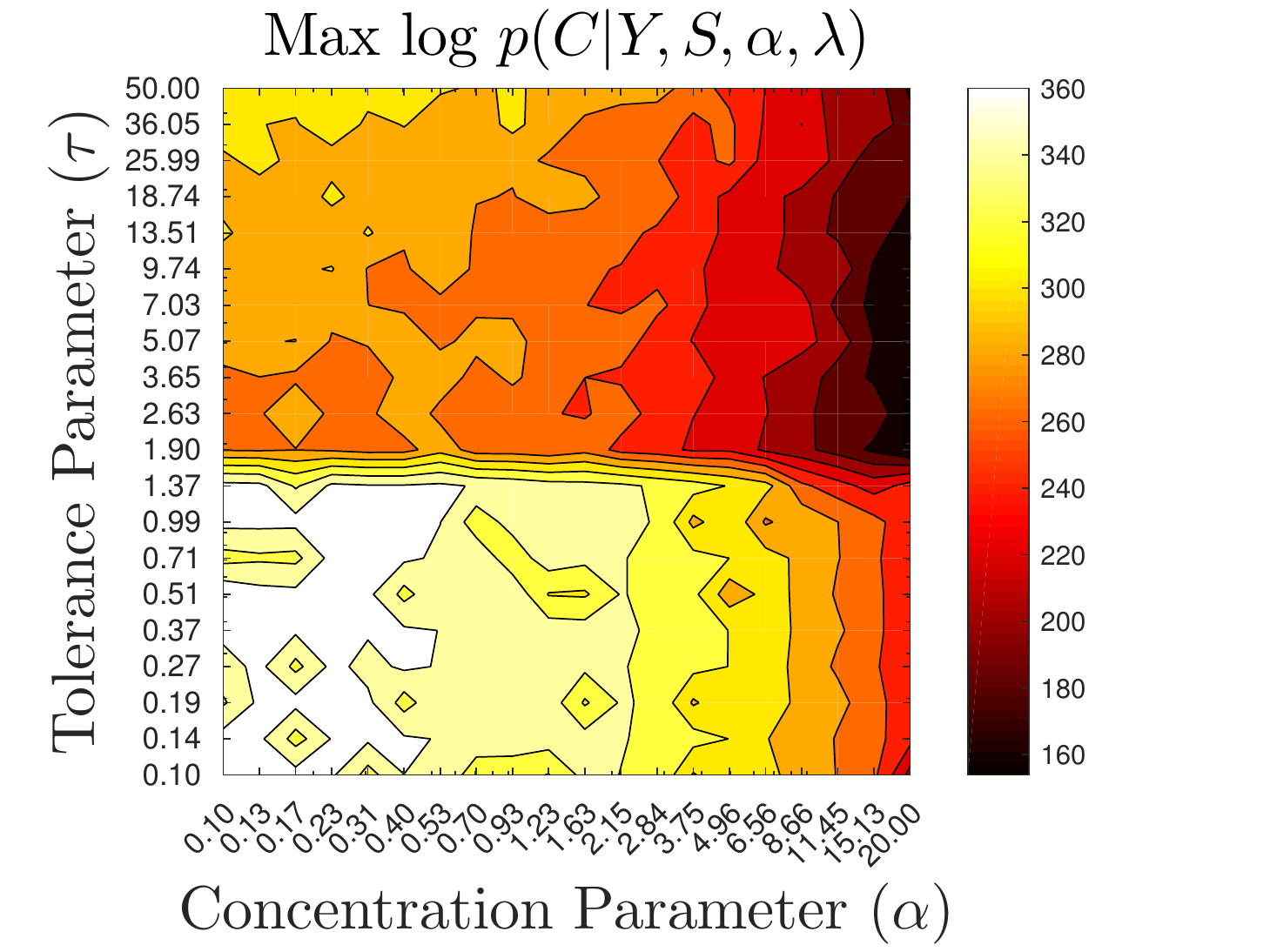}
\end{minipage}
\captionof{figure}{\small Hyper-parameter robustness tests for \textbf{3D Toy Dataset} (left plots) and \textbf{6D Real Dataset} (right plot), setting a log-spaced range of $\alpha=[1,20]$ and $\tau=[1,50] $ \label{fig:hyper_6dReal}}
\end{figure}

To evaluate hyper-parameter sensitivity of our clustering algorithm, we performed a grid search on a log-spaced range of $\tau=[1,50]$ and $\alpha=[0.1,20]$ and recorded the $\mathcal{F}$-measure and $\max \log$ posterior probabilities for two of our datasets, shown in Figure \ref{fig:hyper_6dReal}. We report solely on the \textbf{3D Toy Dataset} and \textbf{6D Real Dataset} as the other datasets exhibited minimal to absolutely no change in performance while increasing $\alpha$ and $\tau$. Our algorithm is not restricted to $\alpha$ taking on values $\geq 1$. For this reason,  we explore a wider range for $\alpha=[0.1,20]$ rather than the natural choice of $\alpha=[0.1,1]$.  A large value for $\alpha>>1$ only imposes a higher probability for observations to not be grouped with other observations; resulting in a higher number of clusters. This can be clearly seen in the $\mathcal{F}$-measure  heatmap, where the $\mathcal{F}$-measure starts to gradually decrease as $\alpha>>2.8$ for the \textbf{3D Toy Dataset}. For the \textbf{6D Real Dataset} we see a similar behavior, the area with highest values of the $\mathcal{F}$-measure is within the range of $\alpha=[0.1,6]$. Hence, we can conclude that our algorithm is not really sensitive to the value $\alpha$ as long as $\alpha < 2$, which is in fact, twice the maximum value of our similarity function \eqref{eq:bspcm}.

Regarding the effect of the tolerance value $\tau$, we can see that for clearly separable datasets of Covariance matrices, as the \textbf{3D Toy Dataset}, this value has no effect whatsoever on the clustering results. However, when the Covariance matrices are not so clearly separable, amplifying the similarities will only deteriorate the clustering results, as Covariance matrices which are not so similar will have higher similarities and hence be projected onto very close coordinates in the spectral embedding.

\subsubsection{Comparison with other methods}
At its core, our proposed clustering algorithm is a mixture model applied to the spectral embedding $\mathbf{Y}\in \mathds{R}^{P \times M}$ induced by a similarity matrix $\mathbf{S} \in \mathds{R}^{M \times M}$. The novelty of our approach is not only that we use a Bayesian non-parametric mixture model on this embedding, but, that we impose a similarity-dependent prior through the very same similarities that were used to create the embedding. Hence, in order to highlight the power of imposing this new prior on the mixture model we compare our approach to two variants: (i) SCPM embedding ($\mathbf{Y}$)+GMM (w/Model Selection) and (ii) SCPM embedding ($\mathbf{Y}$)+$\mathcal{CRP}$-GMM.

Since we propose a non-parametric approach, the first mixture variant is used as a baseline. Our goal is thus, to surpass the performance of the standard $\mathcal{CRP}$-GMM and exhibit better or comparable performance to the classic finite GMM, with parameters optimized through model selection. We ran the three algorithms 10 times for all of our datasets and report mean (std.) of the corresponding clustering metric in Table \ref{tab:spcm-clust-compare}. Numbers in \textbf{bold} indicate best scores. As can be seen, for all datasets, our approach is superior to applying the standard $\mathcal{CRP}$-GMM on $\mathbf{Y}$ and, either superior or comparable to the finite GMM variant.

\begin{table}[!h]
\begin{center}
\footnotesize
\begin{tabular}{cc|ccc}
    \hline
    \hline
    \multicolumn{1}{c}{\multirow{4}{*}{Dataset}} & \multicolumn{1}{c}{\multirow{4}{*}{Metrics}} & \multicolumn{3}{c}{\multirow{2}{*}{Spectral Embedding ($\mathbf{Y}$) + Clustering}}
	\\  
    & & \multicolumn{1}{c}{\multirow{2}{*}{GMM w/MS}} & \multicolumn{1}{c}{\multirow{2}{*}{CRP-GMM}} & \multicolumn{1}{c}{\multirow{2}{*}{\textbf{Our approach}}}    \\ \\ \hline
    
    \multicolumn{1}{c}{\multirow{4}{*}{Toy 3D}} & NMI & 0.807 (0.018) & 0.298 (0.316) & \textbf{1.00} \\
    & Purity & 1.00 & 0.74 (0.135) & \textbf{1.00}\\
    & $\mathcal{F}$ & 0.877 (0.006) & 0.738 (0.111) & \textbf{1.00}\\
    & $K(2)$ & 3 & 1.8 (0.789)& \textbf{2}\\
        \hline
        
    \multicolumn{1}{c}{\multirow{4}{*}{Toy 6D}} & NMI & \textbf{1.00} & 0.633 (0.342) & \textbf{1.00} \\
	& Purity & \textbf{1.00} & 0.633 (0.189) & \textbf{1.00}\\
	& $\mathcal{F}$ & \textbf{1.00} & 0.744 (0.146) & \textbf{1.00}\\
    & $K(3)$  & \textbf{3} & 1.900 (0.568) & \textbf{3}\\
	\hline
	
	\multicolumn{1}{c}{\multirow{2}{*}{6D Task}} & NMI &  0.893 (0.129) & 0.599 (0.417) & \textbf{0.984 (0.050)} \\   
	\multicolumn{1}{c}{\multirow{3}{*}{Ellipsoids}} & Purity &  0.952 (0.080) & 0.759 (0.126)  &  \textbf{0.980 (0.063)}  \\
	& $\mathcal{F}$ &  0.943 (0.100) & 0.794 (0.151) & \textbf{0.987 (0.042)} \\
    & $K(3)$ & \textbf{3} & 1.800 (0.632) & \textbf{2.900 (0.316)} \\
	\hline
	
	\multicolumn{1}{c}{\multirow{1}{*}{3D DTI}} & NMI & \textbf{0.804} & 0.640 (0.089)  &  0.730 (0.102)\\
	\multicolumn{1}{c}{\multirow{2}{*}{Synthetic}} & Purity & \textbf{0.701} & 0.536 (0.087) &  0.620 (0.105) \\
    & $\mathcal{F}$ & \textbf{0.750} & 0.536 (0.087) &  0.679 (0.097) \\
    $(\tau=10)$ & $K (5)$ & 3 & 2.200 (0.422) & 2.700  (0.675) \\
	\hline
	
	\multicolumn{1}{c}{\multirow{1}{*}{3D DTI }} & NMI & 0.585 & 0.502 (0.197) &  \textbf{0.633 (0.039)} \\
	\multicolumn{1}{c}{\multirow{2}{*}{Rat's Hippo.}} & Purity & 0.696 & 0.652 (0.137) &  \textbf{0.745 (0.036)} \\
	& $\mathcal{F}$ & 0.705 & 0.502 (0.197) &  \textbf{0.740 (0.046)} \\
	$(\tau=10)$ & $K (4)$ & 3 & 3.100 (0.994) & \textbf{4.600 (0.699)} \\	
	\hline	
	
	\multicolumn{1}{c}{\multirow{1}{*}{400D Cov.}} & NMI & \textbf{0.541} & 1.800 (0.632) &  0.486 (0.091)\\
	\multicolumn{1}{c}{\multirow{2}{*}{ETH-80}} & Purity & \textbf{0.325} &  0.214 (0.061) &  0.265 (0.038)\\
	& $\mathcal{F}$ & \textbf{0.441} & 0.338 (0.080) &  0.386 (0.038) \\
	& $K (8)$ & 3 & 1.800 (0.632) & 2.889 (0.333) \\
	\hline	
	\hline
\end{tabular}
\end{center}
\caption{Performance Comparison of Covariance Matrix clustering with  \textbf{Our Proposed Approach} vs. GMM w/Model Selection and CRP Mixture Model \textit{(presenting mean (std) of metrics over 10 runs).} \label{tab:spcm-clust-compare}}
\end{table}

\begin{figure*}[!h]
	\centering
	\subfloat[Embedding $\mathbf{Y}$ with true labels]{\includegraphics[trim={1cm 0cm 1.5cm 0cm},clip,width=0.29\linewidth]{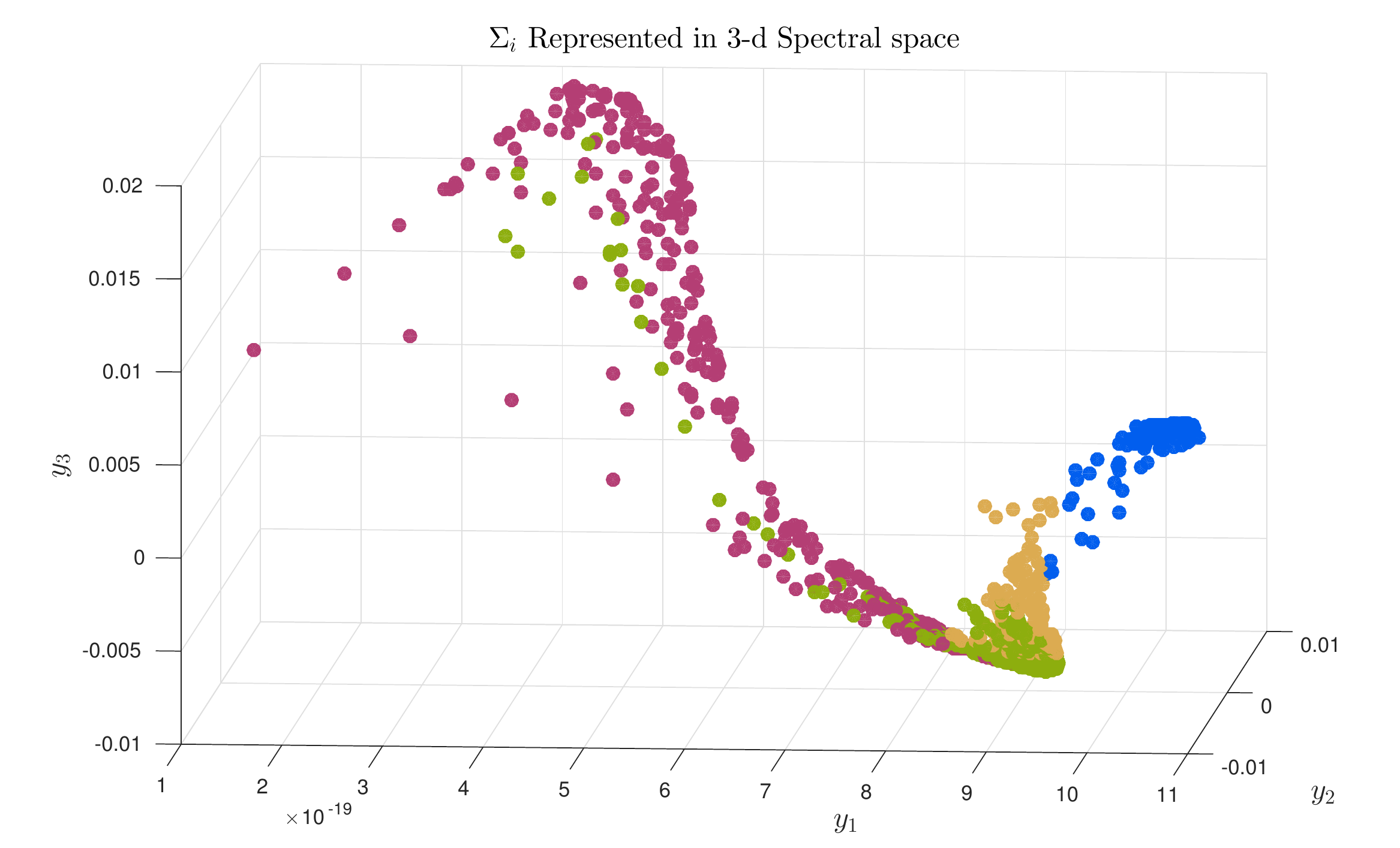}}
	\subfloat[GMM+MS $K=3$]{\includegraphics[trim={0cm 0cm 1.5cm 0cm},clip,width=0.24\linewidth]{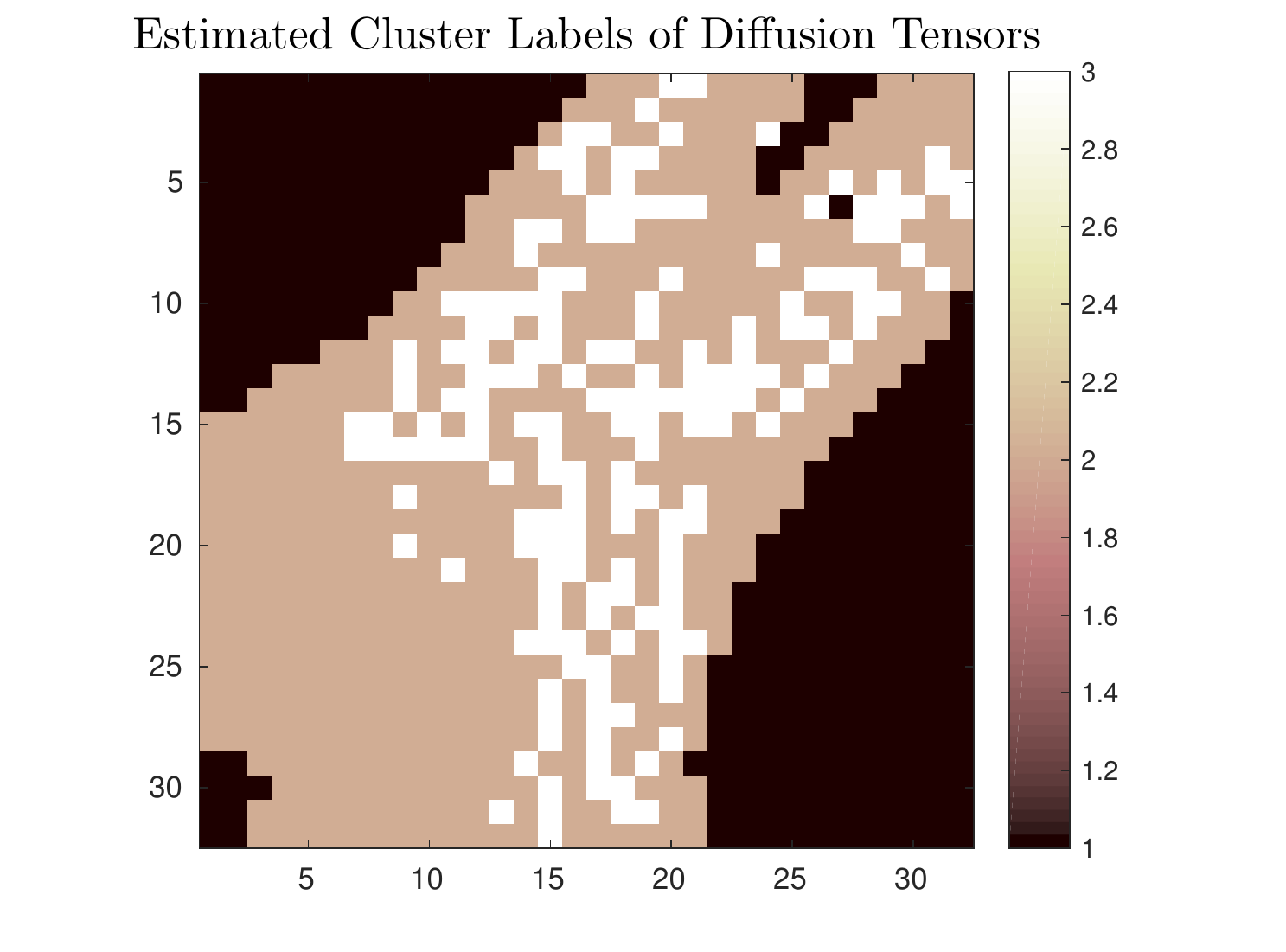}}\subfloat[CRP-MM $K=3$]{\includegraphics[trim={0cm 0cm 1.5cm 0cm},clip,width=0.24\linewidth]{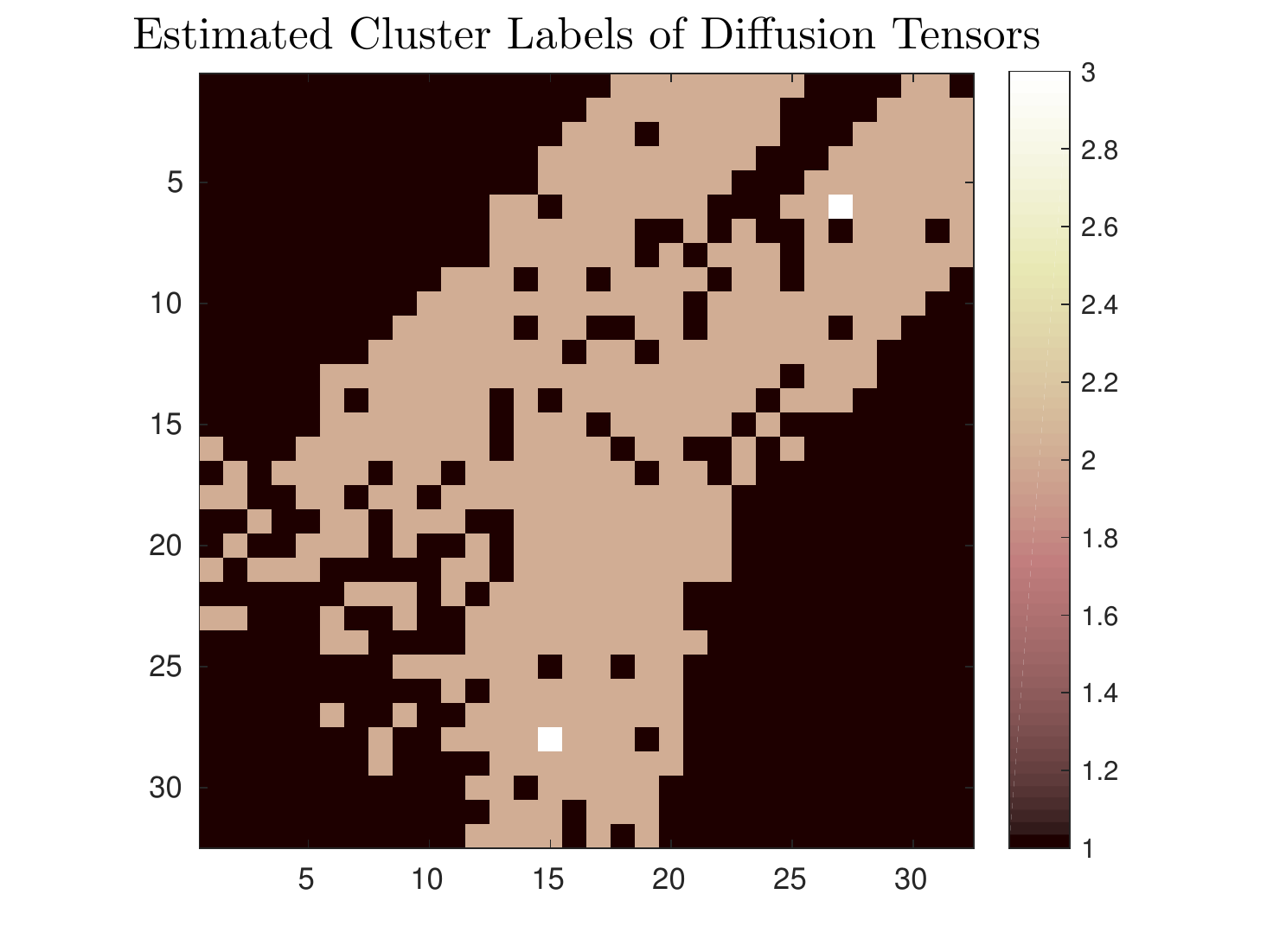}}\subfloat[\textbf{Our Approach} $K=5$]{\includegraphics[trim={0cm 0cm 1.5cm 0cm},clip,width=0.24\linewidth]{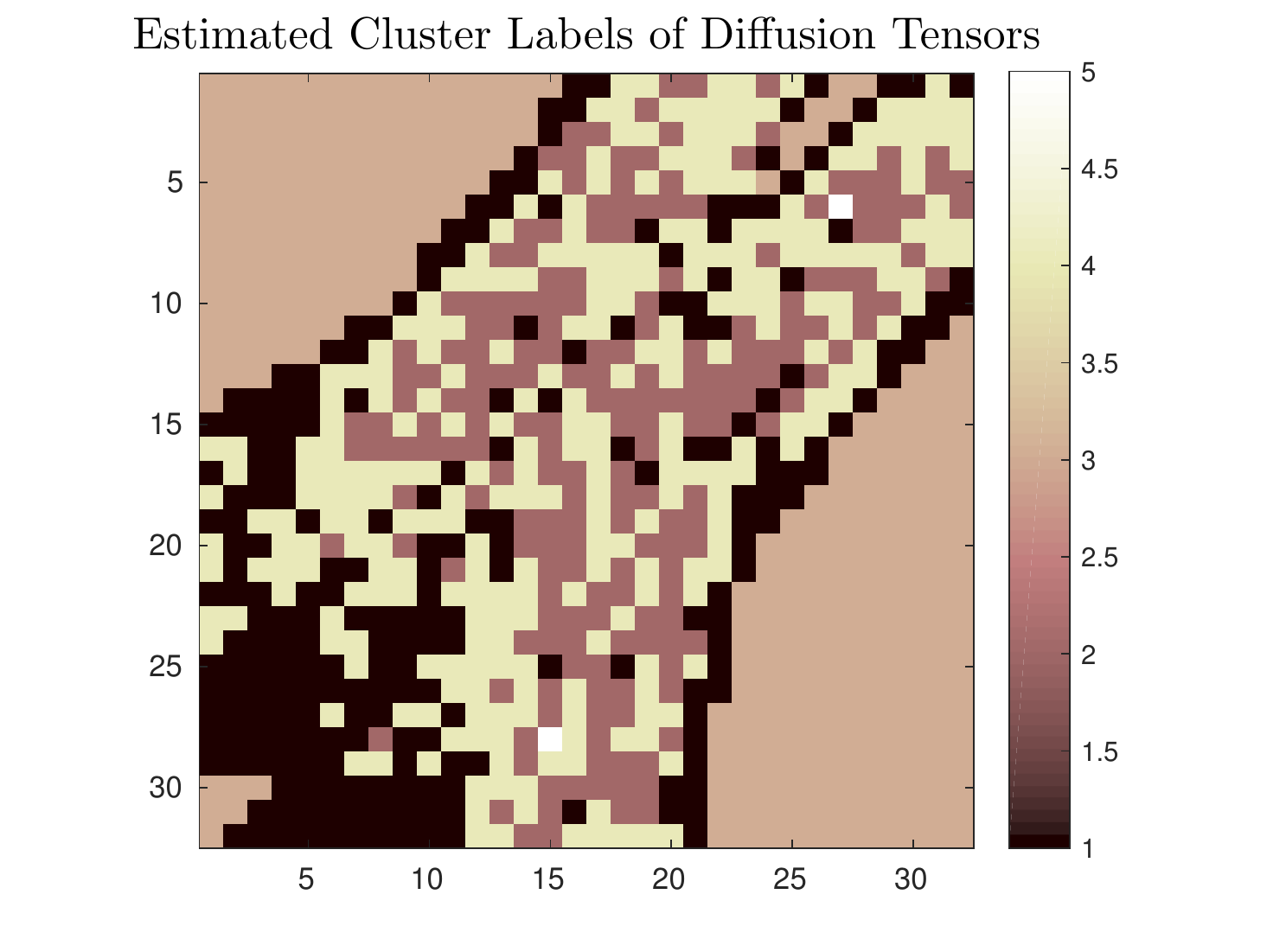}}
	\captionof{figure}{\small Clustering of \textbf{DTI Synthetic Dataset} \textbf{DTI Real Dataset} \label{fig:clustering_dti}}
\end{figure*}

It must be noted that, the optimal $K$ for the finite GMM variant was chosen considering the interpretation of the BIC/AIC curves as well as our prior knowledge of the optimal clusters in each dataset. Moreover, in both $\mathcal{CRP}$-GMM and SPCM-$\mathcal{CRP}$-GMM we set $\alpha=1$. In the $\mathcal{CRP}$, $\alpha$ has an analogous effect on clustering dispersion. In all of our datasets, we see that the $\mathcal{CRP}$-GMM generally produces less clusters than our approach. Moreover, thanks to the similarity-dependent prior, our approach seems to extract more meaningful clusters from the datasets. A clear example of this is shown on the clustering results from all algorithms on the DTI datasets presented in Figure \ref{fig:clustering_dti}. For the Real DTI dataset, which represents a lattice of diffusion tensors captured from a rat's hippocampus, we can see that our approach recovers an image segmentation much closer to the expected segmentation (see Figure \ref{fig:DTI-datasets}) as opposed to the other mixture variants. In this figure, we can also see the automatically generated embeddings $\mathbf{Y}$ with their corresponding true labels. Finally, for the challenging ETH-80 dataset, although low, our approach exhibits comparable performance to the finite mixture variant. Such low performance was expected as this dataset contains extremely overlapping classes which no clustering algorithm is capable of extracting.

\subsection{Datasets \& Metrics for Segmentation Evaluation}
To evaluate the segmentation and action recognition algorithm proposed in Section \ref{sec:ICSC-HMM}, we use the following datasets\footnote{All datasets in this section are available in: \url{https://github.com/nbfigueroa/ICSC-HMM}.\\ They will be uploaded to the UCI Machine Learning Repository upon acceptance.}:\\

\noindent \textbf{Toy 2D Dataset}: A set of 4 synthetic 2-D time-series generated by 2 Gaussian distributions, as the one shown in Figure \ref{fig:problem2}. Within each time-series the two Gaussian emission models switch randomly and are transformed with a random rotation and scaling factor. Hence, the goal is to extract $K = 4$ transform-dependent emission models and group them into $K_{Z} = 2$ transform-invariant state clusters.\\

\noindent \textbf{7D Vegetable Grating Dataset \citep{Pais:IEEE:2015}}: 7-D time-series data of kinesthetic demonstrations on a 7DOF KUKA arm for a Carrot Grating Task. These demonstrations are trajectories of the end-effector of the robot consisting of $M=12$ (7-D) time-series $\mathbf{X}^{(i)} = \{\mathbf{x}_1,\cdots,\mathbf{x}_{T^{(i)}}\}$  for $i=\{1,\dots,12\}$. The 7 dimensions in $\mathbf{x}_t = [\xi_1, \xi_2, \xi_3, q_i, q_j, q_k, q_w]^{T}$ correspond to position and orientation of the robot's end-effector. This task involves 3 actions: (a) reach (b) grate and (c) trash. The interesting feature of this dataset is that 6 of the time-series are recorded in one reference frame (RF); i.e. the base of the robot and the other 6 are recorded in the RF of the grating tool. Hence, the two sets of demonstrations are subject to a rigid transformation. The goal is thus to segment all time-series into $K\approx6$ transform-dependent actions and group them into $K_z = 3$ transform-invariant actions (as in Figure \ref{fig:segmentation_results}).\\

\noindent \textbf{13D Dough Rolling Dataset \citep{Figueroa:HRI:2016}}: 13-D time-series data of kinesthetic demonstrations on a 7DOF KUKA arm for a Dough Rolling Task. The trajectories consist of $M=15$ (13-D) time-series $\mathbf{X}^{(i)} = \{\mathbf{x}_1,\cdots,\mathbf{x}_{T^{(i)}}\}$  for $i=\{1,\dots,12\}$. The 13 dimensions in $\mathbf{x}_t = [\xi_1, \xi_2, \xi_3, q_i, q_j, q_k, q_w, f_x, f_y, f_z, \tau_x, \tau_y, \tau_z]^T$ correspond to position, orientation, forces and torque sensed on the end-effector during kinesthetic teaching. This task involves 3 actions: (a) reach (b) roll and (c) back, which are repeated 2-4 times within each time-series. In this case, the RF is fixed to the origin of the table where the dough is being rolled (see \href{https://www.youtube.com/watch?v=br5PM9r91Fg}{video}), however, two distinctive rolling directions are observed: (i) along the x -axis and (ii) y-axis of the table, as seen in Figure \ref{fig:segmentation_results}. Hence, the goal is to segment all time-series into $K\approx6$ transform-dependent actions and group them into $K_z \approx 3$ transform-invariant actions.\\

\begin{figure*}[!t]
	\begin{minipage}{0.32\textwidth}
		\centering
		\vspace{-5pt}
    \includegraphics[trim={1cm 1cm 1.5cm 2.5cm},clip,width=0.825\linewidth]{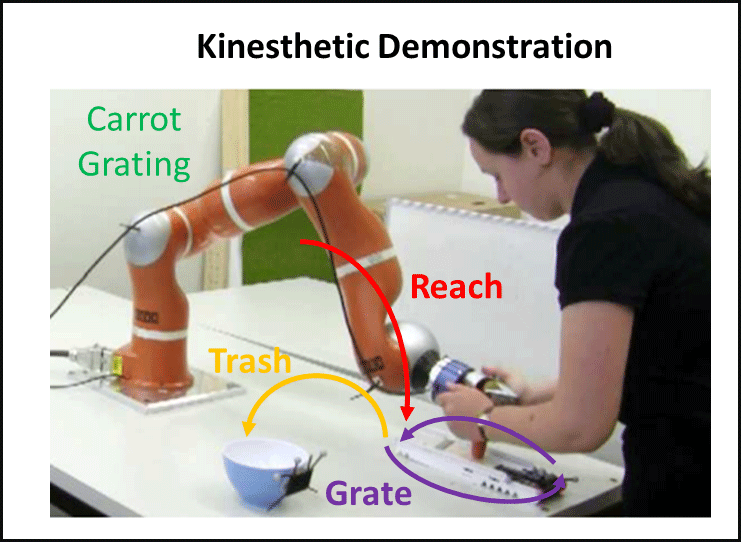}
	\subfloat[Grating Dataset]{\includegraphics[trim={0.65cm 7.3cm 1cm 0.25cm},clip,width=0.95\linewidth]{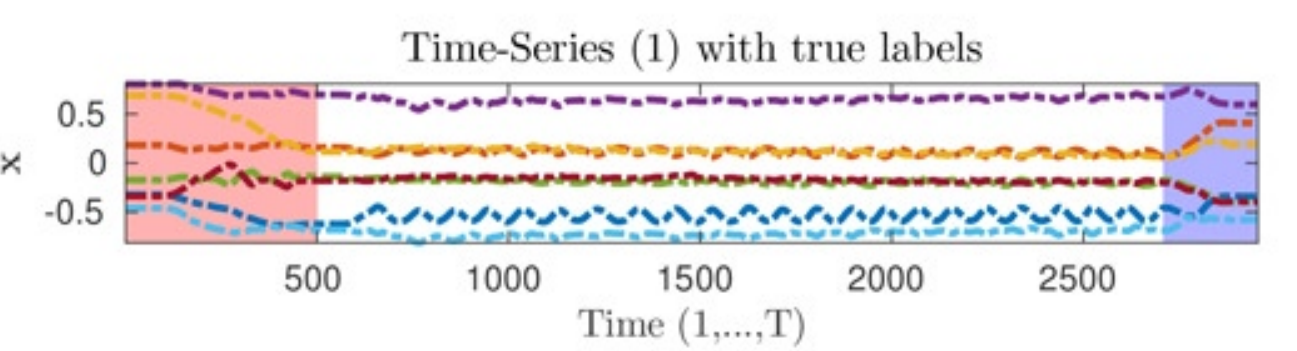}}
	\end{minipage}
\begin{minipage}{0.32\textwidth}
	\centering
    \includegraphics[trim={1cm 0cm 1.5cm 1.5cm},clip,width=0.865\linewidth]{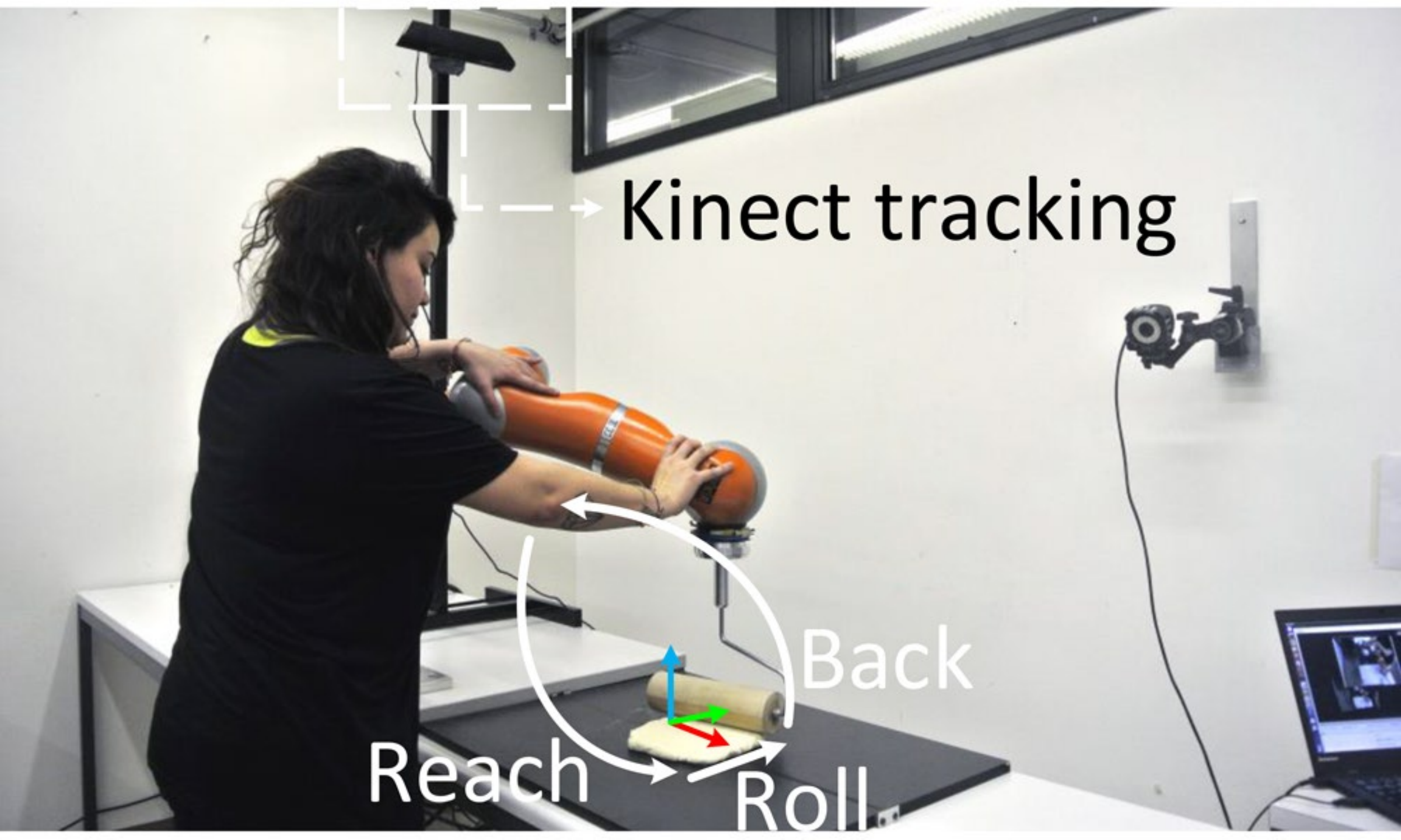}
\subfloat[Rolling Dataset]{\includegraphics[trim={0.95cm 7.35cm 1cm 0.4cm},clip,width=\linewidth]{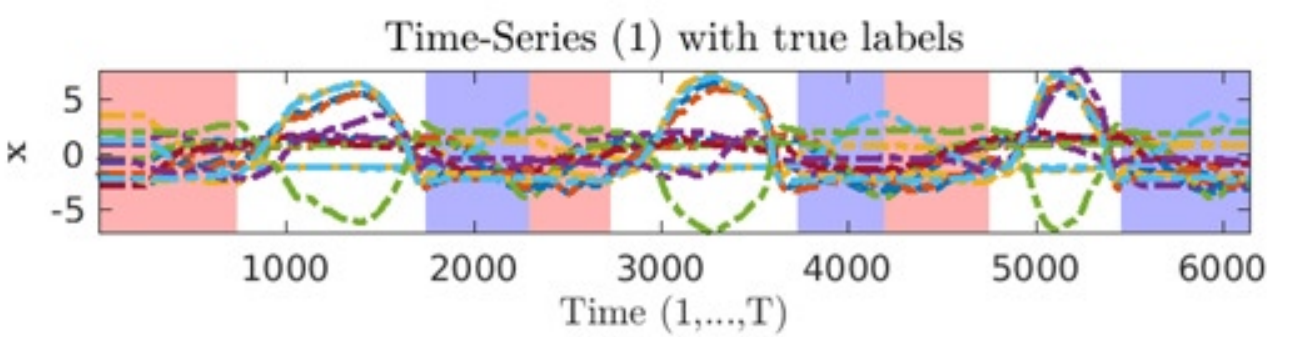}}
\end{minipage}	
\begin{minipage}{0.32\textwidth}
	\centering
    \includegraphics[trim={0cm 4cm 0.85cm 1.75cm},clip,width=0.78\linewidth]{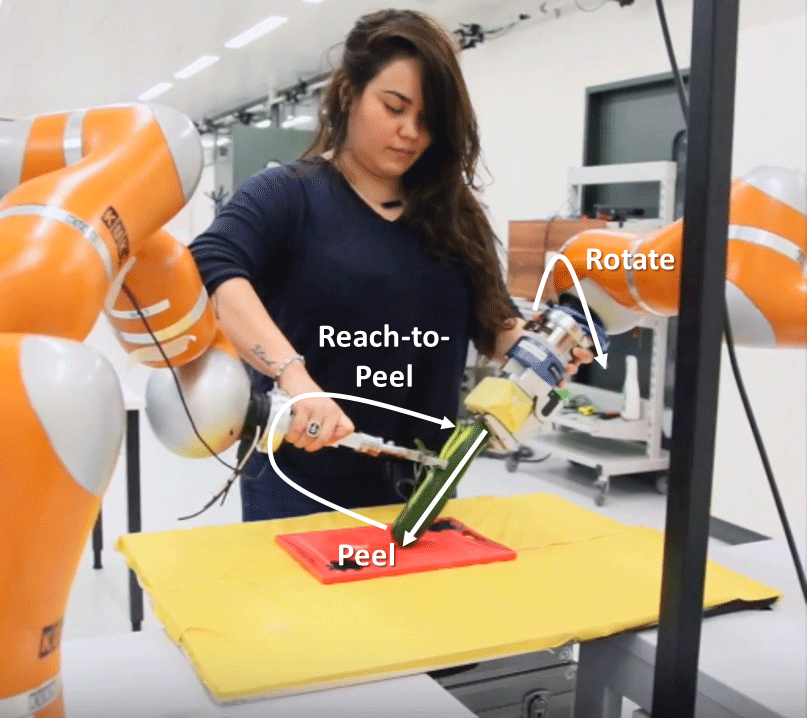}
	\subfloat[Peeling Dataset]{\includegraphics[trim={0.5cm 7.25cm 1cm 0.65cm},clip,width=\linewidth]{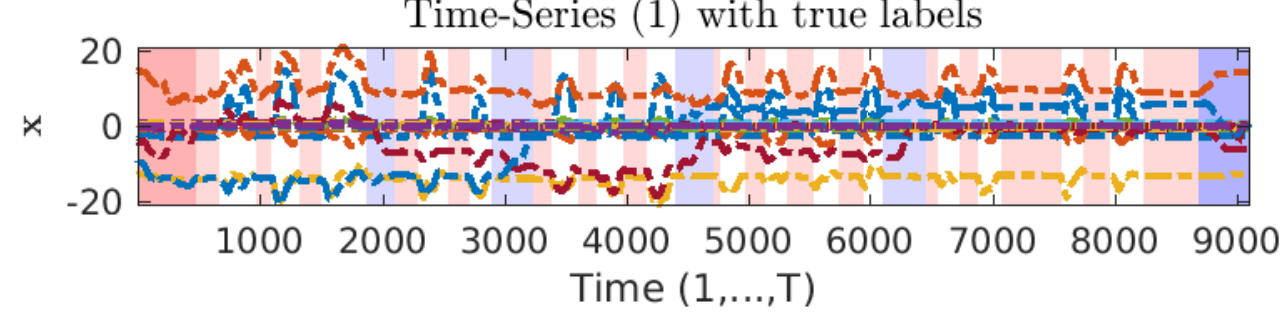}}
\end{minipage}	
	\captionof{figure}{\small Sample time-series from \textbf{real} datasets with ``true labels" defined by human segmentations. Colors indicate \textit{ground truth} segmentation. \label{fig:segmentation_true}}
\end{figure*}

\noindent \textbf{26D Bi-manual Zucchini Peeling Dataset}: 26-D time-series data of kinesthetic demonstrations on a pair of 7DOF KUKA arms for a Zucchini Peeling Task. The trajectories consist of $M=5$ (26-D) time-series $\mathbf{X}^{(i)} = \left\{ [\mathbf{x}_1^{(left)}; \mathbf{x}_1^{(right)}]  ,\cdots,[\mathbf{x}_T^{(left)}; \mathbf{x}_T^{(right)}]  \right\}$  for $i=\{1,\dots,5\}$, where  $\mathbf{x}_t^{(\cdot)}$ corresponds to the position, orientation, forces and torque of an end-effector. This task involves 5 bi-manual actions: (a) \textit{reach}, (b) \textit{reach-to-peel},(c) \textit{peel}, (d) \textit{rotate} and (e) \textit{retract}. Each demonstration begins with a \textit{reach} and ends with a \textit{retract}. Between these two actions a sequence of \textit{reach-to-peel}, \textit{peel} and \textit{rotate} is repeated $\approx3-5$ times. Moreover, an internal sub-sequence of \textit{reach-to-peel} and \textit{peel} is repeated in each case $\approx3-5$ times. As shown in Figure \ref{fig:segmentation_true} the workspace of the robots is limited to the cutting board (in red), nevertheless, we observe some transformations between time-series corresponding to the location and length of the zucchini, as can be seen in Figure \ref{fig:segmentation_results}. Moreover, this dataset is particularly challenging due to its dimensionality and switching dynamics. Hence, the goal is to segment all time-series into $K\approx7$ transform-dependent actions and group them into $K_z \approx 5$ transform-invariant actions.\\

\noindent \textbf{External Segmentation Metrics}
Given that we have the true labels for all of our datasets (from human segmentations), we use the following external segmentation metrics (in addition to the clustering metrics used in the previous section) to evaluate our proposed approach: 
\begin{enumerate}[leftmargin=*]
\item The \textit{\underline{Hamming}} distance measures the distance between two sets of strings (or vectors) as the number of mismatches between the sets \citep{Mulmuley1987}.
\vspace{-5pt}
\item The \textit{Global Consistency Error \underline{(GCE)}} is a measure that takes into account the differences in granularity while comparing two segmentations \citep{MartinFTM01}. 
\vspace{-5pt}
\item \textit{Variation of Information \underline{(VI)}} is a metric which defines the distance between two segmentations as the average conditional entropy of the two segmentations \citep{Meila:2005:CCA}.
\end{enumerate}
\normalsize
Refer to Appendix \ref{app:segmentation_metrics} for computation details of each metric.

\subsection{ICSC-HMM Evaluation}
\subsubsection{MCMC Sampler Convergence}
\begin{figure*}[!t]
\begin{minipage}{0.325\linewidth}
\centering
   \includegraphics[trim={1.25cm 19cm 1.5cm 0cm},clip,width=\linewidth]{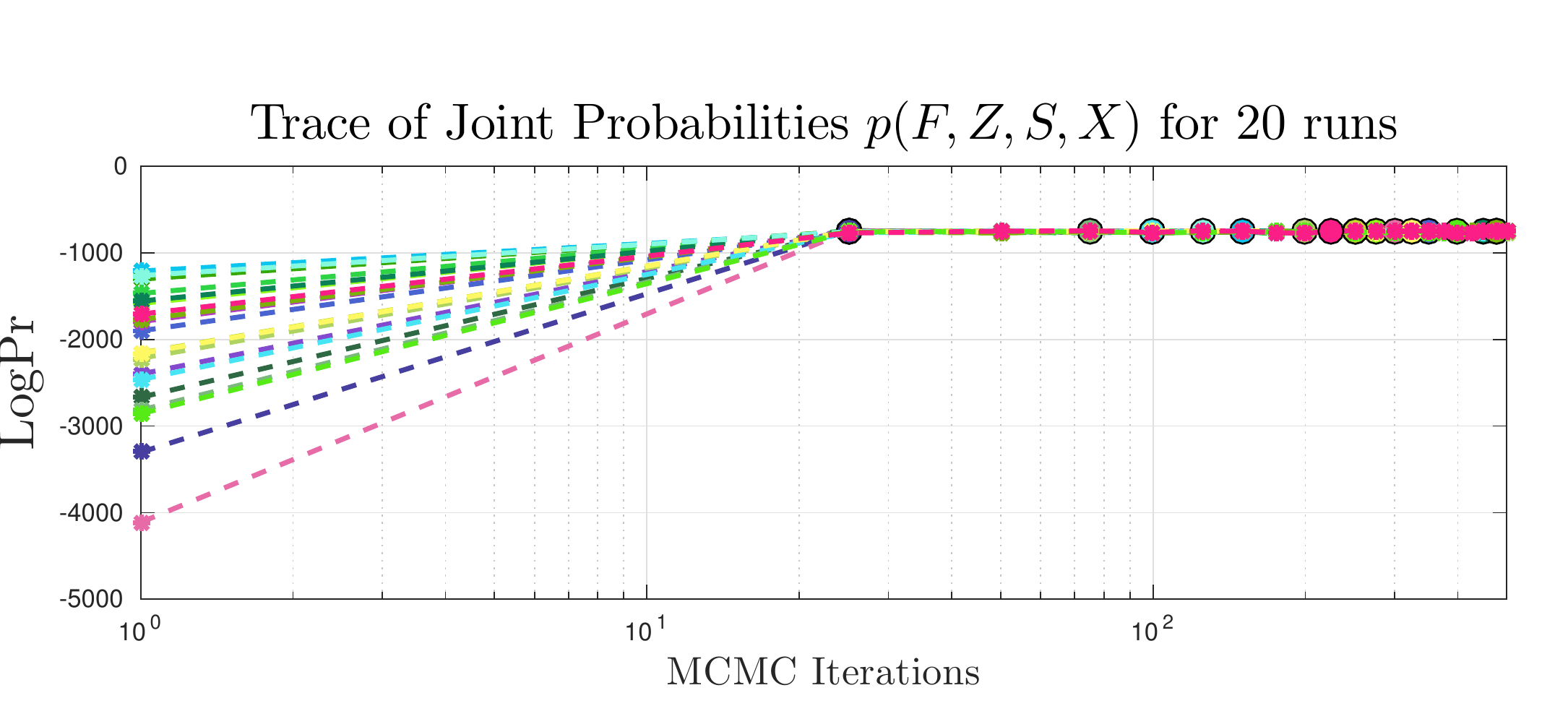}
  \includegraphics[trim={1.5cm 1.8cm 1.5cm 1cm},clip,width=\linewidth]{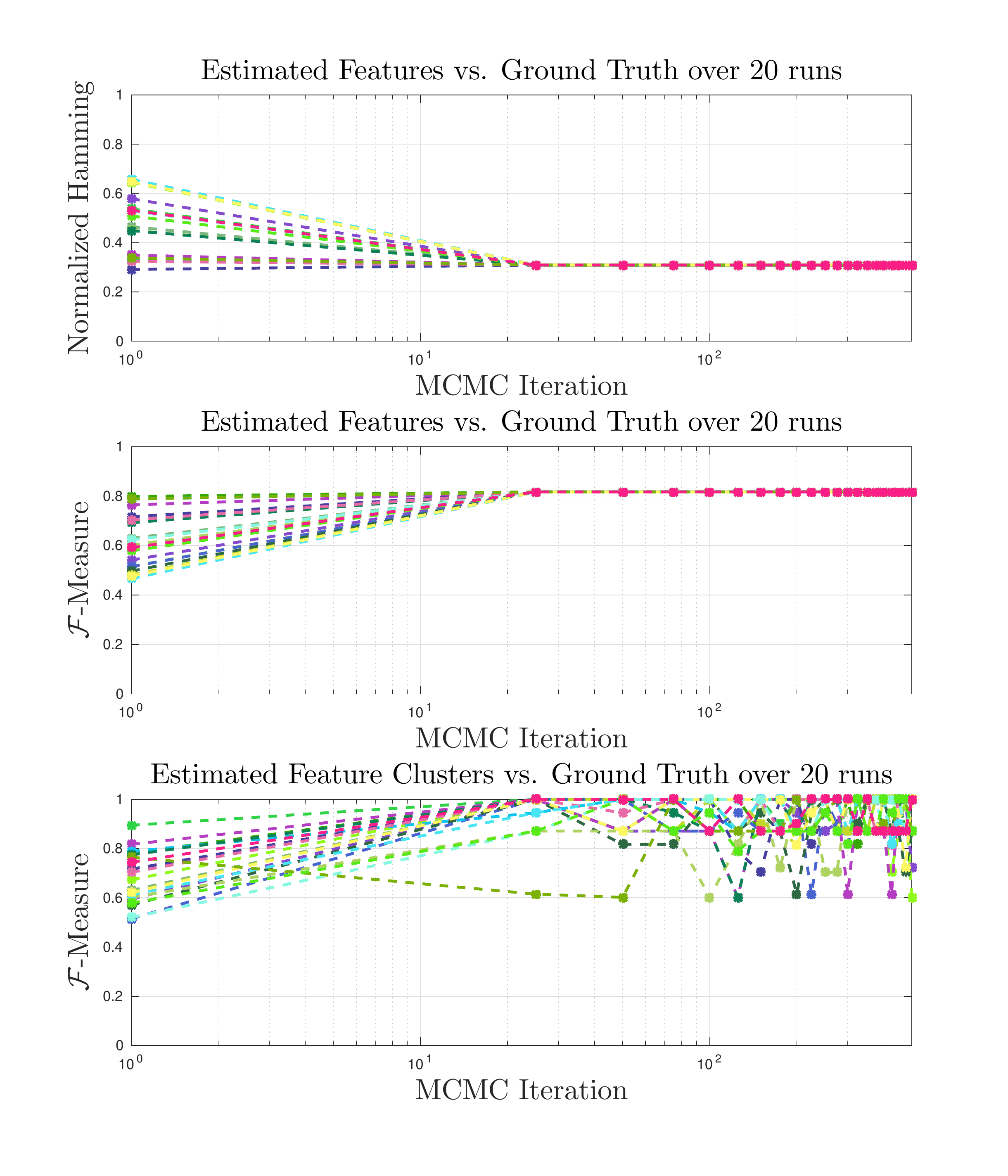}
\end{minipage}\hspace{5pt}\begin{minipage}{0.325\linewidth}
\centering
   \includegraphics[trim={1.25cm 19cm 1.5cm 0cm},clip,width=\linewidth]{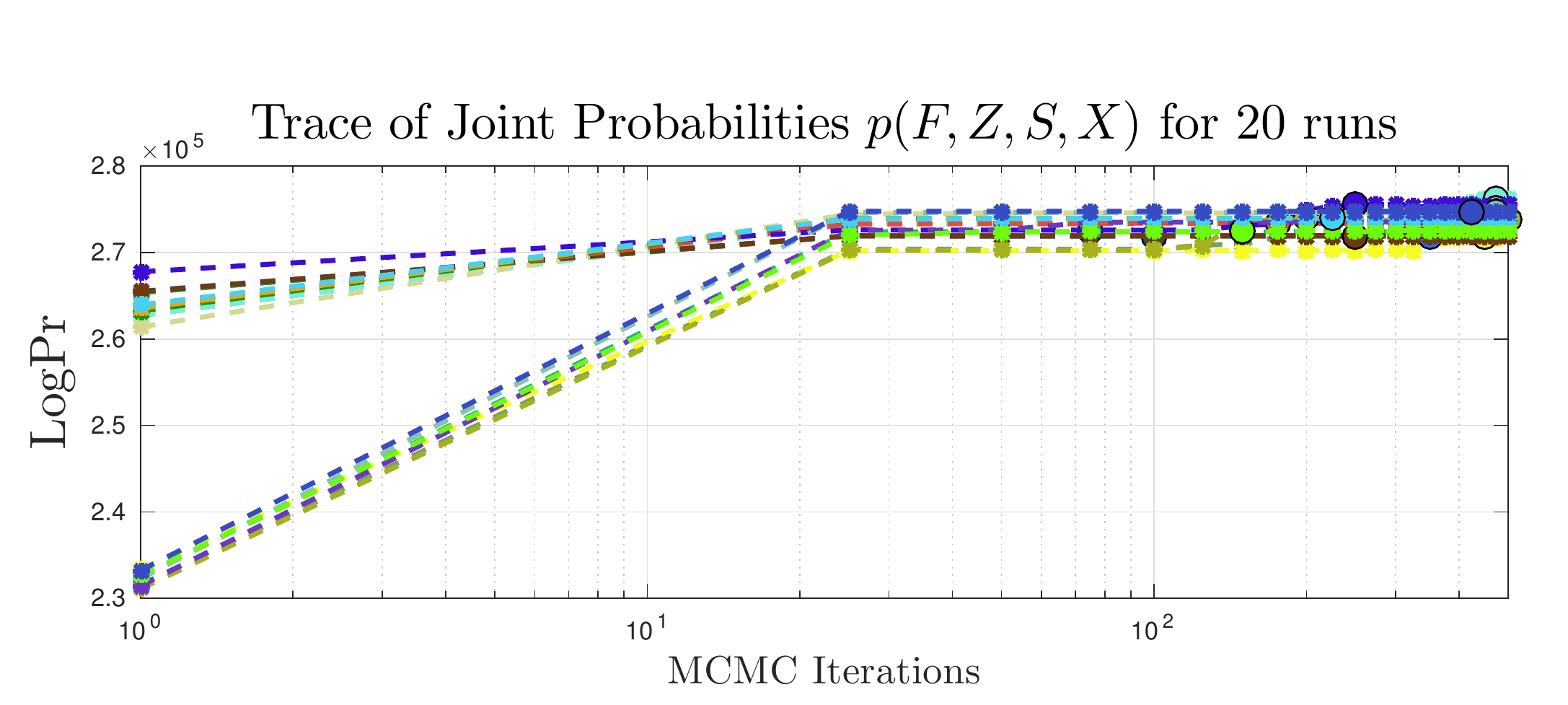}
  \includegraphics[trim={1.5cm 1.8cm 1.5cm 1cm},clip,width=\linewidth]{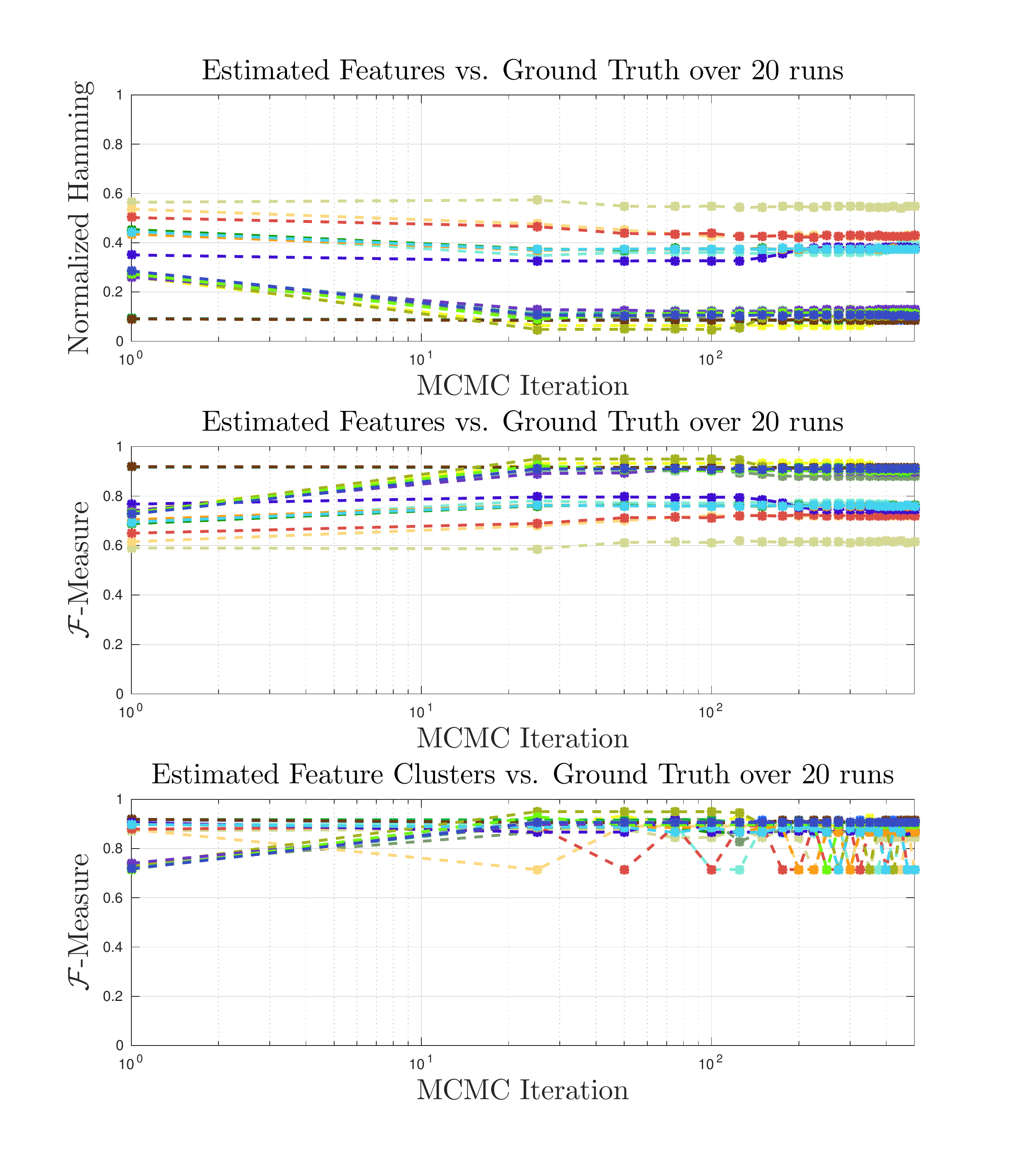}
\end{minipage}\hspace{5pt}\begin{minipage}{0.325\linewidth}
\centering
   \includegraphics[trim={1.25cm 20cm 1.5cm 0cm},clip,width=\linewidth]{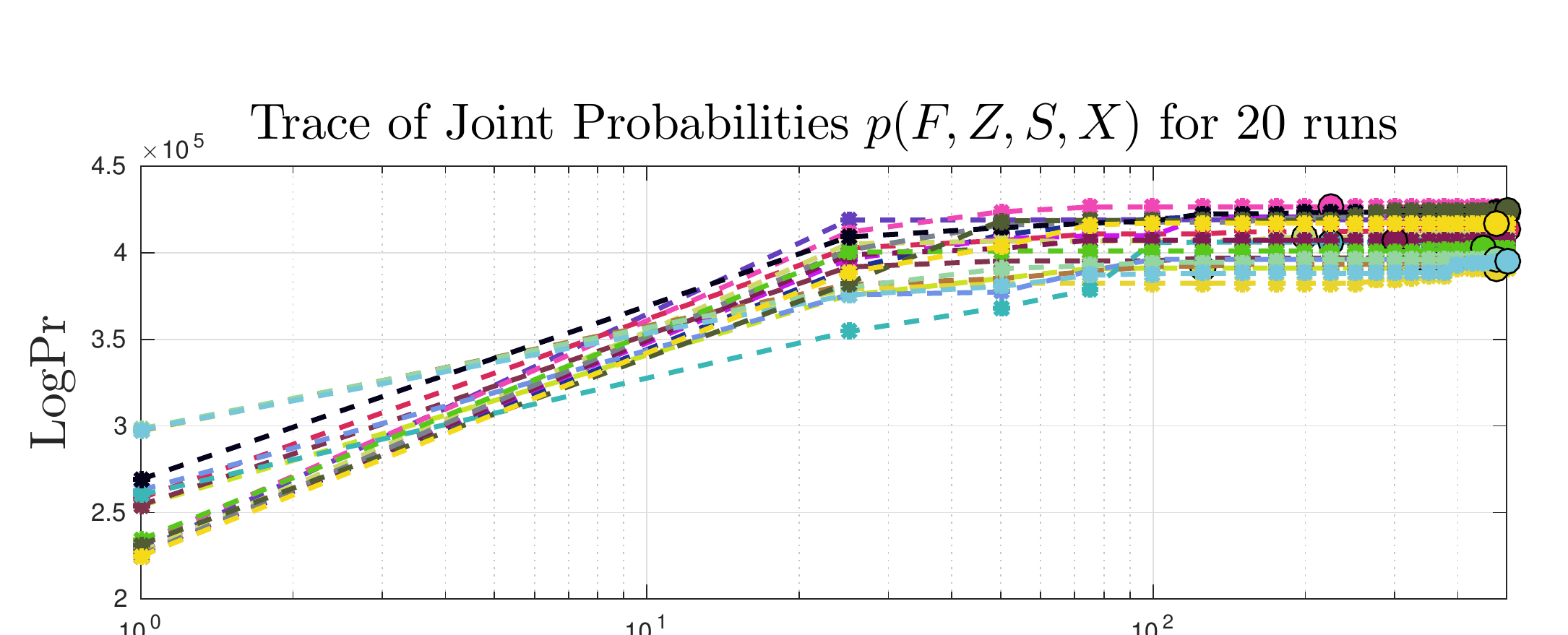}
  \includegraphics[trim={1.5cm 1.7cm 1.5cm 0cm},clip,width=\linewidth]{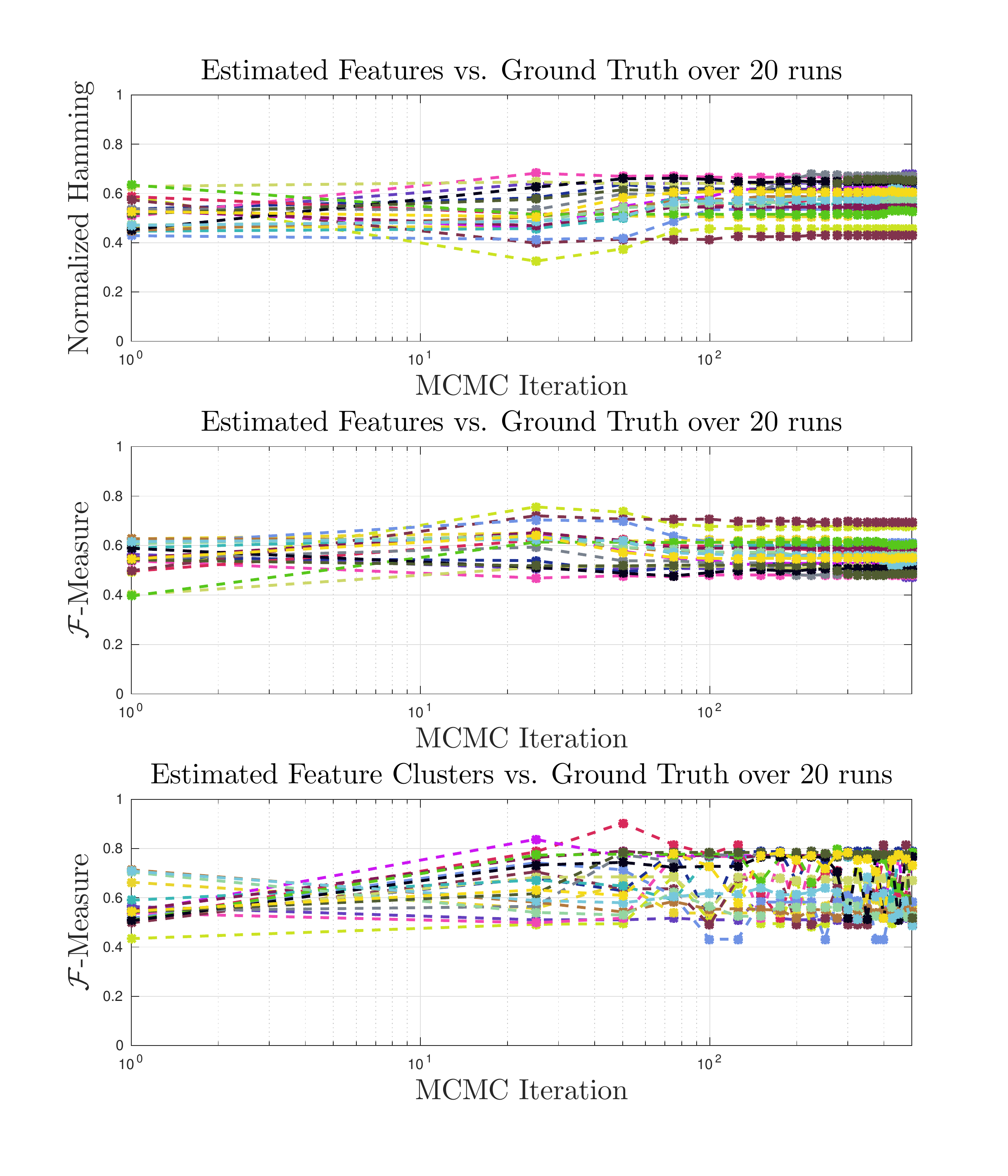}
\end{minipage}
	\captionof{figure}{\small Sampler Tests on (left)\textbf{2D Toy Dataset} (center) and \textbf{Grating Dataset} (right) \textbf{Rolling Dataset}, for 20 random initializations with 500 iterations. \label{fig:sampler_rolling}}
\end{figure*}

To evaluate the convergence properties of the MCMC Sampler for the coupled model we recorded the trace of the log posterior probabilities, accompanied by the Hamming distance (for segmentation evaluation) $\mathcal{F}$-measure (for clustering evaluation) on an Intel® Core™ i7-3770 CPU@3.40GHz$\times$8; for three of our datasets: \textbf{2D Toy Dataset}, \textbf{7D Grating Dataset} and \textbf{13D Rolling Dataset} (see Figure \ref{fig:sampler_rolling}, respectively). 

We devised \textit{sampler tests} where 20 independent chains were run for 500 iterations each. The coupled algorithm has only two hyper-parameters which need to be defined, namely the $\alpha_b$ and $\tau$ pertaining to to the SPCM-CRP prior, these have the same effect on the coupled model as in the SPCM-CRP mixture model. We thus, set $\tau=1$ and randomly sample $\alpha_b$ from a range of $[1,M]$. Moreover, for all runs we randomly sample the initial value of the features $K$ within a range of $[1,M]$ corresponding to the number of shared states for the multiple HMMs and define the initial $K_Z = K$. For all datasets, we observe that the estimated features $F$ and its corresponding \textit{transform-dependent} segmentation (measured through the Hamming distance) stabilize in $< 100 $ iterations. One can also see how the feature clustering $Z$ comes closer to the true clusters as the features $F$ are better estimated. Once features $F$ stabilize, we can see how the SPCM-$\mathcal{CRP}$ mixture explores customer assignments with all chains reaching higher $\mathcal{F}$-measure with $Z$ than with $F$, which is the main desideratum of this algorithm.

\subsubsection{Comparison with other methods}
At its core, our proposed segmentation algorithm can be seen as an extension of a Bayesian non-parametric HMM. The novelty, however, is that we can jointly estimate \textit{transform-invariant} segmentations which are capable of providing a more semantically rich description of a set of time-series data, as opposed to the typical HMM variants. To prove this claim, we compare our approach on the presented datasets against four HMM variants: (i) HMM w/Model Selection, (ii) sticky $\mathcal{HDP}$-HMM, (iii) $\mathcal{IBP}$-HMM and (iv) $\mathcal{IBP}$-HMM + SPCM-$\mathcal{CRP}$ mixture model. The first variant is treated as our baseline, as the expected number of states $K$ is computed through model selection with the AIC/BIC scores. The last variant is simply a decoupled approach where segmentation and feature clustering are performed in a two-step procedure. In this approach, we initially estimate the features $F$ by running the $\mathcal{IBP}$-HMM sampler multiple times and selecting the best iteration for each run. Then, we run the SPCM-$\mathcal{CRP}$ mixture model sampler on the emission models $\Theta$ for 500 iterations and select the best clustering $Z$ based on the highest log posterior. It must be noted that this approach is already biased to yield higher performance as the feature clustering is applied to the ``best features" obtained from the MCMC sampler. Nevertheless, in Table \ref{tab:spcm-segm-compare} we show that our coupled sampler is capable of achieving the same performance and even better in some cases, not only improving feature clustering but also segmentation. 

\begin{table*}[!t]
\begin{center}
\resizebox{\textwidth}{!}{\begin{tabular}{cc|ccc|ccc|cc}
    \hline
    \hline
    \multicolumn{1}{c}{\multirow{4}{*}{Dataset}} & \multicolumn{1}{c}{\multirow{4}{*}{Approaches}} & \multicolumn{6}{c}{\multirow{2}{*}{Metrics (Segmentation | Feature  Clustering)}}
	\\  
    & & \multicolumn{1}{c}{\multirow{2}{*}{Hamming}} & \multicolumn{1}{c}{\multirow{2}{*}{GCE}} & \multicolumn{1}{c}{\multirow{2}{*}{VI}} & \multicolumn{1}{c}{\multirow{2}{*}{Purity}} & \multicolumn{1}{c}{\multirow{2}{*}{NMI}} & 
	\multicolumn{1}{c}{\multirow{2}{*}{$\mathcal{F}$}} & 	\multicolumn{1}{c}{\multirow{2}{*}{$K$}} &
	\multicolumn{1}{c}{\multirow{2}{*}{$K_{Z}$}}
	\\ \\ \hline
    
    \multicolumn{1}{c}{\multirow{4}{*}{Toy 2D}} & HMM w/MS & 0.385 (0.016) & 0.026 (0.082) & 1.025 (0.229) & 0.977 (0.074) & 0.679 (0.109) & 0.752 (0.041)  & 4 & -\\
    
    & sticky HDP-HMM & 0.379 (0.000) & 0.000 (0.001) & 0.956 (0.006) & 1.000 (0.000) & 0.713 (0.002) & 0.765 (0.000) &  4 & - \\
    
    \multicolumn{1}{c}{\multirow{3}{*}{$K_Z$ (2)}}& $\mathcal{IBP}$-HMM & 0.379 (0.000) & 0.000 (0.000) & 0.953 (0.000) & 1.000 (0.000) & 0.714 (0.000) & 0.765 (0.000)  & 4 & - \\
    
    & $\mathcal{IBP}$-HMM + SPCM-$\mathcal{CRP}$  &  \textbf{0.000 (0.000)} & \textbf{0.000 (0.000)} & \textbf{0.000 (0.000)} &  \textbf{1.000 (0.000)} & \textbf{1.000 (0.000)} & \textbf{1.000 (0.000)}  & \textbf{4 (0.0)} & \textbf{2 (0.0)} \\
    & ICSC-HMM & \textbf{0.000 (0.000)} & \textbf{0.000 (0.000)} & \textbf{0.000 (0.000)} &  \textbf{1.000 (0.000)} & \textbf{1.000 (0.000)} & \textbf{1.000 (0.000)}  & \textbf{4 (0.0)} & \textbf{2 (0.0)} \\ \hline
        
    \multicolumn{1}{c}{\multirow{3}{*}{7D Grating}} & HMM w/MS & 0.542 (0.009) & 0.215 (0.022) & 1.827 (0.043) & 0.874 (0.015) & 0.318 (0.037) & 0.596 (0.010) & 5 & -  \\
    \multicolumn{1}{c}{\multirow{3}{*}{Dataset}}    & sticky HDP-HMM & 0.299 (0.190) & 0.150 (0.029) & 1.213 (0.403) & 0.908 (0.023) & 0.502 (0.092) & 0.766 (0.119) & 4.7 (0.8) & - \\
        \multicolumn{1}{c}{\multirow{3}{*}{$K_Z$ (3)}} & $\mathcal{IBP}$-HMM &  0.336 (0.170) & 0.094 (0.029) & 1.208 (0.364) &  \textbf{0.940 (0.024)} & \textit{0.573 (0.066)} & 0.769 (0.100)   & 5.1 (1.28) & - \\
        & $\mathcal{IBP}$-HMM + SPCM-$\mathcal{CRP}$ & \textbf{0.122 (0.044)} & \textit{0.107 (0.045)} & \textit{0.755 (0.114)} & 0.897 (0.050) & 0.572 (0.203) & \textit{0.879 (0.059)} & 5.1 (1.28) & \textbf{3.0 (0.94)}  \\
        & ICSC-HMM & \textit{0.126 (0.075)} & 0.108 (0.031) & \textbf{0.751 (0.231)} & \textit{0.918 (0.022)} & \textbf{0.633 (0.069)} & \textbf{0.890 (0.039)} &   4.6 (1.07) & \textbf{3.4 (1.07)} \\ \hline
	
	\multicolumn{1}{c}{\multirow{3}{*}{13D Rolling}} & HMM w/MS & 0.463 (0.036) & 0.188 (0.024) & 1.897 (0.108) &  \textit{0.865 (0.031)} & \textbf{0.547 (0.026)} & 0.667 (0.036)  & 6 & - \\
	 \multicolumn{1}{c}{\multirow{3}{*}{Dataset}}    & sticky HDP-HMM & 0.630 (0.039) & \textbf{0.176 (0.037)} & 2.611 (0.144)  & \textbf{0.874 (0.031)} & \textit{0.497 (0.027)} & 0.525 (0.038) & 11.9 (1.59) & - \\
	        \multicolumn{1}{c}{\multirow{3}{*}{$K_Z$ (3)}}& $\mathcal{IBP}$-HMM & 0.587 (0.046) & 0.224 (0.022) & 2.507 (0.141) & 0.838 (0.022) & 0.471 (0.021) & 0.555 (0.038) &  9.5 (1.5) & -  \\
	        & $\mathcal{IBP}$-HMM + SPCM-$\mathcal{CRP}$ & \textit{0.362 (0.106)} & 0.226 (0.161) & \textit{1.586 (0.514)}  &  0.687 (0.071) & 0.473 (0.109) & \textit{0.699 (0.093)} & 9.5 (1.5) & \textbf{3.1 (1.6)}\\
	        & ICSC-HMM & \textbf{0.354 (0.083)} & \textit{0.180 (0.099)} & \textbf{1.367 (0.298)} & 0.673 (0.050) & 0.489 (0.102) & \textbf{0.713 (0.076)}  & 9.3 (1.05) & \textbf{3.0 (1.15)} \\ \hline
	
	\multicolumn{1}{c}{\multirow{3}{*}{26D Peeling}} & HMM w/MS & 0.514 (0.030) & 0.400 (0.005) & 2.452 (0.087)  & \textit{0.673 (0.003)} & \textbf{0.420 (0.009)} & 0.545 (0.039) & 6 & - \\
		    \multicolumn{1}{c}{\multirow{3}{*}{Dataset}}    & sticky HDP-HMM & 0.453 (0.055) & 0.386 (0.045) & 2.306 (0.228) & 0.670 (0.018) & 0.392 (0.042) & 0.612 (0.054) & 6.3 (0.95) & - \\
		        \multicolumn{1}{c}{\multirow{3}{*}{$K_Z$ (5)}}& $\mathcal{IBP}$-HMM & 0.413 (0.026) & 0.390 (0.019) & 2.492 (0.071) & \textbf{0.712 (0.017)} & \textit{0.413 (0.019)} & 0.640 (0.018) & 8.4 (0.89) & - \\
		        & $\mathcal{IBP}$-HMM + SPCM-$\mathcal{CRP}$ & \textit{0.397 (0.126)} & \textbf{0.206 (0.154)} & \textbf{1.853 (0.329)} &  0.633 (0.138) & 0.385 (0.207) & \textbf{0.649 (0.112)} & 8.4 (0.89) & \textbf{4.8 (1.92)} \\
		        & ICSC-HMM & \textbf{0.390 (0.088)} & \textit{0.211 (0.108)} & \textit{1.913 (0.183)} & 0.640 (0.090) & 0.375 (0.126) & \textit{0.647 (0.070)} & 6.4 (1.26) & \textbf{4.2 (1.31)}   \\ \hline
	\hline
\end{tabular}}
\end{center}
\vspace{-10pt}
\caption{\small Performance Comparison of Segmentation and state clustering with  \textbf{Our Proposed Approach} vs. HMM w/Model Selection, sticky HDP-HMM, $\mathcal{IBP}$-HMM and decoupled $\mathcal{IBP}$-HMM + SPCM-$\mathcal{CRP}$ \textit{(presenting mean (std) of metrics over 10 runs).} \label{tab:spcm-segm-compare}}
\end{table*}

As in the original $\mathcal{IBP}$-HMM, all dimensions of our time-series have zero-mean. Moreover, in our real-world datasets, the range of each dimension is considerably different, for example the position variables are between $[0.1, 1]$ meters, while the forces range between $[1,25]$, hence, we scale each dimension such that they all lie within a comparable range and the Gaussian distributions estimated by each HMM are not skewed towards a few dimensions. While analyzing the results in Table \ref{tab:spcm-segm-compare} one must take into consideration that these are mere estimates of how well each approach reaches a level of human-segmentation, which in turn might not be the optimal one. Moreover, it is well-known that often such metrics do not fully represent the power of a segmentation algorithm and much papers rely on visual and qualitative evaluations. For this reason, we illustrate segmentations of the 3D end-effector trajectories of each real-world dataset labeled with (i) the ground truth (provided by a human), (ii) \textit{transform-dependent} segmentations taken from $S$ and (iii) \textit{transform-invariant} segmentations taken from $Z$. 

\begin{figure*}[!t]
	\centering
	\begin{minipage}{\textwidth}
	\centering
	\vspace{-10pt}
    \includegraphics[trim={1cm 0.5cm 1.5cm 1cm},clip,width=0.32\linewidth]{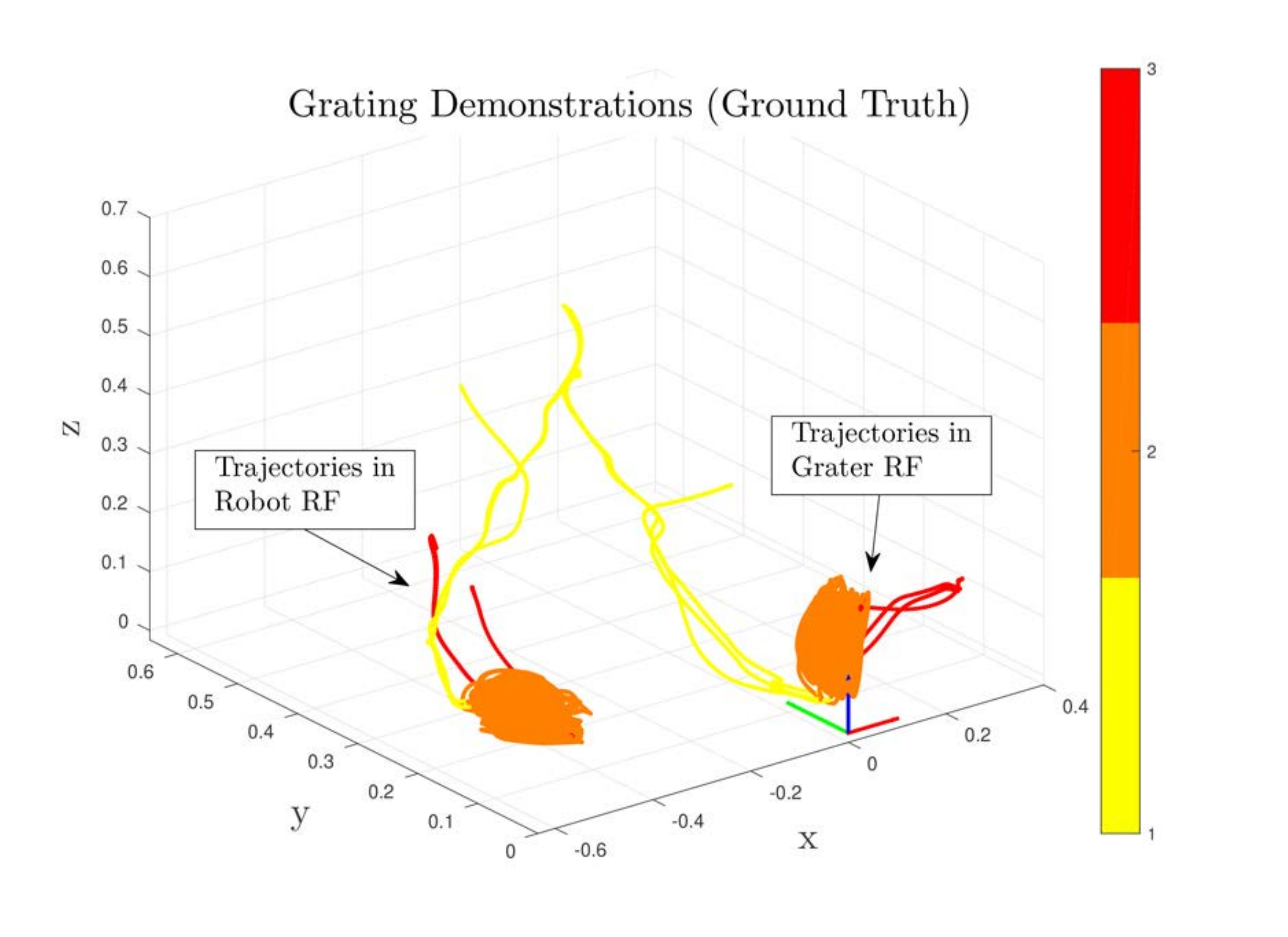}
   \includegraphics[trim={1cm 0.5cm 1.5cm 1cm},clip,width=0.32\linewidth]{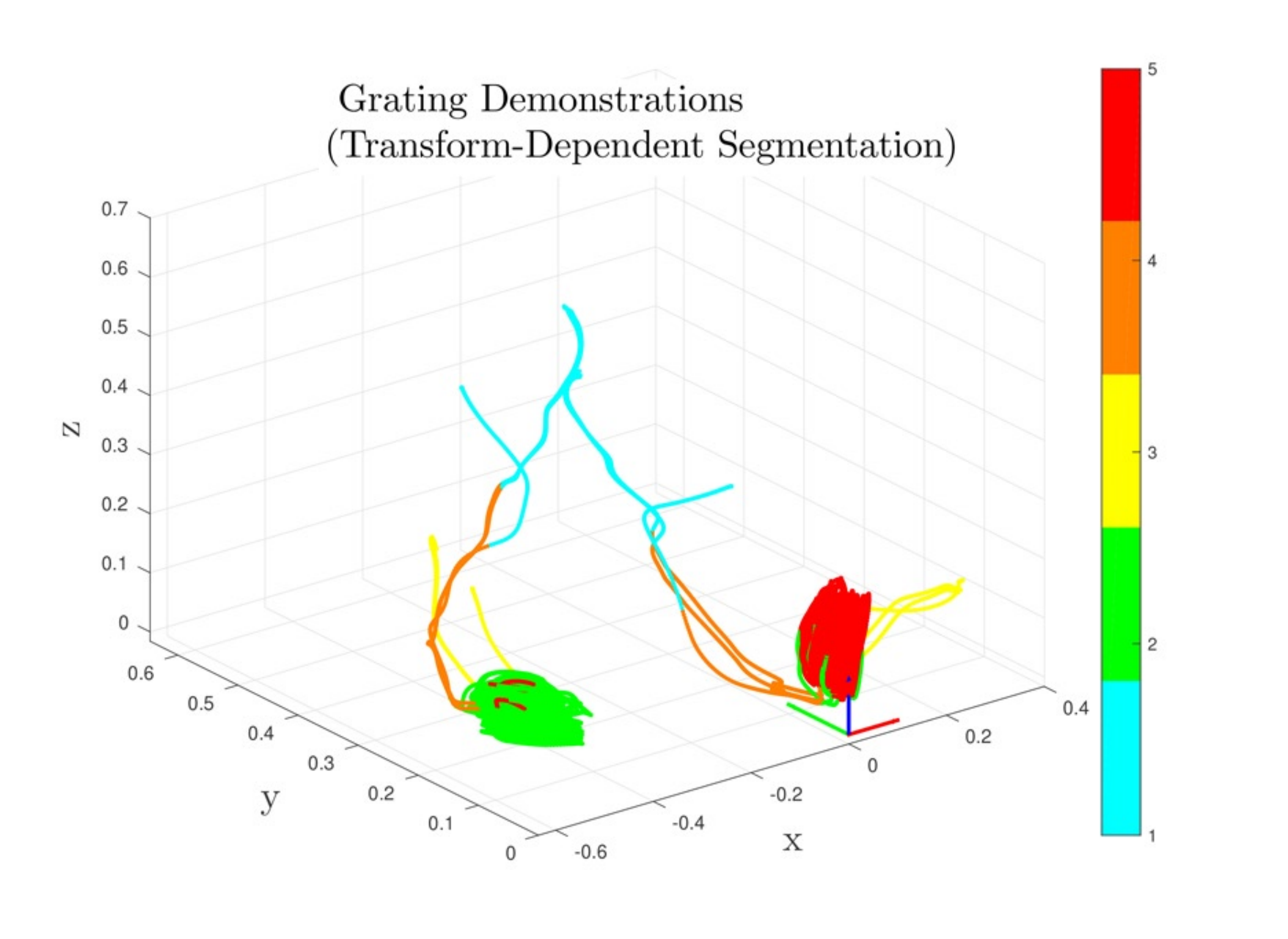}
\includegraphics[trim={1cm 0.5cm 1.5cm 1cm},clip,width=0.32\linewidth]{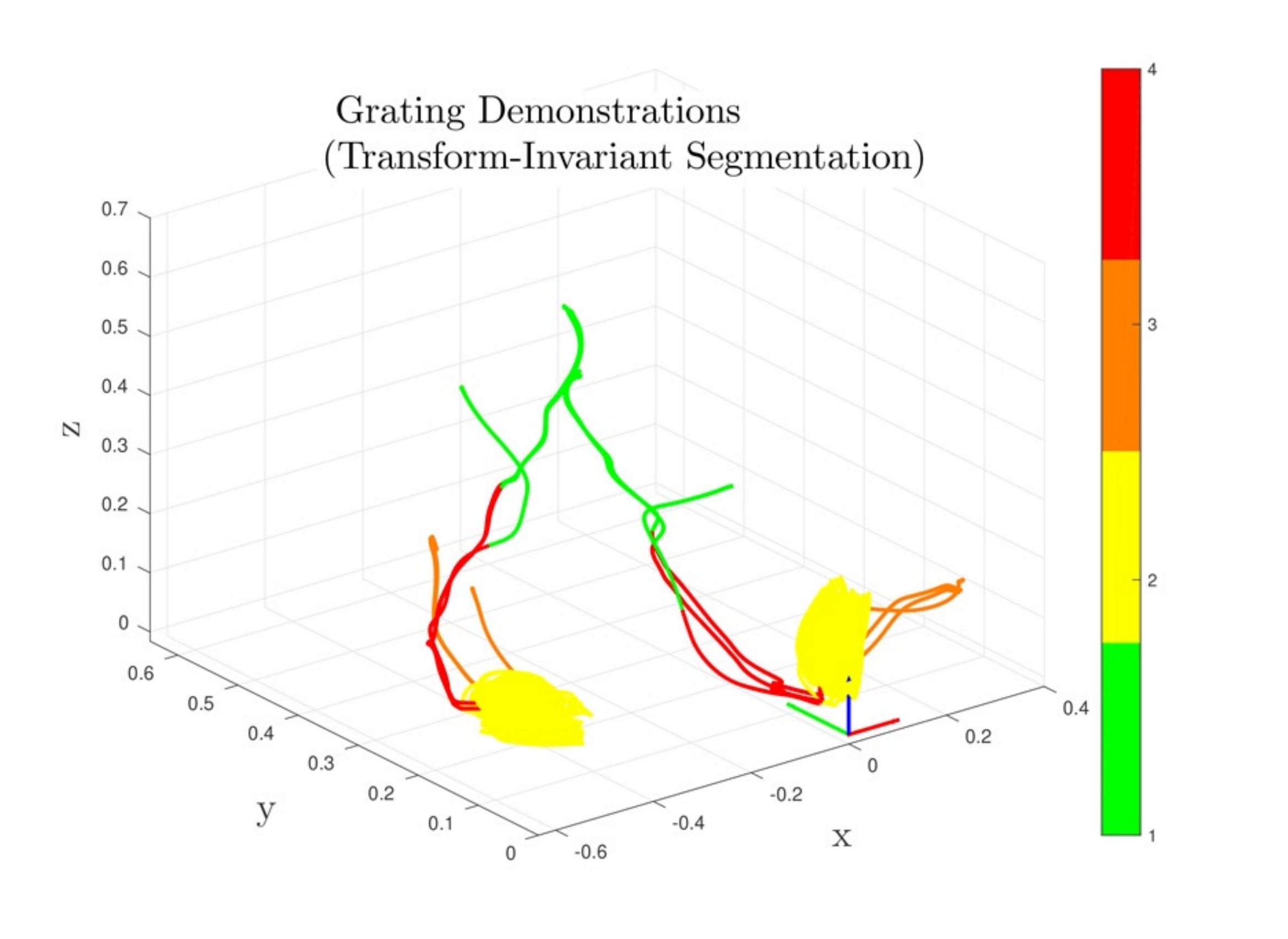}
	\end{minipage}
	
		\begin{minipage}{\textwidth}
		\centering
	    \includegraphics[trim={1cm 0.5cm 1.5cm 1cm},clip,width=0.32\linewidth]{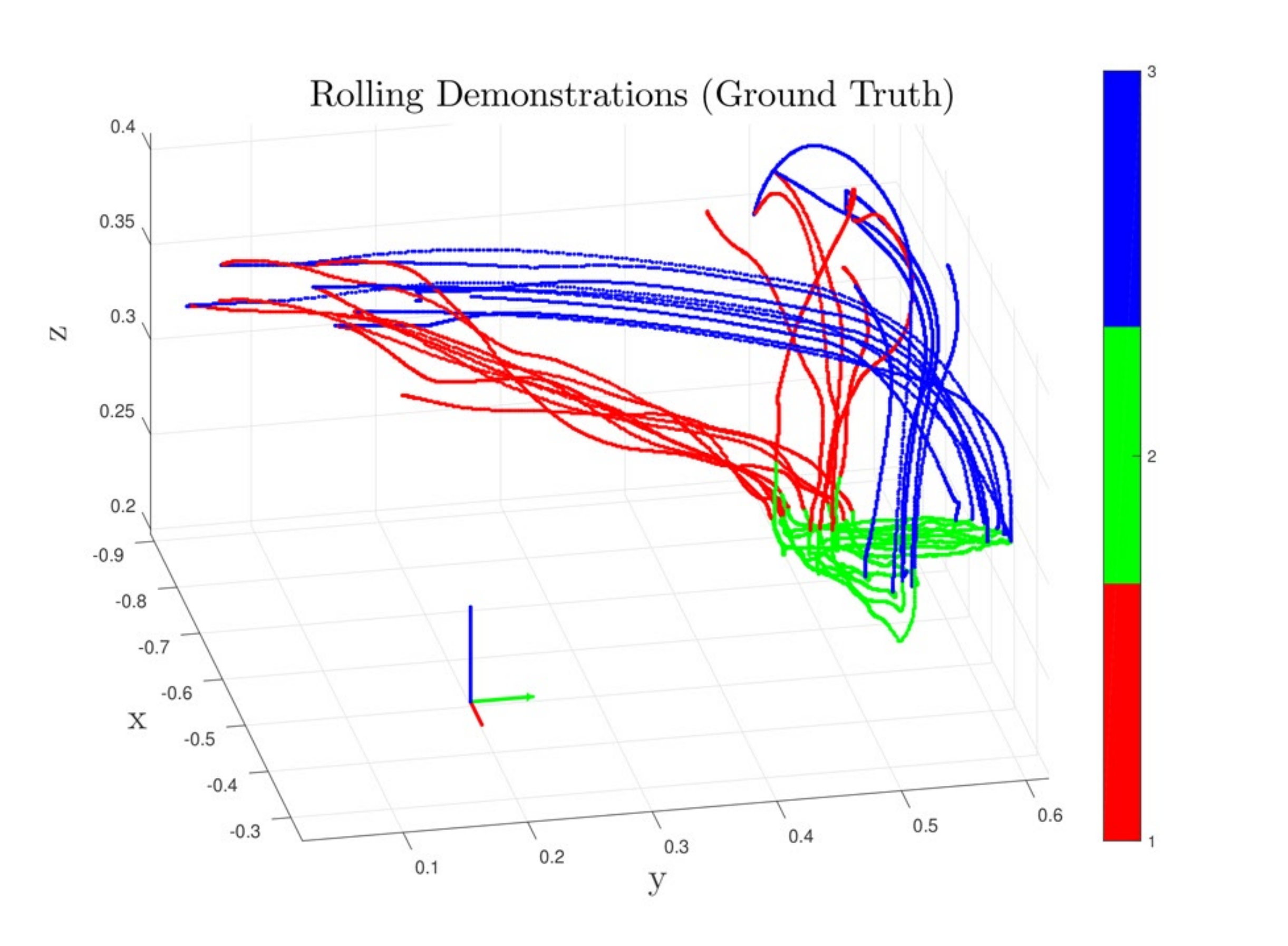}
	   \includegraphics[trim={1cm 0.5cm 1.5cm 1cm},clip,width=0.32\linewidth]{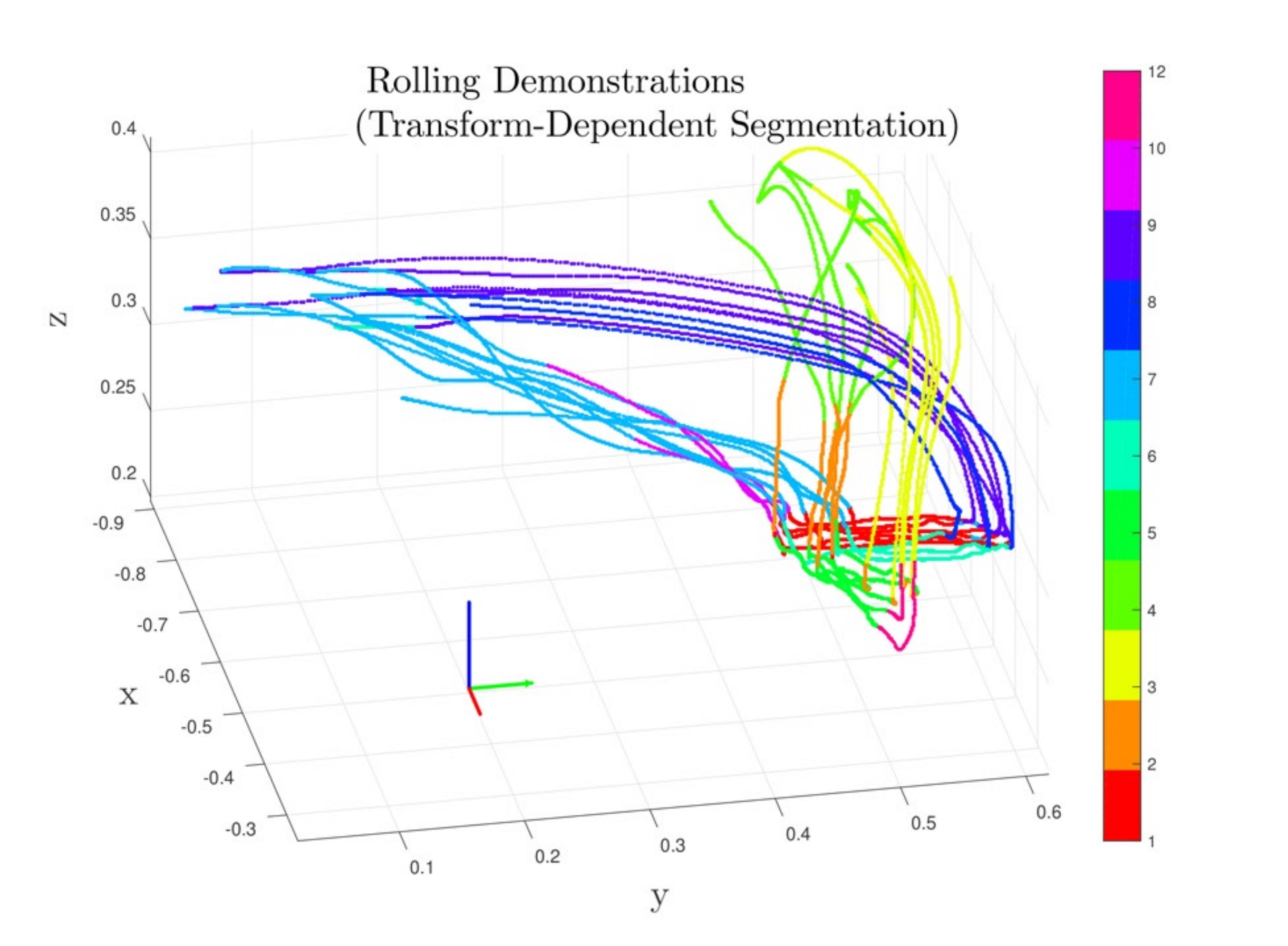}
	\includegraphics[trim={1cm 0.5cm 1.5cm 1cm},clip,width=0.32\linewidth]{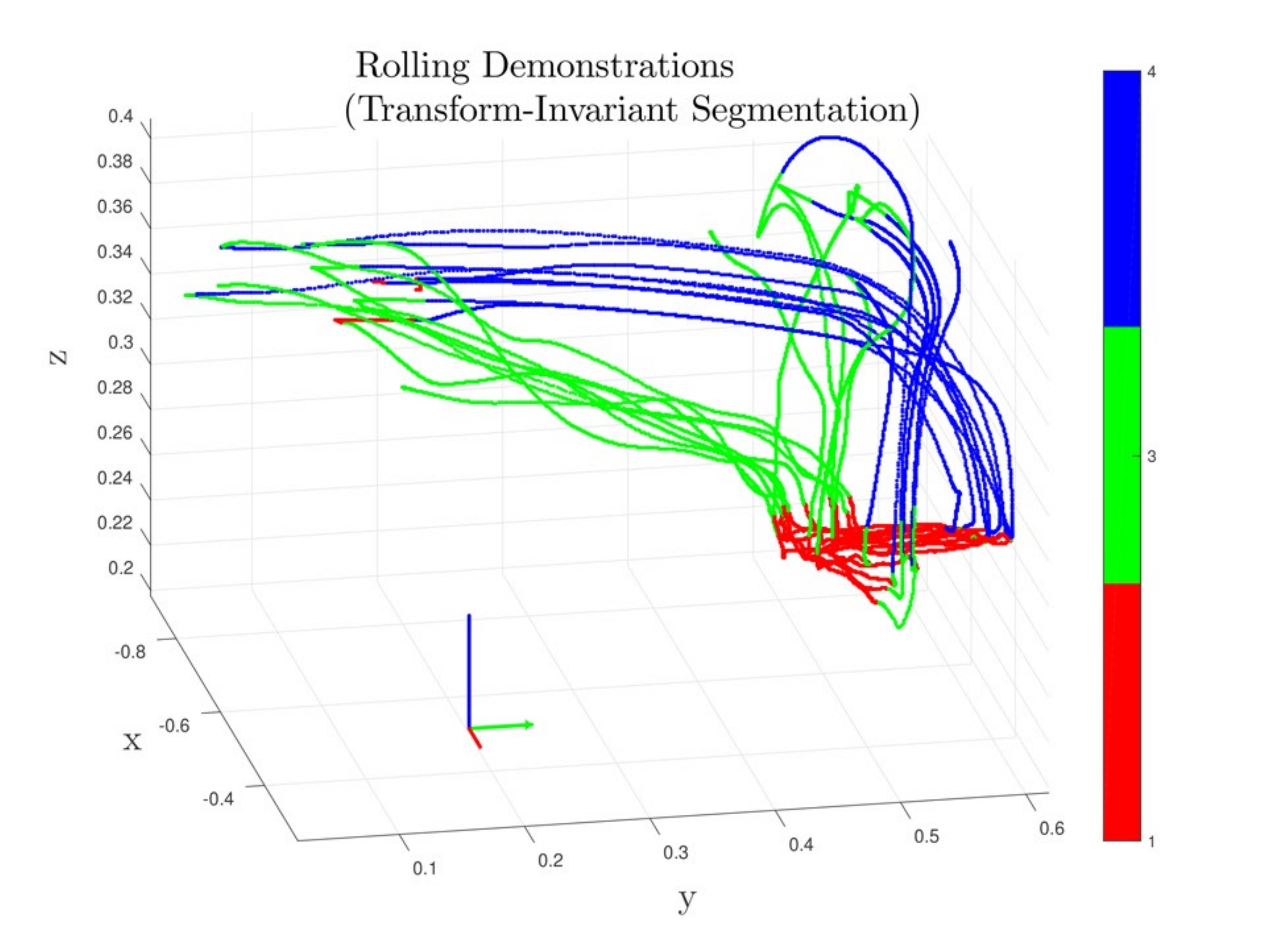}
		\end{minipage}

		\begin{minipage}{\textwidth}
		\centering
	    \includegraphics[trim={1cm 0cm 1.5cm 0.25cm},clip,width=0.32\linewidth]{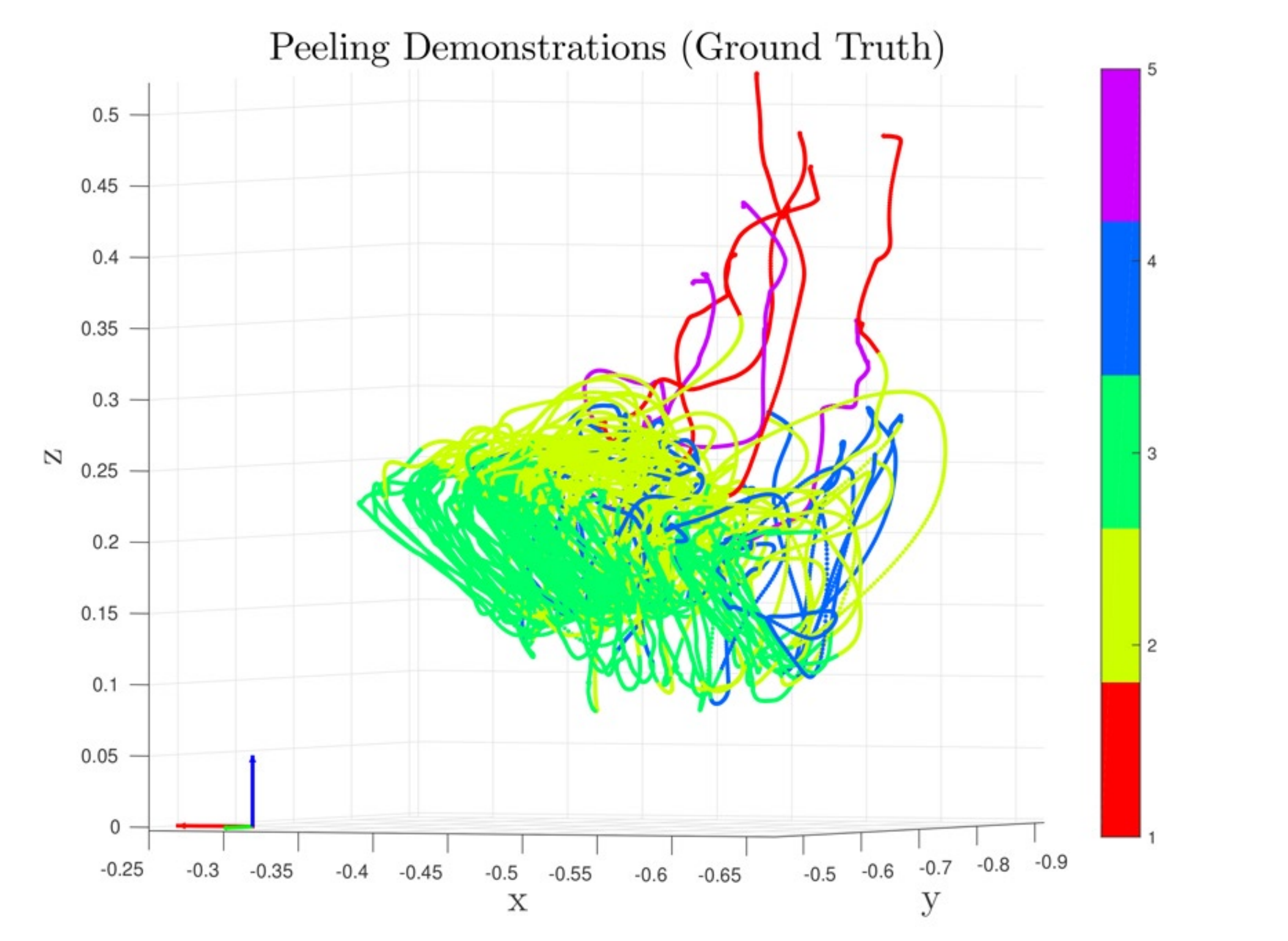}
	   \includegraphics[trim={1cm 0cm 1.5cm 0.25cm},clip,width=0.32\linewidth]{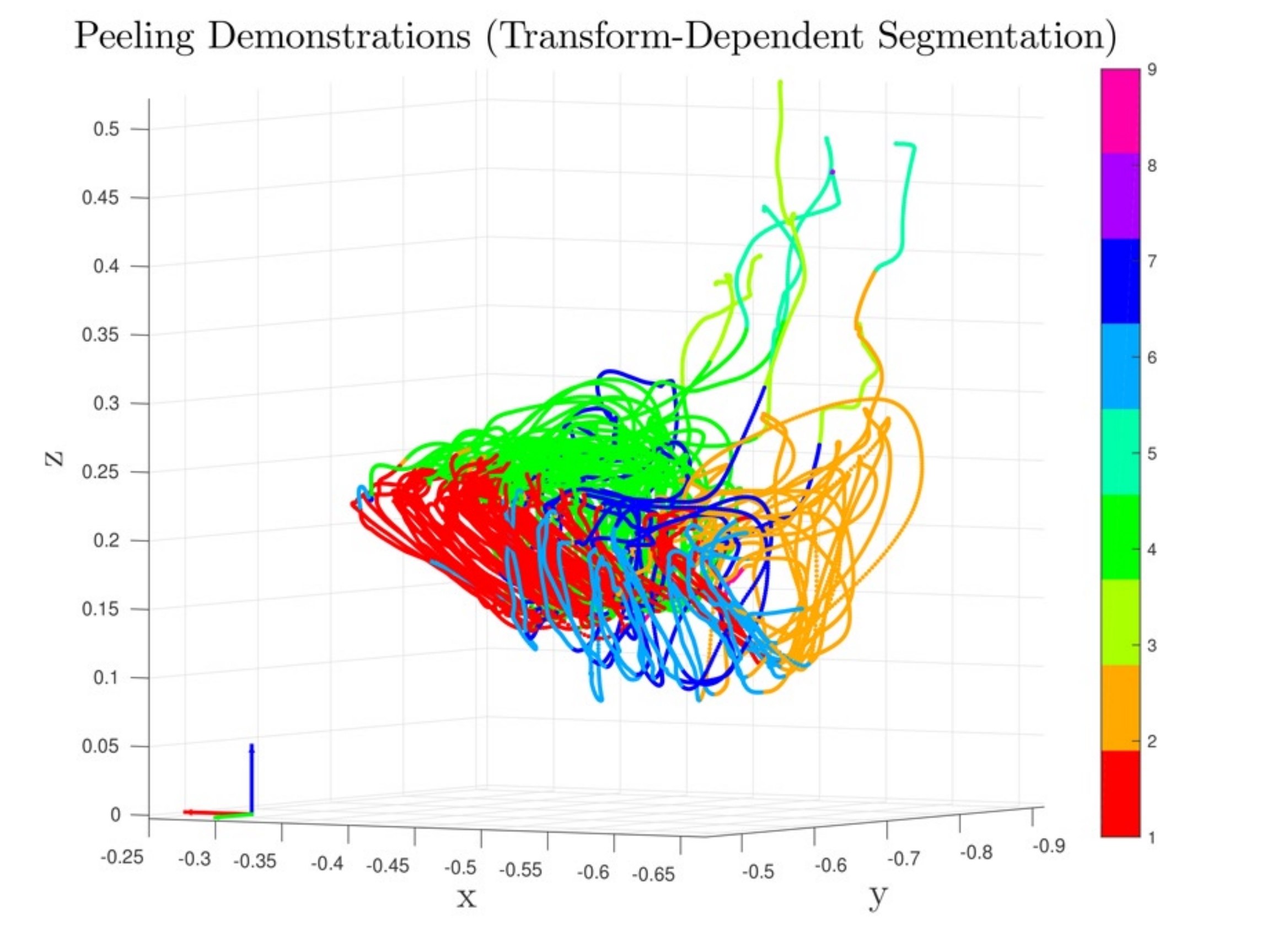}
	\includegraphics[trim={1cm 0cm 1.5cm 0.25cm},clip,width=0.32\linewidth]{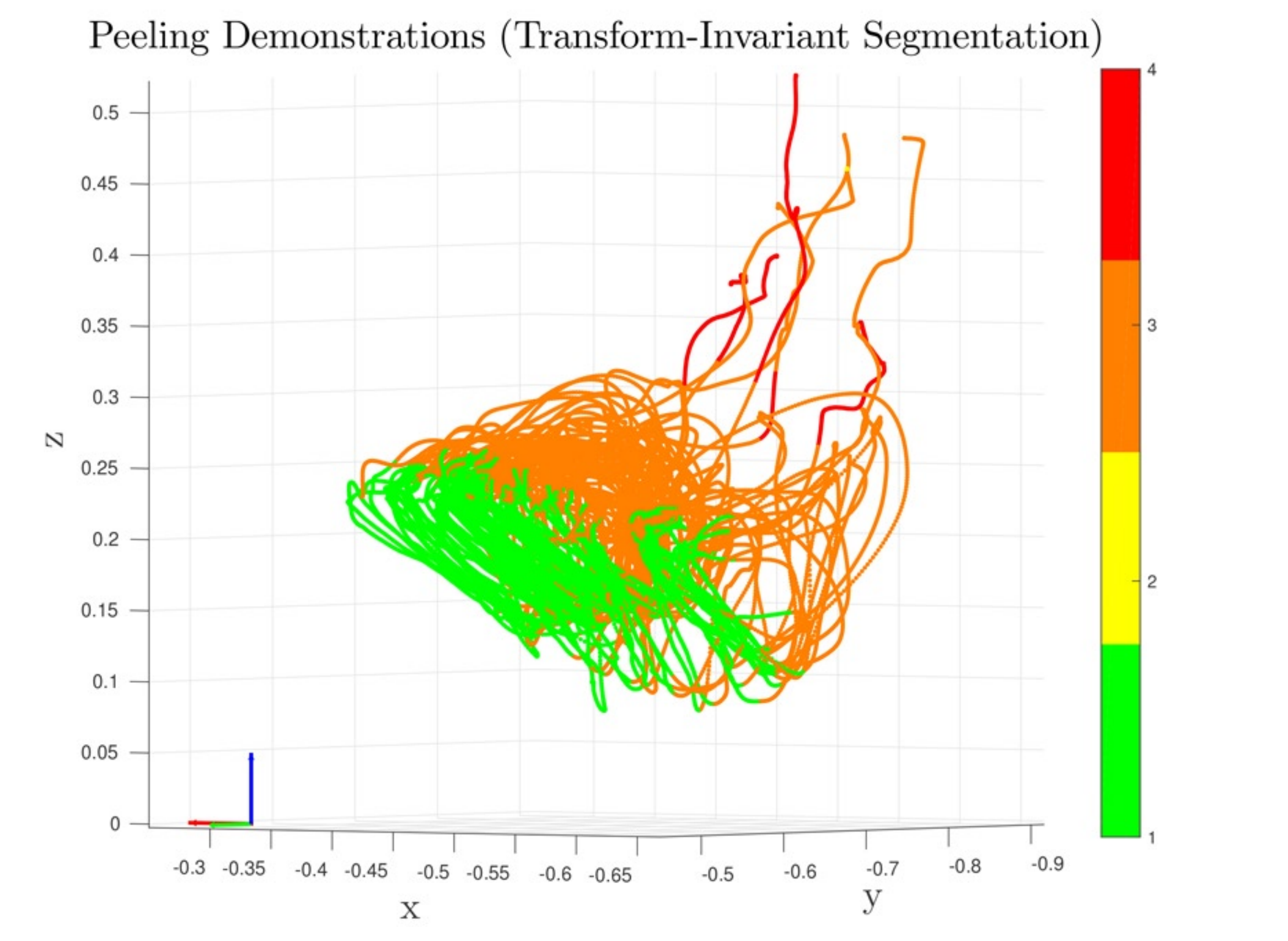}
		\end{minipage}

\vspace{-7pt}
	\captionof{figure}{\small 3D end-effector trajectories with (left) ``true segmentations" defined by humans, (center) \textit{transform-dependent} segmentations from the best selected run of ICSC-HMM over 20 runs with 500 iterations each and the corresponding \textit{transform-invariant} segmentations (right) for grating (top), rolling (middle) and peeling (bottom) datasets; showing the active arm for the latter. \label{fig:segmentation_results}}
	\vspace{-10pt}
\end{figure*}

For the $7D$ Grating Dataset, we see that our coupled approach yields the highest clustering performance as well as comparable segmentation wrt. the decoupled model. However, instead of recovering the 3 expected actions (i) reach (ii) grate and (iii) trash, we recovered 4, this is mainly due to a sub-segmentation of the reaching action into two segments. Such a sub-segmentation is not entirely incorrect, if we analyze the time-series in Figure \ref{fig:segmentation_true}, we can see that there seems to be two phases of the reaching motion: (i) the initial one which varies wrt. it's starting position and (ii) an alignment phase of the vegetable wrt. the grater. To recall, these datasets are intended for robot learning; i.e. through this segmentation algorithm we decompose the task into primitive actions and learn individual dynamics for each, following the approaches in \cite{Pais:IEEE:2015, Figueroa:HRI:2016}. Having an expected primitive being sub-segmented is simply a way of representing the task on a different level of granularity. 

Regarding the $13D$ Rolling dataset, we can see that indeed our coupled model yields the best performance in both feature clustering and segmentation. In this dataset, not only are the time-series transformed, but they are also intersecting each other. This causes the considerably low performance of all of the other HMM variants which assume that all emission models $\Theta$ are used. Given the flexibility of the $\mathcal{IBP}$-HMM we can represent each time-series as a single HMM with different emission models and alleviate this problem. As shown in Figure \ref{fig:segmentation_results}, this yields an excessive number of features. Yet, our coupled model is capable of grouping all of these unique features into a time-series clustering extremely close to the ground truth. 

Finally, in the more challenging but less \textit{transformed} 26D Peeling dataset, we can see that the proposed algorithm yields comparable results to the decoupled approach. Given that each time-series is generated by peeling a different zucchini at a different location we can see how the \textit{transform-dependent} features are excessive, as in the previous dataset. Even though the \textit{transform-invariant} segmentation seems to be quite close to the ground truth the overall metrics yield considerably lower results than the other datasets. This is due to the fact that the \textit{rotate} action (which is extremely similar to the \textit{reach-to-peel}) was not properly extracted as a segment. This however, can be alleviated by adding more features, such as change in color or shape of the manipulated object.

\section{Discussion}
In this work, we presented \textbf{three} main contributions for \textit{transform-invariant} analysis of data. We specifically focused on introducing  (i) a novel measure of similarity in the space of Covariance matrices which allowed us to derive (ii) non-parametric clustering and (iii) segmentation algorithms for applications which seek \textit{transform-invariance}.

The proposed similarity function (B-SPCM) is inspired on spectral graph theory and the geometry of convex sets. Although simply, we proved it's effectiveness on a several of synthetic and real-world datasets with promising results. Even though the presented similarity function is, in fact, specific to Covariance matrices the presented clustering approach (SPCM-$\mathcal{CRP}$ mixture model) is not. In fact, any dataset that has a proper measure of similarity which can generate a matrix $\mathbf{S}$ and a corresponding set of real-valued observations $\mathbf{Y}$, be it a spectral embedding or not, can be clustered with this algorithm. 

Regarding the proposed segmentation and action discovery approach (ICSC-HMM) we presented successful results on difficult datasets which would otherwise need manual tuning and pre-processing. The novelty of our approach does not lie on the fact that we use the $\mathcal{IBP}$-HMM, it lies on the joint estimation of  \textit{transform-invariant} segmentations. One could easily couple the SPCM-$\mathcal{CRP}$ with the $\mathcal{HDP}$-HMM, however, this non-parametric formulation does not provide the flexibility of assuming different switching dynamics which would limit the number of \textit{transform-dependent} states. An interesting extension to this work, would be to handle time-series with missing dimensions. For example, in our last dataset (26-D Peeling), we indeed have two sets of time-series with 6 extra features, corresponding to color change, which help disambiguate two actions for those specific time-series. However, since the features are non-existent in the remaining time-series, we could not use them in this analysis. Nevertheless, since the $\mathcal{IBP}$-HMM provides the flexibility of having unique features per time-series, this can be exploited to model Gaussian emission of different dimensions and a new metric could be introduced which handles such differences. 

\acks{This research was supported by the European Union via the H2020 project \textit{COGIMON}  $ H2020-ICT-23-2014  $.
}
\appendix
\section{Sampling from the $\mathcal{NIW}$ distribution}
\label{app:Sample_NIW}
The $\mathcal{NIW}$ \citep{Gelman:BDA:2003} is a four-parameter $\lambda = \{\mu_0,\kappa_0,\Lambda_0,\nu_0\}$ multivariate distribution generated by
$\Sigma \sim \mathcal{IW}(\Lambda_0,\nu_0), \quad \mu|\Sigma \sim$ and  $\mathcal{N}\left(\mu_0,\frac{1}{\kappa_0}\Sigma\right)$ where $\kappa_0,\nu_0 \in \mathds{R}_{>0}$, and $\nu_0 > P -1$ indicates degrees of freedom of the $P$-dimensional scale matrix $\Lambda \in \mathds{R}^{PxP}$ which should be $\Lambda \succ 0$.
The density of the $\mathcal{NIW}$ is defined by
\begin{equation}
\begin{aligned}
p(\mu, \Sigma \hspace{2pt} | \hspace{2pt} \lambda) & = \mathcal{N}\left(\mu | \mu_0,\frac{1}{\kappa_0}\Sigma\right)\mathcal{IW}(\Sigma \hspace{2pt}|\hspace{2pt} \Lambda_0, \nu_0)\\
& = \frac{1}{Z_0}|\Sigma|^{-[(\nu_0 + d)/2 + 1]} \exp \left\lbrace -\frac{1}{2} \texttt{tr}(\Sigma^{-1}\Lambda_0)  \right\rbrace\\
& \times  \exp \left\lbrace -\frac{\kappa_0}{2} (\mu - \mu_0)^{T}\Sigma^{-1}(\mu - \mu_0) \right\rbrace\\
\end{aligned}
\end{equation}
where $Z_0 =\frac{2^{\nu_0 d/2}\Gamma_d(\nu_0 /2)(2\pi/\kappa_0)^{d/2}}{|\Lambda_0|^{\nu_0/2}} $ is the normalization constant.

A sample from a $\mathcal{NIW}$ yields a mean $\mu$ and covariance matrix $\Sigma$. One first samples a matrix from an $\mathcal{W}^{-1}$ parameterized by $\Lambda_0$ and $\nu_0$; then $\mu$ is sampled from a $\mathcal{N}$ parameterized by $\mu_0, \kappa_0, \Sigma$. Since $\mathcal{N}$ and $\mathcal{NIW}$ are a conjugate pair, the term $\left( \prod_{i \in \mathbf{Z}(C)=k} p\left( \mathbf{y}_{i} \hspace{2pt}|\hspace{2pt} \theta \right) \right)  p\left( \theta\hspace{2pt}|\hspace{2pt}\lambda \right)$ in \eqref{eq:integral} also follows a $\mathcal{NIW}$ \citep{Murphy:CBG:2007} with new parameters $ \lambda_n = \{\mu_n,\kappa_n,\Lambda_n,\nu_n\}$ computed via the following posterior update equations,
\begin{equation}
\label{eq:niw_updates}
\begin{aligned}
p(\mu,\Sigma|\mathbf{Y}_{1:n},\lambda) & = \mathcal{NIW}(\mu,\Sigma | \mu_n,\kappa_n,\Lambda_n,\nu_n)\\
\kappa_n = \kappa_0 + n,  \qquad &\nu_n = \nu_0 + n, \qquad \mu_n = \frac{\kappa_0\mu_0 + n\bar{\mathbf{Y}}}{\kappa_n}\\
\qquad \Lambda_n =  \Lambda_0 & + S + \frac{\kappa_0n}{\kappa_n}(\bar{\mathbf{Y}} - \mu_0)(\bar{\mathbf{Y}} - \mu_0)^T
\end{aligned}
\end{equation}
where $n$ is the number of samples $\mathbf{Y}_{1:n}$, whose sample mean is denoted by $\bar{\mathbf{Y}}$ and $S$ is the scatter matrix, as introduced earlier. 

\section{Split-merge sampler for the $\mathcal{IBP}$-HMM \citep{Hughes:NIPS:2012}}
\label{app:split-merge}
A proposal selects anchor objects and features to split/merge at random to improve upon finding anchors that are \textit{similar}. Following are the algorithmic details of such sampler:
\begin{enumerate}[leftmargin=*]
\item Randomly select a pair of sequences $\{i,j\}$
\item Select candidate feature pair $\{k_i,k_j\}$ by first selecting a random feature $k_i$, then select $k_j$ given the following proposal distribution:
\begin{equation}
\begin{split}
q_k(k_i,k_j) = \text{Unif}(k_i|f_i)q(k_j|k_i,f_j) \quad \text{where}\\
q(k_j|k_i,f_j) \begin{cases}\!
2C_jf_{jk} & \text{if} \quad k = k_i  \\
f_{jk}\frac{m(\mathbf{x}_{k_i},\mathbf{x}_k)}{m(\mathbf{x}_{k_i})m(\mathbf{x}_k)} &  \text{otherwise}. \\
\end{cases}
\label{eq:split-merge}
\end{split}
\end{equation}
Here $m(\cdot)$ is the marginal probability of the data given the state sequence $\{z^{(i)}\}_{i=1}^M$ (collapsing away $\Theta$) and $C_j = \sum_{k_i \neq k_j} f_{jk}m(\mathbf{x}_{k_i},\mathbf{x}_k)/m(\mathbf{x}_{k_i})m(\mathbf{x}_k)$.
\item One then computes the MH Ratio to accept/reject a merge/split with the following distribution:
\begin{equation}
q(\Psi^*|\Psi) = q_k(k_i,k_j)q(k_j|k_i,f_j)
\end{equation}
, where $\Psi$ is the current state of the Markov Chain and $\Psi^*$ is the proposed change.
\end{enumerate}
\eqref{eq:split-merge} encourages choices $k_i = k_j$ for a
feature merge that explains \textit{similar} data via the \textit{marginal likelihood ratio}. A large ratio indicates that the data assigned to both $k_i , k_j$ are better modeled together rather than independently. 

\section{External Clustering Metrics}
\label{app:clustering_metrics}
\begin{enumerate}[leftmargin=*]
\item \underline{\textit{Purity}} is a simple metric that evaluates the quality of the clustering by measuring the number of clustered data-points that are assigned to the same class, as follows,
\begin{equation}
Purity(\mathcal{S},\mathcal{C}) = \frac{1}{N} \sum_{j} \underset{k}{\max}|s_j\cap c_k|
\end{equation}
where $\mathcal{S} = \{s_1,\dots,s_J\}$ is the set of classes and $\mathcal{C} = \{c_1,\dots,c_K\}$ the set of predicted clusters. $s_j$ is the set of data-points in the $j$-th class, whereas $c_k$ is the set of data-points belonging to the $k$-th cluster \citep{Manning:2008:IIR}.
\item \textit{Normalized Mutual Information \underline{(NMI)}}: is an information-theoretic metric, which measures the trade-off between the quality of the clustering  and the total number of clusters, as follows,
\begin{equation}
NMI(\mathcal{S},\mathcal{C}) = \frac{\mathcal{I}(S,C)}{[\mathcal{H}(S) + \mathcal{H}(C)]/2}
\end{equation}
for mutual information $\mathcal{I}(\mathcal{S},\mathcal{C}) =  \sum_{j}\sum_{k}P(s_j\cap c_k) \log\frac{P(s_j\cap c_k)}{P(s_j)P(c_k)}$ and  
and entropy $
\mathcal{H}(\mathcal{C})  = -\sum_{k}P(c_k)\log P(c_k)$
Both $\mathcal{I}(\mathcal{S},\mathcal{C}) $ and $\mathcal{H}(\mathcal{C})$ have closed-form ML estimates \cite{Manning:2008:IIR}.
\item The \underline{$\mathcal{F}$-measure} is a well-known classification metric which represents the harmonic mean between Precision ($P = \frac{TP}{TP+FP}$) and Recall ($R = \frac{TP}{TP+FN}$). In the context of clustering, Recall and Precision of the $k$-th cluster wrt. the $j$-th class are $R(s_j,c_k) = \frac{|s_j \cap c_k|}{|s_j|}$ and $P(s_j,c_k) = \frac{|s_j \cap c_k|}{|c_k|}$, respectively. The $\mathcal{F}$-measure of the $k$-th cluster wrt. the $j$-th class is then, 
\begin{equation}
\mathcal{F}_{j,k} = \frac{2 P(s_j,c_k) R(s_j,c_k)}{P(s_j,c_k) + R(s_j,c_k)},
\end{equation}
and the $F$-measure for the overall clustering is computed as,
\begin{equation}
\mathcal{F}(\mathcal{S},\mathcal{C}) =\sum_{s_j \in \mathcal{S}} \frac{|s_j|}{|\mathcal{S}|}\underset{k}{\max}\{\mathcal{F}_{j,k}\}.
\end{equation}
\end{enumerate}

\section{External Segmentation Metrics}
\label{app:segmentation_metrics}

\begin{enumerate}[leftmargin=*]
\item The \textit{\underline{Hamming}} distance measures the distance between two sets of strings (or vectors) as the number of mismatches between the sets. In our case, the segmentation labels $S^{true} \in F^{K^{true}}, S^{est} \in F^{K^{est}}$ may possess different values. After finding the correspondence between $S^{est} \rightarrow S^{true}$ by solving the assignment problem \cite{Mulmuley1987}, one computes: 
\begin{equation}
d(S^{true},S^{est}) = \sum_{s_{i}^{est} \in S^{est}} \left(\sum_{s_k^{true} \neq s_j^{true},s_k^{true} \cap s_i^{est} \neq 0} \left|s_k^{true} \cap s_i^{est}\right| \right),
\label{eq:hamming}
\end{equation}
which corresponds to the total area of intersections between $S^{est}$ and $S^{true}$\cite{Hamming}.\eqref{eq:hamming} can be normalized as follows: $d(S^{true},S^{est})/|S^{true}|$.

\item The \textit{Global Consistency Error \underline{(GCE)}} is a measure that takes into account the differences in granularity while comparing two segmentations \cite{MartinFTM01}. If one segment is a proper subset of the other it is considered as an area of \textit{refinement}, rather than an error. Thus, for two sets of segmentations $S^{true} \in F^{K^{true}}, S^{est} \in F^{K^{est}}$, the GCE is computed as follows:  
\begin{equation}
\begin{split}
GCE(S^{true},S^{est}) & =  \\ 
\frac{1}{M_s}\min & \left\{ \sum\limits_{i} E( S^{true},S^{est}, s_i), E( S^{est},S^{true}, s_i), \right\}
\end{split}
\label{eq:gce}
\end{equation}
for label location $s_i$ and local refinement error $E(S^1,S^2,s_i)$ which measures the degree to which two segmentations agree at label location $s_i$ and $M_s$ is the size of the segment containing $s_i$.

\item \textit{Variation of Information \underline{(VI)}} is a metric which defines the distance between two segmentations as the average conditional entropy of one segmentation given the other, as follows:
\begin{equation}
VI(S^{true}, S^{est}) = \mathcal{H}(S^{true}) + \mathcal{H}(S^{est}) - 2\mathcal{I}(S^{true},S^{est})
\label{eq:vi}
\end{equation}
where $\mathcal{H}(\cdot)$ and $\mathcal{I}(\cdot,\cdot)$ are computed as in \ref{app:clustering_metrics} Intuitively, \eqref{eq:vi} is a measure of the amount of randomness in one segmentation that cannot be explained by the other \cite{Meila:2005:CCA}.
\end{enumerate}


\vskip 0.2in
\bibliography{Bibliography_long.bib}

\end{document}